%% file: Gaussian_regression.tex
\documentclass[11pt]{amsart}

\usepackage{Style_files/Packages}
\usepackage{Style_files/New_commands}
\usepackage{Style_files/Style}
\usepackage{comment}
\usepackage{etoolbox}

\title{On Learning Gaussian Multi-index Models with Gradient Flow}
\author{Alberto Bietti$^\dagger$, Joan Bruna$^{\ast, \star, \dagger}$ and Loucas Pillaud-Vivien$^{\ast, \dagger}$}
\address{$^\ast$ Courant Institute of Mathematical Sciences,  New York University}
\address{$^\star$ Center for Data Science, New York University}
\address{$^\dagger$ Center for Computational Mathematics, Flatiron Institute}

\begin{document}
\normalem

\begin{abstract}
We study gradient flow on the multi-index regression problem for high-dimensional Gaussian data. Multi-index functions consist of a composition of an unknown low-rank linear projection and an arbitrary unknown, low-dimensional link function. As such, they constitute a natural template for feature learning in neural networks.

We consider a two-timescale algorithm, whereby the low-dimensional link function is learnt with a non-parametric model infinitely faster than the subspace parametrizing the low-rank projection. By appropriately exploiting the matrix semigroup structure arising over the subspace correlation matrices, we establish global convergence of the resulting Grassmannian population gradient flow dynamics, and provide a quantitative description of its associated `saddle-to-saddle' dynamics. Notably, the timescales associated with each saddle can be explicitly characterized in terms of an appropriate Hermite decomposition of the target link function. In contrast with these positive results, we also show that the related \emph{planted} problem, where the link function is known and fixed, in fact has a rough optimization landscape, in which gradient flow dynamics might get trapped with high probability.
\end{abstract}

\maketitle

\tableofcontents

\clearpage

\input{I-Introduction.tex}

\input{0-Notations}

\clearpage

\input{III-Harmonic_analysis}

\input{IV-Grassmannian}

\input{V-Planted_model}

\input{VI-Perspectives}

\input{conclusions}

\bibliographystyle{apalike}

\bibliography{Gaussian_regression.bib}

\clearpage

\input{Appendix}

\end{document}

%% file: I-Introduction.tex
\section{Introduction}

\subsection{Background}

High-dimensional data is prevalent in modern data analysis and machine learning, yet provably extracting useful information out of it remains an outstanding challenge. While learning via classic regularity classes, e.g. Sobolev smoothness, suffers from the curse of dimensionality, Neural Networks (NNs) hint at an efficient alternative, backed by repeated empirical breakthroughs over the last decade. 

A defining empirical characteristic of NNs is their ability to perform \emph{feature learning} via gradient-descent optimization; that is, automatically discovering relevant low-dimensional structure out of complex and high-dimensional data that enables efficient prediction of downstream learning tasks. In other words, a high-dimensional target function $\R^d \ni x \mapsto F(x) \in \R$, is sought to be learnt with a  `compositional' model of the form $\hat{F}(x) = f( \Phi(x))$, where the `features' $\Phi:\R^d \to \R^q$ that map to a lower-dimensional space $q \ll d$ are learnt alongside the `link' function $f$. 

Feature learning is thus a form of adaptivity, reminiscent of non-linear approximation and sparse variable selection in signal processing and statistics on low-dimensional settings. The statistical and approximation benefits of adapting features carry over to the high-dimensional regime, for instance by considering the class of shallow neural networks: A generic non-linear scalar activation function $\sigma$ defines the family of ridge functions $\{ \mathsf{P}_\theta \sigma(x):=\sigma( \theta^\top x )\,;\, \theta \in \R^d\}$, whose span defines a rich function class~\cite{pinkus2015ridge}. 

In this context, the benefits of non-linear approximation, where one can choose  `neurons' $\theta_1, \ldots, \theta_N$ to build an approximation $f=\sum_i \alpha_i \mathsf{P}_{\theta_i} \sigma$, over linear counterparts, where the neurons are chosen a priori, are by now well-understood \cite{barron1993universal,maiorov1998approximation}. Similarly, sparsity assumptions, here encoded with integral representations $\{ f = \int \mathsf{P}_\theta \sigma d\nu(\theta); |\nu|<\infty\}$ parametrized by Radon measures $\nu$ defined over the space of parameters $\theta$, are known to break the curse of dimensionality in the context of statistical learning \cite{bach2017breaking}. The main outstanding  question is therefore computational: how to efficiently discover planted low-dimensional structure, in the face of the non-convexity of the associated learning objectives. 

Mean-field formulations of shallow neural networks \cite{mei2018mean, rotskoff2018parameters, chizat2018global, sirignano2020mean} recover a source of convexity by expressing the associated loss functional in the Radon space of measures, which is leveraged to establish  global convergence of the associated Wasserstein Gradient dynamics in the infinitely-wide mean-field limit. However, such guarantees are qualitative under such conditions, since there is an inherent computational hardness associated with learning over the whole class of integral representations above, as witnessed by SQ-lower bounds \cite{goel2020superpolynomial,diakonikolas2020algorithms} and cryptographic reductions \cite{song2021cryptographic,daniely2020hardness}. 

Positive algorithmic results that explain feature learning thus require further structural assumptions. An appealing and arguably on of the simplest instance of such is given by \emph{linear} feature learning, whereby $\Phi$ is a linear projection. 
By focusing on canonical high-dimensional data distributions, such as the Gaussian distribution, and on the scalar case ($q=1$, referred to as the \emph{single-index} setting), the seminal works \cite{dudeja2018learning,arous2021online} built a harmonic analysis framework of Stochastic Gradient Descent (SGD) leading to quantitative learning guarantees.

More precisely, in the single-index model  a direction $\theta^* \in \mathcal{S}_{d-1}$ and a (generally non-linear) scalar function $f^*$  parametrize the single-index function $x \mapsto \mathsf{P}_{\theta^*} f^*(x) = f^*(x\cdot \theta^*)  $, 
and one attempts to recover it via minimizing the regression on $L(\theta) = \| \mathsf{P}_\theta f^* - \mathsf{P}_{\theta^*} f^*\|^2_{\gamma_d}$, where $\gamma_d$ is the standard Gaussian measure in dimension $d$. Thanks to the rotational symmetry of the Gaussian, this is equivalent to maximazing via some gradient ascent method the correlation $\langle \mathsf{P}_\theta f^*,\mathsf{P}_{\theta^*} f^* \rangle$, with a uniform initialization over the sphere $\theta_0 \sim \mathrm{Unif}(\mathcal{S}_{d-1})$.

In turn, the rotational invariance and stability properties of the Gaussian measure enable a precise control over the population gradient dynamics, through the one dimensional correlation (or \emph{summary statistic}) $m_t := \theta_t \cdot \theta^* $ and the Ornstein-Ulhenbeck semigroup. Due to the high-dimension, the system is uncorrelated at initialization, i.e. $m_0 \sim 1/\sqrt{d}$ and the learning may suffer from \textit{vanishing gradient} during its initial phase. This flatness near initialization is controlled by the number of vanishing moments, or \emph{information exponent} $s$ of $f^*$,  and corresponds to the order of the saddle at the equator $\{ \theta\in \S_{d-1}\,;\, \theta \cdot \theta^*=0 \Leftrightarrow m =0 \}$. As shown in~\cite{arous2021online}, 
this saddle precisely characterizes the overall time complexity to recover the signal $t \sim d^{s/2-1}$, and dictates the SGD sample complexity of $n \sim d^{s-1}$). In essence, the SGD dynamics, when viewed through the prism of the correlation summary statistic, evolve along a diffusive process whose drift is precisely given by the population gradients. 

The goal of this article is to address the natural extension of this model, namely the \emph{multi-index} setting in which $1<q \ll d$. Motivated by the central importance of the deterministic drift in the above discussion, our main focus is on the study of population gradient flow dynamics. As we will uncover next, the main takeaway is that, after a proper parametrization of the learning problem, the gradient flow dynamics evolve in a \emph{benign} non-convex optimization landscape, characterized by multiple saddle points, as opposed to the single saddle for $q=1$. Moreover, due to the high-dimensionality of the setup, we will see that the dynamics will spend most of its time near specific saddle points that can be explicitly described in terms of certain harmonic decompositions of $f^*$. Note that this `saddle-to-saddle' dynamics has first been unveiled in a very similar set-up, yet for specific target link functions, in \cite{abbe2023sgd}.

\subsection{Going beyond the single index model}
\label{sec:regression_model}
Going beyond the problem of learning a one dimensional projection, one can ask  how the system behaves when replacing it by a $r$ dimensional linear subspace. 
We thus replace the direction $\theta$ in the single-index model by a matrix from the \textit{Stiefel manifold}: $\mathcal{S}(d,r):=\{W \in \R^{d \times r}, W^\top W = I_r\}$, that is matrices whose columns form a set of $r$ orthonormal vectors of $\R^d$. On top of that, in all generality, we consider a loss where we do not know a priori the link function $f^*$: 
\begin{align} 
\label{eq:multi-index}
L(f, W) = \frac{1}{2}\E_{\gamma_d}\left[ f (W^\top x) - f^* ({W^*}^\top x)\right]^2 = \frac{1}{2} \|  \P_W f - \P_{W^*} f^* \|^2,
\end{align}
where $W \in \mathcal{S}(d,r)$, $W^* \in \mathcal{S}(d,q)$ and $f \in L^2_{\gamma_r}$,  $f^* \in L^2_{\gamma_q}$, and where have used the change of variable operator, for any matrix $M \in \R^{q \times d}$ ,$\|M\| \leq 1$ defined as 
\begin{align}
\P_M: L^2_{\gamma_d} \to L^2_{\gamma_q},\  \text{ such that:  }\ \P_{M} f(x) := f(M^\top x),    
\end{align}
We can assume without loss of generality that $\mathbb{E}_z [ f^*(z)]=0$.

 Our model thus consists of a linear feature extractor, parametrised as a unitary subspace, followed by a non-linear non-parametric model, that crucially is defined on a \emph{dimension-free} subspace. One can view this multi-index model as a template for more intricate neural network models, that extract low-dimensional features in their inner layers. 

\vspace{0.5em}
\paragraph{Full semi-parametric learning.}
In all generality, the target $F^* = \mathsf{P}_{W^*} f^*$ is a multi-index model of dimension $q$. Since any such target can be also written as $F^* = \mathsf{P}_{\tilde{W}^*} \tilde{f}^*$ with $\tilde{f}^*:\mathbb{R}^{\tilde{q}} \to \mathbb{R}$ and $\tilde{W}^* \in \S(d, \tilde{q})$ for any $\tilde{q}>q$, there is a need to further define the multi-index class via their \emph{intrinsic dimension} (see Definition \ref{def:intrinsic_dim}). The \textit{joint learning} of both $f$ and $W$ will be described in Section~\ref{sec:grassmannian_perspective}.
\paragraph{Planted parametric model.} 
The case where  $q = r$ and $f = f^*$ is known, with intrinsic dimension $q$, will be called the \textit{planted model} where we have a maximum of information to recover the projection. It is treated in Section~\ref{sec:planted_model}.
\subsection{Loss representation}

We start by expanding the population loss as 
\begin{align}
\label{eq:general_loss_representation}
L(f, W) = \frac{1}{2}\| \mathsf{P}_W f - \mathsf{P}_{W_*} f^* \|_{\gamma_d}^2 &= \frac{1}{2}\| f\|_{\gamma_r}^2 + \frac{1}{2}\| f^*\|_{\gamma_q}^2 - \langle \mathsf{P}_W f, \mathsf{P}_{W_*} f^* \rangle_{\gamma_d}~.
\end{align}

We thus need to understand how multi-index models ``interact'' through the correlation term $\langle \mathsf{P}_W f, \mathsf{P}_{W_*} f^* \rangle$. 
After inspecting Eq.\eqref{eq:general_loss_representation}, it is tempting to consider the adjoint of $\mathsf{P}_{W_*}$ in $L^2_{\gamma_q}$ to give a compact representation of the loss, quadratic in $f$ and depending only on $W$ through the correlation operator $\mathsf{P}_{W_*}^\dagger \mathsf{P}_{W}$. We show that this adjoint operator is precisely a Gaussian averaging and can be seen as a natural multivariate extension of the Ornstein-Ulhenbeck semigroup (see Proposition~\ref{prop:OU}):
\begin{definition}[Averaging Operator] 
\label{def:averaging_operator}
Let $M \in \mathbb{R}^{q \times r}$ and $\|M\|\leq 1$. The \emph{averaging} of $f \in L^2_{\gamma_r}$ w.r.t. $M$ is
\begin{align}
  \begin{array}{l|rcl}
\Av_M  : & L^2_{\gamma_r} & \longrightarrow & L^2_{\gamma_q} \\
     & f \hspace*{0.2cm} & \longmapsto &   \Av_M f(z):=\mathbb{E}_{y\sim \gamma_r} \left[f\left( M^\top z + (I_r - M^\top M)^{\frac{1}{2}}y\right)\right] ~.
\end{array}  
\end{align}
where the expectation is to indicate that $y$ is distributed according to $\gamma_r$.
\end{definition}
Going further and anticipating the calculations written in subsection~\ref{subsub:basic_A}, we show that in fact, the loss can be written as a simple function of this averaging operator, with respect to the correlation matrix $M = W_*^\top W \in \R^{q \times r}$.
\begin{proposition}
\label{prop:representation_loss}
Let $M = W_*^\top W \in \R^{q \times r}$. We have the following representation:  
    \begin{align}
    \label{eq:loss_learning_model}
        L(f, W) = \frac{1}{2}\| f\|_{\gamma_r}^2 + \frac{1}{2}\| f^*\|_{\gamma_q}^2 -  \langle  \Av_M f, f^* \rangle_{\gamma_q}~.
    \end{align}
\end{proposition}
 Beyond the technical fact, this proposition has a simple, yet important consequence: we have identified a natural family of dimension-free \emph{summary statistics}, in the language of \cite{arous2022high}, given by the correlation matrix $M$, whose size is independent of the dimension, and the low-dimensional function $f$. Remarkably, in the single-index model case~\cite{arous2021online}, $m = M \in [-1,1]$ is a one-dimensional parameter, and the corresponding operator family $(\mathsf{A}_m)_{m \in [-1,1]}$ has a commutative semi-group structure that makes it \textit{jointly diagonalizable}. The main difficulty we face here is the absence of this property due to the dimension $q \geq 2$. This is why, before delving into the examination of the energy landscape, we first allocate Section~\ref{sec:harmonic_analysis} to a thorough exploration of this operator. In this context, we touch upon its connection to the Ornstein-Uhlenbeck process and introduce multivariate Hermite bases. However, the central focus of this section revolves around the decomposition of the operator into its singular vectors, as demonstrated in Corollary~\ref{cor:SVD_of_A}. Specifically, if we express $M = V \Lambda U^\top$ through its Singular Value Decomposition, we have that
    \begin{equation*}
     \mathsf{A}_M = \sum_{\beta \in \N^q} \lambda_\beta H_\beta(V) \otimes  H_\beta(U)~,  
    \end{equation*}
where $H(U), H(V)$ are multivariate Hermite bases w.r.t $U, V \in \O_q$, and $\lambda_\beta = \prod \lambda_i^{\beta_i}$.

\subsection{Main Results}

\subsubsection{The joint learning}
     
\paragraph{A `fast-slow' procedure} 

While the first properties of the problem have been unveiled in the previous section, Section~\ref{sec:grassmannian_perspective} addresses the question of \textit{jointly learning the summary statistics~$(f^*,W_*)$}. To do so, we consider a two-timescale procedure where for $t>0$, at fixed~$W_t$, the function $f_t$ is optimized perfectly via a non-parametric method, while the matrix $W_t$ follows the Stiefel gradient flow over the resulting loss~$L(f_t, W_t)$. We show how this procedure can be ran in practice via a Reproducing Kernel Hilbert Space method, mimicking a two timescale gradient flow~\cite{berthier2023learning}  on the two ``layers'' of learning (the function $f$, and the projection $W$), reminiscent of the learning of a neural network. This is developed in subsection~\ref{subsec:fast_learning}.

\paragraph{First main result: critical points} 

This procedure translates into a loss that depends only on the PSD matrix $G(W) = M M^\top = V\Lambda^2 V^\top \in \C_q$, the set of $q \times q$ PSD matrices of norm bounded by one:
\begin{align*}
    L(W) =  \| f^*\|_{\gamma_q}^2 - \sum_{\beta \in \N^q} \lambda_\beta^2 \, \langle f^*, H_\beta(V)\rangle^2.  
\end{align*}
Except from the constant term, remarkably, this loss has a structure of a dot product in $\ell_2(\N^q)$ (the space of square summable multivariate sequences) \textit{with constraints}: for all $\beta$,  $\lambda^2_\beta \in [0,1]$ and the coefficients $\langle f^*, H_\beta(V)\rangle^2$ sum to $\|f^*\|^2_{\gamma_q}$. This structure, reminiscent of linear programming, allows us to show the following:
\begin{theorem}[Optimization Landscape of fast-slow joint learning (informal version of Theorem \ref{thm:critical_points})]
The optimization landscape of $L(W)$ has no bad local minima, with a unique global minimiser corresponding to~$G(W)=I_q$. Moreover, all of its saddles are in the boundaries of $\C_q$, satisfying $\lambda_i \in \{0,1\}$ for $i \in \llbracket 1,q \rrbracket $.
\end{theorem}
In other words, gradient flow with generic initialisation will converge to the global solution via standard stable-manifold arguments \cite{pmlr-v49-lee16}. The `nice' geometry of PSD matrices is instrumental in obtaining such benign optimization landscape.  

\paragraph{Second main result: quantitative description of the gradient flow trajectory, time complexity}

The aim of Section~\ref{subsec:gradient_flow} is to describe the Grassmaniann gradient flow $\dot{W}_t = - \nabla^{\mathcal{G}} L(W_t)$, where $W_t$ is initialized uniformly at random on the Grassmaniann manifold. From the previous result, this ODE qualitatively converges to a global optimum. Nonetheless we go further and describe precisely the timescales of the incremental dynamics that occurs. High-dimension plays a prevalent role in this: at initialization the correlation between two random subspaces $W,W_*$ is of order $1/\sqrt{d}$, see Proposition~\ref{lem:eigenvalues_init}: this is a quantitative extension of Borel's Lemma~\cite[Theorem 2.4]{meckes2019random}. Hence, when $d$ is large and $f^*$ has no linear component,  the dynamics is automatically initialized at a saddle point. 

Viewing this gradient flow via a dynamical system lens, we know that it will then spend most of its time near saddle points, joining them via a path that is in the vicinity of heteroclinic orbits~\cite{bakhtin2011noisy}, reducing the index (number of descent direction) of each saddle visited by at least one~\cite{matsumoto2002introduction}, and eventually converging towards the global minimum. Furthermore, we can quantify generically the time spent in each saddle point as its degeneracy (number of vanishing derivatives), also referred to as flatness. The specification of these two quantities in our framework is done in Subsection~\ref{sec:escapemediocrity}: the index of the saddle is related to the so-called \textit{leap dimension} $l_k$ (Definition~\ref{def:leap}),  whereas the degeneracy of each saddle, controlling the overall time complexity of the flow, is referred to as the \textit{leap exponent} $s_k$ (Eq.~\eqref{eq:relative_info}). These quantities were first introduced in \cite{abbe2022merged,abbe2023sgd} and adapted in \cite{dandi2023learning} to analyze (S)GD dynamics for a certain class of multi-index models with the so-called \emph{staircase} property (see Remark \ref{rem:staircase}). The main result of Section \ref{subsec:gradient_flow} is the quantification and precise characterization of the \emph{saddle-to-saddle} gradient flow dynamics:
\begin{theorem}[Saddle-to-Saddle dynamics (informal version of Theorem \ref{thm:coarse-grained})]
The summary statistics~$G_t = M_tM_t^\top$ goes in the vincinity of $K \leq q$ saddles, corresponding to a flag $ \{0\} = W_0 \subset W_1 \subset \dots \subset W_K = \R^q $ explicitly described in terms of the harmonic decomposition of $f^*$. Moreover, the gradient flow escapes each saddle after time $t_k \simeq d^{s_k -1}$, where $(s_k)_{k \leq K}$ are the successive relative information exponents associated to each saddle.  
\end{theorem}
%
\begin{figure}
\hspace*{0.5cm}
\begin{minipage}{.5\textwidth}
    \input{octahedron_saddles_intro}
\end{minipage}%
\hspace*{-2.5cm}
\begin{minipage}{0.6\textwidth}
\includegraphics[width=\textwidth]{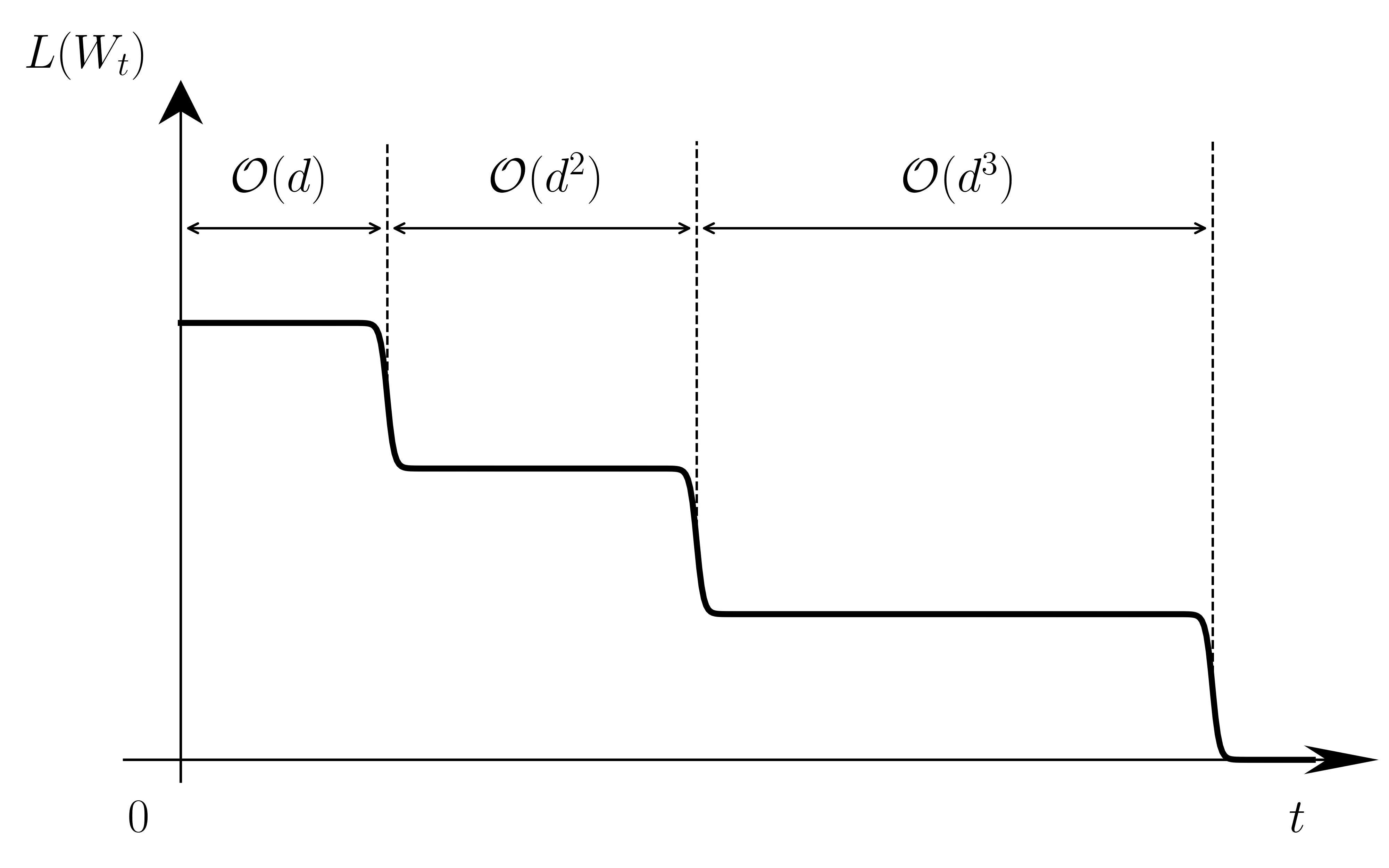}
\end{minipage}
    \caption{Cartoon illustrations summarizing the theorems \ref{thm:critical_points} and \ref{thm:coarse-grained} on the evolution and the geometry of the population loss during the dynamics. \textbf{Left}, the octahedron represents the geometry given by Theorem \ref{thm:critical_points} on the critical points, while the trajectory displayed showcases the successive learning of the subspaces $W_1, W_2, W_3$, which lie on the polytope. The time scales of these learning is illustrated symbolically throrugh the learning curve in the \textbf{right} plot as a succession of three saddles with timescales (or plateaus) of increasing orders $\mathcal{O}(d), \mathcal{O}(d^2)$ and $\mathcal{O}(d^3)$, before eventually showing convergence.}
\end{figure}

This saddle-to-saddle description has been recently analysed in other Machine Learning setups (besides the aforementioned \cite{abbe2023sgd}), see e.g. \cite{jacot2021saddle,li2020towards,berthier2023incremental,pesme2023saddle}, and provides an appealing geometric perspective on incremental feature learning,  consistent with recent `emergent' empirical behavior \cite{nanda2023progress}. 
Note that the description of the function learned at each saddle is explicitly given in Section~\ref{sec:escapemediocrity}, and corresponds to the spectral thresholding $(\mathsf{S}_{W_k} f)_{k \leq K}$, see Definition \ref{def:spectral_thresholding}. 

In particular, the total time complexity to learn the multi-index model is $O(d^{s^* -1})$, where $s^* = \max_k s_k$ is the largest leap exponent. Remarkably, this exponent was already identified in \cite{abbe2023sgd} (so-called $\mathrm{IsoLeap}$) as driving the sample-complexity of learning a certain class of `staircase' models using a variant of SGD; and was conjectured to hold for general multi-index models. Our results partially resolve this conjecture by showing that it indeed drives the time complexity of the underlying population dynamics for any multi-index target. 


\subsubsection{The planted setting}

The \textit{planted setting} corresponds to the setting where the link function $f = f^*$ is known in advance. Quite surprisingly, this set up turns out to be less well conditioned with respect to the subspace learning via gradient flow. Indeed, we show that, while for simple functions the learning is possible (e.g. radial functions), it turns out that there are generic cases of functions that present symmetries for which the gradient flow \textit{will be trapped in bad local minima}, thus not recovering the targeted subspace with high probability. This stands in stark contrast with the one dimensional case where success occurs with probability at least one half~\cite{arous2021online}. While the dynamics for radial functions, described in Theorem~\ref{thm:planted_radial_case}, provably find the global minimiser, the situation is drastically different for certain choices of target function:
\begin{theorem}[Failures of Stiefel Gradient Flow for Planted Problem (informal version of Theorem \ref{thm:planted_failure} and \ref{prop:badmaxima})]
For $q>1$, there exist choices of target function $f^*$ such that Stiefel Gradient Flow converges to bad local optima with high probability.  
\end{theorem}
The negative results of this section contrast with the previous one, and indicate that 
jointly learning may be necessary to avoid trapped gradient flow trajectories; this is in accordance with the folklore that \emph{overparametrisation} overcomes non-convex optimization difficulties. 

\subsubsection{Towards Sample Complexity Guarantees}

Finally, we briefly address the statistical aspects of the model in Section \ref{subsec:sample_complexity}, by discussing how to turn our quantitative time complexity guarantees into sample complexity ones. By instantiating the fast non-parametric learning thanks to Hermite reproducing kernel Hilbert spaces we build, we informally show that by appropriately adjusting the ridge regularisation parameter, the empirical Stiefel gradients are sufficiently concentrated around their population counterparts in the initial mediocrity phase, provided $n \gtrsim d^{s}$. Such preliminary analysis stands in contrast with prior works that relied on frozen second-layer weights in order to escape mediocrity \cite{bietti2022learning,damian2022neural,abbe2023sgd,ba2022high,dandi2023learning}, and suggests that the fast kernel learning might be a good template to analysze finite-sample guarantees.

\subsection{Further Related Works}
\paragraph{Single-Index and Multi-Index models}
The ability of gradient-descent methods to learn single- and multi-index models has been thoroughly studied in prior work. 
There is a long literature, starting at least with \cite{kalai2009isotron,shalev2010learning,kakade2011efficient}, that exploits certain properties of the link function, such as invertibility or monotonicity, under generic data distributions satisfying mild anti-concentration properties \cite{soltanolkotabi2017learning,frei2020agnostic,yehudai2020learning,wu2022learning}.

Besides the works \cite{dudeja2018learning,arous2021online,arous2022high}, several authors have built and enriched the setting to multi-index models~\cite{abbe2022merged,abbe2023sgd,damian2022neural,arnaboldi2023high,dandi2023learning,ba2022high}, addressing the semi-parametric learning of the link function~\cite{bietti2022learning,berthier2023learning,mahankali2023beyond}, closing the gap with vanilla neural network architectures \cite{glasgow2023sgd}, as well as exploring SGD-variants~\cite{arous2022high,barak2022hidden,chen2023learning}, relaxing the Gaussian assumption \cite{bruna2023single}, or extending it to invariant architectures \cite{zweig2023single}. We also highlight works \cite{chen2020learning} that leverage tensor thresholding methods to learn certain families of multi-index models with optimal ($O(d)$) sample complexity. 
Finally, \cite{damian2023smoothing} demonstrates that the sample complexity $n=O(d^{s-1})$ of SGD/GD on single-index models can be improved to $\tilde{O}(d^{s/2})$ by running SGD on a smoothed loss, matching the CSQ (Correlational Statistical Query) lower bound for this problem \cite{abbe2023sgd,damian2022neural}. 

On the statistical side, multi-index models have been long studied in the literature, dating at least to the sliced inverse regression method of \cite{li1991sliced,xia2002adaptive,hristache2001structure}, and were studied also in \cite{dudeja2018learning}. Recently, \cite{follain2023nonparametric} proposed a non-parametric estimation of the planted subspace that leverages some of the Hermite decomposition tools described here, but still exhibits a curse of dimensionality. 

\subsection{Roadmap through the Article}
Throughout the article, we have opted to present the results in a linear and detailed fashion, transferring only a minimal amount of content to the appendix, for the sake of completeness. The drawback of this choice is that some readers may want to take a more direct route to the essentials. We provide a map of the possible shortcuts here: 
\begin{itemize}
    \item Section \ref{sec:harmonic_analysis} develops the harmonic analysis tools needed to study the gradient flow dynamics, and may be used as a reference. The main results required to understand the rest of the article are Corollary~\ref{cor:SVD_of_A} on the loss representation and the notion of \textit{intrinsic dimension} presented in Definition~\ref{def:intrinsic_dim}.
    \item Section \ref{sec:grassmannian_perspective} contains the core positive results for the semi-parametric learning model. Subsection~\ref{sec:fastideal} is instrumental to understand the joint learning procedure under study, resulting in the loss given in Corollary~\ref{cor:loss_psd}. Then, the main result  concerning the loss landscape is Theorem~\ref{thm:critical_points} in Subsection~\ref{subsec:critical_points}. Finally, the notations and tools required to understand the dynamics are given in Subsection~\ref{sec:escapemediocrity}. This Subsection is as self-contained as possible and is the core of the article. The other subsections can be overlooked at first reading. 
    \item Section \ref{sec:planted_model} analyses the planted model, and provides in particular the negative optimization results when there is no joint learning. In particular, the main failure theorem is presented in Subsection~\ref{sec:planted_failure2}.
    \item Finally we conclude with Section \ref{subsec:sample_complexity}, where we informally discuss how to leverage our time complexity guarantees into sample complexity guarantees. 
\end{itemize}


\vspace{1cm}

\paragraph{Acknowledgements} We thank Emmanuel Abbé, Gerard Ben Arouss, Alex Damian, Cédric Gerbelot, Florent Krzakala, Theodor Misiakiewicz, Eric Vanden-Eijnden, Denny Wu, Lenka Zdeborová and Aaron Zweig for helpful discussions while preparing this manuscript. We also thank Raphaël Berthier for his useful feedback on an early version of the manuscript. JB was partially supported by the Alfred P. Sloan Foundation and awards NSF RI-1816753, NSF CAREER CIF 1845360, NSF CHS-1901091 and NSF DMS-MoDL 2134216.

\newpage

%% file: octahedron_saddles_intro.tex
\definecolor{cof}{RGB}{219,144,71}
\definecolor{pur}{RGB}{186,146,162}
\definecolor{greeo}{RGB}{91,173,69}
\definecolor{greet}{RGB}{52,111,72}

\begin{tikzpicture}[thick,scale=5]
\coordinate (A1) at (0,0);
\coordinate (A2) at (0.6,0.2);
\coordinate (A3) at (1,0);
\coordinate (A4) at (0.4,-0.2);
\coordinate (B1) at (0.5,0.5);
\coordinate (B2) at (0.5,-0.5);

\begin{scope}[thick,dashed,,opacity=0.6]
\draw (A1) -- (A2) -- (A3);
\draw (B1) -- (A2) -- (B2);
\end{scope}
\draw[fill=yellow!35,opacity=0.6] (A1) -- (A4) -- (B1);
\draw[fill=cyan!35,opacity=0.6] (A1) -- (A4) -- (B2);
\draw[fill=purple!35,opacity=0.6] (A3) -- (A4) -- (B1);
\draw[fill=olive!35,opacity=0.6] (A3) -- (A4) -- (B2);
\draw (B1) -- (A1) -- (B2) -- (A3) --cycle;

\filldraw[black] (0.25,0.15) circle (0.2pt) {};
\filldraw[black] (0.15,0.02) circle (0.2pt) {};
\filldraw[black] (0.35,0.075) circle (0.2pt) {};
\filldraw[black] (0.35,-0.275) circle (0.4pt) node[anchor=south east]{$\ W_1$};

\filldraw[black] (0.6,-0.275) circle (0.4pt) node[anchor=south west]{$\ W_2$} {};

\filldraw[black] (0.85,0.1) circle (0.2pt) {};
\filldraw[black] (0.65,0.1) circle (0.4pt) node[anchor=north east]{$\ W_3$} {};

\filldraw[black] (B2) circle (0.4pt) node[anchor=north east]{$\ W_0 = \{0\}$};

\filldraw[black] (B1) circle (0.4pt) node[anchor=south east]{$\ W_4 = \{\R^q\}$};

\draw [black, line width=0.35mm] plot [smooth, tension=0.6] coordinates { (B2) (0.35,-0.275) (0.6,-0.275) (0.65,0.1) (B1)};

\end{tikzpicture}

%% file: 0-Notations.tex
\paragraph{General mathematical notations}

 $ $

\noindent \textit{Dot products, matrix and norms}: For $\theta, \theta' \in \R^d$, the notation $\theta \cdot \theta'$ stands for the dot product between vectors, and $\|\theta\|^2 = \theta \cdot \theta$ is the Euclidean norm. For a matrix $M \in \R^{q \times r}$, $\|M\|$ denotes its operator norm and $\|M\|_{\mathrm{Fr}}$ its Froebinius norm. \\
\noindent \textit{Integers}: $\llbracket p, q \rrbracket$ is the set of numbers between $p$ and $q$ \\
\noindent \textit{Eigenvalues, singular values}: The eigenvalues of a PSD matrix are always sorted in non-increasing order and denoted $\lambda_1 \geq  \dots \geq  \lambda_q$. Similarly for the singular values of a matrix but with the letter $\sigma$. The largest, respectively smallest eigenvalue will be noted $\lambda_{\max}(\Sigma)$ and $\lambda_{\min}(\Sigma)$. Similarly for the singular values, that we denote $\sigma_{\max}(M)$ and $\sigma_{\min}(M)$. \\
\noindent \textit{Multi-index}: For $\beta = (\beta_1, \dots, \beta_q) \in \N^q$, we define $|\beta| = \beta_1 + \dots + \beta_q$, and more generally, for $ 1 \leq r \leq s \leq q $, $|\beta|_{r:s} = \beta_r + \dots + \beta_s$. \\
\noindent \textit{Multi-index spaces}: For $q \in \N^*$, and $(u_\beta)_{\beta \in \N^q}$ a multi-index sequence, we define for $i = 1,2$ the spaces $\ell_i(\N^q)=\{ u\ ;\ \text{s.t.} \sum_\beta |u_\beta|^i < + \infty \}$. $\ell_\infty(\N^q)=\{ u\ ;\ \text{s.t.} \sup_\beta |u_\beta| < +\infty \} $ \\
\noindent \textit{Sphere}: $\mathcal{S}_{d-1} = \{ u \in \mathbb{R}^{d}; \| u \| = 1\} $, the $d$-dimensional sphere. \\
\noindent \textit{Orthogonal group}: $\O_q = \{ U \in \mathbb{R}^{q \times q}; U^\top U = U U^\top  = I_q\}$. \\
\noindent \textit{Stiefel manifold}: $\S(d,r):=\{W \in \R^{d \times r},\, W^\top W = I_r\}$, matrices whose columns form a set of $r$ orthonormal vectors of $\R^d$. \\
\noindent \textit{Grassmann manifold}: $\G(d,r) = \S(d,r) / \O_r$. The manifold of $r$-dimensional subspaces in $\mathbb{R}^d$. \\
\noindent \textit{Gaussian Measure}: $\gamma_d = \mathcal{N}(0, I_d)$. \\ 
\noindent \emph{$L^2$ Gaussian space}: $L^2_{\gamma_d} = \{ f : \mathbb{R}^d \to \mathbb{R}\,;\, \mathbb{E}_{\gamma_d}[|f(x)|^2]< \infty \}$. Inner product $\langle f, g \rangle_{\gamma_d}:= \mathbb{E}_{\gamma_d}[f(x) g(x)]$. \\
\noindent \emph{PSD matrices}: For two PSD matrices $A, B$, the notation $ A \preccurlyeq B$ refers to the natural Loewner order and $A^{1/2}$ stands for the unique PSD square root of $A$. We also define the subset of PSD matrices $\C_q := \{ A \in \mathbb{R}^{q \times q}; A^\top = A; A \succeq 0; \|A\|\leq 1\}$.

\vspace{0.5cm}

\paragraph{Operators}

$ $

\noindent For $M \in \R^{q \times r}$, $\|M\| \leq 1 $, the averaging operator is defined by
\begin{align*}
\begin{array}{l|rcl}
\hspace*{3.9cm} \Av_M  : & L^2_{\gamma_r} & \longrightarrow & L^2_{\gamma_q} \\
     & f \hspace*{0.2cm} & \longmapsto &   \Av_M f(z)=\mathbb{E}_{y} \left[f\left( M^\top z + (I_r - M^\top M )^{\frac{1}{2}}y\right)\right] ~.
\end{array}  
\end{align*}
\noindent For $M \in \R^{q \times r}$, $\|M\| \leq 1 $ we define the linear change of variables
\begin{align*}
\begin{array}{l|rcl}
    \P_M  : & L^2_{\gamma_r} & \longrightarrow & L^2_{\gamma_q} \\
     & f \hspace*{0.2cm} & \longmapsto &   \P_M f(z) = f (M^\top z) ~.
\end{array}  
\end{align*}

\noindent For $W \in \mathcal{G}(q,p)$ with parameter $s \in \N^*$, the spectral filtering, w.r.t. $W$ is defined as  
\begin{equation*}
    \mathsf{S}^{s}_W(f) := \sum_{\beta; |\beta|_{p+1:q}\leq s} \langle f, H_\beta \rangle H_\beta~.
\end{equation*}
When $W=\emptyset$, we write $\mathsf{S}^s(f) = \mathsf{S}^{s}_\emptyset(f) $ represents the projection in the subspace of harmonics of maximal exponent $s$.  
\noindent For all linear operator $T: L^2_{\gamma_r}  \rightarrow  L^2_{\gamma_q}$, we denote the operator norm as $ \|T\| = \sup_{\|f\|_{\gamma_r} \leq 1 } \|T(f)\|_{\gamma_q} $, and the Hilbert-Schmidt norm as $\|T\|_{\mathrm{HS}}$.

%% file: III-Harmonic_analysis.tex
\section{Preliminaries}
\label{sec:harmonic_analysis}
The aim of this section is to analyze the properties of the averaging operator. Multivariate basis of Hermite polynomials will play a central role in this study, this is the reason why we start by recalling some known properties, as well as representation properties of functions in these basis. This is the goal of subsection~\ref{subsec:tensorized_hermite}. Section~\ref{subsec:properties_averaging_operator} is devoted to present the properties of the averaging operator. Finally subsection~\ref{subsec:intrinsic_dimension} explains intrinsic properties of functions derived with respect to their interaction with this averaging operator.  

\subsection{Tensorized Hermite Decomposition}
\label{subsec:tensorized_hermite}
\subsubsection{Definitions}
Let $(h_k)_{k \in \mathbb{N}}$ be the one-dimensional orthonormal Hermite basis of $L^2_{\gamma}$, where $\gamma:=\gamma_1$ is the standard Gaussian measure in one dimension. We refer to \cite{szego1939orthogonal} for a comprehensive treatment of such family, but a particularly compact representation formula is given by
\begin{equation}
\label{eq:hermite_univ}
    \tilde{h}_k(x) := (-1)^k \gamma(x)^{-1} \frac{d^{k}}{dx^k} \gamma(x)~,~h_k = \frac{\tilde{h}_k}{\| \tilde{h}_k \|_\gamma}~.
\end{equation}
We can naturally tensorize this family to make it a orthonormal family of $L^2_{\gamma_q}$ by defining for all multi-indices $\beta = (\beta_1, \dots, \beta_q) \in \N^q$, the family, for all $x = (x_1, \dots, x_q) \in \R^q $,
\begin{align}
  H_\beta(x) := h_{\beta_1}(x_1) \dots h_{\beta_q}(x_q).
\end{align}
Now, we can define a family of multivariate tensorized polynomials simply by rotating the canonincal basis. Indeed, any matrix of the orthogonal group $U \in \O_q$ can be associated to an orthogonal basis of $\mathbb{R}^q$ via its columns, $U=(u_1 \ldots u_q)$. We define here multivariate Hermite polynomials with respect to $U$ the following way: for any $\beta \in \mathbb{N}^q$ the polynomial $H_\beta(U) \in L^2_{\gamma_q}$ is defined as
\begin{equation}
    x\mapsto H_\beta(U)(x):=\prod_{i=1}^q h_{\beta_i}( x \cdot u_i )~.  
\end{equation}
This can also be written with the change of variable operator: $\P_U H_\beta = H_\beta (U)$, for all $\beta \in \N^q$. From the orthonormality of both $U$ and $(h_k)_k$, we have the immediate consequence that for all $\beta, \zeta \in \N^q$,
\begin{equation}
    \langle H_\beta(U), H_{\zeta}(U)\rangle_{\gamma_q} = \delta_{\beta = \zeta} 
\end{equation}
Hence, 
any orthogonal matrix $U \in \O_q$ has a natural correspondence into an orthonormal basis of $L^2_{\gamma_q}$, given by $(H_\beta(U))_{\beta \in \N^q}$.
\begin{definition}[Tensorized Hermite Basis]
Given an orthogonal matrix $U \in \O_q$, its associated Tensorized Hermite basis of $L^2_{\gamma_q}$ is given by $( H_\beta(U) )_{\beta \in \mathbb{N}^q}$.   
\end{definition}
When the context is clear, we will omit the explicit dependence in $U$ to lighten notations. Obviously, any function $f \in L^2_{\gamma_q}$ can be decomposed in any of these basis 
\begin{equation}
\label{eq:hermite_expansion_basic}
    f = \sum_{\beta \in \N^q} \langle f, H_\beta(U) \rangle_{\gamma_q} H_\beta(U)~,
\end{equation}
where the sum converges for the $L^2_{\gamma_q}$ norm and as such, we have $\|f\|^2_{\gamma_q} = \sum_\beta \langle f, H_\beta(U) \rangle_{\gamma_q}^2$.
The Tensorized Hermite basis may be viewed as the analog of the Fourier Basis of $L^2(\mathbb{R}^q)$ when replacing the Lebesgue measure with the Gaussian measure. As such, many of the familiar tools of Fourier analysis translate to our setting. In particular, for $R\geq 1$, consider 
\begin{align}
\label{eq:spaces_of_functions}
    \mathcal{T}_{R} &:= \{ f \in L^2_{\gamma_q}\ ;\  \sum_\beta R^{2|\beta|} \langle f, H_\beta \rangle^2 < + \infty  \}~, \\ 
    \mathcal{B}_R &:= \{ f \in L^2_{\gamma_q}\ ;\  \langle f, H_\beta \rangle = 0 \text{ for } |\beta| > R \} ~.
\end{align}
We will refer to these subspaces respectively as \textit{tempered} and \emph{band-limited functions} (note that these latter functions are in fact polynomials of degree smaller than $R$). Tempered functions have faster than exponentially decreasing Hermite coefficients, while band-limited function can be seen as an extreme cases of tempered function for which the coefficients are strictly zero from a certain point. For all $ 1 \leq R \leq R'$, we have the relationship 
\begin{align}
    \mathcal{B}_{R} \subset \mathcal{B}_{R'} \subset \mathcal{T}_{R'} \subset \mathcal{T}_{R} \subset  \mathcal{T}_{1} = L^2_{\gamma_q}.
\end{align} 
A straightforward but important fact is that these spaces are well-defined, i.e. they are both independent of the choice of orthonormal basis, as shown by the following fact.
\begin{lemma} 
\label{claim:hermite_wellposed}
Let $(H_\beta(U))_\beta$,  $(H_{\zeta}(V))_{\zeta}$ be two Tensorized Hermite basis associated with orthogonal matrices $U$ and $V$, respectively. Then 
$\langle H_\beta(U), H_{\zeta}(V) \rangle = 0$ whenever $|\beta| \neq |\zeta|$. 
\end{lemma}
The proof of this lemma follows from elementary properties of Hermite polynomials, and we include it for the sake of completeness in Lemma \ref{claim:deghomv2}. 

\subsubsection{Spectral filtering}

We now introduce a natural class of shrinkage, akin to spectral filters in classic harmonic analysis. For that purpose, given a subspace $W \in \mathcal{G}(q,p)$ with $p \leq q$, consider an orthonormal basis $U=(u_1, \ldots, u_q) \in \O_q$ adapted to $[W; W^\perp]$, meaning that $(u_1, \ldots, u_p)$ is an orthonormal basis of $W$ and $(u_{p+1}, \ldots, u_q)$ is an orthonormal basis of $W^\perp \in \mathcal{G}(q,q-p)$. Let $(H_\beta)_\beta$ be its associated Tensorised Hermite Basis, where we decide to drop the $U$ dependency for the sake of clarity. Given $\beta \in \mathbb{N}^q$ and $I \subseteq \llbracket 1,q \rrbracket$, we denote $|\beta|_I = \sum_{i \in I} \beta_i$, with the shorthand $|\beta|= |\beta|_{1:q}$
. Observe that $|\beta|$ corresponds precisely to the degree of the polynomial $H_\beta$.

\begin{definition}[Spectral Thresholding]
\label{def:spectral_thresholding}
The \emph{spectral filtering} of $f \in L^2_{\gamma_q}$, relative to $W \in \mathcal{G}(q,p)$ with parameter $s \in \N^*$~is 
\begin{equation}
\label{eq:spectral_thresholding}
    \mathsf{S}^{s}_W(f) := \sum_{\beta; |\beta|_{p+1:q}\leq s} \langle f, H_\beta \rangle H_\beta~.
\end{equation}
When $W=\{0\}$, we write $\mathsf{S}^s(f) = \mathsf{S}^{s}_{\{0\}}(f)$ which, in fact, is the orthogonal projection onto  $\mathcal{B}_s$.  
\end{definition}
We verify that the spectral thresholding is a well-defined bounded linear operator, i.e. it satisfies $\mathsf{S}^{s}_W(f) \in L^2_{\gamma_q}$ and it does not depend on the choice of orthonormal basis, as per Lemma \ref{claim:hermite_wellposed}. Naturally, the spectral thresholding is bounded as it is a contraction, that is, for all $f \in L^2_{\gamma_q}$, $\|\mathsf{S}^{s}_W(f)\|_{\gamma_q} \leq \|f\|_{\gamma_q}$,
for all $s \in \N^*$ and subspaces $W \in \G(q,p)$.

\subsubsection{Relationship with Tensor Formulation}

Similarly as the definition of univariate Hermite polynomials from (\ref{eq:hermite_univ}), one can define for each $k\in \mathbb{N}$ the $k$-th normalized Hermite Tensor in dimension $q$ as 
\begin{align}
    \mathbf{H}_k(x) &:= (k!)^{-1/2} (-1)^k \frac{\nabla^k \gamma_q(x)}{\gamma_q(x)}~. 
\end{align}
This tensor is often used in the analysis of multi-index models \cite{damian2022neural,damian2023smoothing}. It verifies analogous orthogonality relationships across orders, after taking into account the symmetries inherent to the tensor representation. Specifically, for any $k$-tensor $A$ and any $k'$-tensor $B$, it holds 
\begin{align}
    \mathbb{E}_{x \sim \gamma_q} \left[\langle \mathbf{H}_k(x), A \rangle \cdot \langle \mathbf{H}_{k'}(x), B \rangle \right] = \delta_{k,k'} \langle \mathrm{Sym}(A), \mathrm{Sym}(B) \rangle~,
\end{align}
where $\mathrm{Sym}(T)_{i_1, \ldots i_k} = \frac{1}{k!} \sum_{\sigma \in S_k} T_{\sigma(i_1), \ldots \sigma(i_k)}$ is the symmetrization of a $k$-tensor along all permutations of its indices. 
Given $f \in L^2_{\gamma_q}$ and $k \in \mathbb{N}$, we can build the $k$-tensor containing the $k$-th harmonic coefficients $\mathbf{C}_k := \mathbb{E}_{z \sim \gamma_q} [f(z) \mathbf{H}_k(z) ]~,$
leading to the representation
\begin{equation}
\label{eq:hermite_expansion_tensor}
f(x) = \sum_{k \geq 0} \frac{1}{k!} \langle \mathbf{C}_k, \mathbf{H}_k(x) \rangle~.    
\end{equation}
As one can readily suspect, the representations in Equations (\ref{eq:hermite_expansion_basic}) and (\ref{eq:hermite_expansion_tensor}) are equivalent. Indeed, let $\mathcal{I}_k = \{ \beta \in \mathbb{N}^q; |\beta|=k \}$, and consider the \emph{aggregation} mapping 
\begin{align}
    \mathrm{Agg}: [1,q]^{\otimes k} &~~ \to ~~\mathcal{I}_k \nonumber \\
    (i_1, \ldots, i_k) &~~ \mapsto ~~ ( \#\{i_j=l , j=1,\ldots k \}  )_{l=1,\ldots q}~.
\end{align}

We then verify that 
\begin{align}
    (\mathbf{H}_k)_{i_1, \ldots, i_k} &= H_{\mathrm{Agg}(i_1, \ldots, i_k)}~.
\end{align}
The representation (\ref{eq:hermite_expansion_basic}) will turn out to be more convenient for our purposes than (\ref{eq:hermite_expansion_tensor}), so from now on we will adopt it. 

\subsection{The Averaging Operator}
\label{subsec:properties_averaging_operator}

We have already introduced a spectral shrinkage operator $\mathsf{S}^s_W$ on $L^2_{\gamma_q}$, performing the familiar low-pass filtering in $L^2_{\gamma_q}$. We now introduce the central object of the paper, which can be considered as another class of shrinkage operator, but which also allows to rotate the Hermite basis and rescale Hermite coefficients --see Corollary~\ref{cor:SVD_of_A} for more details. It has already been introduced in Definition \ref{def:averaging_operator} and serves in the low-dimensional representation of our objective function as shown in Proposition~\ref{prop:representation_loss}. Let us recall its definition here. Let $M \in \mathbb{R}^{q \times r}$ be such that $\|M\|\leq 1$, and $f \in L^2_{\gamma_r}$, then for all $z \in \R^q$, we have
\begin{align*}
\Av_M f(z)=\mathbb{E}_{y} \left[f\left( M^\top z + (I_r - M^\top M)^{\frac{1}{2}}y\right)\right]~,
\end{align*}
where the expectation is to indicate that $y$ is distributed according to $\gamma_r$. We begin this section by discussing the link between the averaging operator and the multivariate Ornstein-Uhlenbeck process. Let $\C_q$ be the set of $q\times q$ PSD matrices of norm bounded by one. Let $\Sigma \in \C_q$ and let $(X_t)_{t\geq 0}$ follow the following Ornstein-Uhlenbeck process in $\R^q$ 
\begin{align*}
    d X_t = - \Sigma X_t \mathrm{d} t + \sqrt{2}\Sigma^{1/2} \mathrm{d} B_t~,
\end{align*} 
where $(B_t)_{t \geq 0}$ is a standard Wiener process. Then let us define for all $t \geq 0$, the semigroup $P_t(f)(z) := \E \left[ f(X_t) | X_0 = z \right]$, where the expectation is taken w.r.t. the path of the Brownian motion. Then we have the following link with the average operator.
\begin{proposition}
\label{prop:OU}
    For all $t \geq 0$, we have then relationship $P_t = \Av_{\exp(-\Sigma t)}$.  
\end{proposition}
\begin{proof}
    This representation formula is due to Kolmogorov~\cite{Kolmogorov1934}. For more details on the multivariate  Ornstein Uhlenbeck semi-group, we also refer to~\cite{daprato1995ornstein} --in English this time.
\end{proof}
This representation formula of the operator provides a good intuition on its action. However, even though the comparison seems tempting, the family of operators at hand $(A_M)_{M \in \R^{q \times r}}$ goes beyond this semi-group representation. Even in the simpler case, where $M \in \C_q$, all such $M$ do not belong to the one parameter group $\{\exp{(-\Sigma t) },  t \in \R_+ \}$. This is in striking contrast with the single index model~\cite{arous2021online} where such link was one-to-one. Finally, we will see that in the particular case where the matrices $M$ are diagonal that such a structure persists.
\subsubsection{Basic properties of \texorpdfstring{$\mathsf{A}$}{A}}
\label{subsub:basic_A}
Let us begin with the very first general properties of the operator $\Av_M$. Indeed, we derive an multiplicative property reminiscent of the semi-group representation given in Proposition~\ref{prop:OU}.
\begin{restatable}[Multiplicative properties.]{lemma}{lemmamultiplicativeproperties}
\label{lem:averaging_operator_multiplicative}
We have the following properties: 
\begin{enumerate}[label=(\roman*)]
    \item For $M \in \mathbb{R}^{r \times p }$, $N \in \mathbb{R}^{p \times q }$ such that $\|M\|, \|N\| \leq 1$, we have $\Av_{MN} = \Av_M \Av_N$ and similarly $\P_{MN} = \P_M \P_N $.
    \item If $M \in \S(r,q)$, then the operators coincide, i.e. $\Av_M = \P_M$. 
\end{enumerate}
\end{restatable}
\begin{proof}
    We refer to Section~\ref{subsubsecapp:first_properties} of the Appendix for a proof of this result. 
\end{proof}
Next, the following proposition compute their adjoints, which is an important feature regarding the shape of the loss.
\begin{restatable}[Adjoint]{lemma}{lemmaadjoint}
\label{lem:averaging_operator_standard}
For all $M \in \R^{d \times r}$, $\Av_M^\dagger = \Av_{M^\top}$.
\end{restatable}
\begin{proof}
    We refer to Section~\ref{subsubsecapp:first_properties} of the Appendix for a proof of this result. 
\end{proof}
The computation of the adjoint for Stiefel matrices additionally to the Lemma~\ref{lem:averaging_operator_multiplicative} leads to the following calculation,
\begin{align*}
\mathsf{P}^\dagger_{W_*}  \mathsf{P}_W = \Av^\dagger_{W_*}  \Av_W = \Av_{W^\top_*}  \Av_W = \Av_{W^\top_* W } = \Av_{M}~,
\end{align*}
which gives the proof of Proposition~\ref{prop:representation_loss}.

\subsubsection{Representation in the tensorized Hermite basis}
We now examine how the averaging operator is represented in an appropriate Tensorised Hermite basis. We first begin by a simple result: on the subspace of diagonal matrices, we have the following nice property.

\begin{restatable}[Averaging Operator on diagonal matrices]{proposition}{proavediagonal}
\label{prop:ave_diagonal}
Let $\mathcal{D} = \{ \Lambda \in \mathbb{R}^{q \times q} ; \Lambda = \mathrm{diag}(\lambda_1, \ldots, \lambda_q); 0\leq \lambda_j \leq 1 \}$, then $( \mathsf{A}_\Lambda)_{\Lambda \in \mathcal{D}}$ jointly diagonalises in the tensor Hermite basis $( H_{\beta}(I) )_{\beta \in \mathbb{N}^q} \in L^2_{\gamma_q}$. 
       More precisely, $\forall \beta \in \N^q$,
\begin{align}
     \mathsf{A}_\Lambda H_\beta = \lambda_\beta H_\beta~, 
\end{align}   
where we have defined $\lambda_\beta =  \prod_i \lambda_i^{\beta_i}$, for all $\beta \in \N^q$.
\end{restatable}
\begin{proof}
The proof of the proposition can be found in Subsection~\ref{subsubsecapp:hermite_averagin} of the Appendix.
\end{proof}
As made concrete in this proposition, throughout the article for a $q$-tuple $\lambda =(\lambda_1, \dots, \lambda_q) \in \R^q$, we associated the multi-index sequence, defined for all $\beta \in \N^q$,  $\lambda_\beta = \prod_i \lambda_i^{\beta_i}$. The fact that  $\Av_\Lambda$ has such a simple representation when $\Lambda$ is diagonal leads naturally to perform the Singular Value Decomposition (SVD) of $M$ to represent $\Av_M$. In the following theorem, we show that the SVD of $M$ leads quite nicely to the SVD of $\Av_M$. In all the rest of the article, we fix notation and consider 
\begin{align}
    \label{eq:SVD_M}
    M = W_*^\top W = V \Lambda U^\top \in \R^{q \times r}~,
\end{align}
where $V \in \O_q$, $\Lambda \in \R^{q \times q}$ and $ U \in \S(r, q)$.
\begin{corollary}[SVD of $\mathsf{A}$]
\label{cor:SVD_of_A}
    Let $M=V \Lambda U^\top$ be the SVD of $M$, with $\Lambda = \mathrm{diag}(\lambda_1, \dots, \lambda_q)$ and $\|M\| \leq 1$,
    \begin{equation}
    \label{eq:SVD_of_A}
     \mathsf{A}_M = \sum_{\beta \in \N^q} \lambda_\beta H_\beta(V) \otimes  H_\beta(U)~,   
    \end{equation}
    where we recall  $\lambda_\beta =  \prod_i \lambda_i^{\beta_i}$ and we have defined, for $f \in L^2_{\gamma_q}, g \in L^2_{\gamma_r}$ the tensor product $f \otimes g : L^2_{\gamma_r} \to L^2_{\gamma_q}$ such that $(f \otimes g)(h) = \langle g, h \rangle_{\gamma_r} f $.
    \end{corollary}
\begin{proof}
    We have, thanks to the multiplicative property of $\Av$ that $ \mathsf{A}_M =  \mathsf{A}_V \mathsf{A}_\Lambda \mathsf{A}_{U^\top}$. Hence, for all $f \in L^2_{\gamma_r}$, we can decompose $\Av_{U^\top} f$ according to the canonical tensorized Hermite basis, that is, $\Av_{U^\top} f = \sum_\beta \langle \Av_{U^\top} f, H_\beta \rangle H_\beta$, so that 
    \begin{align*}
        \Av_M f = \Av_V \Av_\Lambda \left[\sum_\beta \langle \Av_{U^\top} f, H_\beta \rangle H_\beta\right] = \Av_V\left[\sum_\beta \lambda_\beta \langle \Av_{U^\top} f, H_\beta \rangle H_\beta\right] = \sum_\beta \lambda_\beta \langle f, \P_U H_\beta  \rangle \P_V H_\beta~,
    \end{align*}
    which is exactly the formula given in the corollary.
\end{proof}
This formula enables us to develop a convenient representation of the loss as a function of the SVD triplet $(\Lambda, U, V)$. We will use this representation to analyse the loss landscape and the gradient flow as presented in Sections~\ref{sec:grassmannian_perspective} and \ref{sec:planted_model}. Before this, in this section we show important properties of the averaging operator that will be needed later on. 

\subsubsection{Orthogonal Subspace Projections}
There is a one-to-one map between Grassmannians $W \in \G(r,q)$, with $r \geq q$, and the orthogonal projection onto $W$, written as $W W^\top \in \R^{r \times r}$. Remarkably, Corollary~\ref{cor:SVD_of_A} shows that $\Av_{W W^\top}$ is also an orthogonal projection of $L^2_{\gamma_r}$ onto the space of functions that depend intrinsically only on the coordinates $(w_1 \cdot x, \dots, w_q \cdot x)$. We denote $\Pi_W$ this orthogonal projection. Indeed, completing the orthonormal vectors of $W$ as a orthonormal basis of $\R^r$, i.e. $O = [W, W^\perp] \in \O_r$, with $W^\perp \in \G(r, r - q)$, we have the representation
\begin{align}
\label{eq:shrinkage_space}
    \Pi_W := \Av_{W W^\top} = \sum_{\beta\,;\, |\beta|_{q+1:r} = 0} H_\beta(O) \otimes H_\beta(O).
\end{align}
We will study this more deeply in the next subsection~\ref{subsec:intrinsic_dimension}. 
We now state a commutation property between thresholding operators and the averaging with respect to a PSD matrix. 
\begin{proposition}
    Let $W \in \mathcal{G}(q,p)$ and $G \in \C_q$. If the orthogonal projector $WW^\top$ and $G$ commute, then the operator $\mathsf{S}_W^s$ and $\mathsf{A}_G$ commute. In particular $\mathsf{S}^s$ and $\mathsf{A}_G$ commute for any matrix $G \in \C_q$. 
\end{proposition}
\begin{proof}
    If $WW^\top$ and $G$ commute, this means that $w_1, \dots, w_p$, the column vectors of the matrix $W$ are eigenvectors of $G$. Hence, the two operators $\mathsf{S}_W^s$ and $\mathsf{A}_G$ diagonalize in the basis $(H_\beta(U) \otimes H_\beta(U))_{\beta \in \N^q}$ where $U = [W, W^\perp] \in \O_q$ any orthogonal completion of $W$. 
\end{proof} 
We conclude this subsection by showing that the averaging operator is closed within each harmonic, and furthermore it can be expressed as a simple change of variables.
\begin{proposition}[Harmonic Representation of $\Av$]
\label{prop:ave_closed}
For all $M \in \R^{q \times q}$, such that $\|M\| \leq 1$, we have
    \begin{align}
        \Av_M = \sum_{k \geq 0}\mathsf{\Pi}^k \P_M \mathsf{\Pi}^k~, 
    \end{align}
    where for all $k \geq 0$, $\mathsf{\Pi}^k = \mathsf{S}^k - \mathsf{S}^{k-1}$, is the orthogonal projection onto the subspace of Harmonics function of degree $k$ exactly\footnote{with the convention $\mathsf{S}^{-1} \equiv 0$}.
\end{proposition}
The proof of this follows from Proposition~\ref{prop:local_representation} stated in the Appendix in Section~\ref{subsubsecapp:link_aveargin_hermite}. More generally, this calculation is a consequence of a general change of (orthonormal) basis formula for Hermite polynomials that is presented in Appendix~\ref{subsecapp:change_of_coordinates}.

\subsection{Intrinsic Dimension}
\label{subsec:intrinsic_dimension}

Besides the averaging operator, the other key object that will be needed later is the notion of 
intrinsic dimension. 
\subsubsection{Definition of the intrinsic dimension}
Informally, it corresponds to the smallest number of independent Gaussian variables needed to represent $F(x)$ for any $x$. 

\begin{definition}[Intrinsic Dimension]
\label{def:intrinsic_dim}
The \emph{intrinsic dimension} of $F \in L^2_{\gamma_d}$ is 
$$\mathsf{d}(F) = \inf\{\, q\,;\, \exists W \in \G(d,q) \ s.t.\ \Pi_{W} F = F \, \}~. $$
Furthermore, we say that $W$ is the \textit{support} of $F$ if $\,{\Pi}_W F = F$ and $W \in \G(d, \mathsf{d}(F))$.  
\end{definition}
Thanks to this definition, we show that each function can be represented uniquely by a basis function defined on its support.
\begin{proposition}
\label{prop:representation_intrinsic}
    The support of $F \in L^2_{\gamma_d}$, noted $W[F]$, is defined uniquely in $\G(d, \mathsf{d}(F))$. Furthermore, we can define the minimal representation of $F$ as the couple $(W[f], f)$, where $f \in L^2_{\gamma_{\mathsf{d}(F)}}$, is defined up to rotation such that $\P_{W[f]} f = F$ and $\mathsf{d}(f) = \mathsf{d}(F)$. 
\end{proposition}
\begin{proof}
    Suppose there exist two $W,W' \in \G(d, \mathsf{d}(F))$, such that $\Pi_{W} F = \Pi_{W'} F = F $. Denote $W = [w_1, \dots , w_{\mathsf{d}(F)}]$ and $W = [w'_1, \dots , w'_{\mathsf{d}(F)}]$. If there exists one vector in $W$ but orthogonal to $W'$, say wlog $w_{\mathsf{d}(F)}$, then the subspace $[w_1, \dots , w_{\mathsf{d}(F)-1}]$ would support $F$ as well. But this contradicts the minimality of $\mathsf{d}(F)$, hence the contradiction. Noting $f = \Av_{W^\top} F \in L^2_{\gamma_{\mathsf{d}(F)}}$, we have that $\P_{W[f]} f = F$, and if there exists $\bar{W} \in \G(\mathsf{d}(f), \mathsf{d}(f) - 1)$ such  that $\Pi_{\bar{W}} f = f$, then we have $\Pi_{W \bar{W}} F = F$, with $W \bar{W} \in \G(d, \mathsf{d}(f) - 1)$ but this contradicts the fact that $W$ has minimal rank. Hence, $\mathsf{d}(f)  =\mathsf{d}(F)$.
\end{proof}
Note that, for every non-zero $F \in L^2_{\gamma_d}$, we have the inequality  $1 \leq \mathsf{d}(F) \leq d$.
An immediate consequence of the definition is also that the intrinsic dimension is invariant to rotations:
\begin{proposition}[Rotational Invariance]
\label{prop:intrinsic_rotations}
    Let $f \in L^2_{\gamma_q}$ and $U \in \mathcal{O}_q$. Then $\mathsf{d}(f) = \mathsf{d}( \mathsf{P}_U f)$  and $W[ \mathsf{P}_U f] = U W[f]$.
\end{proposition}

\begin{example} To illustrate the notion, we derive the intrinsic dimension and support of the following classes of functions:
\begin{itemize}
    \item \emph{Single-index.} If $f(z) = g( z \cdot \theta )$, then $f$ has intrinsic dimension $1$ with support $\mathrm{span}(\theta)$. 
    \item \emph{Radial functions.} If for all $z \in \R^q$, $f(z) = \phi(\|z\|)$, then $f$ has intrinsic dimension $q$ with support $\R^q$.
    \item \emph{Pure Harmonics.} If $f = H_{\gamma}$ then the intrinsic dimension of $f$ is $q = | \text{supp}(\gamma)| = \left| \{j ; \, \gamma_j > 0 \}\right|$ with support $\mathrm{span}(e_j,\,s.t.\ \gamma_j > 0)$. 
\end{itemize}
\end{example}

Given $F\in H^1_{\gamma_q}$, the gradient Gram matrix is given by $\mathsf{G}_F = \mathbb{E}_x [ \nabla F(x) \nabla F(x)^\top] \in \mathbb{R}^{d \times d}$.
This matrix has been considered in prior works, e.g. \cite{hristache2001structure}, as a natural procedure to estimate the support of multi-index models. Unsurprisingly, the intrinsic dimension and support of $F$ are completely determined by~$\mathsf{G}_F$.
\begin{proposition}
    We have $W[F]=\mathrm{span}(\mathsf{G}_F)$. In particular, it holds $\mathsf{d}(F) = \mathrm{rank}(\mathsf{G}_F)$. 
\end{proposition}
\begin{proof}
Observe first that the gradient outer product is equivariant to changes of basis: if $V \in \mathcal{O}_d$ and $\tilde{F} = \mathsf{P}_V F$, then $\mathsf{G}_{\tilde{F}}=V^\top \mathsf{G}_F V$. In particular, we have that $\mathrm{span}( \mathsf{G}_{\tilde{F}}) = V \mathrm{span}(\mathsf{G}_F)$.
Since this rotation equivariance is also verified by $W[F]$, we can assume 
without loss of generality that $W[F]=[e_1, \ldots e_{\mathsf{d}(F)}]$ is aligned with the canonical basis, so $F(x_1, \ldots, x_d) = f(x_1, \ldots, x_{\mathsf{d}(F)})$. 
We verify immediately that any $e_j$, $j > \mathsf{d}(F)$ is in $\mathrm{span}(\mathsf{G}_F)^\perp$, showing that $W[F]^\perp \subseteq \mathrm{span}(\mathsf{G}_F)^\perp$ and hence $\mathrm{span}(\mathsf{G}_F) \subseteq W[F]$. 

Reciprocally, let $\mathsf{G}_F = V^\top \Lambda V$ be the eigendecomposition of $\mathsf{G}_F$, a symmetric psd matrix, and assume again w.l.o.g. that $V$ is the canonical basis. Define $j \in [1,d]$ such that $\lambda_{j'} = 0$ for $j'\geq j$. For $j'\geq j$, it follows that $\mathbb{E}_x [ (\partial_{x_{j'}} F(x))^2 ]=0$, so $F$ is $\gamma_d$-a.e. constant along the direction $e_{j'}$. Therefore, if $\alpha_\beta = \langle F, H_\beta \rangle$ is the expansion of $F$ in the canonical Hermite tensorized basis, we have $\alpha_\beta = 0$ for any $\beta$ such that $\beta_{j'} > 0$ for $j'\geq j$. By denoting $W := \mathrm{span}(e_1, \ldots, e_{j-1}) = \mathrm{span}(\mathsf{G}_F)$, it follows that 
$$\Pi_W F = \sum_{\beta; |\beta|_{(j:d)}=0} \langle F, H_\beta \rangle H_\beta = \sum_{\beta } \langle F, H_\beta \rangle H_\beta = F~.$$
Since $W \subseteq W[F]$ and $W[F]$ is by definition the minimal dimensional subspace with this property, we conclude that necessarily $W = W[F]$.
\end{proof}

\subsubsection{Minimal energy and stability properties of the intrinsic dimension}

For a function $f \in L^2_{\gamma_q}$, for any $W \in \mathcal{G}(q, p)$, with $p \leq q$, then we obviously have that the intrinsic dimension of $\Pi_W f$ is at most $p$. This observation leads to the following definition.
\begin{align}
\label{eq:energap}
    \mathcal{E}_p(f) := \inf_{W \in \mathcal{G}(q, p)}{\| f  -  \Pi_W f \|^2 }.
\end{align}
Hence, $\mathcal{E}_p(f)$ stands for a square distance between $f$ and the subset of functions of $L^2_{\gamma_q}$ that have intrinsic dimension $p$ ($\simeq L^2_{\gamma_p}$ with Proposition~\ref{prop:representation_intrinsic}). From the definition, note that the intrinsic dimension of $f$ is characterised by the smallest integer $p$ such that $\mathcal{E}_p(f)=0$, hence $\mathcal{E}_{ \mathsf{d}(f)}(f) = 0$ and $\forall p < \mathsf{d}(f)$, $\mathcal{E}_{p}(f) > 0$. 
\begin{definition}[Minimal energy]
    We define the minimal energy of a function as $\mathcal{E}(f) := \mathcal{E}_{ \mathsf{d}(f) - 1}(f) > 0$.
\end{definition}
\begin{proposition}
    For $f \in L^2_{\gamma_q}$, and $p \leq q$, we have the equalities
    \begin{align}
        \mathcal{E}_p(f) = \| f \|^2  - \sup_{W \in \mathcal{G}(q,p)}  \|\Pi_{W} f\|^2 = \inf_{U \in \mathcal{O}_q} \sum_{\beta_{p+1}> 0, \dots, \beta_{q} > 0} \langle f, H_\beta(U) \rangle^2  ~.
    \end{align}
    In particular, $\mathcal{E}(f) =  \| f \|^2 - \sup_{v \in \S_{q-1}}  \|\Av_{I-vv^\top}f\|^2$.
\end{proposition}
\begin{proof}
    The first inequality comes from the fact that $\Pi_W$ is an orthogonal projection and Pythagorean theorem. The second inequality results from the representation given in Eq.\eqref{eq:shrinkage_space} of $\Pi_W$. Finally the last consequence comes from the fact that there the space of Grassmannians $\G(q, q-1)$ is one-to-one with the space of orthogonal projector on $\mathrm{span}(v)^\perp$, which writes $\{I - v v^\top, \  \text{for } v \in \S_{d-1}\}$. 
\end{proof}
This proposition gives a clear interpretation of the energy $\mathcal{E}_p(f)$. It corresponds to the \textit{minimal} energy that the function $f$ carries over $q - p$ directions. We verify immediately that it is invariant to rigid rotations $\mathcal{E}(f) = \mathcal{E}(\mathsf{P}_U f) $, in consistency with Proposition \ref{prop:intrinsic_rotations}.
We now verify that the intrinsic dimension of a function cannot increase by averaging, and is preserved for full-rank averaging matrices. 
\begin{restatable}[Intrinsic dimension and energy after averaging]{proposition}{propintrinsicaveraging}
\label{prop:intrinsic_averaging}
Let $W[f] \in \mathcal{G}(q,p)$ be the support of $f \in L^2_{\gamma_q}$ with minimal representation $\bar{f} \in L^2_{\gamma_p}$ and $M \in \mathbb{R}^{q \times q}$ with $\| M \| \leq 1$. Then 
 \begin{enumerate}[label=(\roman*)]
    \item We have $\mathcal{E}(f) = \mathcal{E}(\bar{f})$.
     \item $\mathsf{A}_M f$ has intrinsic dimension $\mathrm{rank}(MW[f]) \leq q$, supported in $\mathrm{span}(MW[f])$. In particular, if $M$ is invertible then $\mathsf{A}_M f$ has the same intrinsic dimension as $f$. 
     \item For $s\in\N^*$, if $f \in \mathcal{B}_s$ then, letting $\mu = \sigma_{\min}(MW[f])$, we have $\mathcal{E}(\mathsf{A}_M f) \geq \mu^{2s} q^{-s} \mathcal{E}(f)$~.
 \end{enumerate}
\end{restatable}
Finally, we verify that similarly, spectral thresholding cannot increase the intrinsic dimension. 
\begin{restatable}[Intrinsic dimension and energy after spectral thresholding]{proposition}{propintrinsicspectral}
\label{prop:intrinsic_spectral}
    For any $f\in L^2_{\gamma_q}$ and $s\in \mathbb{N}$,  we have  
    $\mathsf{d}(\mathsf{S}^{s}(f)) \leq \mathsf{d}(f)$, $\mathcal{E}(\mathsf{S}^{s}(f)) \leq \mathcal{E}(f)$, and $W[\mathsf{S}^{s}(f)] \subseteq W[f]$. 
    Moreover, there exists a minimal $r \in \mathbb{N}$ such that $\mathsf{d}(\mathsf{S}^{r}(f)) = \mathsf{d}(f)$ and $W[\mathsf{S}^{r}(f)] = W[f]$.
\end{restatable}
The proofs of these two results can be found in Appendix~\ref{subsubsecapp:properties_idim}.

\subsection{Further properties of \texorpdfstring{$\Av$}{A}}

\paragraph{A Hilbert-Schmidt operator}%

We begin this section by providing some structure of $\Av$ seen as a bounded operator.

\begin{restatable}{proposition}{propcompactness}
    \label{prop:compactness}
    For all $M \in \mathbb{R}^{q \times r}$ such that $\|M\| \leq 1$, 
    \begin{enumerate}[label=(\roman*)]
        \item  $\Av_M$ is bounded linear operator with operator norm: $\| \mathsf{A}_M \| = \|M\|$.
        \item If $\|M\| < 1$,  $\Av_M$ is a Hilbert-Schmidt operator with norm $\|\mathsf{A}_M\|_{\mathrm{HS}} = \prod_{i=1}^q (1 - \lambda_i^2)^{-1/2}$.
        \item Finally, for $s\in\N^*$, we have that $\Av_M \circ \mathsf{S}^s$ is a finite rank operator and $\sigma_{\min}(\Av_M \circ \mathsf{S}^s) = \sigma_{\min}(M)^s$.
    \end{enumerate}
\end{restatable}
\begin{proof}
    We refer to Section~\ref{subsubsubsecapp:final_properties} of the Appendix for a proof of this result. 
\end{proof}
A standard consequence of this fact is that the operator is compact whenever $\|M\| < 1$ thanks to property (ii) that he is Hilbert-Schmidt. Hence, it can be decomposed thanks to its singular value decomposition~\cite[Chapter 11, Theorem 1]{birman2012spectral}, with countably many singular values that goes to $0$. This is explicitly the representation given by Corollary~\ref{cor:SVD_of_A}. 

\paragraph{Analytic Extension}

This paragraph leverages the fact that, expressed as its singular value decomposition given in Corollary~\ref{cor:SVD_of_A}, Eq.\eqref{eq:SVD_of_A}, the operator $\Av_M$ does not really require $\|M\| \leq 1$ to exist. This is in contrast with its initial definition given in Definition \ref{def:averaging_operator}, for which it appeared to be necessary. In fact, if the Hermite coefficients of $f$ are decreasing fast enough, it is legitimate to define the analytic extension $\overline{\mathsf{A}}$ of $\mathsf{A}$ for general matrices $M$. We formalize it in this section and show that this extension still satisfies the properties given in the previous sections --notably the multiplicative semigroup property. 

Recall first the definition of \textit{tempered} functions, given in Eq.$\eqref{eq:spaces_of_functions}$,  $\mathcal{T}_{R} := \{ f \in L^2_{\gamma_q}\ ;\, \sum_\beta R^{2 |\beta|} \langle f, H_\beta \rangle^2 < + \infty  \}$. And let's equip it with a norm, defined naturally for all $f \in \mathcal{T}_{R}$, as $\|f\|^2_{\mathcal{T}_R} = \sum_\beta R^{2 |\beta|} \langle f, H_\beta \rangle^2$. This makes $(\mathcal{T}_{R}, \|\cdot\|_{\mathcal{T}_R})$ a Banach space.

For any $R > 0$, we define $\Av_M$ for all $M\in \mathbb{R}^{q \times r}$, such that $\|M\|\leq R$. Indeed, let $M = V \Lambda U^\top$ be its singular value decomposition, we define the operator $\overline{\mathsf{A}}_{M}$ with domain $\mathcal{T}_{R}$, such that  
    \begin{align}
        \overline{\mathsf{A}}_{M} :=  \sum_{\beta} \lambda_\beta  H_\beta(V) \otimes H_\beta(U)~.
    \end{align}
As announced, we have the following properties.
\begin{proposition}
    \label{prop:Aextprod}
    The operator $\overline{\mathsf{A}}_{M}:\mathcal{T}_{R} \to L^2_{\gamma_q} $ is a well-defined and bounded, with norm $\|\overline{\mathsf{A}}_{M}\| \leq 1$. Furthermore, for any pair $M \in \mathbb{R}^{r \times p }$, $N \in \mathbb{R}^{p \times q }$, such that $\|M\|, \|N\|\leq R$ we have  $\overline{\mathsf{A}}_{M} \overline{\mathsf{A}}_{N} = \overline{\mathsf{A}}_{M N}$.
\end{proposition}
    \begin{proof}
        For $f \in \mathcal{T}_{R}$, we have, $\overline{\mathsf{A}}_{M} f = \sum_{\beta} \lambda_\beta \langle f,  H_\beta(U) \rangle  H_\beta(V)$, so that $$\|\overline{\mathsf{A}}_{M} f\|^2_{L^2_{\gamma_q}} = \sum_{\beta} \lambda_\beta^2 \langle f,  H_\beta(U) \rangle^2 \leq \sum_{\beta} R^{2 |\beta|} \langle f,  H_\beta(U) \rangle^2 =\| f \|^2_{\mathcal{T}_R} ~.$$
        Hence the first point. Now, let call $\lambda, \gamma$ the $q$-tuples of singular vectors of $M,N$. The second point follows from the fact that the function $(\lambda, \gamma) \to \overline{\mathsf{A}}_{M} \overline{\mathsf{A}}_{N} - \overline{\mathsf{A}}_{M N}$ is analytic and cancels on the infinity ball of radius $1$ by Lemma~\ref{lem:averaging_operator_multiplicative}. Hence by the isolated zero principles, it is uniformly zero on whole its domain.
    \end{proof}

This analytic extension will come in helpful to extend some operations later in the article, and find a direct application in the next section to define the inverse of $\Av_G$ when $G$ is PSD.

\paragraph{Properties of $\Av$ in the PSD cone}

Here we detail a bit more the study of $\Av_G$ when $G \in \R^{q \times q}$ is a PSD matrix. Following the discussion of the previous subsections, we can assert that for $\|G\| \in \C_q$, $\|G\| < 1$, we have that $A_G$ is a compact (even Hilbert-Schmidt) self-adjoint operator and therefore diagnonalizable in an orthonormal basis. Furthermore, Corollary~\ref{cor:SVD_of_A} specifies a natural basis of diagonalization: the one of Hermite polynomials associated to the eigenvectors of $G$. Indeed, considering the Eigenvalue decomposition of $G := U \Lambda U^\top $, we have the representation
\begin{align}
\label{eq:representation_A_PSD_order}
    \Av_G := \sum_\beta \lambda_\beta H^U_\beta \otimes H^U_\beta ~.
\end{align}
Now, we can pursue and extend $\Av$ as been done in previous subsections. In fact, in this case we can also define properly a bounded inverse of the operator. Together with this result, we also see that $G \to \Av_G$ is also monotonic for the Loewner order on PSD matrix/operators.
\begin{proposition}[Loewner Order Preservation]
\label{prop:loewner}
    Let $G, H$ two PSD matrices of $\R^{q \times q}$, such that  $G \preccurlyeq H$ and $R^{-1} \leq \lambda_{\min}(G) \leq \lambda_{\max}(H) \leq R $, then
    \begin{enumerate}[label=(\roman*)]
        \item On the domain $\mathcal{T}_R$, we have the equality $\Av_G^{-1} = \Av_{G^{-1}}$.
        \item We have that $\mathsf{A}_G \preceq \mathsf{A}_H$ as operator with domain $\mathcal{T}_R$. 
    \end{enumerate}
\end{proposition}
\begin{proof}
    The first equality follows from the definition and the fact that operators are well defined thanks to the conditions on the spectrum of $G$. For the point (ii), we begin to observe that since $G \preceq   H$, we have $ H^{-1/2} G H^{-1/2} \preceq I_q $. But this implies that $ \mathsf{A}_{H^{-1/2} G H^{-1/2}}  = \mathsf{A}_H^{-1/2} \mathsf{A}_G \mathsf{A}_H^{-1/2}   \preceq \mathrm{Id}(L^2_{\gamma_q})$ and inverting the inequalities again we have $\mathsf{A}_G \preceq \mathsf{A}_H$. Note that every inversion is valid thanks to the spectrum constraints. 
\end{proof}
\subsection{Partial conclusion}

We have derived all the important properties that we will later need to study the learning procedure associated with the loss. Before moving forward, let us summarize shortly the main information on what we have built so far. First, we have shown that the overall loss only depends on \textit{dimension-free} objects: it is quadratic in the low-dimensional function~$f \in L^2_{\gamma_q}$, and depends on the subspace correlation $M = W_*^\top W \in \R^{q \times r} $ through the averaging operator $\Av_M$. Then, we have shown that this operator had a simple representation with respect to the SVD of $M$ and rotated Hermite polynomials (according to the left and right eigenvectors of $M$). Finally, we have defined a notion of energy with respect to lower-dimensional filters on the function $f$: this corresponds to the intrinsic dimension that the function $f$ carries over subspaces that will carry the dynamics throughout the dynamics that we present in the following section.

%% file: IV-Grassmannian.tex
\section{Jointly learning the link function: the semi-parametric model}
\label{sec:grassmannian_perspective}

Let us recall the results of Section~\ref{sec:regression_model} and notably the loss given in Proposition~\ref{prop:representation_loss}. The models corresponds to learning both the function $f \in L^2_{\gamma_r}$, and the subspace $W \in \G(d, r)$ via the loss  
\begin{align}
\label{eq:loss_learning_model_ref}
L(f, W) = \frac{1}{2}\| f\|_{\gamma_r}^2  + \frac{1}{2}\| f^*\|_{\gamma_q}^2 -  \langle  \mathsf{A}_{M} f, f^* \rangle_{\gamma_q}~,
\end{align} 
where $ M = W_*^\top W \in \R^{q \times r}$ and $f^* \in L^2_{\gamma_q}$.

This corresponds to a \emph{semi-parametric} problem in which one wants to solve both a high-dimensional parameter recovery (the subspace $W_*$), as well as a non-parametric recovery of a low-dimensional function (the target $f^*$). 
As discussed earlier, the ability of gradient descent methods to solve this statistical inference task fundamentally depends on how efficiently one can optimize the associated population loss given by (\ref{eq:loss_learning_model_ref}). 

This section presents a gradient-flow learning procedure, and establishes quantitative time complexity guarantees $T=O( d^{s^*-1/2} + \log \epsilon^{-1})$ to reach a certain target accuracy $\epsilon$, where $s^*=s^*(f)$ is a suitable generalization of the information exponent from single-index models.

\subsection{Fast non-parametric learning}
\label{subsec:fast_learning}
\input{IV-Grassmann_fast/Fast_learning}

\subsection{Analysis of the loss landscape} 
\label{subsec:critical_points}

\input{IV-Grassmann_fast/Critical_points}

\subsection{Grassmanian Gradient Flow dynamics}
\label{subsec:gradient_flow}

In this section we describe the optimization dynamics that follows $W$ when optimizing the loss function $W \to L(W)$, given by Eq.\eqref{eq:lossgrassm0} (or, equivalently, the ridge regularised loss given by Eq. \eqref{eq:rkhs_grassloss}). Starting from a random matrix in the Grassmannian manifold $W_0 \sim \mathrm{Unif}(\G(d,r))$, we learn the subspace $W$ according to the gradient flow:
\begin{align}
    \label{eq:gradient_flow_on_W}
    \dot{W}_t = \nabla_W^{\G} L (W_t) = (I - W W^\top) \nabla_W L (W_t) ~, 
\end{align}
where we recall that the loss is the correlation given by Eq.\eqref{eq:lossgrassm0}, expressed trough the averaging operator:  $L(W) = \langle \Av_{G_W} f, f \rangle$, where $G_W = M M^\top = W_*^\top W W^\top W_* \in \R^{q \times q}$.

\input{IV-Grassmann_fast/Gradient_flow}

%% file: IV-Grassmann_fast/Fast_learning.tex
\subsubsection{Two-timescale idealized optimization procedure}
\label{sec:fastideal}

The joint optimization of $L$ over $(f, W) \in L^2_{\gamma_r} \times \mathcal{G}(d,r)$ has the appealing property that the loss $L(f, W)$ is quadratic and strongly convex with respect to its first argument $f$. This motivates an idealized local descent procedure whereby the convex structure is `optimized away'; that is, for 
each $t \geq 0$, we solve the optimization problem  $f_{W}  := {\mathrm{argmin}_{f \in L^2_{\gamma_r}}}\  L(f, W)$, and then locally optimize $W_t$ according to a steepest descent on the loss $W \to L(f_W, W)$. Since $L(f,W)$ is a quadratic function of $f$, we can build $f_W$ is closed form. Indeed, 
\begin{align}
\label{eq:lossgrass_f}
f_W  := \underset{\ \  f \in L^2_{\gamma_r}}{\mathrm{argmin}}\  L(f, W) =  \underset{\ \  f \in L^2_{\gamma_r}}{\mathrm{argmin}}\    \frac{1}{2}\|f \|_{\gamma_r}^2 -  \langle f, \Av_{M^\top} f^*\rangle_{\gamma_r} = \Av_{M^\top} f^*.
\end{align}
Hence, plugging $f_W$ back into the loss, if we denote $G = M  M^\top\!\!  = V \Lambda^2 V^\top\!\! \in \C_q$, with a slight abuse of notation, the resulting loss writes
\begin{align}
\label{eq:lossgrassm0}
{L}(W) := L(f_W, W) =  \frac12\|f^*\|_{\gamma_q}^2 - \frac12\langle \mathsf{A}_{G} f^*, f^* \rangle_{\gamma_q}~.
\end{align}

Notice that this loss only depends on $W$ through the projection $WW^\top$, hence, up to a change of basis in the $r$-dimensional subspace, i.e. we have the equivalence relation $W \simeq W'$ if $W' = WU$ for $U \in \mathcal{O}_r$. This is precisely the Grassmann manifold $\mathcal{G}(d,r)$, and so we will consider the optimization inherited in this quotient manifold structure for the remainder of the section. 

Given the eigenrepresentation of the operator $\Av_G$ given in Eq.\eqref{eq:representation_A_PSD_order}, the loss measures the norm of $f^*$ after applying a shrinkage operator $\Av_{\Lambda^2}$ over a  $V$-rotation of the tensorized Hermite basis. Naturally, in the generic case, one expects this norm to be maximised when the shrinkage operator converges to the identity, i.e. $\Lambda \to I_q \Leftrightarrow G \to I_q$.  Note also that from now on we consider for simplicity $r = q$, as there is no difference due to overparametrization $r > q$ at this point.

\subsubsection{A practical method: Fast RKHS Learning}
\label{sec:rkhs_learning}

The procedure described above considers an idealised scenario where the non-parametric regression is performed on the whole ambient space 
$L_{\gamma_q}^2$. Since we view the optimization of the loss $L$ as the population limit of an underlying estimation problem, such $L^2_{\gamma_q}$ regression is unpractical when faced with finite number of data. Instead, 
we perform this non-parametric regression on a dense Reproducing Kernel Hilbert Space $\mathcal{H} \subset L_{\gamma_q}^2$, written as RKHS in short, see \cite{scholkopf2002learning} or \cite{schaback2006kernel} for an overview. 
The main takeaway of this Section is that, for appropriate choices of isotropic kernels, \emph{the resulting kernel regression will respect the geometry of the loss \eqref{eq:lossgrassm0} given in the previous idealised scenario}.  
Let us detail this below. 

\paragraph{An isotropic Hermite RKHS} 

First, we construct such a RKHS space, adapted to the Hermite structure of our problem. To do this let us define 
$(\mathsf{c}_k)_{k \in \N} \in \ell_2(\N)$ a positive sequence.  
Furthermore, we ask that $\sum_\beta \mathsf{c}_{|\beta|} =  \sum_k  \binom{k + q - 1}{k} \mathsf{c}_k < +\infty $, that is to say that $\mathsf{c}$ has some fast decay at infinity, e.g. $\mathsf{c}_k = O(k^{-(q+1)})$ and anything faster as $\binom{k + q - 1}{k} \sim k^{q-1}$. Then, we use the following construction.
\begin{proposition}
    Let $\{H_\beta\}_\beta$ be the Tensorized Hermite basis associated to an arbitrary basis of $\mathbb{R}^q$. 
    For all sequences $(\mathsf{c}_k)_{k \in \N}$ defined as previously, the subspace     %
\begin{align}
    \mathcal{H} := \big\{ f \in L^2_{\gamma_q} \cap \mathcal{C}(\R^q)\ ; \ \text{such that } \| f\|^2_{\mathcal{H}} := \sum_{\beta} \mathsf{c}_{|\beta|}^{-1} \langle f, H_\beta \rangle^2  < +\infty \big\} 
\end{align}
    defines a dense RKHS with kernel function $K: \R^q \times \R^q \to \R$, 
\begin{align}
\label{eq:ourkernel}
    \forall x,y \in \R^q, \quad K(x, y) = \sum_{\beta} \mathsf{c}_{|\beta|} H_\beta(x) H_\beta(y)~, 
\end{align}
where the equality stands in a point-wise sense. Finally, we also define, for any $f,g \in \mathcal{H}$, the inner product $\langle f, g \rangle_{\mathcal{H}} := \sum_\beta \mathsf{c}_{|\beta|}^{-1} \langle f, H_\beta \rangle \langle g, H_\beta  \rangle $.
\end{proposition}
Note that similar RKHSs have already been considered for numerical computation of expectation with respect to Brownian paths, e.g.~\cite{irrgeher2015high} and statistical estimation of multi-index problems~\cite{follain2023nonparametric}. We refer to the first cited paper for the construction as well as more on smoothness properties of such RKHSs.

The RKHS norm $\| f  \|_{\mathcal{H}}^2$ is thus of the form $\sum_\beta \tilde{\mathsf{c}}_\beta \langle f, H_\beta \rangle^2 $, and therefore defines a weighted Sobolev space where the spectral weights are precisely given by $\tilde{\mathsf{c}}_\beta = \mathsf{c}_{|\beta|}$. 
The kernel $K(x,y)$ has an associated integral operator $\mathsf{Q}: L^2_{\gamma_q} \to L^2_{\gamma_q}$, given by 
\begin{equation}
\label{eq:kernel_integral_op}
    \mathsf{Q}f := \int K(\cdot, y) f(y) \mathrm{d}\gamma_q(y)~,
\end{equation}
that naturally defines $Q^{1/2}$ as an isometry between $\mathcal{H}$ and $L^2_{\gamma_q}$ as $\langle Q f, g \rangle_{\mathcal{H}} = \langle f, g\rangle_{\gamma_q} $.

By definition, the pure harmonics $\text{span}\{ H_\beta; |\beta|=k \}$ are eigenspaces of $\mathsf{Q}$, with associated eigenvalues $\mathsf{c}_k$, with multiplicity  $\binom{k + q - 1}{k}$, yielding the spectral representation $\mathsf{Q} = \sum_\beta \mathsf{c}_{|\beta|} H_\beta \otimes H_\beta$.  Note that Mercer theorem gives back the expression of $K$.   

A crucial remark is that since $\tilde{\mathsf{c}}_\beta$ only depends on the order $|\beta|$, the resulting kernel is isotropic, and therefore independent of the choice of basis:
\begin{proposition}
    The RKHS $\mathcal{H}$ defined above is well-defined: it is closed under unitary transformations, i.e.  $\mathsf{P}_U f \in \mathcal{H}$ whenever $f \in \mathcal{H}$ for any $U \in \mathcal{O}_q$, with $\| \mathsf{P}_U f\|_{\mathcal{H}} = \|f \|_{\mathcal{H}}$. Moreover, the associated kernel function is rotationally invariant: $K( U x, U y) = K(x, y)$ for any $x,y \in \mathbb{R}^q$ and $U \in \mathcal{O}_q$. 
\end{proposition}
\begin{proof}
    We immediately verify from Lemma \ref{claim:hermite_wellposed} that  
    \begin{align*}
     \| \mathsf{P}_U f\|_{\mathcal{H}}^2 &= \sum_k \mathsf{c}_k^{-1} \sum_{|\beta|=k} \langle \mathsf{P}_U f, H_\beta \rangle^2 = \sum_k \mathsf{c}_k^{-1} \sum_{|\beta|=k} \langle f, H_\beta \rangle^2 \nonumber = \|f \|_{\mathcal{H}}^2~.   
    \end{align*}
    Let us now verify that $K(x,y)$ is rotationally invariant. For each $U \in \mathcal{O}_q$, let $\mathsf{P}_U K(x,y) = K(Ux, Uy)$ and $\mathsf{Q}_U$ the associated integral operator. But since $\text{span}\{ H_\beta; |\beta|=k \} = \text{span}\{ H_\beta(U); |\beta|=k \} $ we have that $\mathsf{Q}_U \equiv \mathsf{Q}$, hence $\mathsf{P}_U K = K$ $\gamma_q$-a.e, which extends to any $x,y$ since $K$ is continuous.
\end{proof}
By Weyl's first fundamental theorem for $\mathcal{O}_q$, the kernel $K(x,y)$ is thus of the form $K(x,y) = \varphi( x \cdot y, \|x\|, \|y\|)$ for a certain function $\varphi: \R^3 \to \R$ parametrized by the spectrum $(\mathsf{c}_k)_k$. It is therefore tempting to construct our RKHS starting from a given $\varphi$. Such a function $\varphi: \R^3 \to \R$ is an admissible choice for a kernel of the form of Eq.\eqref{eq:ourkernel} if, and only if, the following reproducing condition holds:
for any $\beta \in \N^q$ and $x \in \R^q$, 
\begin{align}
     \mathbb{E}_{y \sim \gamma_q} \left[ \varphi( x \cdot y, \|x\|, \|y\|) H_\beta(y) \right] &= \mathsf{c}_{|\beta|} H_\beta(x)~.
\end{align}

\paragraph{Random Feature Expansion}
\label{par:random_features}
    The kernel, being expressed as an infinite sum, might not be tractable in practice. However, one can use a random process to express it, as it has been already pinned as a \textit{random feature expansion of the kernel}~\cite{rahimi2007random}. Indeed, setting $\|\mathsf{c}\|_1 = \sum_k \mathsf{c}_k$, and defining the probability on $\N^q$ according to $p(\beta) := \|\mathsf{c}\|_1^{-1} \mathsf{c}_{|\beta|}$, we have for all $x, y \in \R^q$,
\begin{align}
    \label{eq:kernel_random_features}
    K(x, y) &= \|\mathsf{c}\|_1 \E_{\beta \sim p}\left[ H_\beta(x) H_\beta(y) \right]~,
\end{align}
    which provides an efficient practical alternative via the random feature expansion
    \begin{align}
    \label{eq:RKHSrandomfeature}
        \hat{K}(x,y) &= \frac{\|\mathsf{c}\|_1}{N} \sum_{i=1}^N H_{\beta_i}(x) H_{\beta_i}(y)~,
    \end{align}
    where $\beta_1,\ldots, \beta_N$ are $N$ i.i.d. samples from $p$.  %

\paragraph{Optimization over the RKHS} 

As will be discussed more deeply in Section~\ref{subsec:sample_complexity}, this construction enables us to derive a proper two-step algorithm: alternating between (i) one step of Grassmannian gradient descent for $W$ on the training loss (that is replacing expectation with respect to $\gamma_d$ by an empirical average) and (ii) solving perfectly the non-parametric on~$f$. For statistical purposes, step (ii) is often conducted while adding a regularizer $\mu \|f\|^2_\mathcal{H}/2$ for some $\mu > 0$. In this case the regularized loss function that we consider is
\begin{align}
\label{eq:loss_learning_model_ref_regularize}
L_{\mu}(f, W) := \frac{1}{2}\| f\|_{\gamma_q}^2  + \frac{1}{2}\| f^*\|_{\gamma_q}^2 -  \langle  \mathsf{A}_{M} f, f^* \rangle_{\gamma_q} + \frac{\mu}{2} \|f\|^2_\mathcal{H}~,
\end{align} 
over $W \in \G(d,q)$ and $f \in \mathcal{H}$. Thanks to the isotropic properties of the RKHS, the minimiser with respect to $f \in \mathcal{H}$ is still explicit, recovering an analog of Eq. \eqref{eq:lossgrassm0}:
\begin{proposition}[Fast Kernel Grassmann Loss]
\label{prop:kernelopt}
 We have $f^\mu_W  := \mathrm{argmin}_{f \in \mathcal{H}}\ L_{\mathcal{H}}(f,W) = [\mathsf{Q}+\mu I]^{-1} \mathsf{Q} \mathsf{A}_{M^\top} f^*$,
and therefore
\begin{equation}
\label{eq:rkhs_grassloss}
    L_{\mu}(W):= L_{\mu}(f^{\mu}_W, W) = \frac{1}{2}\| f^*\|_{\gamma_q}^2 -  \frac{1}{2} \langle  \mathsf{A}_{G} f^\mu_*, f^\mu_* \rangle_{\gamma_q}~,
\end{equation}
where we denoted $f^\mu_* := [\mathsf{Q} + \mu I]^{-1/2} \mathsf{Q}^{1/2} f^*$.
\end{proposition}
\begin{proof}
Observe that for any $g \in L^2_{\gamma_q}$, we have $\mathsf{Q} g \in \mathcal{H}$, and, given 
$f \in \mathcal{H}$, by definition we have the identity 
$\langle f, g \rangle_{\gamma_q} = \langle f, \mathsf{Q} g \rangle_{\mathcal{H}}$. Observe that by Proposition \ref{prop:ave_closed}, the averaging operator $\Av_M$ and the kernel integral operator $\mathsf{Q}$ commute, and thus we have the identity $\langle  \mathsf{A}_{M} f, f^* \rangle_{\gamma_q} = \langle   f,  \mathsf{A}_{M^\top} \mathsf{Q} f^* \rangle_{\mathcal{H}}$, which, in turn transforms the loss into a quadratic form with respect to $f \in \mathcal{H}$; minimizing it explicitly over the RKHS space yields 
\begin{align}
\label{eq:kernel_minimiser}
f^\mu_W  &:= \underset{\ \  f \in \mathcal{H}}{\mathrm{argmin}}\ \frac{1}{2}\langle\mathsf{Q}f,f\rangle_{\mathcal{H}}  + \frac{1}{2}\| f^*\|_{\gamma_q}^2 -  \langle   f, \mathsf{Q} \mathsf{A}_{M^\top} f^* \rangle_{\mathcal{H}} + \frac{\mu}{2} \|f\|^2_\mathcal{H} \nonumber \\
&=  [\mathsf{Q} + \mu I]^{-1} \mathsf{Q} \mathsf{A}_{M^\top} f^* \nonumber \\
&= \mathsf{A}_{M^\top} [\mathsf{Q} + \mu I]^{-1} \mathsf{Q} f^*~.
\end{align} 
Finally, this results in the loss to minimize: 
\begin{align*}
    L(f^\mu_W, W) &= \frac{1}{2}\langle(\mathsf{Q} + \mu I )f^\mu_W,f^\mu_W\rangle_{\mathcal{H}}  + \frac{1}{2}\| f^*\|_{\gamma_q}^2 -  \langle   f^\mathcal{H}_W, \mathsf{Q} \mathsf{A}_{M^\top} f^* \rangle_{\mathcal{H}} \nonumber\\
    &= \frac{1}{2}\| f^*\|_{\gamma_q}^2 + \frac{1}{2}\langle(\mathsf{Q} + \mu I )\mathsf{A}_{M^\top} [\mathsf{Q} + \mu I]^{-1} \mathsf{Q} f^*,\mathsf{A}_{M^\top} [\mathsf{Q} + \mu I]^{-1} \mathsf{Q} f^*\rangle_{\mathcal{H}}  \\
    & \hspace{5.5cm}-  \langle   \mathsf{A}_{M^\top} [\mathsf{Q} + \mu I]^{-1} \mathsf{Q} f^*, \mathsf{Q} \mathsf{A}_{M^\top} f^* \rangle_{\mathcal{H}} \nonumber\\
    &=\frac{1}{2}\| f^*\|_{\gamma_q}^2 -  \frac{1}{2} \langle  \mathsf{A}_{G} \mathsf{Q} f^*, [\mathsf{Q} + \mu I]^{-1} \mathsf{Q} f^* \rangle_{\mathcal{H}} \nonumber\\
    &=\frac{1}{2}\| f^*\|_{\gamma_q}^2 -  \frac{1}{2} \langle \mathsf{Q}  \mathsf{A}_{G} [\mathsf{Q} + \mu I]^{-1/2} \mathsf{Q}^{1/2} f^*, [\mathsf{Q} + \mu I]^{-1/2} \mathsf{Q}^{1/2} f^* \rangle_{\mathcal{H}} \nonumber \\
     &=\frac{1}{2}\| f^*\|_{\gamma_q}^2 -  \frac{1}{2} \langle  \mathsf{A}_{G} f^\mu_*, f^\mu_* \rangle_{\gamma_q}~,
\end{align*}~
where we use for the last line the isometry induced by $\mathsf{Q}$ to map the dot product back to $L^2_{\gamma_q}$. 
\end{proof}
In conclusion, the optimization problem arising from the fast kernel learning is analogous to the ambient fast learning from Eq. \eqref{eq:lossgrassm0}, 
where we have replaced $f^*$ by $f^\mu_* = [\mathsf{Q} + \mu I]^{-1/2} \mathsf{Q}^{1/2} f^*$, which is a spectral shrinkage of the initial target $f^*$. Since $\| [\mathsf{Q} + \mu I]^{-1/2} \mathsf{Q}^{1/2} \| < 1$, the resulting population loss now carries the approximation error $\|f^*\|^2 - \|f_*^{\mu} \|^2 > 0$. Crucially, this approximation error is \emph{independent of $W$}, so it does not affect the geometry of the landscape nor the gradient flow dynamics, after replacing $f^*$ by the regularised target $f^\mu_*$. In the following, for the sake of clarity, we will rename this target as $f^*$.

\paragraph{Neural Network Implementation} The isotropic kernel from Eq.~\eqref{eq:ourkernel} provides a seamless adaptation of the idealized method from Section \ref{sec:fastideal} with statistical and computational guarantees, thanks to its random feature expansion from Eq.~\eqref{eq:RKHSrandomfeature}. 
One can `interpret' the resulting model as a certain type of neural network, where the first layer performs a low-rank linear projection $x \mapsto W^\top x $ parametrized by $W \in \mathcal{S}(d,q)$, the second layer performs a (fixed) random feature expansion $z\in \R^q \mapsto \Phi(z)=( H_{\beta_i}(z) )_{i=1\ldots N} \in \R^N$, with $\beta_i \sim p$, and finally the third layer performs linear regression. In summary, 
$F_\theta(x) = a^\top \Phi( W^\top x)$ is our `student' network, with trainable parameters $\theta=\{W, a\},\, W \in \mathcal{S}(d,q)\,, a\in \R^N$.  

Admittedly, this architecture is not standard, mainly due to the choice of non-linearities, here given by a random family of Hermite polynomials of diverging degree. Such a choice is motivated to preserve the geometric structure of the problem, for arbitrary choices of regularization parameter $\mu$. We mention however that the choice of kernel could be relaxed, as long as it defines a dense RKHS with `preference' for smooth functions; see Section \ref{subsec:sample_complexity} for a discussion that highlights the importance of the kernel ridge regularisation.

\paragraph{Link with two time-scale gradient flows} 
We decided to present a two step procedure where the non-parametric learning on the quadratic cost Eq.~\eqref{eq:loss_learning_model_ref_regularize} is solved explicitly in closed form. However, at an informal level, there is a way to get rid of this procedure and consider only gradient flow learning jointly on the two variables $(f, W)$. The crux is to force the system to decouple the time scales of learning between $f$ and $W$, as we want to learn $f$ infinitely faster than $W$. This can be seen informally as a \textit{fast-slow gradient flow}: for a small parameter $0<\varepsilon \ll 1$
\begin{align}
    \label{eq:fast_slow}
    \frac{d}{dt} f_t = \frac{1}{\varepsilon} \nabla_f L(f_t, W_t) \quad \text{ and } \frac{d}{dt} W_t =  \nabla^{\G}_W L(f_t, W_t)~.
\end{align}
Such scale decoupling between the learning of the two components echoes some studies of neural network training that optimize the outer linear layer at each step while freezing the inner weights~\cite{marion2023leveraging}, and is conjectured to be more adaptive to the problem than learning jointly the two layers~\cite{bach2021}.

\paragraph{Correlation loss}
We end this subsection with the following convention: 
\begin{remark}
    From now on, we redefine the loss as $L(W) = \langle \mathsf{A}_G f , f \rangle_{\gamma_q}$, dropping the $^*$ superscript because there is no ambiguity. Note that now we seek to maximise this correlation function. All statements about this loss are immediately transferred (by switching the roles of maxima with minima) to the original squared loss. 
\end{remark}

\subsubsection{Preliminary remarks on the geometry of the loss}

Equation ~\eqref{eq:lossgrassm0} expresses the loss as a quadratic function associated with the operator $\Av_G$. Hence, from the representation in Eq.~\eqref{eq:representation_A_PSD_order}, we have the following expression of the loss through the matrix $G = MM^\top \in \C_q $.
\begin{corollary}
\label{cor:loss_psd}
Let $\alpha_\beta(V) = \langle f, H_\beta(V) \rangle$ be the decomposition of $f$, then 
\begin{align}
\label{eq:loss_psd_folded}
       L(W) =  \sum_{\beta} \alpha^2_\beta(V)  \lambda_\beta^2 .
\end{align}   
\end{corollary}
Remark that the equation is given with respect to the pair $(\lambda, V)$, which are functions of $W$ as they are respectively the singular values and left eigenvectors of $W_*^\top W$. Note that it is also possible to unfold the formula given by equation~\eqref{eq:loss_psd_folded}, as we have a closed form expression of $\alpha_\beta(V)$ given by Lemma~\ref{lem:hermitechange}. 

Let us emphasize that the loss has a specific form of a correlation. Indeed, we notice here that the action of the rotation $V$ and the subspace learning via $\Lambda$ are decoupled nicely. In fact, if we consider the two sequences $(\alpha_\beta(V), \lambda_\beta)_{\beta \in \N^q}$  of $\ell_2(\N^q)$, then, $L(W) =   \langle \alpha^2(V), \lambda^2  \rangle_{\ell_2(\N^q)}$.
It can be useful to stratify the space $\ell_2(\N^q)$ with respect to its harmonics of fixed degree  $\ell_2(\N^q) = \bigoplus_{s \geq 0} \ell_2^s(\N^q), $  where $\ell_2^s(\N^q) = \{ u \in \ell_2(\N^q), \text{ such that } u_\beta = 0 \text{ as soon as } \beta_1 + \dots + \beta_q \neq s \}$. Obviously $\ell_2^s(\N^q)$ is isomorphic to $\R^{c^q_s}$, where $c^q_{s} = \binom{s + q - 1}{s}$ is the number of $q$-tuples of $\N$ that sum to $s$. Finally we denote $\pi_s$ the orthogonal projection of $\ell_2(\N^q)$ onto $\ell_2^s(\N^q)$ and $\alpha^s(V) = \pi_s [\alpha(V)]$. If we also denote $\|\cdot \|_{\ell_1}$ the natural $\ell_1$-norm of multi-index sequences, we have
\begin{align}
       L(W) =  \langle \alpha^2(V), \lambda^2  \rangle_{\ell_2(\N^q)} = \sum_{s \geq 0} \langle \alpha^s(V)^2, \lambda^2  \rangle_{\ell_2(\N^q)},
\end{align}  
and where for all $s \geq 0$, $\|\alpha^s(V)^2 
 \|_{\ell_1} = \| \mathsf{S}^s f\|^2_{\gamma_q}$, does not depend on $V$.
Note that we also have that, structurally, $\lambda \in \mathcal{B}_\infty(0,1)$, the closed unit ball of $\ell_\infty(\N^q)$.

\paragraph{Conclusion on the loss geometry} 
The fact that we can express the loss as a dot product with $\ell_1$ constraints (on $\alpha(V)^2$) points towards the intuition that the loss landscape and optimization organize according to the faces of the simplex of the multi-index sequences. This is what we confirm theoretically in the following: Section~\ref{subsec:critical_points} depicts the critical points of the loss landscapes as the vertices of a polytope while Section~\ref{subsec:gradient_flow} shows how the Grassmannian gradient flow movement corresponds to a displacement in the vicinity of the edges of this polytope, quite reminiscent of the simplex algorithm in linear programming.

%% file: IV-Grassmann_fast/Critical_points.tex
\paragraph{Preliminaries} We begin with some preliminary remarks and calculations on the Grassmaniann gradient. As said before, we notice that the energy $L(W)$ can be expressed solely in terms of the left eigenvectors/singular values of $W_*^\top W = M(W) = V(W) \Lambda(W) U^\top(W)$, where we put emphasis on the fact that these quantities are in fact functions of $W$. As there is no ambiguity, we decide to drop the writing as an explicit function of $W$ and define
$$L(W)  = \langle \Av_{\Lambda^2}\P_V f, \P_V f \rangle_{\gamma_q} =: \ell(V, \Lambda). $$
In this section we  describe the critical points of the loss 
\begin{equation}
\label{eq:critical_points}
    \mathrm{Crit}(L) = \left\{W \in \mathcal{G}\,;\,  \nabla_W^{\mathcal{G}} L(W) = 0 \right\}~,
\end{equation}
where $\nabla_W^{\mathcal{G}} L(W)$ is the Grassmann gradient, that is~$\nabla_W^{\mathcal{G}} L(W) = (I - W W^\top) \nabla_W L(W)$. For some complementary results on the Grassmaniann geometry, see Section~\ref{secapp:grassmann_manifold} of the Appendix and~\cite{bendokat2020grassmann}. From the chain rule, we have the following characterization of the Grassmann gradient $\nabla_W L(W)$, with respect to derivatives of $\ell$.
\begin{lemma}[Chain Rule for Grassmann gradients]
\label{lem:chainrule_grassmann}
We have, for all $W \in \S(d,r)$
\begin{equation}
    \nabla_W^{\mathcal{G}} L(W) =  ( I - W W^\top) W_* V  \Xi U^\top~,
\end{equation}
%
where $\Xi := \overline{\Lambda} +  (F \circ \overline{V}^a) \Lambda $, with $\overline{V}^a = V^\top \overline{V} -\overline{V}^\top V$ and we have defined 
$\overline{\Lambda} := \mathrm{diag}( \partial_{\lambda_i} \ell(V, \Lambda) )$, $\overline{V} := \nabla_{V} \ell( V, \Lambda)$ and $F_{i,j} := (\lambda_j^2 - \lambda_i^2)^{-1}$ for $i,j$ such that $\lambda_i \neq \lambda_j$, and $F_{i,j}=0$ otherwise.
\end{lemma}
\begin{proof}
    The loss in also a function of $M$, that we call $\tilde{\ell}$. Hence, the chain rule gives $ \nabla_W^{\mathcal{G}} L(W) = ( I - W W^\top) W_* \nabla_M \tilde{\ell}(M)$, and given the chain rule for eigenelements (see \cite{townsend2016differentiating}), we have the result given in the lemma.
\end{proof}
We have the following characterization of critical points, establishing that the landscape has no bad local minima/maxima.

\paragraph{Description of the critical points.}

As it turns out, the critical points of $L(W)$ for $W \in \mathcal{G}(d,r)$ are characterized by critical points of a related, \emph{underparametrised} family of correlations. 
For $k \in \llbracket 0, q \rrbracket$, consider $W_k \in \mathcal{G}(q, k)$ and 
\begin{equation}
\label{eq:banana}
    L_k( W_k) := \| \mathsf{A}_{W_k} f \|^2_{\gamma_{k}} = \langle \Pi_{W_k} f, f\rangle_{\gamma_q}~,
\end{equation}
where we recall that we defined the projection $\Pi_{W_k} = \Av_{W_k W_k^\top}$. 
It is the same loss as $L$ but now defined over a different domain: $\mathcal{G}_q:=\{W_k W_k^\top; \text{ for } W_k \in  \mathcal{G}(q, k) \} \subset \mathcal{C}_q$, which is the set of orthogonal projectors onto subspaces $W_k$.
Hence, we should interpret Eq.\eqref{eq:banana} as a restriction of the original loss $L$. In words, this measures the energy that $f$ carries in the subspace $W_k$.  For $k \in \llbracket 1, q \rrbracket$, let 
\begin{equation}
    \Gamma_k(f) := \mathrm{Crit}( L_k) = \{ W_k \in \mathcal{G}(q,k)\,;\, \nabla^{\mathcal{G}} L_k(W_k) = 0\}~
\end{equation}
denote the critical points of the underparametrised objective Eq.\eqref{eq:banana} in the Grassmann manifold $\mathcal{G}(q,k)$. 
We also define the subset of $\Gamma_k(f)$ given by global minimisers, for $k \in \llbracket 0, q-1 \rrbracket$:
\begin{equation}
    \Gamma_k^{0}(f) := \{W_k \in \mathcal{G}(q,k) \,;\, L_k(W_k) = 0 \} ~.
\end{equation}
For $G \in \mathcal{C}_q$, given its eigendecomposition $G = V \Lambda V^\top$, $V=[v_1,\, \ldots,\, v_q] \in \O_q$, we consider the following subspaces:
 \begin{align}
\mathrm{Ess}(G) := \text{span}\{ v_i\, ;\, 0<\lambda_i<1 \} \in \mathcal{G}&( q, \tau')~, \qquad~\mathrm{Sp}(G) := \text{span}\{ v_i\, ;\, \lambda_i =1 \} \in \mathcal{G}(q, \tau)~, \nonumber \\
\mathrm{Jt}(G) &:= \mathrm{Ess}(G) \oplus \mathrm{Sp}(G)~,    
 \end{align}
 where we defined $\tau' :=\left|\{i\,;\, 0<\lambda_i<1 \}\right|$ and $\tau:=\left|\{i\,;\, \lambda_i = 1 \}\right|$.
For $W \in \mathcal{G}(d, r)$, we set as before $G_W = M M^\top  \in \mathcal{C}_q$, with $M = W_*^\top W $.

\begin{theorem}[Critical points of $L$]
\label{thm:critical_points}
The critical points of $L(W)$ in $\mathcal{G}(d,r)$ defined in \eqref{eq:critical_points} are given by 
\begin{equation}
    \mathrm{Crit}(L) = \left\{W \in \mathcal{G}(d,r)\,;\, \mathrm{Sp}(G_W) \in \Gamma_{\tau}(f) \text{ and } \|\mathsf{A}_{\mathrm{Sp}(G_W)}f\| = \|\mathsf{A}_{\mathrm{Jt}(G_W)}f\| \right\}~.
\end{equation}
Moreover:
\begin{enumerate}[label=(\roman*)]
    \item There is a unique local maxima. It is hence global and corresponds to $W = W_*$. 
    \item All local minima are also global, given by 
    $$\{W; \mathrm{Sp}(G_W) \in \Gamma_\tau^0(f) \text{ and } \|\mathsf{A}_{\mathrm{Sp}(G_W)}f\| = \| \mathsf{A}_{\mathrm{Jt}(G_W)}f\| \}~.$$
    \item Therefore, all the saddle points are given by 
    $$\bigcup_{\tau = 1}^{q-1} \left\{W ; \mathrm{Sp}(G_W) \in \Gamma_{\tau}(f)\setminus \Gamma_\tau^0(f) \text{ and } \|\mathsf{A}_{\mathrm{Sp}(G_W)}f\| = \|\mathsf{A}_{\mathrm{Jt}(G_W)}f\| \right\}~,$$
    where for each $ \tau \in \llbracket 1 , q-1 \rrbracket$, each set in the union has at least $\tau$ ascent directions and at least $ q - \tau$ descent ones.
\end{enumerate}
\end{theorem}
\begin{proof}
    We postpone the proof to Appendix~\ref{appsec:critical_points}.
\end{proof}
\begin{remark}[Degenerate condition]
    In all cases, the condition $\|\mathsf{A}_{\mathrm{Sp}(G_W)}f\| = \|\mathsf{A}_{\mathrm{Jt}(G_W)}f\|$ is  present.  This captures the situation where $\mathsf{A}_{\mathrm{Jt}(G_W)}f$ is degenerate, in the sense that its intrinsic dimension $\tau$ is strictly smaller than $\tau+\tau'$, and moreover all of its energy is captured by $\mathsf{A}_{\mathrm{Sp}(G_W)}f$; in other words, it is supported in $\mathrm{Sp}(G_W)$. Notice that if $f$ is \emph{generic}, i.e such that $\mathsf{A}_W f$ has intrinsic dimension equal to $q'$ for any $W \in \mathcal{G}(q, q')$, then we can remove this case and conclude the $\mathrm{Ess}(G) = 0 $ or equivalently $\tau' = 0$. Hence, for the interpretations below, we get rid of this term by assume this generic situation. 
\end{remark}

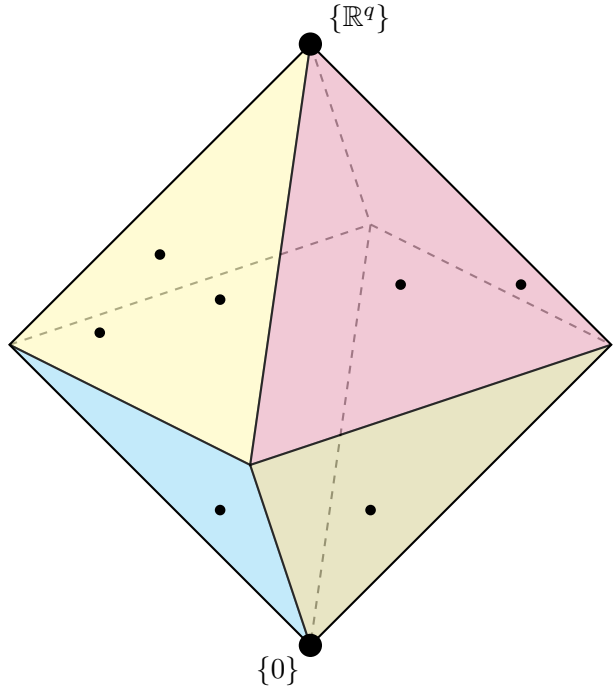
\begin{figure}
    \centering
\input{octahedron}
    \caption{Illustration of the geometry of the domain $\mathcal{C}_q$ of summary statistics underlying Theorem \ref{thm:critical_points}. Its boundary contains matrices $G$ such that either $0= \lambda_{\min}(G)$ or $1= \lambda_{\max}(G)$, but we illustrate here the only the boundaries of the form $\lambda_i(G) \in \{0,1\}$ relevant to the presence of critical points. Each facet should be thought as a Grassmann manifold, e.g. ${\color{cyan} \G(q,1)}$ represents the {\color{cyan} cyan} facet, ${\color{olive} \G(q,2)}$ represents the {\color{olive} olive} facet, etc. Two particular Grassmannians are represented differently: $\G(q,0) \simeq \{0\}$ and $\G(q,q) \simeq \{\R^q\}$ are naturally drawn as dots. and  The critical points of $L$ (illustrated for simplicity as dots), expressed in terms of their canonical summary statistics $G$, will be on the facets of this domain. Moreover, for generic targets (where the condition $\|\mathsf{A}_{\mathrm{Sp}(G_W)}f\| = \|\mathsf{A}_{\mathrm{Jt}(G_W)}f\|$ can be assumed to hold), the inclusion also goes the other way: any critical point of $L$ when restricted to the Grassmann manifold $G \in \G(q,\tau)$ will also be a critical point of $L(G)$ in $\mathcal{C}_q$. 
    }
    \label{fig:critical_points}
\end{figure}

This result can be interpreted as follows. In the positive semi-definite cone section $\mathcal{C}_q$, there are special `corners', given by $\Gamma_{\tau}(f)$ (and whose eigenvalues are either $0$ or $1$). These corners are stratified according to the dimension of the subspace (or equivalently $\text{rank}(G_W) = \sum_i \lambda_i$), which plays a role similar to the index of the critical point -- although in general these saddles won't be strict. In that sense, it is interesting to note that the underparametrised landscape (corresponding to $k < q$) is in fact controlling the saddle point structure of the overparametrised problem ($r \geq q$). 
Hence, in the $(\lambda, V)$ space, the eigenvalues $\lambda$ belong to the polytope $[0,1]^q$, and Theorem~\ref{thm:critical_points} states that the natural critical points are located \emph{at the vertices of this polytope}, that is $\lambda \in \{0,1\}^q$, while $V$ is to be a critical point of the energy in the corresponding Grassmann space. This is illustrated in Figure \ref{fig:critical_points}. 

In the next section, we show that, similarly to what happens in the simplex algorithm, the dynamics will consist of displacements through the edges of the $\lambda$-polytope, staying long times in the vincinity of critical points that we describe. 

%% file: octahedron.tex
\pgfmathsetmacro{\escala}{3}

\definecolor{cof}{RGB}{219,144,71}
\definecolor{pur}{RGB}{186,146,162}
\definecolor{greeo}{RGB}{91,173,69}
\definecolor{greet}{RGB}{52,111,72}

\begin{tikzpicture}[thick,scale=8]
\coordinate (A1) at (0,0);
\coordinate (A2) at (0.6,0.2);
\coordinate (A3) at (1,0);
\coordinate (A4) at (0.4,-0.2);
\coordinate (B1) at (0.5,0.5);
\coordinate (B2) at (0.5,-0.5);

\begin{scope}[thick,dashed,,opacity=0.8]
\draw (A1) -- (A2) -- (A3);
\draw (B1) -- (A2) -- (B2);
\end{scope}
\draw[fill=yellow!35,opacity=0.6] (A1) -- (A4) -- (B1);
\draw[fill=cyan!35,opacity=0.6] (A1) -- (A4) -- (B2);
\draw[fill=purple!35,opacity=0.6] (A3) -- (A4) -- (B1);
\draw[fill=olive!35,opacity=0.6] (A3) -- (A4) -- (B2);
\draw (B1) -- (A1) -- (B2) -- (A3) --cycle;

\filldraw[black] (0.25,0.15) circle (0.2pt) {};
\filldraw[black] (0.15,0.02) circle (0.2pt) {};
\filldraw[black] (0.35,0.075) circle (0.2pt) {};
\filldraw[black] (0.35,-0.275) circle (0.2pt) {};

\filldraw[black] (0.6,-0.275) circle (0.2pt) {};

\filldraw[black] (0.85,0.1) circle (0.2pt) {};
\filldraw[black] (0.65,0.1) circle (0.2pt) {};

\filldraw[black] (B2) circle (0.5pt) node[anchor=north east]{$\ \{0\}$};

\filldraw[black] (B1) circle (0.5pt) node[anchor=south west]{$\ \{\R^q\}$};

\end{tikzpicture}

%% file: IV-Grassmann_fast/Gradient_flow.tex
\subsubsection{Gradient computations and Bures-Wasserstein gradient flow on \texorpdfstring{$G$}{G}}

Before entering into the details of the dynamics, let us state an important structural property of the learning.
Let us introduce some intermediate calculations and results. Let us denote $g : W \to G_W = M  M^\top= W_*^\top W W^\top W_* $, and define $\ell : G \to \ell(G)$ such that $ L(W) = \ell (g(W))$. Then we have an explicit calculation of the gradient
\begin{proposition}[Correlation Gradients]
\label{prop:gradient_loss_G}
    For $f \in H^1_{\gamma_q}$, the gradient of the loss with respect to $G$ writes 
    \begin{align}
     \nabla_G \ell (G) &= (\mathsf{A}_M \nabla f) \otimes (\mathsf{A}_M \nabla f)  
     =  \nabla f \otimes (\mathsf{A}_G \nabla f)
     = \E_z \left[ \nabla_z f(z) [\Av_G \nabla_z f]^\top(z) \right].   
    \end{align}
    In particular, $\nabla \ell_G (G) \succcurlyeq 0 $  
    is positive semi definite.
\end{proposition}
From the gradient flow on $W$, following for all $t \geq 0$, $\dot{W}_t = \nabla_W^\G L(W)$, from basic chain rules, we can derive the dynamics on  $(G_t)_{t \geq }0$ with respect to $\overline{G}:=\nabla_G \ell $. We have that 
\begin{lemma}
\label{lem:dotGlemma}
    For all $t \geq 0$, G follows the dynamics
    \begin{align}
    \label{eq:dotGeq}
        \dot{G} &= 2 \left( G  \overline{G}  + \overline{G} G - 2 G  \overline{G} G  \right)~,
    \end{align}
    where the expression of $\overline{G}=\nabla_G \ell$ is given in Proposition~\ref{prop:gradient_loss_G}. 
\end{lemma}

We prove the previous proposition and lemma in Section~\ref{subsecapp:gradient_flow_G} of the Appendix. We also provide more interpretation of this calculation in the same section of the Appendix (see e.g. Eq.\eqref{eq:random_G}). 

This Ordinary Differential Equation (ODE) on $G$ has two distinct terms: one is $G  \overline{G}  + \overline{G} G$ and the other one is $ -2 G  \overline{G} G$. The second one corresponds to a ``boundary term" to prevent $G$ from growing outside of $\C_q$. The first term is the one that is responsible for the growth of $G$. Remarkably, it corresponds to a gradient term for the \emph{Bures-Wassertein} metric on PSD matrices: indeed $\nabla_G^{BW} \ell = G \nabla_G \ell + \nabla_G \ell G$, with  the natural notation that $\nabla_G^{BW} \ell$ corresponds to the gradient field induced by the Bures-Wassertein metric.

\subsubsection{Spectral Summary statistics}

The gradient flow on $W$ or the ODE on $G$ translate to a flow on the summary statistics $(\Lambda, V)$. We will use $\lambda$ as a notation for the vector corresponding to the diagonal of $\Lambda$. Their flow is  a simple consequence of the chain rule, and is presented in the following lemma.
\begin{lemma}
\label{lem:sumstatdyn}
    The summary statistics $\lambda, V$ follows the dynamics:
    \begin{align}
    \label{eq:movement_lambda_V}
        \dot{\lambda}_t &= (1 - \lambda_t^2) \odot \nabla_\lambda \ell(\lambda_t, V_t) \\
        \dot{V}_t &= V_t \left[C_{\lambda_t} \circ \overline{V}^a_t \right]
    \end{align}
    where we defined $[C_{\lambda}]_{ij} = \frac{\lambda_i \lambda_j ( 2 - (\lambda_i^2 + \lambda_j^2))}{(\lambda_i^2 - \lambda_j^2)^2}$ and we recall that $\overline{V}^a = V^\top \nabla_V \ell - \nabla_V \ell^\top V $.
\end{lemma}
\begin{proof}
    The proof of this fact follows from the chain rule and is given in Appendix~\ref{subsecapp:gradient_flow_lambda_V}.
\end{proof}
There is an obvious but important consequence of the equation governing the dynamics of $(\lambda_t)_{t \geq 0}$:
\begin{corollary}
\label{coro:lambda_increase}
 $t\mapsto \lambda_t$ is a coordinate-wise non-decreasing function.     
\end{corollary}

\begin{remark}
\label{rem:time_scales}
    We will see that, thanks to Lemma~\ref{lem:eigenvalues_init}, we have $\lambda \sim d^{-1/2}$, hence, anticipating  the definition of the information exponent, $s$, presented next in Eq.~\eqref{eq:information_expo}, we have $\ell \sim d^{-s/2}$, thus $\nabla_\lambda \ell \sim d^{-s +1/2}$ and $\nabla_V \ell \sim d^{-s}$,  where $s$ is the index of the smallest non-zero pure Hermite component in the decomposition of $f$. As finally $C_\lambda \sim d$, we have the estimates $\dot{\lambda}  = \Theta(d^{-s + 1/2}),~ \text{and} ~ \dot{V}  = \Theta(d^{-s + 1})$.
    Hence, in the limit of large $d$, $V$ is infinitely faster than $\lambda$ and a slow fast-dynamics occurs~\cite{berglund2006noise}.
\end{remark}
As it turns out, the spectral parametrization $(\lambda, V)$ is the best suited to describe the dynamics. This is what we do in the following section. 
\subsubsection{Initialization}
As we study the system of ODEs on the summary statistics given by Lemma~\ref{lem:sumstatdyn}, an important aspect of the system is the initialization. In fact, quite similarly to the single-index case~\cite{arous2021online}, the correlation between the subspace $W_0 \sim \mathrm{Unif}(\G(d, q))$ and the target $W_*$ is small with high probability. More precisely, the singular values of $M = W_*^\top W$, also called the \textit{principal angles} between the subspaces $W$ and $W_*$, follow the following multivariate distribution~\cite[Equation (1)]{absil2006largest}. A simple square root change of variable gives the distribution of the $\lambda$'s:
\begin{align}
\label{eq:multivariate_law_eigenvalues}
    p(\lambda_1, \cdots, \lambda_q) = Z_q^{-1} \prod_{i < j} |\lambda_i^2 - \lambda_j^2| \prod_{ i = 1}^q (1 - \lambda_i^2)^{( d- 2q - 1)/2} \mathds{1}_{ 0 \leq \lambda_q \leq \dots \leq \lambda_1 \leq 1}~, 
\end{align}
with normalization constant $Z_q = \frac{\Gamma_q^2(q/2)}{\pi^{q^2 / 2}} \frac{\Gamma_q((d-q)/2)}{\Gamma_q(d/2)}$.
\begin{lemma}[Concentration of Eigenvalues at Initialization]
\label{lem:eigenvalues_init}
For any $a, b > 0$ we have
\begin{align}
\mathbb{P}\left( \frac{b}{ \sqrt{d}} \leq \lambda_q \leq \lambda_1 \leq \frac{a}{\sqrt{d}}\right) \geq 1 - n_q b - m_q e^{-a^2/4}~,
\end{align}
where we define $m_q = \frac{\pi^q e^{2q}}{2^{q}\Gamma^2(q/2)}$ and $n_q =  \frac{2\pi^{2q}  e^{3q} \Gamma(2q) }{ \Gamma^4((q-1)/2)}$.
\end{lemma}
\begin{proof}
    We defer the proof of this lemma to subsection~\ref{subsubsecapp:proof_initialization} of the Appendix.
\end{proof}
Let us comment this result. It says that at initialization all the principal angles are to be found in a band of $1/\sqrt{d}$. As $q$ is to be thought of as $\Theta_d(1)$, $n_q, m_q$ are also $\Theta(1)$. When the dimension is large, it is expected that the distribution of $(\sqrt{d} \lambda_1, \cdots, \sqrt{d} \lambda_q)$ becomes the tensorized distribution of a $q$-dimensional i.i.d. Gaussian vector\footnote{While we did not find this result written as such, we believe it is folklore in the literature on high-dimensional probability over compact matrix groups~\cite{meckes2019random}}, hence it should not be surprising that the bound on the upper spectrum shows a Gaussian tail and the one on the lower bound is linear, as expected by \textit{anti-concentration}.

A remarkable feature of this is that if the function $f$ carries no linear term (Hermite of order $1$), then the initialization of the gradient flow will occur in a saddle point. We anticipate this fact in the next section.
\subsection{Description of the dynamics}
\label{sec:escapemediocrity}

We first introduce the notions that enable us to describe the dynamics of the gradient flow. We then state the main result on how the target function is \textit{sequentially} learnt.

\subsubsection{Information exponent and decomposition relative to a subspace}

In the seminal paper of Ben Arous et al., a crucial quantity was the information exponent of the function $f$. This was a key analytic property of the target function driving the sample and time complexity of SGD \cite{arous2022high,arous2021online} in the single-index setting, and given by the number of vanishing moments. A natural multi-index extension of this quantity would be in our case
\begin{equation}
\label{eq:information_expo}
    s(f) := \inf\left\{ |\beta|=\sum_i \beta_i; \langle f, H_\beta\rangle \neq 0 \right\}~.
\end{equation}
As explained in the aforementioned papers, it also corresponds (as in this one dimensional case) to the degeneracy of the origin as a saddle point of the loss. We see in this section that we will need more information than this information exponent. As in fact, the dynamics will travel in the vicinity of saddle points, we need to introduce information exponent \textit{with respect to each of these saddle points}. As described in the previous section by Theorem~\ref{thm:critical_points}, the saddle point are described by Grassmanianns, hence, we need a description of the degeneracy  of the saddle respective to the Grassmanianns.

Indeed, given $f \in L^2_{\gamma_q}$ and $W \in \mathcal{G}(q, q')$, we define its \emph{relative information exponent} $s(f; W)$ as follows: Let $U \in \mathcal{O}_q$ be of the form $U = [W; W^\perp]$ ; and consider $\{H_\beta(U)\}_\beta$ the tensorised Hermite basis associated with $U$. Let $I=\llbracket q'+1,q \rrbracket$ and recall the notation  $|\beta|_\Omega = \sum_{i \in \Omega} \beta_i$. 
Then
\begin{equation}
\label{eq:relative_info}
    s(f; W) := \inf\{ |\beta|_I; \langle f, H_\beta(U) \rangle \neq 0\},
\end{equation}
and we argue the following result:
\begin{claim}
\label{claim:relative_info}
    The relative information exponent is well-defined; i.e. it does not depend on the choice of basis for $W$ and $W^\perp$. We have $s(f, \emptyset) = s(f)$, $s(f, I_q) = 0$. Moreover, if $f$ has intrinsic dimension $q$ and $\text{dim}(W)<q$, then $s(f, W)>0$. 
\end{claim}
\begin{proof}
    The fact that this exponent is well defined comes from~Lemma~\ref{claim:hermite_wellposed}. Moreover, if $\text{dim}(W)<q$, then $\Pi_{W^\perp} f \neq 0 $ and hence, $s(\Pi_{W^\perp} f) \geq s(f, W) > 0$. 
\end{proof}

Given $f \in L^2_{\gamma_q}$ and $W \in \mathcal{G}(q,q')$, its associated relative information exponent $s(f;W)$ defines a `canonical' spectral filtering, given by $\mathsf{S}_W^{s(f;W)}f$, that we denote simply by $\mathsf{S}_W f$ when the context is clear. From the previous claim, the relative smoothing operator is well-defined (it does not depend on the choice of Hermite basis). 

We now specialize this operator in the context of the incremental learning. Consider the isotropic 
filtering $f_1 := \mathsf{S}_\emptyset f$ by the information exponent, and let $W_1 \in \mathcal{G}(q, \mathsf{d}(f_1))$ be its support. Let $f_2 := \mathsf{S}_{W_1} f$. We want to compare the intrinsic dimensions and supports of $f_1$ and $f_2$. 

\begin{claim}
\label{claim:relative_smoothing}
We have $\mathsf{d}(f_1) \leq \mathsf{d}(f_2)$, with supports satisfying $W_1 \subseteq W_2$. Moreover, if $s(f;W_1)>0$, then the inequalities are strict. 
\end{claim}
\begin{proof}
    Observe that $f_1 = \mathsf{S}^{s(f)} f =  \mathsf{S}^{s(f)} f_2$. From Claim \ref{prop:intrinsic_spectral} we thus have that $\mathsf{d}(f_1) \leq \mathsf{d}(f_2)$ and $W_1 \subseteq W_2$. Finally, if $s(f;W)>0$, then by definition we have $\|f_2 \| > \| \mathsf{A}_{W_1} f_2 \|$, showing that $\mathsf{d}(f_1) < \mathsf{d}(f_2)$.
\end{proof}

\subsubsection{Description of the incremental subspace learning}
\label{sec:incremental_learning}

\paragraph{A fine-grained description} Let $W_0 = \emptyset$ and set $p_0=0$. Define $s_1=s(f, W_0)=s(f)>0$, and let $p_1>0$ be the intrinsic dimension of $\mathsf{S}_{W_0}$, whose support is denoted $W_1$. 
If $p_1 < q$, then from Claim \ref{claim:relative_info} we have that $s_2=s(f; W_1)>0$, and thus from Claim \ref{claim:relative_smoothing} we have that $\mathsf{S}_{W_1}$ has intrinsic dimension $p_2 > p_1$. 
By iterating this construction, we will eventually reach a subspace $W_K$ of dimension $q$. 
We have thus found a strictly increasing sequence $0=p_0<p_1<\ldots <p_K = q$ with associated information exponents $s_k = s(f; W_{k-1})$. We have the following definition of \emph{leaps}.
\begin{definition}[Leaps]
\label{def:leap}
    Let $f \in L^2_{\gamma_q}$ of intrinsic dimension $q$, and consider the previously built flag $W_0 = \emptyset \subset W_1 \subset \ldots \subset W_K = I_q$. Its \emph{leap dimensions} are $l_k := p_k - p_{k-1}$ for $k=1\ldots K$, its \emph{leap exponents} are $s_k := s(f; W_{k-1})$
    Finally, we define $s^* = \max_k s_k $ as the \emph{largest} leap exponent.  
\end{definition}

\begin{remark}[Leap complexity]
\label{rem:isoleap}
Our definition of the largest leap exponent corresponds precisely to the rotationally-invariant version of the leap complexity, the so-called $\mathrm{isoLeap}$ \cite[Appendix B.2]{abbe2023sgd}, presented in the Gaussian formulation of the cited paper. Notably, the exponent is intrinsic to the target $f$, ie it does not depend on the choice of orthonormal basis of the Hermite tensor decomposition. 
\end{remark}

\begin{remark}[Kernel Regularisation]
    Observe that since $f^\mu$ (defined in Proposition \ref{prop:kernelopt}) is an isotropic spectral shrinkage of $f^*$, it has the same cascade decomposition as $f$, with same information exponents and subspace supports; therefore, we can without loss of generality consider $f$ and transfer the conclusions (with different numerical constants). 
\end{remark}

 Given a target function $f$, we set $f_0 \equiv 0$, and
 have defined recursively a sequence of functions $f_k:= \mathsf{S}_{W_{k-1}} (f)$, where $W_{k-1} \in \mathcal{G}(q, p_{k-1})$ is the support of $f_{k-1}$, with intrinsic dimension $p_{k-1}$ that increases at each step, from $p_0=0 < p_1 < \dots < p_L = q$.
 \begin{align*}
\begin{array}{ll||ccccccc}
\text{Functions}: \hspace{1cm} & f_{k+1} = \mathsf{S}_{W_{k}}(f) & 0 = f_0 & \rightarrow & f_1 & \rightarrow & \dots & \rightarrow &  f_K = f   \\
\text{Inf. exponent}: \hspace{1cm} & s_{k+1} = s(f, W_{k})  & 0 = s_0 & \rightarrow & s_1 & \rightarrow & \dots & \rightarrow &  s_K  \\
\text{Supports}: \hspace{1cm} & W_{k+1} = \mathrm{supp}(f_{k+1})  & 0 = W_0 & \subset & W_1 & \subset & \dots & \subset &  W_K = W^* \\
\text{Dimensions}: \hspace{1cm} & p_{k+1} = \mathsf{d}(f_{k+1})  & 0 = p_0 & \leq  & p_{1} & \leq  & \dots & \leq  &  p_K = q  \\
\text{Grassmannians}: \hspace{1cm} & \mathcal{G}_{k+1} = \mathcal{G}(p_{k+1}, q)  & 0 = \mathcal{G}_0 & \rightarrow  & \mathcal{G}_1 & \rightarrow  & \dots & \rightarrow  &  \mathcal{G}_K = I_q 
\end{array}
\end{align*}

Let us show a concrete example of such a sequence.
\begin{example}[Fine-grained sequence]
\label{ex:sequential_cascade_polynomial}
    Let $(v_1, v_2, v_3, v_4)$ an orthonormal basis of $\R^4$. Define, 
    \begin{align}
    \label{eq:polynomial_sequence}
        f(x) = h_2(v_1 \cdot x) + h_4(v_2 \cdot x) + h_6(v_1 \cdot x) h_1(v_3 \cdot x) + h_3(v_1 \cdot x) h_5(v_3 \cdot x) h_3(v_4 \cdot x),    
    \end{align}

    then, the fine-grained sequence of polynomials learnt is

\begin{align*}
\begin{array}{l||ccccccc}
\ k  & k = 1 & \vert &   k = 2 & \vert &    k = 3 & \vert &  k = 4 = K  \\
\hline
f_k & h_2(v_1 \cdot x) & \boldsymbol{+} &  h_6(v_1 \cdot x) h_1(v_3 \cdot x) & \boldsymbol{+} &   h_3(v_1 \cdot x) h_5(v_3 \cdot x) h_3(v_4 \cdot x) & \boldsymbol{+} & h_4(v_2 \cdot x) = f  \\
s_k & s_1 = 2 & \rightarrow &  s_2 = 1 & \rightarrow &  s_3 = 3 & \rightarrow &  s_4 = 4  \\
W_k & \mathrm{span}(v_1) & \subset & \mathrm{span}(v_1, v_3) & \subset &  \mathrm{span}(v_1, v_3, v_4) & \subset &   \R^4 \\
\end{array}
\end{align*}

\end{example}

\vspace{0.25cm}

This incremental structure of $f$ suggests a `saddle-to-saddle' dynamics in $K$ phases. During each phase, the intuition is that a block of $l_k$ eigenvalues escape from the vicinity of $0$ at a time that, we will see, scales like $\tau_k \simeq d^{s_k-1}$ (assuming $s_k>1$). At the end of  phase $k$, we have identified the support $W_{k}$. Finally, the overall time to learn $f$ is of order $\tau^* \simeq d^{s^*-1}$.

\paragraph{Regrouping the cascades.} 
In the above fine-grained description of the dynamics, if two successive relative information exponents are in non-increasing order, e.g.~$s_k \geq s_{k+1}$, then phases~$k$ and~$k+1$ are indistinguishable because with the previous proposition $\tau_k \gtrsim \tau_{k+1}$. More concretely, in Example~\ref{ex:sequential_cascade_polynomial}, we expect that the learning of the first component $f_1 = h_2(v_1 \cdot x)$ provokes at its time-scale~$d$, a instantaneous \textit{cascade} of learning $f_2$, because $s_1 >s_2$. Moreover the \textit{true} time-order of learning $f_1, f_2$ may be ill-defined (or simply of combinatorial difficulty to describe). Hence, we circumvent this by \textit{regrouping} the subspace learning by blocks. To do this, denoting $c_k$ the number of induced cascades at index $k \geq 1$, we re-index the sequence starting with $t_0 = 1$ and for all $k =  1 \dots \tilde{K}$, $t_k := c_{t_{k-1}} + t_{k-1}$ where $\tilde{K}$ is such that $t_{\tilde{K}} = K $. Thanks to this we define  for $k \in 1 \dots \tilde{K}$,
\begin{align}
    \{\tilde{f}_k, \tilde{W}_k, \tilde{l}_k, \tilde{s}_k, \tilde{p}_k, b_k\}= \{f_{t_k}, W_{t_k}, l_{t_k}, s_{t_{k}}, p_{t_k}, c_{t_k}\},
\end{align}
where we added the sequence $(b_k)$ for the number of cascades occurring at time $k$. This results naturally in an incremental structure where now $\tilde{s}_{k+1} > \tilde{s}_k$. Note that $\max_k s_k = \max_{{k}} \tilde{s}_k$ and thus $s^*$ is not affected by this regrouping. Taking again Example~\ref{ex:sequential_cascade_polynomial}, with this merging, the sequentially learnt sequence is expected to be:
\begin{example}[Coarse-grained sequence]
\label{ex:true_sequential_cascade_polynomial}
    Take the  same  polynomial as the one defined in Eq.\eqref{eq:polynomial_sequence}, then we have the merged sequence
 \begin{align*}
\begin{array}{l||ccccc}
\ k  & k = 1  & \vert & k = 2  & \vert &  k = 3 = \tilde{K} \\
\hline
\tilde{f}_k & h_2(v_1 \cdot x) +  h_6(v_1 \cdot x) h_1(v_3 \cdot x) & \boldsymbol{+} &  h_3(v_1 \cdot x) h_5(v_3 \cdot x) h_3(v_4 \cdot x) & \boldsymbol{+} & h_4(v_2 \cdot x) = f  \\
\tilde{s}_k & \tilde{s}_1 = 2  & < &  \tilde{s}_2 = 3  & < &  \tilde{s}_3 = 4  \\
\tilde{W}_k &  \mathrm{span}(v_1, v_3) & \subset &   \mathrm{span}(v_1, v_3, v_4) & \subset & \R^4 \\
b_k &  b_1 = 2 \mathrm{\ cascades\ } &  &   b_2 = 1 &  &   b_3 = 1.
\end{array}
\end{align*}
\end{example}
 \begin{remark}[Subspaces and \emph{Staircase} property]
 \label{rem:staircase}
    The nested sequence of subspaces $\{W_k\}_k$ generalises the sequential learning structures 
   described in prior works that study certain classes of multi-index models.  
    Specifically, with a different algorithm, \cite[Appendix D.2]{abbe2023sgd} studies the learning of \emph{leap} functions (an extension of \emph{staircase functions} from \cite{abbe2022merged}), whose Hermite decompositions obey certain lacunary structure across successive harmonics. Precisely, there exists an orthonormal basis $V$ of $\R^q$ such that, for each $k$, $\mathsf{\Pi}^k f = a_k H_{\beta_{(k)}}(V)$ consists of a pure harmonic. In this setting, the associated subspaces are given by $W_k=\text{span}(v_i; \beta_{(k)}[i] >0 )$. Similarly, \cite[Definition 4 and Theorem 3]{dandi2023learning} describes, through their so-called `subspace conditioning', a setting where the associated leap exponent is $1$, where the saddle-to-saddle structure simplifies into a single step $\tilde{W}_1=I_q$.   
\end{remark}
\begin{example}[An example with basis recombination]
We use here an example showing that the rotation invariance of the Gaussian can lead to tricky examples of the learnt incremental sequence. Indeed, take the polynomial of $\R^2$, $f(x_1,x_2) = \frac{1}{\sqrt{2}} \left[h_1(x_1) + h_1(x_2)\right] + \frac{1}{2} \left[ h_2(x_1) + h_2(x_2) - 2 h_1(x_1) h_1(x_2) \right] $. \textit{In appearance}, it seems that the learned sequence will be fast, with $s^* = 1$, thanks to the presence of the degree one monomials  $h_1(x_1)$ and $h_1(x_2)$ spanning the two directions of $\R^2$. However, defining the basis $(v_1, v_2)$ as the rotation of the canonical basis with angle $\pi/4$, it is possible to rewrite $f(x_1,x_2) = h_1(v_1 \cdot x) + h_2(v_2 \cdot x) $. This recombination corresponds to the incremental learning that we describe and provides the correct time scale for the learning given by $s_1 = 1$ for the first direction $v_1$ and $s_2 = 2 = s^* $ for the second direction $v_2$.
\end{example}
\begin{example}[\emph{Teacher} with $q$ neurons]
Consider $\theta_1, \ldots, \theta_q \in \mathcal{S}_{d-1}$ in general position, and an activation function $\sigma:\R \to \R$. The function $F= \sum_{j=1}^q \alpha_j \mathsf{P}_{\theta_j} \sigma \in L^2_{\gamma_d}$ generated by these $q$ `teacher' neurons is a multi-index model of the form $F(x) = f(W_*^\top x)$, 
where $W_*= \text{span}(\theta_j; j=1\ldots q)$ and $f=\sum_{j=1}^q \alpha_j \mathsf{P}_{z_j} \sigma$, with $z_j = W_*^\dagger \theta_j \in \R^q$. 
In this case, the cascade structure is simplified: Consider the Hermite expansion $\sigma = \sum_i \alpha'_i h_i = \alpha_{s}' h_{s} + \alpha_{\tilde{s}}' h_{\tilde{s}} + \sum_{i > \tilde{s}} \alpha'_i h_i $ of the activation function. If the information exponent of $\sigma$ is $s(\sigma)=s=1$, then $W_1 = \text{span}(\sum_j \alpha_j \theta_j)$, the linear subspace carrying the `average' of the neurons, will be learnt first with $s_1 = 1$; followed by $W_2 = W_*$, with associated relative information exponent $s_2=\tilde{s}-s$, the second non-zero term in the Hermite expansion of $\sigma$. If $s(\sigma)>1$, the cascade `collapses' to $W_1=W_*$, with $s_1 = s(\sigma)$. Notice that our analysis does not address the internal learning dynamics within the subspace, i.e. the order in which the `neurons' $\theta_j$ will be identified. 
\end{example}

We are now in position to state the main result of this section. Recall that to each $W \in \G(d, r)$, we associate $G_W = W_*^\top W W^\top W_* \in \C_q$. Finally let define $\Cu$ a \textit{dimensionless} constant depending solely on the function $f \in L^2_{\gamma_q}$ and numerical constants.
\begin{theorem}[Incremental Learning, Main Result]
\label{thm:coarse-grained} Assume $f \in H^{s^*+q}(\gamma_q)$. The Grassmaniann gradient flow on the Loss~\eqref{eq:lossgrassm0} written in Eq.\eqref{eq:gradient_flow_on_W} initialized in $W_0 \sim \mathrm{Unif}(\G(d,r))$, is such that, with probability greater than $1-\delta$ over $W_0$, 
\begin{enumerate}[label=(\roman*)]
    \item $(G_{W_t})_{t \geq 0}$ passes in a neighborhood of size at most $\Cu d^{-1/2}$ of the $\tilde{K} - 1$ strict saddle-points corresponding to the subspaces $\tilde{W}_k$, for $k \leq \tilde{K} - 1$. Each saddle is escaped within a time-scale $\tau_k \leq \mathcal{C}_{f,\delta,s} (d^{\tilde{s}_k - 1}\vee \log(d))$ for $k \leq \tilde{K} - 1$.
    \item We have $\displaystyle \lim_{t \to \infty} W_t = W_*$, and, for any $\eta > 0$, after time $t \geq \mathcal{C}_{f,\delta,s^*} [(d^{s^* - 1} \vee \log(d)) + \log(1/\eta)]$, 
    \begin{align}
    \label{eq:thm_quantitative_convergence}
        \|W_t W_t^\top - W_* W_*^\top\|^2 \leq 4 \eta~,
    \end{align}
\end{enumerate}
where we have defined $\mathcal{C}_{f,\delta,s} = \delta^{-2(s-1)} \Cu$ for $s>1$ and $\mathcal{C}_{f,\delta,1} = \Cu + \log(1/\delta)$.    
\end{theorem}
\begin{figure}[ht]
    \centering
    \input{octahedron_saddles}
    \caption{Cartoon illustration of the `saddle-to-saddle' optimization dynamics in the octahedron representation of the critical points for a case where $\tilde{K}=4$. The trajectory of the dynamics \textit{selects} some of the critical points as precised by Theorem~\ref{thm:coarse-grained}. These are represented marked as the subspaces $\tilde{W}_k$, for $k \in \llbracket 0,4 \rrbracket$.}
    \label{fig:saddles}
\end{figure}
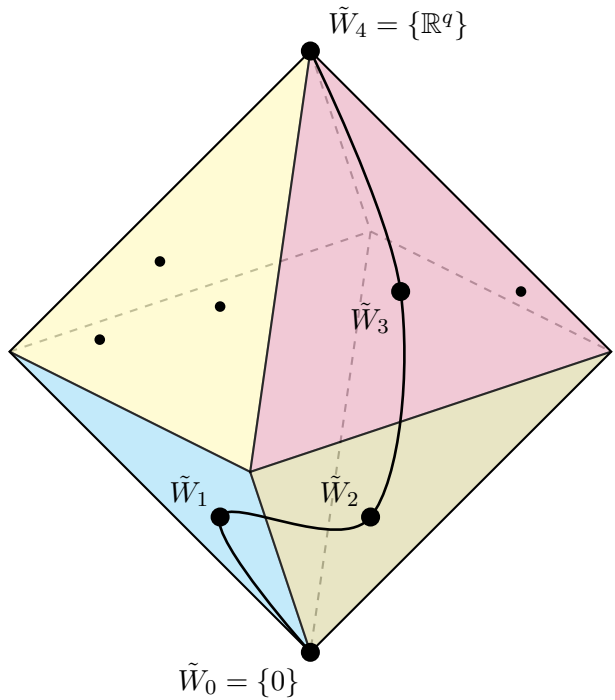
This theorem provides a description of the dynamics and of the time-scales at which the important events (saddle-point escapes) happen. In fact, due the high dimension, the subspaces $W$ and $W_*$ are nearly uncorrelated at initialization. Hence, the initialization happens near the origin, which is a critical point of the loss. From a dynamical system perspective, it means that all the time spent for the dynamics will occur near critical points~\cite{bakhtin2011noisy}, and as the dynamical system at hand is a gradient flow it eventually converges to a local maximum of the correlation. Theorem~\ref{thm:critical_points} provided the structures \textit{of all the critical points}. The purpose of this  result, and of this section, is to indicate clearly what are the critical points \textit{visited by the gradient flow} and what are the time-scales at which the dynamics escapes these sticky points. 

These saddle points have a clear geometric interpretation arising from the harmonic analysis of $f$: each subspace $\tilde{W}_k$ carries the lowest available frequencies, in the orthogonal complement of the already `captured' energy. This picture is thus consistent with several empirical findings and heuristics in neural networks, e.g. \cite{rahaman2019spectral,cao2019towards} that identified an incremental learning structure, whereby `simplest' features of the target are learnt first. Finally, in terms of gradient dynamics, the saddle-to-saddle dynamics described by Theorem \ref{thm:coarse-grained} paint a similar picture than previous works on different non-convex learning models, such as \cite{jacot2021saddle, pesme2023saddle}, and, closer to our setting, the dynamics of \cite{abbe2023sgd}. 

\begin{remark}[Smoothness assumption]
    The Sobolev assumption $f \in H^{s^*+q}(\gamma_q)$ is a sufficient technical condition used to establish an upper bound on the correlation growth (Lemma \ref{lem:correl_upperbound}), and it may be weakened by improving the analysis. We remark that this lemma is used to establish that learning cannot happen ``too quickly". On the other hand, the upper bound on $\tau_{\tilde{K}}$, dictating the overall time complexity, does not require such assumption (only $f \in H^1$ to ensure that gradients are well-defined). 
\end{remark}

\begin{figure}
\includegraphics[width=0.7\textwidth]{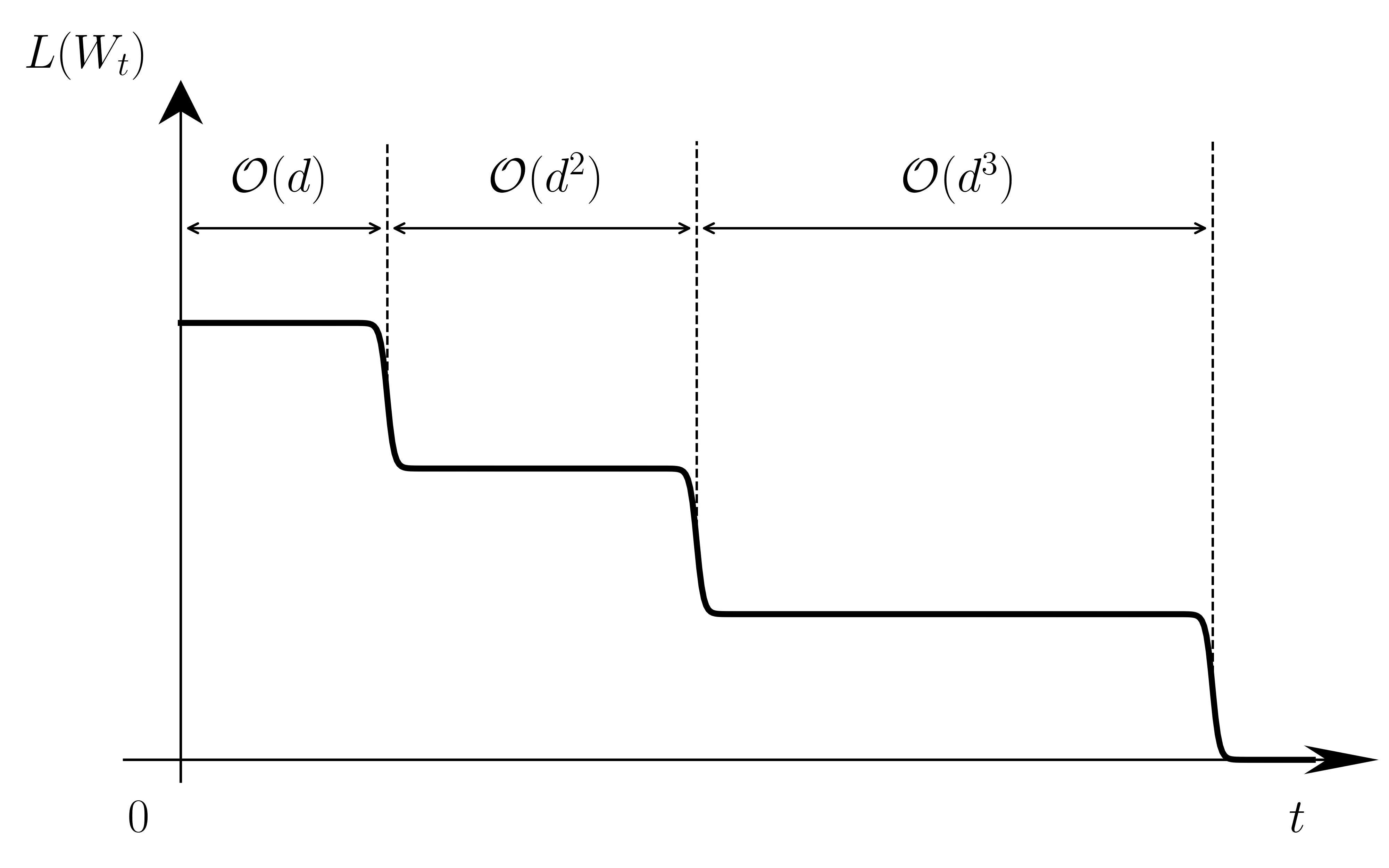}
\caption{Cartoon illustration of the evolution of the population loss during the dynamics. The plot represents the loss evolution according to the problem depicted in Example~\ref{ex:true_sequential_cascade_polynomial}. In this example, after the initialization point, which is a first saddle associated with the timescale $\mathcal{O}(d)$, the loss dynamics has two other plateaus of time scales respectively $\mathcal{O}(d^2)$ and $\mathcal{O}(d^3)$, before eventually converging.}
\label{fig:loss_saddles}
\end{figure}

\subsection{Proof of Theorem \ref{thm:coarse-grained}}

Let us assume once and for all that the eigenvalues are ordered non-increasingly for all $t \geq 0$, $\lambda_1 \geq \dots \geq \lambda_q$. 
From the initialisation $W_0 \sim \mathrm{Unif}(\G(d,r))$, and given Lemma~\ref{lem:eigenvalues_init}, we have that all eigenvalues $\lambda_i(0)$ are in the interval $\left(b d^{-1/2},a d^{-1/2} \right)$ with probability $1 - m_q b - n_q \exp(-a^2/4)$.
We assume from now on that we are in this likely event. 

Recall from Section \ref{sec:incremental_learning} that we introduced a fine-grained cascade structure $\{ f_k, W_k, l_k, p_k, s_k\}_{k \leq K}$ and a regrouped cascade structure $\{ \tilde{f}_k, \tilde{W}_k, \tilde{l}_k, \tilde{s}_k, \tilde{p}_k\}_{k \leq \tilde{K}}$ with increasing exponents $\tilde{s}_{k+1} > \tilde{s}_k$. 
For technical reasons, it is necessary to add an additional element in the definition of the cascades. For each $k=2,\ldots K$, we identify low-order harmonics in $f_k$ that give raise to the same information exponent and support. For that purpose, from Claim \ref{prop:intrinsic_spectral} there exists $\tilde{r}_k \in \mathbb{N}$ such that 
$\mathsf{d}(\mathsf{S}^{r} f_k) = \mathsf{d}(f_k)$ and $W[\mathsf{S}^{r} f_k]=W[f_k]$ for $r\geq \tilde{r}_k$. We define $r_k=\max(2 s_k, \tilde{r}_k)$ as the \emph{acquired} exponent of $f_k$ relative to $W_{k-1}$. Observe that we can express $\mathsf{S}^{r_k} f_k$ in coordinates using an orthonormal basis of $[W_k; W_k^\perp]$ as follows:
\begin{equation}
\label{eq:acqexp}
\mathsf{S}^{r_k} f_k = \sum_{\substack{\beta; |\beta|_{\mathsf{d}(f_k)+1:q} \leq s(f;W_k); \\ |\beta|\leq r_k}} \langle f, H_\beta \rangle H_\beta~.    
\end{equation}

We first show that, using the fine-grained structure, we can explicitly lower bound the number of eigenvalues that have grown after certain time. For that purpose, for $k=1\ldots K$ and  
a fixed parameter $\eta \in (0, 1)$, we define the escape times 
 \begin{align}
     \label{eq:escape_times}
       \tau_k(\eta) := \inf\{ t \geq 0\, ;\,  \lambda_{p_k} \geq 1 - \eta\}~.
 \end{align}
We will divide the growth of the eigenvalues in two distinct steps: {\it (i)} times until $\tau_k(1/2)$ at which we will say that the ``eigenvalues have escape mediocrity" (any $\Theta(1)$ constant could hold in place of $1/2$), and {\it (ii)} times between $\tau_k(1/2)$ and $ \tau_k(\eta)$, for which the growth is exponential. Note that most of the time is spent near in a band of size $d^{-1/2}$.   
Precisely, we prove the following key lemma in Appendix \ref{sec:prooflemmaseqescape}. 
\begin{restatable}[Sequential Escape from Mediocrity]{lemma}{lemmaescape}
\label{lem:sequential_escape}
Assume $f \in H^1(\gamma_q)$. With probability greater than $1-\delta$, for each $k=1, \ldots K$, after time $\tau_k (\eta) = \Cu \left[\sum_{k' \leq k} (\delta^{-2(s_{k'} -1)} d^{s_{k'}-1} \vee \log(d /\delta)) + \log(1/\eta) \right]$ 
at least $p_k$ eigenvalues are larger than $1 - \eta $. 
\end{restatable}
We recall from Corollary \ref{coro:lambda_increase} that the eigenvalues $\lambda_j(t)$ measuring the subspace correlation are non-decreasing functions of time. 
The main ingredient behind Lemma \ref{lem:sequential_escape} is a lower bound on $\frac{d}{dt}\lambda_j$ that quantifies, via a Gronwall-type differential inequality, the escape time of order $d^{s_k -1}$ for suitable packets of eigenvalues (given by the cascade dimensions $p_k$).

From Lemma \ref{lem:sequential_escape} we can already deduce the last conclusion of Theorem \ref{thm:coarse-grained} in Eq (\ref{eq:thm_quantitative_convergence}). 
Indeed, observe that after time $t\geq \tau_{K}(\eta)$ we have that all eigenvalues are greater than $1-\eta$, and therefore  
\begin{align}
    \|W(t) W(t)^\top - W_* W_*^\top \|^2 & \leq 2 - 2 \lambda_{\text{min}}(W(t) W(t)^\top W_* W_*^\top) \nonumber \\
    &= 2( 1 - \lambda_{q}^2(t))\nonumber \\
    & \leq 4 \eta~. \label{eq:counded_W}
\end{align}

Next, we establish the fact that the dynamics have a `saddle-to-saddle' structure, where the saddles are precisely given by the subspaces $\tilde{W}_k$ determined by the regrouped cascade $\{ \tilde{f}_k, \tilde{W}_k, \tilde{l}_k, \tilde{s}_k\}_{k \leq \tilde{K}}$. 
In order to establish such structure, we need a matching upper bound for the subspace correlation growth discussed earlier. This then leads to the following characterization, proven in Appendix \ref{sec:prooflemmalowergroup}. 
\begin{restatable}[Visiting saddles]{lemma}{lemmaregrouped}
\label{lem:lowergroup}
    Assume that $f\in H^\Xi$, where $\|f\|_\Xi$ is the Sobolev norm with spectral weights $\Xi(k) := k^{s^*} {c_{q,k}}\simeq k^{s^*+q}$.
    With probability greater than $1-\delta$, for each $k=1,\ldots \tilde{K}$, at timescale \\
    $\tilde{\tau}_k(\eta):=\mathcal{C}_{f,\delta,\tilde{s}_k} \left[(d^{\tilde{s}_k-1} \vee \log(d)) + \log(1/\eta) \right]$ there are exactly $\tilde{p}_k$ eigenvalues that have reached level $1- \eta$. More precisely,
    \begin{align}
    \label{eq:decoupling_saddle}
        \hspace{1.5cm} \lambda_j(\tilde{\tau}_k(\eta)) &\geq 1 - \eta~~~~~, \hspace{3cm} \text{ for $j \in \llbracket 1 , \tilde{p}_k \rrbracket $}~, \hspace{1cm} \text{ and, }  \\
        \hspace{1.5cm} \lambda_j(\tilde{\tau}_k(\eta)) &\leq \Cu d^{-1/2}~, \hspace{3cm} \text{ for $j \in \llbracket \tilde{p}_k + 1, q \rrbracket $.}
    \end{align}    
    Moreover, $V_{\tilde{p}_k}$ is aligned with the support $\tilde{W}_k$ of $\tilde{f}_k$, in the following sense: for $\tilde{\tau}_k \ll t \ll \tilde{\tau}_{k+1}$, we have
    \begin{equation}
    \label{eq:alignment}
        \| V_{\tilde{p}_k}(t) V_{\tilde{p}_k}(t)^\top  - \tilde{W}_k \tilde{W}_{k}^\top \|^2 =O(d^{-1})~.
    \end{equation}
\end{restatable}
Finally, let us address the convergence of the gradient flow: as $(W_t)_{t\geq 0}$ is bounded, up to extraction, it converges. From the fact that $W$ follows a gradient flow, the only possible points of accumulation are the critical points of $L$. The loss landscape analysis provided in Theorem~\ref{thm:critical_points} states that, for $\eta$ small enough, the only critical point such that $\|W W^\top - W_* W_*^\top \|^2 \leq 4 \eta$ is in fact $W_*$. Hence, the only accumulation point of $(W_t)_{t\geq 0}$ is $W_*$ and we deduce that $W_t \to W_*$.
This concludes the proof of Theorem \ref{thm:coarse-grained}.

%% file: octahedron_saddles.tex
\definecolor{cof}{RGB}{219,144,71}
\definecolor{pur}{RGB}{186,146,162}
\definecolor{greeo}{RGB}{91,173,69}
\definecolor{greet}{RGB}{52,111,72}

\begin{tikzpicture}[thick,scale=8]
\coordinate (A1) at (0,0);
\coordinate (A2) at (0.6,0.2);
\coordinate (A3) at (1,0);
\coordinate (A4) at (0.4,-0.2);
\coordinate (B1) at (0.5,0.5);
\coordinate (B2) at (0.5,-0.5);

\begin{scope}[thick,dashed,,opacity=0.6]
\draw (A1) -- (A2) -- (A3);
\draw (B1) -- (A2) -- (B2);
\end{scope}
\draw[fill=yellow!35,opacity=0.6] (A1) -- (A4) -- (B1);
\draw[fill=cyan!35,opacity=0.6] (A1) -- (A4) -- (B2);
\draw[fill=purple!35,opacity=0.6] (A3) -- (A4) -- (B1);
\draw[fill=olive!35,opacity=0.6] (A3) -- (A4) -- (B2);
\draw (B1) -- (A1) -- (B2) -- (A3) --cycle;

\filldraw[black] (0.25,0.15) circle (0.2pt) {};
\filldraw[black] (0.15,0.02) circle (0.2pt) {};
\filldraw[black] (0.35,0.075) circle (0.2pt) {};
\filldraw[black] (0.35,-0.275) circle (0.4pt) node[anchor=south east]{$\ \tilde{W}_1$};

\filldraw[black] (0.6,-0.275) circle (0.4pt) node[anchor=south east]{$\ \tilde{W}_2$} {};

\filldraw[black] (0.85,0.1) circle (0.2pt) {};
\filldraw[black] (0.65,0.1) circle (0.4pt) node[anchor=north east]{$\ \tilde{W}_3$} {};

\filldraw[black] (B2) circle (0.4pt) node[anchor=north east]{$\ \tilde{W}_0 = \{0\}$};

\filldraw[black] (B1) circle (0.4pt) node[anchor=south west]{$\ \tilde{W}_4 = \{\R^q\}$};

\draw [black, line width=0.35mm] plot [smooth, tension=0.6] coordinates { (B2) (0.35,-0.275) (0.6,-0.275) (0.65,0.1) (B1)};

\end{tikzpicture}

%% file: V-Planted_model.tex
\section{The planted model}
\label{sec:planted_model}

In this section, following~\cite{arous2021online}, the aim is to understand the subspace learning in the case where the function $f^*$ is already known. Hence, this corresponds to $f = f^*$ in Eq.\eqref{eq:loss_learning_model} with $r = q$.  We refer to this as the \textit{planted model}, and the loss function now reads, $ L(W) = \frac{1}{2}\E_{\gamma_d}( f (W^\top x) - f ({W_*}^\top x))^2 = \| f\|_{\gamma_q}^2 -  \langle  f, \mathsf{A}_{M} f \rangle_{\gamma_q}$,
where we still define $M = W_*^\top W \in \R^{q \times q}$, for $W,W_* \in \S(d,q)$ and $f \in L^2_{\gamma_q}$. Once again, as the norm of the function $\|f\|^2$ is a fixed quantity, we will redefine, with a slight abuse of notation the loss as the correlation  
\begin{align}
\label{eq:loss_planted_model}
L(W) =  \langle  f, \mathsf{A}_{M} f \rangle_{\gamma_q}~,
\end{align} 
that we seek to maximize. We train the model with the canonical gradient structure inherited by the Stiefel manifold, that we note $\nabla^\S$.  

At first glance, the planted model appears to be an easier estimation problem than the semi-parametric problem, since there is more information available to the learner. Indeed, in the single-index setting $q=1$, the optimization of Eq. \eqref{eq:loss_planted_model} has a simple geometry, depending on the parity of $f$: if $f$ is even, then gradient flow always recovers the planted direction, while if $f$ is not even this is only guaranteed whenever $\theta(0) \cdot \theta_* >0 $, which occurs with probability $1/2$ \cite{arous2021online}.

As we show next, the situation for $q>1$ is fundamentally different: 
we will see that the loss landscape has, for certain target functions, bad critical points that trap gradient flow in sub-optimal solutions with arbitrarily high probability. This reveals an inherent advantage of performing hierarchical learning in the semi-parametric problem.

\subsection{Loss and summary statistics}

As previously, the loss has a nice representation in terms of the SVD of  $M = W_*^\top W = V \Lambda U^\top$, where $U,V \in \O_q$ and $\Lambda$ is the diagonal matrix of singular values.  Indeed, from the results of the section~\ref{sec:harmonic_analysis}, we can write the loss as
\begin{align}
\label{eq:loss_planted}
       L(W) =  \langle  \P_{V^\top} f, \Av_\Lambda \P_{U^\top}  f \rangle =  \sum_{\beta} \alpha_\beta(V) \alpha_\beta(U) \lambda_\beta~,
\end{align}   
where we defined, for all $\beta \in \N^q$, the coefficients  $\alpha_\beta(V) = \langle f, H_\beta(V) \rangle$ and $\alpha_\beta(U) = \langle f, H_\beta(U) \rangle$ of $f$ in two differently \textit{rotated basis} of Hermite polynomials. This reveals once again the natural summary statistics of the problem: the diagonal (non-negative) operator $\Lambda \in \mathcal{D}$ of singular values, and the unitary operators $U, V \in \O_q$. They encode respectively the alignment between the Grassmanian projections of $W$ and $W_*$ (ie the two orthogonal projections defined by the subspaces) and the alignment between the frames. 
Indeed, from Eq.\eqref{eq:loss_planted} we can verify that the energy $L$ is maximised when two alignments occur: alignment in terms of the Grassmanian, since $P_\Lambda$ is a contraction for any $\Lambda \neq I_r$,
and alignment in terms of the basis, measured through the correlation $\langle \P_V f, \P_U f \rangle$ in the case $\Lambda =  I_r$. 

An important remark that makes the fast-learning model and the planted model fundamentally different is that now, the coefficients of $L$ as a multivariate power series in $\lambda$ \textit{are no longer non-negative}. When trained via the Stiefel gradient flow, 
\begin{align}
\label{eq:stiefel_flow_planted}
\dot{W}_t = \nabla^\S_W L(W_t)~,
\end{align}
where $W_0 \sim \mathrm{Unif}(\S(d,q))$. This will have the consequence that $t \to \Lambda_t$ is no longer non-decreasing. For the sake of clarity, we decided to show the equation of motion of $(U,V,\Lambda)$ resulting from Eq.\eqref{eq:stiefel_flow_planted} in Lemma~\ref{lem:stiefel_summary} of the Appendix. Finally, following \cite[Theorem 3.2.1]{chikuse2003statistics}, the eigenvalues in the diagonal of $\Lambda$ at initialization are distributed similarly as in the Grassmann case, see Eq.\eqref{eq:multivariate_law_eigenvalues}, and hence the resulting Lemma~\ref{lem:eigenvalues_init} still holds to describe their magnitude.

In the following we study typical cases of success and failure of the learning of $W_*$. 
\subsection{The radial case}
We assume in this section that the function $f$ is radial, i.e., for all $O \in \O_q$, we have $\P_O f = f$. Let us recall that, from its rotation invariance, we know that necessarily the function has intrinsic dimension $q$ and can be written $f = \sum_\beta \alpha_\beta H_\beta$ in any  Hermite basis. We call $s$ the \textit{first} information exponent as defined in Eq.\eqref{eq:information_expo}: $s(f) := \inf\left\{ |\beta|=\sum_i \beta_i; \langle f, H_\beta\rangle \neq 0 \right\}$. In this radial case, this exponent alone qualifies the dynamics as will be described below.
We can write the loss as a function of $\Lambda$ only, i.e. $L(W) = \sum_\beta \alpha_\beta^2 \lambda_\beta:= \ell(\Lambda)$. Hence, for all $t \geq 0$, we have $M_t = W_*^\top W_t = V_0 \Lambda_t U_0^\top$, and the eigenvalues verify in this case the ODE,
\begin{align}
\label{eq:lambda_stiefel_flow_planted}
\dot{\Lambda}_t = (1 - \Lambda_t^2) \nabla_\Lambda \ell (\Lambda_t)~.
\end{align}
As a result, we can show that the Stiefel gradient flow leads to perfect recovery on a quantitative fashion. Indeed, we have the following result:
\begin{theorem}
    \label{thm:planted_radial_case}
    The Stiefel gradient flow followed by $(W_t)_{t \geq 0}$ as Eq.\eqref{eq:stiefel_flow_planted} and initialized uniformly in the Stiefel manifold is such that,
    \begin{enumerate}[label=(\roman*)]
        \item We have the following convergences: $L(W_t) \xrightarrow[]{t \to \infty} 0$ as well as $ W_t W_t^\top \xrightarrow[]{t \to \infty} W_* W_*^\top$
        \item For any $\delta,\eta > 0$, with probability at least $1 - \delta$, after time $t \geq \overline{\mathcal{C}}_{f, \delta, s} [(d^{s/2-1} \vee \log d) + \log(1/\eta)]$, 
    \begin{align}
        \|W_t W_t^\top - W_* W_*^\top\|^2 &\leq 4\eta~,
    \end{align}
    where we have defined $\overline{\mathcal{C}}_{f, \delta, s} = \delta^{-(s-2)} \Cu$ for $s>2$ and $\mathcal{C}_{f,\delta,2} = \Cu + \log(1/\delta)$.   
    \end{enumerate}
\end{theorem}
\begin{proof}
    See Section \ref{app:radial_case} for the proof of this theorem.
\end{proof}
We see here that the radial case enables the global convergence of the Stiefel gradient flow in this planted model. This is a particular case that happens to behave well: the eigenvalues still have the non-decreasing property that the previous section leverages. On top of this, the fact that there is no \textit{rotation movement} of the singular vectors really simplifies the dynamics: in this case, the gradient flow escapes only one saddle, similarly to the single index model.     

However, we show in the next section that this benign behavior is not always to be expected, and that gradient flow on the planted model does not always lead to global convergence. 

\subsection{A failure case: lack of global convergence of the Stiefel gradient flow}
\label{sec:planted_failure1}
\begin{figure}[ht]
    \centering
    \includegraphics[width=.6\textwidth]{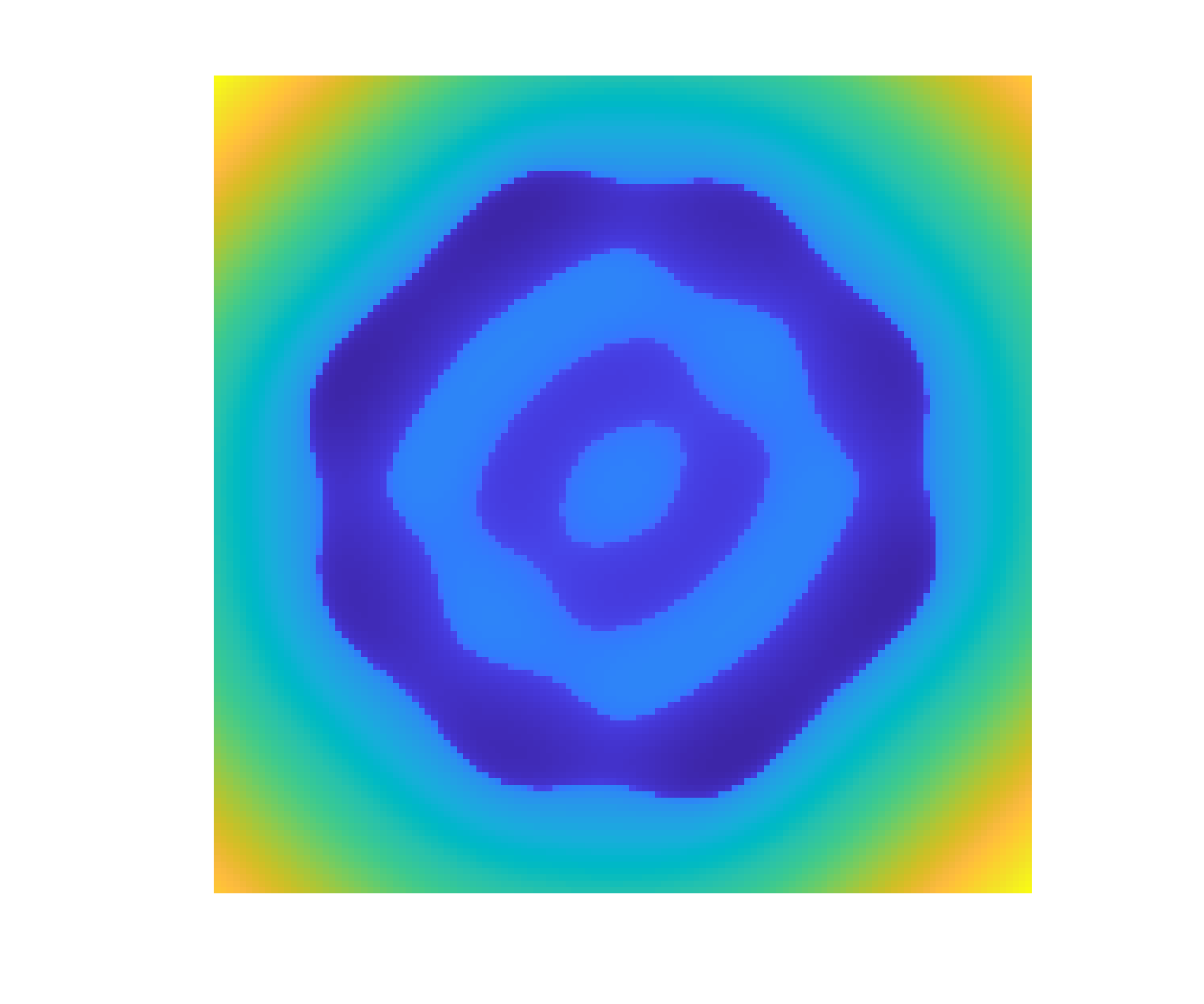} 
    \caption{Example of target function $f\in L^2_{\gamma_2}$ that leads to failures in the Stiefel Gradient Flow, corresponding to Theorem \ref{thm:planted_failure}. }
    \label{fig:planted_examples}
\end{figure}

We consider a target function of intrinsic dimension~$ q= 2$ of the following form, for~$x \in \R^2$:
\begin{equation}
\label{eq:plantedmodelcounter}
f^* := f + \epsilon g, \quad f(x) := \frac{1}{2}\|x\|^2 - 1~,\text{ and } g := 2 \mathsf{P}_{w_0} h_s + \sum_{j=1}^{N-1} \mathsf{P}_{w_j} h_s~,
\end{equation}
where for $j = \llbracket 0, N-1 \rrbracket $, we have $w_j = (\cos(2 \pi j /N), \sin(2 \pi j /N))$ the $j$-th of the $N$-th roots of unity, with $N,s \geq 3$, $s$ even and~$\epsilon > 0$ to be fixed later. The function $f^*$ is illustrated in Figure \ref{fig:planted_examples}. We will use the following properties.
\paragraph{Correlation loss.} Because the functions $f$ and $g$ belong to different harmonics, they act separately on the loss so that for all $W \in \S(d,q)$, such that the SVD of the correlation writes $M = W_*^\top W = V \Lambda U^\top$, we have
\begin{align}
\label{eq:planted_failure}
    L(W) = \langle \Av_M f, f \rangle  + \epsilon \langle \Av_M g, g \rangle = \frac{1}{2}(\lambda_1^2 + \lambda_2^2 ) + \epsilon \langle \Av_M g, g \rangle.
\end{align}
The reason behind building this function as a sum of a radial part and the function $g$ that carries these symmetries is to trap the Stiefel gradient flow in a bad local maximum of the correlation. More concretely, as seen in the previous section, the radial part will make the eigenvalues grow to $I_2$ without any movement of $(V,U)$, and we build $g$ to have bad local maxima when $\Lambda$ is close enough to the identity. This is due to the rotational symmetries in $g$, as we show now. To set up notations, given the two connex components of $\O_2$, we write $U = A R_\theta$ and~$V = R_{\theta'}$, where $R_\theta$ denotes the rotation with angle $\theta \in [0,2\pi)$, and with either $A = I_2$ if $\det(M) > 0$ and either $A = \text{diag}(1, -1)$ if $\det(M) < 0$.  In the two cases, we have the following representation of the correlation that $g$ is responsible for, when $\Lambda= I_2$. 
\begin{lemma}
\label{lem:correlation_planted_model_symmetries}
    Assume that $\Lambda = I_2$, then we have
    \begin{align}
    \label{eq:autocorrel}
        \langle \Av_M g, g \rangle = \varphi(\theta - \theta')~,
    \end{align}
    defining  $ \varphi(u) := 4 \cos(u)^s + 2 \sum_{j=1}^{N-1} \cos(u + \frac{2\pi j}{N})^s + 2 \sum_{j=1}^{N-1} \cos(-u + \frac{2\pi j}{N})^s+ \sum_{j,j'=1}^{N-1} \cos(u + \frac{2\pi}{N}(j - j'))^s$.
\end{lemma}
\begin{proof}
    Indeed, we have $ \langle \Av_M g, g \rangle = \E_z[g(A R_\theta z) g(R_{\theta'}z)] = \E_z[g(R_\theta z) g(R_{\theta'}z)] 
     = \E_z[g(R_{\theta - \theta' }z) g(z)] = \varphi(\theta - \theta') $, thanks to the fact that $\E_z[h_s(v \cdot z) h_s(v' \cdot z)] = (v \cdot v')^s$ for~$\|v\| = \|v'\| = 1$.
\end{proof}
\paragraph{Properties of~$\varphi$.} As being an auto-correlation function, $\varphi$ is $\pi$-periodic, even, and achieves its global maximum at~$0$. Further, the main idea behind building this function is that $0$ is a global maximizer, but there are also $N-1$ \textit{strictly suboptimal} peaks near each $u_k = k \pi/N$ for $k = \llbracket 1, N \rrbracket$, when~$s$ is large enough (we consider~$s$ such that~$\cos(\pi/10N)^s \leq 1/10 N^2$, see Appendix~\ref{sec:app_failure_stiefel}).
\paragraph{A negative result}
We consider the Stiefel gradient flow, $\dot{W}_t = \nabla^{\S} L(W_t)$, associated with the loss Eq.\eqref{eq:planted_failure}, from a uniformly drawn initialization $W_0 \sim \mathrm{Unif}(\mathcal{S}(d,r))$. As announced in the title of this section, we show that with high probability the Stiefel gradient flow does not converge to a global maximizer of the correlation $L$. This is what we quantitavely state in the following theorem.

\begin{theorem}[Stiefel Gradient Flow trapped with high-probability]
\label{thm:planted_failure}
    Fix any~$\epsilon > 0$ small enough such that
    \[
\arccos(1 - 2 \epsilon N \sqrt{2\log N} (1 + \log(c_{f,g} / \epsilon))) \leq \frac{\pi}{10N},
\]
    with~$c_{f,g} = 1 + 2\|f\|^2 + s\|g\|^2$.
    With probability larger than ~$1 - O(1/N)$ with respect to the initialization, the dynamics on~$f^*$ converges to a bad local maximum, i.e. we have $W_t \xrightarrow[t \to \infty]{} W_\infty$ where $ L(W_\infty) \leq  L_{\mathrm{max}} - 2/3 $, where $L_{\mathrm{max}}$ is global maximum of $L$. 
\end{theorem}

In this theorem we show that for general functions $f$, the Stiefel gradient flow in the planted setting may not converge to the global optimum. This is in contrast from the situation of the one-dimensional case, where the probability of success was \textit{at worst} $1/2$: this highlights the differences and difficulties brought by the multiple dimensions of the model. To comment on the function built that prevents global convergence, Lemma~\ref{lem:correlation_planted_model_symmetries} shows that when $\Lambda = I_2$ (i.e. $M \in \mathcal{O}_2$), there are many local failure modes, i.e. local optima for the correlation loss that are not global optima. This is due to the many rotational symmetries inherent to the function $g$ (the radial function obviously also carries these symmetries). The proof's strategy crucially leverages this: as described in the previous section, the radial function pushes $\Lambda_t$ to $I_2$ while not moving $V, U$, which are only affected by the function $\epsilon g$, hence moving only in a $O(\epsilon)$ neighborhood. Eventually, when $\Lambda$ is large enough, the function $g$ comes into play and traps the dynamics in a local maximizer. The whole process is described precisely in the proof of the theorem that can be found in Section~\ref{sec:app_failure_stiefel} of the Appendix.

\begin{figure}[ht]
    \centering
    \includegraphics[width=.49\textwidth,trim=1.5in 0in 1.5in 0in, clip]{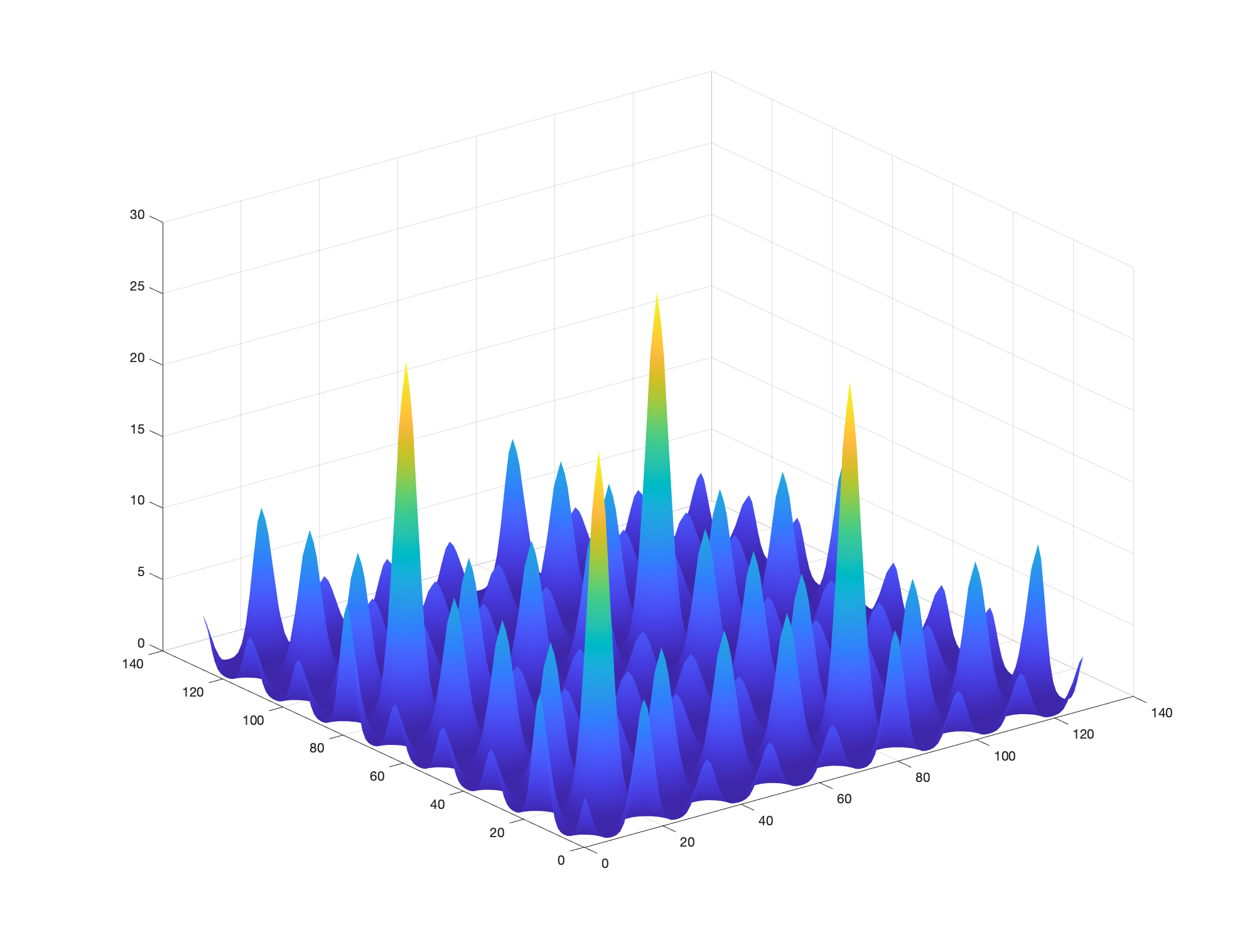} 
    \includegraphics[width=.49\textwidth,trim=1.5in 0in 1.5in 0in, clip]{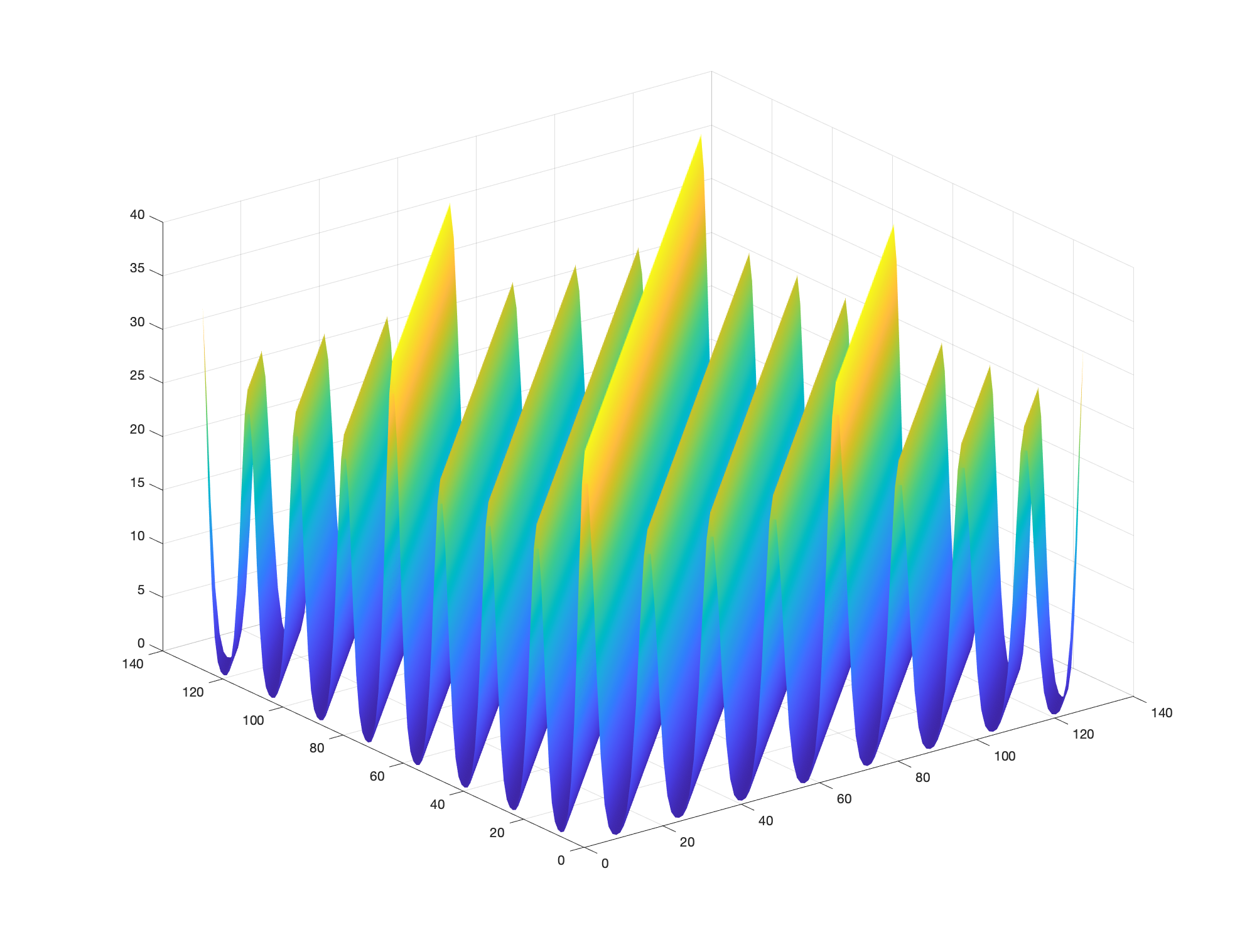} 
    \caption{Landscape associated with the target function of Figure \ref{fig:planted_examples}. We represent $L(W) = \ell(\theta, \eta, \lambda_1, \lambda_2)$, where $M = U \Lambda V^\top$ and $U = R_\theta$, $V = R_\eta$. We plot the $(\theta, \eta)$ plane for different values of $(\lambda_1, \lambda_2)$. {\it left:} $(\lambda_1, \lambda_2) = (1, 10^{-3})$, {\it right:} $(\lambda_1, \lambda_2) = (1, 1)$. Observe that in this latter case the loss is precisely $\varphi(\theta-\eta)$ from Eq \eqref{eq:autocorrel}, and exhibits bad local maxima.}
    \label{fig:planted_examples_b}
\end{figure}

\subsection{Lack of subspace convergence}
\label{sec:planted_failure2}

In the previous example, even though the loss converges towards a suboptimal value, all eigenvalues converge to $1$, and thus  
$WW^\top \to W_* W_*^\top$. In other words, the Stiefel gradient flow managed to recover the planted subspace, albeit not with the proper basis alignment. 
One can thus ask whether whether this is true in general, or whether there are target functions $f^*$ for which the loss $L(W)$ has local maximisers with $W W^\top \neq W_* W_*^\top$, so even the subspace might fail to be recovered. 

We now show that the latter is true, by modifying the previous example (\ref{eq:plantedmodelcounter}) with a different target function with more pervasive adversarial properties. 

\paragraph{From discrete to unitary correlation} For $N \in \N$, given a discrete sequence $Z=(Z_0, \ldots Z_{N-1}) \in \R^N$, we consider its autocorrelation as well as its self-convolution
\begin{align}
    \varphi_Z(k) &= \sum_{l} Z_l Z_{l+k \text{ mod } N} ~,\\
    \overline{\varphi}_Z(k) &= ( Z \ast Z)(k) = \sum_l Z_l Z_{k-l \text{ mod } N}~,
\end{align}
where $\ast$ is the cyclic convolution modulo $N$. 
Similarly as in Lemma \ref{lem:correlation_planted_model_symmetries}, we will now build $g \in L^2_{\gamma_2}$ with specific autocorrelation properties from a suitable discrete sequence $Z$. 
Consider as before $w_j \in \S_1$, $j=0,\ldots N-1$ the $N$-th roots of unity, $s$ even, and define  
\begin{align}
    g_Z &:= \sum_{j=0}^{N-1} Z_j \mathsf{P}_{w_{j}} h_s~.
\end{align}
We verify analogously that the autocorrelation of $g_Z$ in $\mathcal{O}_2$ is directly related to the autocorrelation and self-convolution of $Z$:
\begin{restatable}[discrete-to-unitary autocorrelation]{fact}{discretecontcorrel}
\label{fact:discrete_unitary_correl}
    For $U \in \mathcal{O}_2$, we have 
    \begin{align}
        \varphi_g(\theta) &:= \langle \mathsf{P}_U g_Z, g_Z \rangle = \begin{cases}
            (\varphi_Z \ast \cos^s)(\theta) & \text{if} \det(U)=+1 ~,\\
            (\overline{\varphi}_Z \ast \cos^s)(\theta) & \text{if} \det(U) = -1~,
        \end{cases}
    \end{align}
    where $U = A R_\theta$ as described in the previous section, and where we abused notation to write $\varphi_Z(\theta) = \sum_{j} \varphi_Z(j) \delta(\theta - w_{j})$ the natural embedding of the discrete autocorrelation on $\S_1$. 
\end{restatable}

In words, the autocorrelation of $g_Z$ on the unitary group is a convolved version of the discrete autocorrelation of $Z$, modulo the mirror reflection which changes $\varphi_Z$ to $\overline{\varphi}_Z$. The smoothing kernel has bandwidth $\simeq s^{1/2}$ controlled by the information exponent $s$ of $g$.  
We will now design a discrete sequence $Z$ leading to desirable (i.e. adversarial) properties on the underlying correlation $\varphi_g$. 

\paragraph{Sequences with nearly negative autcorrelation}
In order to prevent the Stiefel gradient flow from recovering the subspace, we want to create a `push-back' force to discourage the singular values $\lambda_i$
 from growing to $1$. This amounts to considering a sequence $Z$ such that both its auto-correlation and self-convolution are `as negative as possible'. Since $\varphi_Z(0) = \| Z \|^2 > 0$, it is not possible that $\varphi, \overline{\varphi}$ are strictly negative functions; however we can contruct sequences with a single positive autocorrelation and self-convolution value (after accounting for the even symmetry):
 \begin{restatable}[nearly negative autocorrelation]{lemma}{negativecorrel}
 \label{lem:negauto}
     There exists an even sequence $Z^* \in \R^N$ such that $\varphi_Z(k) = \overline{\varphi}_Z(k) < 0$ for any $k \neq 0,N/2$. 
  \end{restatable}

\paragraph{Yet Another Negative Result}

\begin{figure}[ht]
    \centering
    \includegraphics[width=.5\textwidth]{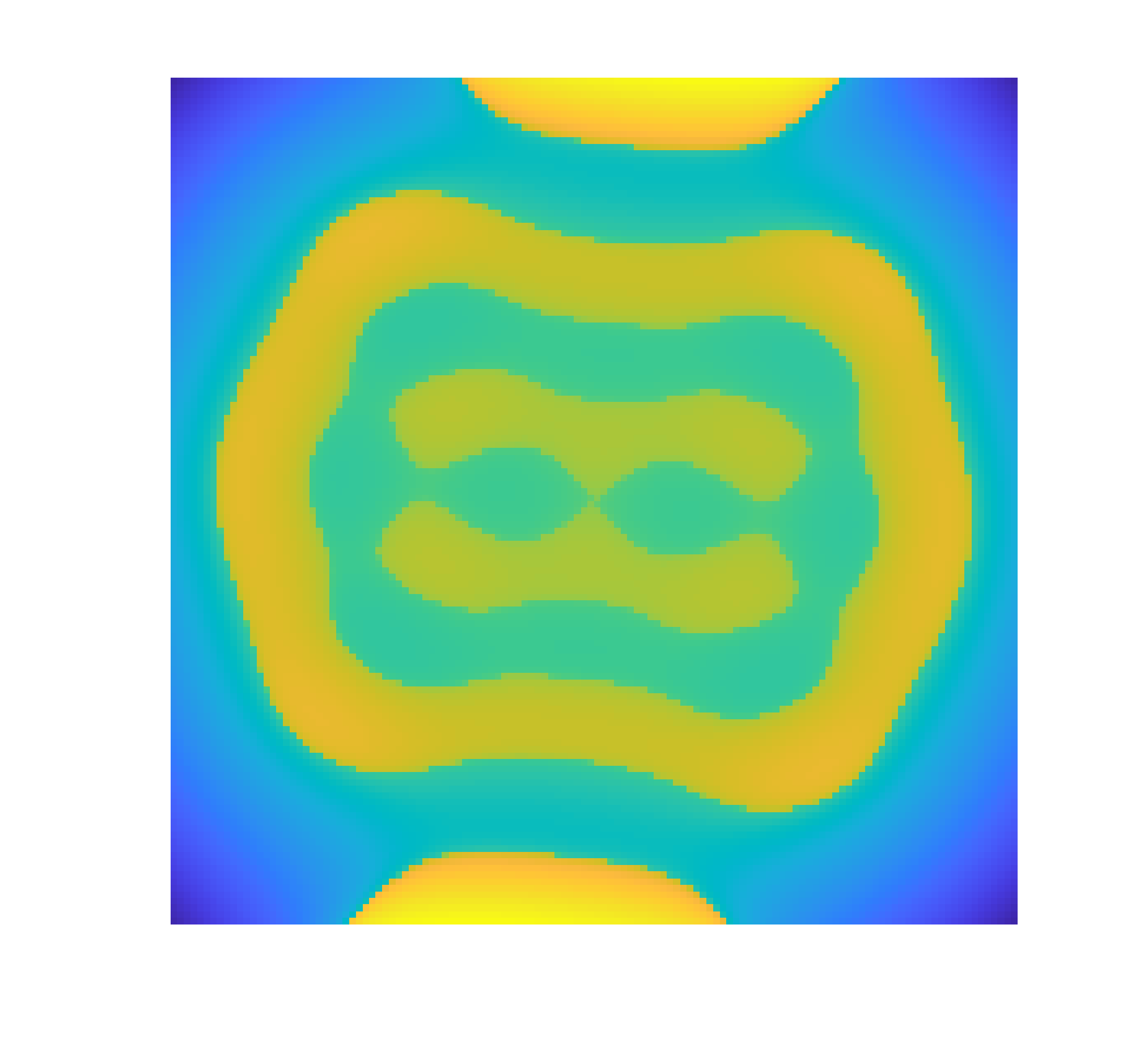}
    \caption{Example of target function $f\in L^2_{\gamma_2}$ that leads to failures in the Stiefel Gradient FLow. Example corresponding to Proposition \ref{prop:badmaxima}.}
    \label{fig:planted_example_2}
\end{figure}

    Take $s\gg N^2$ even, and consider a target $f^* = g_Z$ with $Z$ even and built to have nearly-negative autocorrelation, as per Lemma \ref{lem:negauto}. We illustrate one such example in Figure \ref{fig:planted_example_2}. 

\begin{restatable}[Existence of bad subspace maximisers]{proposition}{badmaxim}
\label{prop:badmaxima}
    The Landscape $L(W)$ contains local maxima at points $W$ such that $\lambda_2 < 1$. 
\end{restatable}
\begin{proof}
The proof of this proposition is in Appendix \ref{app:planted_failure2}. The main idea of the proof is to consider a point of the form $M^*=(w_{a^*}, w_{b^*}, 1, \lambda^*)$, where $a^*\neq b^*$ are such that $Z_{a^*} Z_{b^*} > 0$ and $\lambda^* < 1$, and we recall that $w_l = (\cos 2\pi l /N, \sin 2\pi l/N)$ are $N$-th roots of unity. Thanks to the structural properties of $Z$, we show that, for appropriate values of $\lambda^*<1$, $M^*$ cannot be continuously connected to the boundary $\mathcal{O}_2$ without decreasing the correlation loss, indicating that gradient flow initialized in a neighborhood of $M^*$ will be trapped.  
\end{proof}

Proposition \ref{prop:badmaxima} thus confirms that the topology of the planted optimization landscape is fundamentally different as one goes from the single-index $q=1$ to the multi-index $q>1$ setting. For $q=1$ the landscape has a degenerate saddle at $M=0$; while it cannot be escaped due to a topological obstruction (since it is defined on a one-dimensional domain), it can be avoided with probability $1/2$. On the other hand, for $q>1$ the planted model \emph{can} exhibit bad local maxima outside the boundary $M \in \mathcal{O}_2$ --- which are thus local attractors of the Stiefel gradient flow dynamics. This stands in contrast with the Grassmann gradient flow from Section \ref{sec:grassmannian_perspective}, which is guaranteed to converge, and with explicit rates of convergence. 
Finally, while Proposition \ref{prop:badmaxima} does not quantify the probability that Stiefel gradient flow will be trapped in these bad local maxima from their typical initialisation, we suspect it can be made arbitrarily large by playing with the model parameters $N, s$ --- in particular thanks to the fact that the loss nearby $M \in \mathcal{O}_2$ is negative except for a subset of relative measure $O(1/ N)$.

\begin{figure}[ht]
    \centering
    \includegraphics[width=.49\textwidth,trim=1.5in 0in 1.5in 0in, clip]{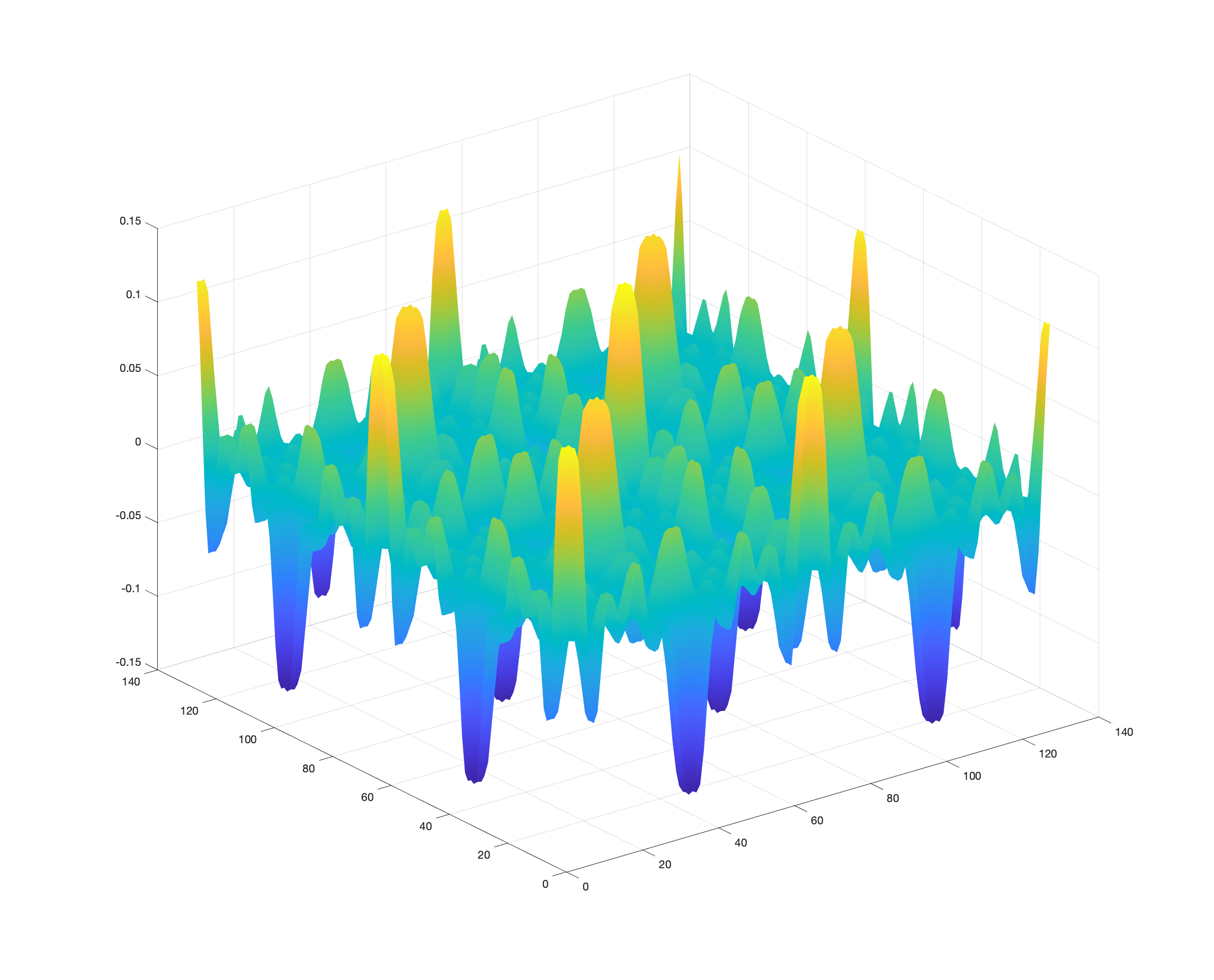} 
    \includegraphics[width=.49\textwidth,trim=1.5in 0in 1.5in 0in, clip]{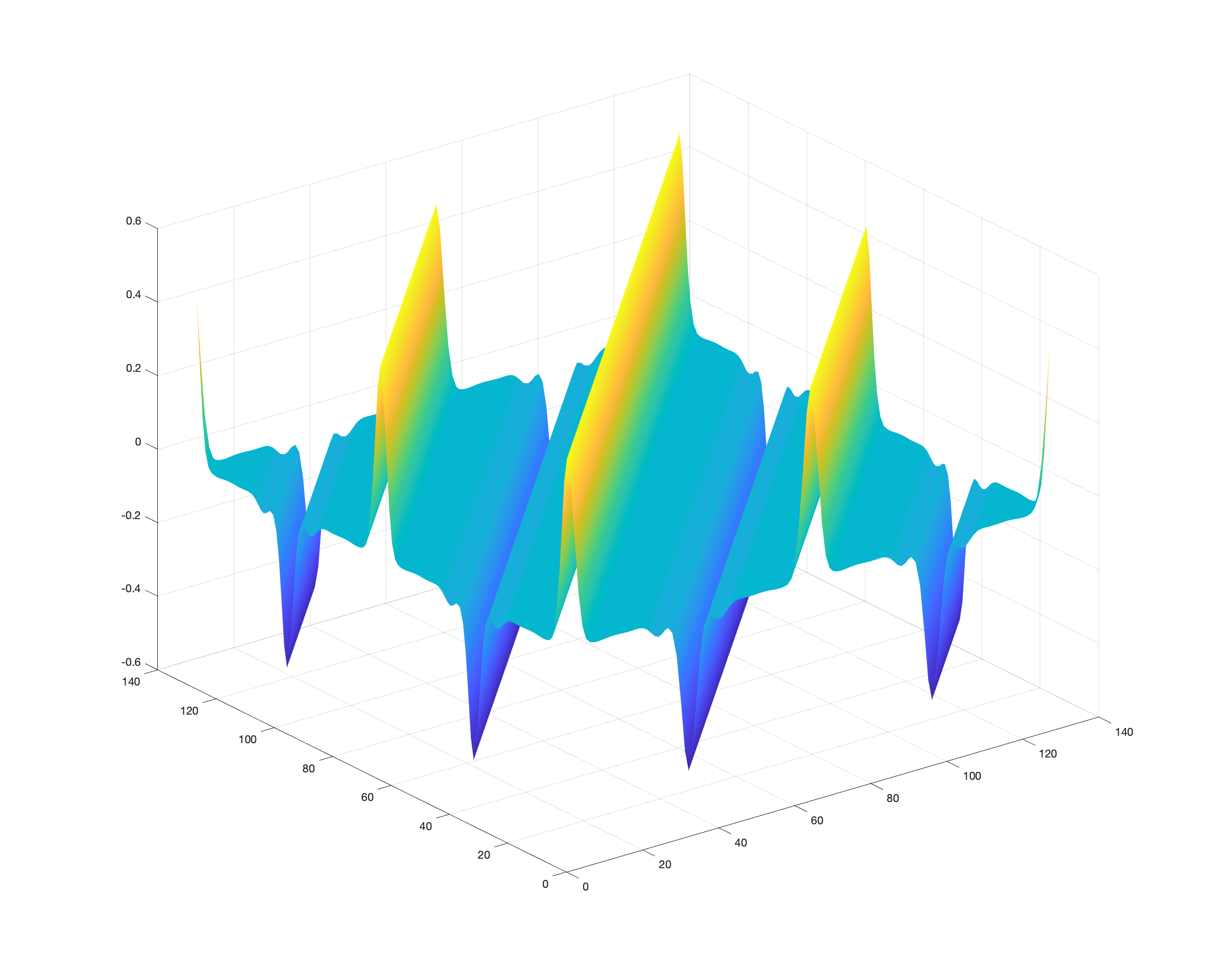}
    \caption{Slices of the optimization landscape $\ell(\theta, \eta, \lambda_1, \lambda_2)$ corresponding to the example from Figure \ref{fig:planted_example_2}. {\it left:} Slice $\ell(\theta, \eta, 1, 10^{-3})$; {\it right:} Slice $\ell(\theta, \eta, 1, 1)$. Observe that the landscape is a function of $\theta-\eta$, and is negative outside a small neighborhood of $\theta=\eta \,\mathrm{mod } \pi$. }
    \label{fig:planted_examples_landsscape2}
\end{figure}

\subsection{From Planted to Random Recovery}
In Sections \ref{sec:planted_failure1} and \ref{sec:planted_failure2} we have established  failures of Stiefel Gradient Flow to solve the optimization of the planted loss in Eq \eqref{eq:loss_planted_model}, while the Grassmannian joint learning objective is free of bad local extrema. 
We conclude this section by discussing how these negative results might relate to existing recent works that analyze the ability of gradient-descent methods to learn multi-index models. 

A popular template to analyze gradient dynamics using shallow neural networks for this problem is `layer-wise' training \cite{bietti2022learning,abbe2023sgd,dandi2023learning,damian2022neural,mahankali2023beyond,ba2022high}, where the first-layer weights of the network are updated during the initial phase of training while the second-layer weights remain frozen. In our language, this would translate into the following basic procedure: one draws a random function $\bar{f}$ from a certain probability measure $\nu$ over $L^2_{\gamma_q}$, and considers the (random) optimization problem 
\begin{equation}
\label{eq:random_landscape}
\bar{L}(W) := \langle \mathsf{P}_W \bar{f}, \mathsf{P}_{W_*} f_* \rangle = \langle \bar{f}, \mathsf{A}_M f_* \rangle~.    
\end{equation}
The natural question, beyond the scope of the present paper, is to understand whether such `random' recovery problem enjoys the benign optimization properties of our joint learning from Section \ref{sec:grassmannian_perspective}, or rather it suffers from similar roughness as our planted setting from Sections \ref{sec:planted_failure1} and \ref{sec:planted_failure2}. On that respect, we note that \cite{dudeja2018learning} considered a particular version of \eqref{eq:random_landscape} with $\bar{f}=\mathsf{P}_{\theta} h_s$ a pure harmonic single-index model, and showed\footnote{Up to the zero-level set.} \cite[Theorem 8]{dudeja2018learning} the absence of bad local maxima outside the set $\|M\|=1$. In other words, the single-index student is able to recover \emph{a} direction contained in the subspace $W_*$. The question is thus whether a different procedure could guarantee, with non-vanishing probability (in $q,d$), that \emph{all} directions of $W_*$ are identified.

%% file: VI-Perspectives.tex
\section{Towards Sample Complexity Guarantees}
\label{subsec:sample_complexity}
We just established quantitative learning guarantees in terms of the time-complexity to reach a certain correlation level on the population loss. We would like to understand to what extent such guarantees translate into sample-complexity bounds. In other words, if one now considers the empirical counterpart of the problem where the multi-index function is to be estimated from samples $\{(x_i, y_i = f^*(W_* x_i) )\}_{i=1\ldots n}$, with $(x_i)_{1\ldots n}$ drawn i.i.d. from $\gamma_d$, 
how should the required amount of samples $n$ scale in terms of the problem parameters $d, q, f$ to guarantee a desired target generalisation error.   

Such `translation' between time and sample-complexity has been addressed in several prior works, under different training algorithms and approximation models; for instance, online Stochastic Gradient Descent \cite{arous2021online,arous2022high,barak2022hidden} and its variants \cite{abbe2022merged,abbe2023sgd}, or batch Gradient Descent \cite{dudeja2018learning,bietti2022learning} using certain shallow neural network architectures. 
While the complete answer to this question is beyond the scope of the present paper, we now give an informal argument that suggests that the fast kernel learning algorithm of Section \ref{sec:rkhs_learning} will be amenable to a similar analysis. 

\paragraph{Known resuls in the single-index planted setting}

Before discussing our specific setting with the fast kernel learning, let us describe first the principles of such sample complexity on the planted single-index model. Here, we recall that a direction $\theta^* \in \mathcal{S}_{d-1}$ parametrizes the single-index function $\mathsf{P}_{\theta^*} f (x) = f( x \cdot \theta^*)$, where $f$ has information exponent $s$. 

As mentioned in the introduction, in the online setting of \cite{arous2021online}, this correlation is iteratively maximised using Stochastic Gradient Descent. 
Starting from a random initialization $\theta_0 \sim \mathrm{Unif}(\mathcal{S}_{d-1})$, one considers iterates of the form 
$$\theta_{t+1} = \frac{\theta_{t} + \eta g(\theta_t; x_t, y_t) }{\|\theta_{t} + \eta g(\theta_t; x_t, y_t)\|} \in \mathcal{S}_{d-1}~,$$
where $g( \theta; x, y) = y f'( x \cdot \theta) \left( x - (x \cdot \theta) \theta\right)$ is the stochastic spherical gradient, $\eta$ is the step-size, and $x_t \sim \gamma_d$, $y_t = f( x_t \cdot \theta^*)$. 
The resulting stochastic process $(\theta_t)_t$ can be analysed in terms of an `entropic' phase, where $\theta_t$ wanders around the equator $\{ \theta; |\theta \cdot \theta^* | \simeq 1/\sqrt{d} \}$ (referred to as `mediocrity'), and a determinstic/ballistic phase, after $\theta_t$ has escaped the mediocrity of initialization. 
The crucial property of the stochastic process $(\theta_t)_t$ is that its associated correlation $m_t = \theta_t \cdot \theta^*$ evolves according to a deterministic drift (given precisely by $\nabla^{\mathcal{S}} \bar{L}(\theta_t) \cdot \theta^*$), `polluted' by a diffusive martingale $\left[g(\theta_t; x_t, y_t) - \mathbb{E} g(\theta_t; x_t, y_t)\right]\cdot \theta^*$ and by the spherical renormalization. In other words, the correlation $m_t$ evolves along a perturbation of the population dynamics, whereby the relative strength of the perturbations is controlled by the step-size $\eta$. For appropriate choices~$\eta \sim d^{-s/2}$, \cite{arous2021online} shows that these perturbations are `tamed', leading to a tight sample complexity bound of $n \sim d^{s-1}$.  

Alternatively, one may consider the offline setting \cite{dudeja2018learning,bietti2022learning}, and view the 
resulting empirical loss landscape as a `noisy' version of the population landscape \cite{mei2016landscape}. In this context, the main question becomes whether the `signal' strength in the empirical gradients dominates over the typical statistical fluctuations near the initialization, so that the summary statistics undergo similar dynamics. Using again the planted single-index setting as an illustrative example, this amounts to measuring the typical fluctuations of $\nabla^{\mathcal{S}}\hat{L}(\theta)\cdot \theta^*=\frac1n \sum_{i=1}^n g(\theta; x_i, y_i) \cdot \theta^*$ around its expectation $\nabla^{\mathcal{S}}L(\theta)\cdot \theta^*$. Near the initialisation, one has $\nabla^{\mathcal{S}}L(\theta)\cdot \theta^* \simeq (\theta_0 \cdot \theta^*)^{s-1} \simeq d^{-(s-1)/2}$, while 
$\sup_\theta |\nabla^{\mathcal{S}}(\hat{L}(\theta)- L(\theta))\cdot \theta^* | \simeq \sqrt{d/n}$ w.h.p., leading to a required sample complexity of $n \sim d^{s}$ samples, ignoring log factors. This basic idea has been considered beyond the planted setting by `freezing' the second-layer weights in shallow neural architectures \cite{bietti2022learning,arnaboldi2023high,mahankali2023beyond}.

\paragraph{Fast Kernel Learning}
We now discuss how these techniques based on gradient concentration may apply to our setting. Recall that while the planted correlation loss is given by $W \mapsto \langle \mathsf{A}_M f, f \rangle$, with $M =  W_*^\top W$, the fast learning objective becomes $W \mapsto \langle \mathsf{A}_G f, f \rangle $, where $G = M M^\top$. 
At first glance, the fast learning method appears to incur in a 
`doubling' of the information exponent, since the order of the saddles is doubled when going from $M$ to $G$, resulting in an effective attenuation of the signal gradient `strength' around initialisation, from $\Theta(d^{-(s-1)/2})$ to $\Theta(d^{-s+1/2})$, where~$s$ is the information exponent relative to the first subspace (denoted $s(f)$ in Section~\ref{sec:escapemediocrity}). 

However, let us derive here a non-rigorous argument that such doubling is not necessarily incurring in worse sample complexity guarantees, e.g. in the offline setting.  
As depicted in Section~\ref{sec:fastideal}, the empirical risk minimisation associated with the fast learning method operates in two stages: given the current subspace $W$, we first perform kernel ridge regression to estimate the optimal current link function
\begin{align}
\label{eq:kernelridge_emp}
    \hat{f}^\mu_W &:= \arg\min_{\tilde{f} \in \mathcal{H}}\  \frac1n \sum_{i=1}^n ( \tilde{f}(W^\top x_i) - y_i)^2 + \frac{\mu}{2} \|\tilde{f}\|_{\mathcal{H}}^2~,
\end{align}
where we recall that $\|f\|_{\mathcal{H}}^2 = \langle f, \mathsf{Q}^{-1} f \rangle_{\gamma_q}$, with integral operator $\mathsf{Q}$ defined in Eq. \eqref{eq:kernel_integral_op}, and $y_i = f(W_*^\top x_i)$. 
Then we take an (infinitesimal) Stiefel gradient step in $W$ on the empirical loss 
\begin{align}
\label{eq:empstiefel_loss}
\hat{L}(\hat{f}^\mu_W,W) &= \frac1n \sum_{i=1}^n (\hat{f}^\mu_W(W^\top x_i) - y_i)^2 ~.   
\end{align}
To understand the order of magnitude of the sample complexity required, let us focus on the initial learning stage, where all the eigenvalues $\lambda_i$ of $M$ are at scale $O(1/\sqrt{d})$ thanks to Lemma~\ref{lem:eigenvalues_init}, and let us ignore the statistical dependencies across iterates\footnote{Typically one uses fresh samples to avoid such dependencies in a rigorous analysis; eg \cite{bietti2022learning,bing2021prediction}.}.  
We write $z_i = W^\top x_i$, and we decompose $y_i = \mathsf{A}_{M^\top} f(z_i) + \varepsilon_i$, with $\varepsilon_i = f(W_*^\top x_i) - \mathsf{A}_{M^\top} f(z_i)$. 
Observe that $\mathbb{E} [\varepsilon_i | \, z_i] = 0$ and 
$\mathbb{E}[ \varepsilon_i^2 |\, z_i] := \sigma^2 \leq \|f\|^2~.$
Using the basic properties of kernel ridge regression (e.g. \cite[Chapter 7]{bach2021learning}) we obtain 
\begin{align}
\label{eq:ridge_solution}
    \hat{f}^\mu_W &= \left( \hat{\mathsf{Q}} + \mu I\right)^{-1} \left[\hat{\mathsf{Q}} \mathsf{A}_{M^\top} f + \frac1n \sum_{i=1}^n \varepsilon_i \varphi(z_i) \right]~,
\end{align}
where $\varphi(z_i) = \sum_\beta {\mathsf{c}_{|\beta|}} H_\beta(z_i) H_\beta$ is the RKHS feature map, and $\hat{\mathsf{Q}} = \frac1n \sum_{i=1}^n \varphi(z_i) \otimes \varphi(z_i)$
is the empirical covariance operator (defined from $\mathcal{H}$ to $\mathcal{H}$). 

Recalling the population loss $L(\tilde{f}, W) = \| \mathsf{P}_W \tilde{f} - \mathsf{P}_{W_*} f \|^2$, our goal is to control the gradient fluctuations 
\begin{align}
\Delta(n, \mu):= \| [\nabla_W^{\mathcal{S}} \hat{L}(\hat{f}^\mu_W, W)]^\top W_* - [\nabla_W^{\mathcal{S}} L(f_W, W) ]^\top W_* \|~,
\end{align}
where $f_W = \arg\min_{\tilde{f}} L(\tilde{f},W) = \mathsf{A}_{M^\top} f$ as seen in Eq.~\eqref{eq:lossgrass_f}. 

Let us now argue that, by properly tuning $\mu$, the typical fluctuations between the empirical and population Stiefel gradients during this initial phase satisfy $\Delta(n, \mu) \ll \| [\nabla_W^{\mathcal{S}} L(f_W, W) ]^\top W_* \|$ whenever $n \gg d^{s-1}$. 

We decompose 
\begin{align*}
    \Delta^2(n, \mu) & \leq  \underbrace{2\left\| \left[\nabla_W^{\mathcal{S}} \left(\hat{L}(\hat{f}^\mu_W, W) - L(\hat{f}^\mu_W, W)\right) \right]^\top W_* \right\|^2}_{\mathcal{A}} + \underbrace{2\left\| \left[\nabla_W^{\mathcal{S}} \left({L}(\hat{f}^\mu_W, W) - L({f_W}, W)\right) \right]^\top W_* \right\|^2}_{\mathcal{B}}~.
\end{align*}
The first term measures the Stiefel gradient fluctuations when using the estimated link function $\hat{f}^\mu_W$, while the second term measures the error in the population gradient due to the link function estimation error. 
Let us first study the term $\mathcal{A}$. By definition, we have
\begin{align*}
    [\nabla_W^{\mathcal{S}} \hat{L}(\hat{f}^\mu,W)]^\top W_*  =\frac1n \sum_{i=1}^n (\hat{f}^\mu_W(W^\top x_i) - y_i) \nabla \hat{f}^\mu_W(W^\top x_i) x_i^\top W_* := \frac1n \sum_{i=1}^n \upsilon_i~,
\end{align*}
where $\upsilon_i$ are iid random vectors with $\mathbb{E} \upsilon_i = [\nabla_W^{\mathcal{S}} L(\hat{f}^\mu_W, W)]^\top W_*$. 
Their empirical average uniformly concentrates over the Stiefel manifold at rate given by $\widetilde{O}\left(\sqrt{d\mathbb{E}[\| \upsilon\|^2]/n}\right)$. Under mild regularity assumptions\footnote{So that we can appeal to concentration of $\upsilon(x)$ thanks to the Gaussianity of $x$.}, and as $\mathcal{H}$ has kernel with spectral decay at least $\mathsf{c}_{|\beta|} = O( |\beta|^{-(q+1)})$, one can thus expect 
\begin{align}
\mathcal{A} &\lesssim \frac{d \mathbb{E}( \| \upsilon\|^2)}{n} \lesssim \frac{d}{n}(\|f\|^2 + \| \hat{f}^\mu_W \|^2) \cdot \| \nabla \hat{f}^\mu_W \|^2 \lesssim \frac{d\| \nabla \hat{f}^\mu_W \|^2}{n} \leq \frac{d\| \hat{f}^\mu_W \|^2_{\mathcal{H}}}{n}~.
\end{align}
On the other hand, denoting $\overline{M}(g) = \partial_M [\langle g, \mathsf{A}_M f\rangle]$, we have by Remark \ref{rem:gradient_loss_M} that $\overline{M}(g) =  \nabla g \otimes \mathsf{A}_M \nabla f$, and the term $\mathcal{B}$ becomes 
\begin{align}
    \mathcal{B} &\lesssim \| \overline{M}(\hat{f}^\mu_W) - \overline{M}(\mathsf{A}_{M^\top} f) \|_F^2 \nonumber \\
    &= \| (\nabla \hat{f}^\mu_W - \nabla \mathsf{A}_{M^\top} f) \otimes \mathsf{A}_M \nabla f\|_F^2 \nonumber \\
    &\lesssim \|  \mathsf{\Pi}^{s-1}[\nabla \hat{f}^\mu_W - \nabla \mathsf{A}_{M^\top} f] \otimes \mathsf{\Pi}^{s-1} [\mathsf{A}_M \nabla f]\|_F^2 \nonumber \\
    & \lesssim d^{-s+1} \|  \hat{f}^\mu_W -  \mathsf{A}_{M^\top} f \|^2~, 
\end{align} 
where we use the fact that the spectral decomposition of $\mathsf{A}_M f$ has exponential decay $\| \mathsf{\Pi}^{k} \mathsf{A}_M f\| \simeq \lambda^k$, with rate $\lambda\sim d^{-1/2}$. 
By decomposing the excess risk into bias and variance terms, as in \cite[Prop. 7.3]{bach2021learning}, and adapting it to unbounded data with sub-exponential tails \cite[Lemma D.17]{bietti2022learning}, we obtain
\begin{align}
    \|  \hat{f}^\mu_W -  \mathsf{A}_{M^\top} f \|^2 &\lesssim \frac{\sigma^2}{\mu n}\text{Tr}((\mathsf{Q}+\mu I)^{-1}\mathsf{Q}) + \mu \langle \mathsf{A}_{M^\top} f, ( \mathsf{Q} + \mu I)^{-1}\mathsf{Q} \mathsf{A}_{M^\top} f \rangle_{\mathcal{H}}~.
\end{align}

Suppose now that $n \gg d^{s}$. Then, recalling that near initialization we have $\|\Av_{M^\top} f\|^2 \sim d^{-s}$, by taking $\mu=\Theta(1) \ll 1$ 
we obtain
\begin{align}
    \|  \hat{f}^\mu_W -  \mathsf{A}_{M^\top} f \|^2 &\lesssim \frac{1}{\mu n} + \mu \|\mathsf{A}_{M^\top} f\|^2 \simeq \frac{1}{\mu n} + \mu d^{-s} \ll \| \mathsf{A}_{M^\top} f \|^2~,
\end{align}
and similarly $\| \hat{f}^\mu_W\|_{\mathcal{H}}^2 \lesssim d^{-s} + \frac{1}{n\mu^2} \lesssim d^{-s}$.
 Therefore  
\begin{align}
    \Delta^2(n,\mu) & \lesssim \frac{d^{-s+1}}{n} + d^{-s+1} \| \hat{f}^\mu_W -  \mathsf{A}_{M^\top} f \|^2 \ll d^{-2s+1}\simeq \|[\nabla_W^{\mathcal{S}} L(f_W, W) ]^\top W_* \|^2~.
\end{align}
In other words, the gradient fluctuations are such that $\Delta \ll \|[\nabla_W^{\mathcal{S}} L(f_W, W) ]^\top W_* \| $ 
whenever $n \gg d^s$, indicating that the empirical gradient flow dynamics will escape the initial mediocrity phase.  

In summary, the kernel ridge regression is aggressively regularized in the mediocrity phase with  $\mu =\Theta(1)$, as the signal-to-noise ratio $d^{-s}\|f\|^{-2}$ is vanishingly small. Thus, when $n\gtrsim d^{s}$ the link function $\mathsf{A}_{M^\top} f$ can be estimated with small relative error, which then translates into a sufficiently accurate empirical Stiefel gradient. 
The next natural step is to make the concentration argument rigorous, and extend it to all the successive phases of learning described in Section \ref{sec:escapemediocrity}. 

Finally, another relevant question is to carry out the online SGD analysis for the fast learning method. By appropriately defining it on mini-batches (where the kernel ridge regression could be solved using fresh samples), the natural device to compensate for the `doubling' exponent would be to consider larger step-sizes than the prescribed $\eta \sim d^{-s/2}$. We note that this procedure could potentially improve the rate to $n = O(d^{s-1})$ from the previous heuristics.

%% file: conclusions.tex
\section{Conclusions}

\paragraph{Summary.}

Extending the Gaussian single-index learning to the multi-index setting 
comes with both positive and negative results. The multivariate nature of the problem
now breaks the commutative semigroup structure of the associated Gaussian smoothing operator ---and this has important implications in the 
geometry to the learning problem. In short, the main difference comes from the fact that the loss can typically travel near many saddle points before finding the target space, unveiling an \textit{incremental learning} where subspaces are learned sequentially.

In the semi-parametric setting, the common strategy of prior works of freezing the second-layer 
weights while optimizing the first ones \cite{bietti2022learning,abbe2023sgd,ba2022high,damian2022neural,arnaboldi2023high,mahankali2023beyond,dandi2023learning} might not readily work for general $q$-dimensional target link functions, as illustrated by the negative results for the planted setting. Instead, we consider a joint learning method (fast kernel) that maps the optimization into the Grassmann manifold, where gradient flow dynamics enjoy quantitative global convergence guarantees.  

Finally, although out of the scope of the present paper, we hint at the possibility of turning our time complexity guarantees into sample complexity guarantees of the form $n = \Theta(d^{s_*})$, by properly adjusting the kernel ridge regularisation parameter at each stage of the dynamics. 

\paragraph{Limitations and Future Work.}

We only addressed the learning problem in continuous time, focusing on the population limit, assuming a true multi-index target (without noise), and using an isotropic Hermite kernel. We are hence neglecting all approximation and statistical aspects of the learning problem, and studying `only' its optimization aspects.  
In order to obtain non-asymptotic guarantees, it is thus necessary to introduce discretization both in time, in samples, and in `neurons' (through the random feature expansion described in Section \ref{par:random_features}), as well as consider the noisy setting where the target might be well approximated by a multi-index model. While the random feature discretization should follow from standard concentration arguments, and only exhibit polynomial dependencies in the ambient dimension (as in \cite{bietti2022learning}), the remaining discretization aspects are technically challenging and deserve further investigation, and our initial arguments from Section \ref{subsec:sample_complexity} should be formalized. 

Next, our techniques fundamentally exploit the Gaussian distribution. Whether the benign optimization dynamics we established may be carried over to other data distributions or not is an intriguing question. 
Finally, the connections between our model and neural networks are currently a bit `academic', in the sense that our neural network architecture is tailored to the multi-index class: its first-layer weights consist of a low-rank orthogonal matrix, whereas its second-layer activations are given by (frozen) random Hermite features. Closing the gap with more standard neural architectures, such as those studied, e.g in \cite{glasgow2023sgd} for the XOR problem, is a tantalizing question for future work. 

%% file: Appendix.tex
\appendix

\begin{center}
{ \scshape \Large  Organization of the Appendix}
\end{center}

\vspace{0.5cm}

\ref{sec:app_harmonic_analysis}. \textit{Harmonic Analysis for Gaussian: proofs of Section \ref{sec:harmonic_analysis}}

\begin{adjustwidth}{0.9cm}{2.5cm}
\noindent Where the proofs of Section~\ref{sec:harmonic_analysis} are written. This includes the basic properties of the averaging operator (Subsection~\ref{subsecapp:basic_A}), change of coordinates formula for the Hermite polynomials (Subsection~\ref{subsecapp:change_of_coordinates}) and finally proving further properties on $\Av$ (Subsection~\ref{subsecapp:further_prop}). 
\end{adjustwidth}

\vspace{0.25cm}

\ref{appsec:critical_points}. \textit{Critical point of the Grassmannian loss: proof of Theorem~\ref{thm:critical_points}}

\begin{adjustwidth}{0.9cm}{2.5cm}
\noindent Where the proof of Theorem~\ref{thm:critical_points}, on the critical points on the Grassmaniann objective, is written.
\end{adjustwidth}

\vspace{0.25cm}

\ref{appsec:gradient_flow}. \textit{Grassmaniann Gradient flow equations}

\begin{adjustwidth}{0.9cm}{2.5cm}
\noindent Where the gradient flow equations for the joint learning model are given. Gradient flow is expressed with respect to the $G$ variable (Subsection~\ref{subsecapp:gradient_flow_G}), and then with respect to the summary statistics $\Lambda, V$ (Subsection~\ref{subsecapp:gradient_flow_lambda_V}).  
\end{adjustwidth}

\vspace{0.25cm}

\ref{appsec:description_dynamics}. \textit{Description of the saddle dynamics: proof of Theorem~\ref{thm:coarse-grained}}

\begin{adjustwidth}{0.9cm}{2.5cm}
\noindent Where we give the proof of the saddle dynamics of the gradient flow. First, we upper bound the number eigenvalues' that have escaped the initialiaztion at the correct time scale: proof of the lemma~\ref{lem:sequential_escape} (Subsection~\ref{sec:prooflemmaseqescape}). Then, we show that the upper bound of possible eigenvalue espace is tight, proving Lemma~\ref{lem:lowergroup} (Subsection~\ref{sec:prooflemmalowergroup}).
\end{adjustwidth}

\vspace{0.25cm}

\ref{appsec:planted_model}. \textit{The Planted Model}

\begin{adjustwidth}{0.9cm}{2.5cm}
\noindent Where we give an explicit form of the dynamical equations of the Stiefel gradient flow on the summary statistics $(U, V, \Lambda)$ (Subsection~\ref{subsecapp:stiefel_equations}). The study of the proof of Theorem~\ref{thm:planted_radial_case} is provided (Subsection~\ref{app:radial_case}). Finally, the proof of Theorem~\ref{thm:planted_failure} on the failure case is given (Subsection~\ref{sec:app_failure_stiefel}).
\end{adjustwidth}

\vspace{0.25cm}

\ref{secapp:grassmann_manifold}. \textit{Some useful results on the Special manifolds}

\begin{adjustwidth}{0.9cm}{2.5cm}
\noindent Where we provide statistical results on the distribution of the summary statistics given the uniform distribution on the Special manifoldds (Grassmannian and Stiefel, see Subsection~\ref{subsecapp:statistics_special}), and recall standard result on their structure (Subsection~\ref{subsecapp:representation_special} and~\ref{appsec:gradients}).
\end{adjustwidth}

\vspace{0.25cm}

\ref{secapp:ODE_technical}. \textit{Some technical results for ODEs}

\begin{adjustwidth}{0.9cm}{2.5cm}
\noindent Where we prove some technical result on differential inequalities \textit{à la Gronwall}.  
\end{adjustwidth}

\section{Proofs of Section \ref{sec:harmonic_analysis}}
\label{sec:app_harmonic_analysis}
\input{Appendix/App_harmonic_analysis}

\section{Critical point of the Grassmannian loss: proof of Theorem~\ref{thm:critical_points}}
\label{appsec:critical_points}
\input{Appendix/App_critical_points}

\section{Gradient flow}
\label{appsec:gradient_flow}
\input{Appendix/App_gradient_flow}

\section{Description of the saddle dynamics: proof of Theorem~\ref{thm:coarse-grained}}
\label{appsec:description_dynamics}
\input{Appendix/App_dynamics_description}

\section{The planted model}
\label{appsec:planted_model}
\input{Appendix/App_planted_case}

\section{Useful results on the special manifolds}
\label{secapp:grassmann_manifold}
\input{Appendix/App_Grassmannians}

\section{Auxiliary results for ODEs}
\label{secapp:ODE_technical}
\input{Appendix/App_ODE_technical}

%% file: Appendix/App_harmonic_analysis.tex
We begin this section by proving the basic properties of the averaging operator $\Av$.

\subsection{Basic properties of \texorpdfstring{$\Av$}{A}}
\label{subsecapp:basic_A}

For the sake of clarity, we decide to rewrite explicitly and prove all the results below the restated lemmata and propositions. We recall first all the definitions of the operators: for $M \in \R^{q \times r}$, the averaging operators and the linear change of variable write:
\begin{align*}
\Av_M f(z) &=\mathbb{E}_{y} \left[f\left( M^\top z + (I_r -  M^\top M)^{\frac{1}{2}}y\right)\right] ~, \\ 
\P_M f(z) &= f (M^\top z) ~.
\end{align*}
\subsubsection{First properties}
\label{subsubsecapp:first_properties}
We restate and proof successively the different lemmata.

\lemmamultiplicativeproperties*
\begin{proof}
    For the point (i), take $M \in \R^{r \times p}$, $N \in \R^{p \times q}$ such that $\|M\|, \|N\| \leq 1$. For all $f \in L^2_{\gamma_r}$ and $z \in \R^q$, we have:
\begin{align*}
\Av_M \Av_N (f)(z) = \Av_M (\Av_N (f)(z)) &= \E_{y\sim \gamma_p} \left[ \Av_N (f)(M^\top z + (I_r - M^\top M)^{\frac{1}{2}} y) \right] \\
&= \E_{y\sim \gamma_p,y'\sim \gamma_r} \left[ f\left(N^\top (M^\top z + (I_r - M^\top M)^{\frac{1}{2}} y) + (I_r - N^\top N)^{\frac{1}{2}} y'\right) \right] \\
&= \E_{y,y'} \left[ f((M N)^\top z + N^\top(I_r -  M^\top M)^{\frac{1}{2}} y + (I_r - N^\top N)^{\frac{1}{2}} y') \right].
\end{align*}
And $u:=N^\top(I_r - M^\top M)^{\frac{1}{2}} y + (I_r -  N^\top N)^{\frac{1}{2}} y'$ is a centred Gaussian of $\R^r$ of covariance 
\begin{align*}
\E[uu^\top] = N^\top (I_r - M^\top M) N +  (I_r - N^\top N) = I_r -  (MN)^\top MN.
\end{align*}
 Hence, finally, 
 $$\Av_M \Av_N (f)(z) = \E_{u} \left[ ((M N)^\top z + (I_r -(MN)^\top MN )^{\frac{1}{2}} u \right] = \Av_{MN} (f)(z).$$ 
 The second part on the change of variable operator is easier:  $$ \P_M \P_N (f)(z) = \P_M(f(N^\top z)) = f(N^\top M^\top z) = \P_{MN} (f)(z). $$
 Finally, the proof of (ii) is trivial as for $M \in \S(r,q)$, $I_r - M^\top M = 0$. 
\end{proof}
We now prove the lemma calculating the adjoint of $\Av$. 
\lemmaadjoint*
\begin{proof}
Let $M \in  \R^{d \times r}$ and $f\in L^2_{\gamma_r},\  g \in L^2_{\gamma_d}$ test functions, we have 
$$\langle \Av_M f, g \rangle_{\gamma_d} = \E_{y\sim \gamma_d ,z\sim \gamma_r} \left[ f(M^\top y + (I_r - M^\top  M)^{\frac{1}{2}} z ) g(y) \right].$$ 
Let $u :=M^\top y + (I_r - M^\top  M)^{\frac{1}{2}} z \in \R^r$, then the law of $(u,y)$ is the joint Gaussian law of covariance matrix: 
$$\Sigma= \begin{bmatrix}I_r & M \\ M^\top & I_q\end{bmatrix}.$$ 
Hence,
 \begin{align*}
 \langle \Av_M f, g \rangle_{\gamma_d} = \E_{(u,y)\sim\mathcal{N}(0,\Sigma)}[f(u) g(y)] = \E_{(y,u)\sim\mathcal{N}(0,\Sigma^\top)}[g(y)f(u)] =  \langle f, \Av_{M^\top}g \rangle_{\gamma_r},
 \end{align*}
 and this concludes the proof of the first part. Noting that $\P$ and $\Av$ are the same operators when acting on the Stiefel manifold, the adjoint property of $\P$ follows instantly in this case.    
\end{proof}
%
%

\subsubsection{Interaction of the averaging operator with the tensorized Hermite basis}
\label{subsubsecapp:hermite_averagin}

\proavediagonal*

\begin{proof}
    Let $\Lambda \in \mathcal{D}$, and $\beta \in \N^q$, as everything is separable, and the $(h_{\beta_j})_{1\leq j\leq q}$ are eigenfunctions of the one-dimensional counterpart $\Av_\lambda$, for $\lambda \in \R$, we have the calculation:
 \begin{align*}
\Av_\Lambda H_\beta (y) &= \E_{(z_1,\hdots, z_q)}\left[ H_\beta\left(\lambda_1 y_1 + \sqrt{1 - \lambda_1^2} z_1, \hdots, \lambda_r y_r+ \sqrt{1 - \lambda_q^2} z_r\right) \right] \\
&= \E_{(z_1,\hdots, z_q)}\left[ \prod_{j=1}^q h_{\beta_j}(\lambda_j y_j + \sqrt{1 - \lambda_j^2} z_j) \right] \\
&= \prod_{j=1}^q \E_{Z}\left[h_{\beta_j}(\lambda_j y_j + \sqrt{1 - \lambda_j^2} Z) \right]\\
&= \prod_{j=1}^q P_{\lambda_j}(h_{\beta_j})(y_j)\\
&= \prod_{j=1}^q \lambda_j^{\beta_j}h_{\beta_j}(y_j)\\
&= \prod_{j=1}^q \lambda_j^{\beta_j} \prod_{j=1}^r h_{\beta_j}(y_j)\\
&= \left(\prod_{j=1}^q \lambda_j^{\beta_j}\right) H_\beta (y).
 \end{align*}
This concludes the proof of the lemma.
\end{proof}

\subsection{Change of coordinates for Hermite polynomials}
\label{subsecapp:change_of_coordinates}

\subsubsection{Change of coordinates in the tensorized Hermite basis}
\begin{lemma}[Unitary change of variables]
\label{lem:hermitechange}
Let $U \in \mathcal{O}_q$ 
Then for any $\beta, \gamma \in \mathbb{N}^q$, 
\begin{equation}
\label{eq:vanilla_change}
       \langle \P_U H_\gamma, H_\beta \rangle = \delta_{|\beta|=|\gamma|} \cdot \sum_{{\Tau} \in \Pi(\beta, \gamma)} Q({ \Tau}; \beta, \gamma)^{1/2} \left(\prod_{i,j=1}^r U_{i,j}^{\Tau_{i,j}}\right) ~,
\end{equation}   
where $\Pi(\beta, \gamma) = \mathbb{N}^{q \times q} \cap \{A \in \mathbb{R}^{q \times q}; \, A{\bf 1}=\gamma ,\, {\bf 1}^\top A = \beta\} $ 
   are the lattice points of the transportation polytope and 
   $$Q({\Tau}; \beta, \gamma) = \prod_{j=1}^q \binom{\beta_j}{\Tau_j} \binom{\gamma_j}{\Tau^j}~,$$
    where $\Tau_j$ and $\Tau^j$ is respectively the $j$-th row and column of $\Tau$. 
\end{lemma}

The proof is based on carefully book-keeping the effect of rotations, sequentially across coordinates.
We will use elementary properties of Hermite polynomials, summarized in the next two lemmas.

\begin{lemma}[Additive and Multiplicative properties of Hermite polynomials]
\label{lem:hermsum_mult}
\begin{equation}
\label{eq:hermite_sum0}
 \sqrt{s!} h_s(x+y) =  2^{-s/2} \sum_{j=0}^s \binom{s}{j} \sqrt{j!} \sqrt{(s-j)!} h_j(\sqrt{2} x) h_{s-j}(\sqrt{2}y)~,\text{ and }    
\end{equation}
\begin{equation}
\label{eq:hermite_mult0}
\sqrt{s!} h_s (\gamma x) = \sum_{i=0}^{\lfloor s/2 \rfloor} \sqrt{(s-2i)!}\gamma^{s-2i} 2^{-i} (\gamma^2 -1)^i \binom{s}{2i} \frac{(2i)!}{i!} h_{s-2i}(x)~.    
\end{equation}
\end{lemma}

\begin{lemma}[Product of Hermite polynomials \cite{luo2006wiener}--Lemma 2.2]
\label{lem:prodherm}
For any nonnegative integers $l$ and $m$, we have 
\begin{equation}
    h_l(x) h_m(x) = \sum_{p \leq \min(l,m)} B(l,m,p) h_{l+m-2p}(x)~,
\end{equation}
where $B(l,m,p) = \sqrt{\binom{l}{p} \binom{m}{p}\binom{l+m-2p}{l-p}}$.
\end{lemma}

\begin{lemma}[Degree Homogeneity]
\label{claim:deghomv2}
If $U \in \mathcal{O}_q$, then 
\begin{equation}
    \langle \mathsf{P}_U H_\alpha, H_\beta \rangle \neq 0 \Rightarrow |\alpha|=|\beta|~.
\end{equation}    
\end{lemma}
\begin{proof}[Proof of Lemma \ref{claim:deghomv2}]
    Suppose first that $|\alpha| < |\beta|$. From Lemmas \ref{lem:hermsum_mult} and \ref{lem:prodherm} we verify that the LHS is in the span of (tensorised) Hermite polynomials of degree at most $|\alpha|$, therefore $\langle P_U H_\alpha, H_\beta \rangle = 0$. 
    Suppose now that $|\alpha| > |\beta|$. Then observe that 
    $\langle P_U H_\alpha, H_\beta \rangle = \langle  H_\alpha, P_{U}^* H_\beta \rangle = \langle  H_\alpha, P_{U^\top} H_\beta \rangle$, so we can reduce to the previous case. 
\end{proof}

We write $H_\beta(z)=H_\beta(z_1, \ldots z_r ) = h_{\beta_1}(z_1) H_{\bar{\beta}}(\bar{z})$.  
\begin{claim}[Induction Step]
\label{claim:basichermsingle_v3}
Assume $|\beta| = |\gamma|$. Then 
\begin{equation}
\label{eq:main_induc}
  \langle P_U H_\gamma, H_\beta \rangle =  2^{-(|\gamma|-\beta_1)/2} \sum_{\tau \in D(\beta_1;\gamma)} q(\tau;\beta_1, \gamma)^{1/2} \left(\prod_{j=1}^r U_{1,j}^{\tau_j}\right) \langle P_{\sqrt{2} \bar{U}}H_{{\bar{\gamma}-\bar\tau}}, H_{\bar{\beta}} \rangle~,    
\end{equation}
where for $l \in \mathbb{N}$ and $\gamma \in \mathbb{N}^r$, $D(l,\gamma):=\left\{ \tau \in \mathbb{N}^r;\, \sum_j \tau_j = l;\, \tau_j \leq \gamma_j \,\forall j\in[r]\right\}~,$  and $\displaystyle q(\tau, l, \gamma) := \binom{\beta_1}{\tau}\prod_{j=1}^r \binom{\gamma_j}{\tau_j}~.$
\end{claim}
\begin{proof}[Proof of Claim \ref{claim:basichermsingle_v3}]

From (\ref{eq:hermite_sum0}), we have 
\begin{align}
\label{eq:bopp}
    P_U H_\gamma (z) &= H_\gamma(U z) = \prod_{j=1}^r h_{\gamma_j}(\langle z, U_j \rangle) \nonumber \\
    &= \prod_{j=1}^r h_{\gamma_j}(z_1 U_{1,j} + \langle \bar{z}, \bar{U}_j \rangle) \nonumber \\
    &= 2^{-|\gamma|/2} \prod_{j=1}^r \left(\sum_{\tau_j=0}^{\gamma_j} {\binom{\gamma_j}{\tau_j}}^{1/2} h_{\tau_j}(\sqrt{2} U_{1,j} z_1)h_{\gamma_j - \tau_j}(\sqrt{2} \langle \bar{z}, \bar{U}_j\rangle) \right)   \nonumber \\
    &= 2^{-|\gamma|/2} \sum_{\tau_1, \ldots, \tau_r=0}^{\gamma_1, \ldots, \gamma_r }\prod_{j=1}^r  {\binom{\gamma_j}{\tau_j}}^{1/2} h_{\tau_j}(\sqrt{2} U_{1,j} z_1)h_{\gamma_j - \tau_j}(\sqrt{2} \langle \bar{z}, \bar{U}_j\rangle) ~.
\end{align}

Let $H_\beta(z) = h_{\beta_1}(z_1) H_{\bar{\beta}}(\bar{z})$. From (\ref{eq:bopp}) we write 
\begin{align}
\label{eq:bopp2}
   &  \langle P_U H_\gamma, H_\beta \rangle  = \nonumber \\
   & = 2^{-|\gamma|/2} \sum_{\tau_1, \ldots, i_r=0}^{\gamma_1, \ldots, \gamma_r }{\prod_{j=1}^r {\binom{\gamma_j}{\tau_j}}}^{1/2} \left \langle \prod_{j=1}^r  h_{\tau_j}(\sqrt{2} U_{1,j} z_1), h_{\beta_1}(z_1) \right\rangle \cdot \left \langle \prod_{j=1}^r h_{\gamma_j - \tau_j}(\sqrt{2} \langle \bar{z}, \bar{U}_j\rangle) , H_{\bar{\beta}}(\bar{z}) \right \rangle
\end{align}
From Lemmas \ref{lem:hermsum_mult} and \ref{lem:prodherm}, we verify that $\prod_{j=1}^r  h_{\tau_j}(\sqrt{2} U_{1,j} z_1)$ lies in the span of Hermite polynomials of degree at most $\sum_j \tau_j$, and analogously that $\prod_{j=1}^r h_{\gamma_j - \tau_j}(\sqrt{2} \langle \bar{z}, \bar{U}_j\rangle)$ lies in the span of (tensorised) Hermite polynomials of degree at most $\sum_j (\gamma_j - \tau_j) = |\gamma| - \sum_j \tau_j = |\beta| - \sum_j \tau_j$. As a result, the only non-zero terms necessarily satisfy $\sum_j \tau_j = \beta_1$, ie, $\tau \in D(\beta_1, \gamma)$. 

Using Lemmas \ref{lem:hermsum_mult} and applying \ref{lem:prodherm} repeatedly, when collecting the terms of degree $\beta_1$ we finally obtain 
\begin{align*}
    \left \langle \prod_{j=1}^r  h_{\tau_j}(\sqrt{2} U_{1,j} z_1), h_{\beta_1}(z_1) \right\rangle &= 2^{\beta_1/2} \left(\prod_{j=1}^r U_{1,j}^{\tau_j}\right) \left[ \binom{\tau_1 + \tau_2}{\tau_1} \binom{\tau_1+\tau_2+\tau_3}{\tau_1+\tau_2} \cdots \binom{\beta_1}{\beta_1-\tau_r}\right]^{1/2} \\
   &=  2^{\beta_1/2} \left(\prod_{j=1}^r U_{1,j}^{\tau_j}\right) {\binom{\beta_1}{\tau}}^{1/2}~.
\end{align*}
Plugging into (\ref{eq:bopp2}), and noting that $\prod_{j=1}^r h_{\gamma_j - \tau_j}(\sqrt{2} \langle \bar{z}, \bar{U}_j\rangle) = P_{\sqrt{2} \bar{U}} H_{\bar{\gamma}-\bar\tau}(\bar{z})$, we obtain (\ref{eq:main_induc}).
\end{proof}

\begin{proof}[Proof of Lemma \ref{lem:hermitechange}]
    Applying repeatedly Claim \ref{claim:basichermsingle_v3}, we obtain 
\begin{align}
    \langle P_U H_\gamma, H_\beta \rangle &= \sum_{ \tau_1 \in D(\beta_1, \gamma)}\sum_{\tau_2 \in D(\beta_2, \gamma-\tau_1)} \dots \sum_{\tau_r \in D(\beta_r, \gamma-\sum_{j<r}\tau_{j})} \left(\prod_{j=1}^r q\left(\tau_j; \beta_j, \gamma-\sum_{j'<j} \tau_{j'}\right) \right)^{1/2}\left(\prod_{i,j=1}^r U_{i,j}^{\tau_{i,j}} \right)~.
\end{align}
Observe that the integer matrix $\Tau:=[\tau_1; \tau_2; \dots; \tau_r]$ satisfies $\Tau {\bf 1} = \gamma$ and ${\bf 1}^\top \Tau = \beta$, 
thus $\Tau \in \Pi(\beta, \gamma)$. 
Reciprocally, if $\Tau'$ is any matrix in $\Pi(\beta, \gamma)$, then its rows $\tau'_1, \ldots, \tau'_r$ satisfy $\tau'_j \in D(\beta_j, \gamma - \sum_{j'<j} \tau'_{j'})$. 

Finally, denote by $\Tau_j=\tau_j$ the rows of $\Tau$, and by $\Tau^j$ its columns. We have
\begin{align*}
\prod_{j=1}^r q\left(\tau_j; \beta_j, \gamma-\sum_{j'<j} \tau_{j'}\right) &= 
    q(\Tau_1; \beta_1, \gamma) q(\Tau_2; \beta_2, \gamma-\Tau_1) \cdots q(\Tau_r, \beta_r, \gamma - \sum_{j'<r} \Tau_{j'} ) \\
    &= \binom{\beta_1}{\Tau_1}\left(\prod_{j=1}^r \binom{\gamma_j}{\Tau_{1,j}}\right) \binom{\beta_2}{\Tau_2}\left(\prod_{j=1}^r \binom{\gamma_j-\Tau_{1,j}}{\Tau_{2,j}}\right) \cdots \\
&= \prod_{j=1}^r \binom{\beta_j}{\Tau_j} \binom{\gamma_j}{\Tau^j}~\\
&= Q(\Tau; \beta, \gamma)~,
\end{align*}
and the proof is complete.
\end{proof}
\begin{remark}
\label{rem:hermite_change_general}
    Notice that the proof of Lemma \ref{lem:hermitechange} does not require that $U$ is unitary in the case $|\beta|=|\gamma|$. In other words, Eq. \eqref{eq:vanilla_change} remains valid when $U \in \mathcal{O}_q$ is replaced by $U \in \mathbb{R}^{q \times q}$ and $|\beta|=|\gamma|$. However, when $|\beta|\neq |\gamma|$, we have that $\langle \mathsf{P}_U H_\beta, H_\gamma \rangle =0$ for $U \in \mathcal{O}_q$, but this is  not true for general $U \in \mathbb{R}^{q \times q}$. 
\end{remark}

\subsubsection{Link between the change of coordinates and the averaging operator}
\label{subsubsecapp:link_aveargin_hermite}
\begin{proposition}[Local Representation of $\Av$]
\label{prop:local_representation}
For any $M \in \mathbb{R}^{q \times q}$ and $\beta, \gamma \in \mathbb{N}^q$ such that $|\beta|=|\gamma|$, it holds
\begin{align}
\langle \Av_M H_\beta, H_\gamma \rangle = \langle \P_M H_\beta, H_\gamma \rangle~.
\end{align}
\end{proposition}
\begin{proof}
Let us consider the SVD of $M$ written as $M = U \Lambda V^\top$. Denote $Z = U \Lambda$. Since $\Av_{Z} =\Av_{\Lambda} \P_{U} $, from Lemma \ref{lem:hermitechange} and Remark \ref{rem:hermite_change_general} and the definition of $\Av_\Lambda$ we deduce that for any $\beta, \gamma$ s.t. $|\beta| = |\gamma|$, 
    \begin{align*}
       \langle  \Av_{Z} H_\gamma, H_\beta \rangle & = \langle  \Av_{\Lambda} \P_{U} H_\gamma, H_\beta \rangle \\
       & = \langle \P_{U} H_\gamma,  \Av_{\Lambda} H_\beta \rangle \\
       &= \left(\prod_{i=1}^q \lambda_i^{\beta_i} \right) \left[ \sum_{{\Tau} \in \Pi(\beta, \gamma)} Q({ \Tau}; \beta, \gamma)^{1/2} \left(\prod_{i,j=1}^r U_{i,j}^{\Tau_{i,j}}\right)\right] \\
       &= \left(\prod_{i=1}^q \lambda_i^{\beta_i} \right) \left[\sum_{{\Tau} \in \Pi(\beta, \gamma)} Q({ \Tau}; \beta, \gamma)^{1/2} \left(\prod_{i,j=1}^r Z_{i,j}^{\Tau_{i,j}}\right) \left(\prod_{i,j=1}^q \lambda_j^{-\Tau_{i,j}} \right)\right] \\
       &= \left(\prod_{i=1}^q \lambda_i^{\beta_i} \right) \left[\sum_{{\Tau} \in \Pi(\beta, \gamma)} Q({ \Tau}; \beta, \gamma)^{1/2} \left(\prod_{i,j=1}^r Z_{i,j}^{\Tau_{i,j}}\right) \left(\prod_{j=1}^q \lambda_j^{-\sum_i \Tau_{i,j}} \right) \right] \\
       &= \left(\prod_{i=1}^q \lambda_i^{\beta_i} \right) \left[\sum_{{\Tau} \in \Pi(\beta, \gamma)} Q({ \Tau}; \beta, \gamma)^{1/2} \left(\prod_{i,j=1}^r Z_{i,j}^{\Tau_{i,j}}\right) \left(\prod_{i=1}^q \lambda_i^{-\beta_i} \right) \right] \\
       &= \sum_{{\Tau} \in \Pi(\beta, \gamma)} Q({ \Tau}; \beta, \gamma)^{1/2} \left(\prod_{i,j=1}^r Z_{i,j}^{\Tau_{i,j}}\right) \\
       &= \langle \P_{Z} H_\gamma, H_\beta \rangle~.
    \end{align*}
Now, by decomposing $\Av_{M} = \P_{V^\top} \Av_{Z}$, hence $\langle \Av_{M} H_\gamma, H_\beta \rangle =  \langle  \Av_{Z} H_\gamma, \P_{V} H_\beta \rangle~.$
But there exists $C_V$, such that $\P_{V} H_\beta = \sum_{|\beta'|=|\beta|} C_{V}(\beta, \beta') H_{\beta'},$ therefore, by using the previous equality, we obtain
\begin{align*}
   \langle \Av_{M} H_\gamma, H_\beta \rangle & = \langle \Av_{Z} H_\gamma, P_{V} H_\beta \rangle \\
   &= \sum_{|\beta'|=|\beta|} C_{V}(\beta, \beta') \langle \Av_{Z} H_\gamma, H_{\beta'} \rangle \\
   &= \sum_{|\beta'|=|\beta|} C_{V}(\beta, \beta') \langle \P_{Z} H_\gamma, H_{\beta'} \rangle \\
    &= \left\langle {\P}_{Z} H_\gamma, \sum_{|\beta'|=|\beta|} C_{V}(\beta, \beta')  H_{\beta'} \right \rangle \\
    &= \langle {\P}_{Z} H_\gamma, \P_{V} H_\beta \rangle \\
    &= \langle \P_{V^\top} {\P}_{Z} H_\gamma,  H_\beta \rangle \\
    &= \langle \P_M H_\gamma,  H_\beta \rangle~,
\end{align*}
as claimed. 
\end{proof}

\subsection{Further properties of the averaging operator \texorpdfstring{$\Av$}{A}}
\label{subsecapp:further_prop}

\subsubsection{Properties on the intrinsic dimension and energy}
\label{subsubsecapp:properties_idim}

\propintrinsicaveraging*

\begin{proof}
    Point \textit{(i)} is a direct consequence of the definition. 
    
    We now prove \textit{(ii)}. Let us first assume that $r=q$, so $W=I_q$. Notice that if $f \in H^1$, then by the previous argument we have that $\mathsf{d}(\mathsf{A}_Y f)=\text{rank}(Y^\top \mathsf{G}_f Y)= \text{rank}(Y)$. We now give an alternative proof for general $f$. 
    Assume towards contradiction that $\mathsf{d}(\mathsf{A}_Y f) < q$, 
    so $\mathsf{A}_Y f = \mathsf{P}_{Q} g$ with $Q \in \mathcal{G}(\mathsf{d}(\mathsf{A}_\Lambda f),q)$. Since $Y$ is invertible, we have $f = \overline{\mathsf{A}}_{Q Y^{-1}} g$. Observe that $\tilde{Q}=Q Y^{-1}$ has rank $\mathsf{d}(\mathsf{A}_\Lambda f) < q$. 
    Let $\tilde{Q} = \tilde{U} \tilde{\Lambda} \tilde{V}^\top$ be its SVD. 
    Expressing $g$ in the coordinate system given by left eigenvectors, we have 
    $$f = \sum_{\beta; |\beta|_{\mathsf{d}(\mathsf{A}_\Lambda f)+1:q}=0} \left(\prod_{i=1}^q \tilde{\lambda_i}^{\beta_i}\right) \langle g, H_\beta(\tilde{V}) \rangle H_\beta(\tilde{U})~.$$
    This shows that $\mathsf{d}(f) \leq \mathsf{d}(g) < q$, which is impossible. The case where $r < q$ is treated analogously, by replacing $Y$ with $W Y$.

    Let us finally prove \textit{(iii)}. Assume again wlog that $W = I_q$, 
    and let $g = \mathsf{A}_Y f$ as before. 
    Assume w.l.o.g. that $Y$ is diagonal, since from the SVD of $Y = U \Lambda V^\top$ we can replace $f$ by $\mathsf{P}_{U} f$ and $g$ by $\mathsf{P}_{V^\top} g$ and use the invariance property of Proposition \ref{prop:intrinsic_rotations}. 
    For $\|v\|=1$, let $\mathcal{E}(f; v) = \|f\|^2 - \| \mathsf{A}_{I - vv^\top} f \|^2 = \langle f, (I - \mathsf{A}_{I-vv^\top})f \rangle$, so that $\mathcal{E}(f) = \inf_{\|v\|=1} \mathcal{E}(f; v)$.  

    Consider $v = e_j$ for any $j \in [1,q]$. Then since $Y$ and $I-vv^\top$ commute, so do $\mathsf{A}_{Y}$ and $\mathsf{A}_{I - vv^\top}$, and therefore we have, for any $j$, 
    \begin{align}
    \label{eq:vail}
      \mathcal{E}(g; e_j) &= \langle g, (I - \mathsf{A}_{I - e_j e_j^\top}) g \rangle \nonumber \\
      &= \langle \mathsf{A}_{{Y}} f, (I - \mathsf{A}_{I - e_j e_j^\top}) \mathsf{A}_{{Y}} f \rangle  \nonumber \\
      &= \langle f, (I - \mathsf{A}_{I - e_j e_j^\top}) \mathsf{A}_{{Y} {Y}^\top} f \rangle \nonumber \\
      &= \langle (I - \mathsf{A}_{I - e_j e_j^\top})^{1/2} f, \mathsf{A}_{{Y} {Y}^\top} (I - \mathsf{A}_{I - e_j e_j^\top})^{1/2}f \rangle \nonumber \\
      &\geq \lambda_{\text{min}}(\mathsf{A}_{{Y}{Y}^\top})  \mathcal{E}(f; e_j) \nonumber \\
      &\geq \lambda_{\text{min}}(Y)^{2s} \mathcal{E}(f)~,      
    \end{align}

    Let $v_*=\arg\min_{\|v\|=1} \mathcal{E}(g; v)$. 
    Let us now relate $\mathcal{E}(g) = \mathcal{E}(g; v_*)$ with the previously computed $\mathcal{E}(g; e_j)$. 
    Since $\{e_j\}$ is an orthonormal basis, we have that 
    \begin{equation}
    \label{eq:vail2}
    \sup_j | v_* \cdot e_j | \geq 1/\sqrt{q}~.    
    \end{equation}

    Let $B_w$ be an orthogonal basis of $[w ; w^\perp]$. $\mathsf{A}_{I - w w^\top}$ is an orthogonal projection onto the subspace
    $$K_w := \text{span}\{ H_\beta(B_w); \beta_1=0; |\beta|_{2:q}=s \} \subset \mathcal{F}_R~,$$
    of dimension $\chi(q,R) - \chi(q-1, R)$. 
    By definition we have $\mathcal{E}(g) = \| P_{K_{v_*}^\perp} g\|^2$ and $\mathcal{E}(g; e_j) = \| P_{K_{e_j}^\perp} g\|^2$, and 
    thus 
    \begin{align}
    \label{eq:brings}
        \mathcal{E}(g) &\geq \sup_j \lambda_{\min}( K_{v_*}^\perp, K_{e_j}^\perp)^2 \mathcal{E}(g; e_j)~,
    \end{align}
    where $\lambda_{\min}(W,\tilde{W})$ is the smallest singular value of $W^\top \tilde{W}$. 
    We now claim the following:
    \begin{claim}
    \label{claim:angleharmonics}
        For any pair of unit vectors $v, w$, we have
            $\lambda_{\min}( (K_{v}^\perp)^\top K_{w}^\perp) = (v \cdot w)^s~.$
    \end{claim}

   Plugging Claim \ref{claim:angleharmonics} into (\ref{eq:vail}), (\ref{eq:vail2}) and (\ref{eq:brings}), we finally obtain  
    \begin{align}
        \mathcal{E}(g) & \geq q^{-s} \inf_j \mathcal{E}(g; e_j) \geq q^{-R} \lambda_{\text{min}}(Y)^{2s} \mathcal{E}(f)~.
    \end{align}
\end{proof}

\begin{proof}[Proof of Claim \ref{claim:angleharmonics}]
We recall the variational characterization of the largest angle between subspaces:
$$\lambda_{\min}(W, \tilde{W}) = \inf_{w \in W, \|w\|=1} \sup_{\tilde{w} \in \tilde{W}, \|\tilde{w}\|=1} w \cdot \tilde{w}~.$$
We first observe (e.g. \cite[Property 2.1.1]{ZhuKnyazev2013}) that for $W, \tilde{W} \in \mathcal{G}(p,q)$, it holds that 
\begin{equation}
\label{eq:cream}
\lambda_{\min}(W^\perp, \tilde{W}^\perp)=\lambda_{\min}(W, \tilde{W})~.    
\end{equation}

In $\mathcal{F}_R$, for $V, \tilde{V} \in \mathcal{G}(p,q)$, we have 
\begin{align}
\lambda_{\min}(\text{span}(\mathsf{A}_V), \text{span}(\mathsf{A}_{\tilde{V}})) &= \inf_{f \in \mathcal{F}_{R,p}, \|f\|=1} \sup_{\tilde{f} \in \mathcal{F}_{R,p}, \|\tilde{f}\|=1} \langle \mathsf{P}_V f, \mathsf{P}_{\tilde{V}} \tilde{f} \rangle \nonumber \\
&= \inf_{f \in \mathcal{F}_{s,p}, \|f\|=1} \langle \mathsf{A}_{V^\top \tilde{V}} f, {f} \rangle \nonumber \\
&= \lambda_{\min}(V^\top \tilde{V})^s~. 
\end{align}
Using (\ref{eq:cream}) twice, it follows that
\begin{align}
     \lambda_{\min}( (K_{v}^\perp)^\top K_{w}^\perp) &=  \lambda_{\min}( (K_{v})^\top K_{w}) \nonumber \\
     &= \lambda_{\min}( v^\perp, w^\perp)^{s} \nonumber \\
     &= \lambda_{\min}( v,  w)^{s} \nonumber \\
     &= (v \cdot w)^s~.
\end{align}

\end{proof}

\propintrinsicspectral*
\begin{proof}
Let $p = \mathsf{d}(f)$, and $\bar{f} \in L^2_{\gamma_{p}}$, $W \in \mathcal{G}(p,q)$ such that $f = \mathsf{P}_W \bar{f}$. 
Let $B$ be an orthonormal basis of $[W; W^\perp]$, and $(H_\beta)_\beta$ its associated Tensorized Hermite basis. Observe that, for any $k$,  
$$\mathsf{S}^k f = \sum_{\beta; |\beta|_{p+1:q}=0; |\beta|\leq k} \langle f, H_\beta\rangle H_\beta ~,$$
showing that $\mathsf{d}(\mathsf{S}^{k}(f)) \leq \mathsf{d}(f)$ and $W[\mathsf{S}^{k}(f)] \subseteq W[f]$. 
Moreover, observe that, for any basis $B$ as above, 
\begin{align}
    \mathcal{E}(f) &= \inf_{B} \sum_{\beta; |\beta|_{p+1:q}=0; \beta_p>0} \langle f, H_\beta \rangle^2 \nonumber \\
    &\geq \inf_{B} \sum_{\beta; |\beta|_{p+1:q}=0; \beta_p>0; |\beta|\leq k} \langle f, H_\beta \rangle^2 \nonumber \\ 
    &= \mathcal{E}(\mathsf{S}^k f)~.
\end{align}
Finally, observe that $\mathsf{S}^k(f) = \mathsf{S}^k( \mathsf{S}^{k'} f)$ for any $k \leq k'$, showing that $p_k = \mathsf{d}(\mathsf{S}^{k}(f))$ is a non-decreasing sequence in $\mathbb{N}$. Since we just showed that $p_k \leq p$ for all $k$, it is also bounded. By monotone convergence, we must have that $p = \lim_{k \to \infty} p_k = \sup p_k$, so there must exist $r$ such that $p_{r'} = p$ for all $r' \geq r$. 
\end{proof}

\subsubsection{Final general properties of $\Av$}
\label{subsubsubsecapp:final_properties}

\propcompactness*

\begin{proof}
Let us prove point (i). Let $M \in \R^{q \times r}$ such that $\|M\| \leq 1$. Moreover, let us fix $G = V \Lambda U^\top$ the reduction of $M$ with elements of $\Lambda$ sorted non increasingly. For any $f \in L^2_{\gamma_r}$, we have $\langle \Av_M f, f  \rangle = \langle \Av_\Lambda \P_{V^\top} f, \P_{U^\top} f\rangle$, and as $\|\P_{U^\top} f\|_{\gamma_q} = \|\P_{V^\top} f\|_{\gamma_r} \| f \|_{\gamma_r}$, finding the supremum of $\langle \Av_M f, f  \rangle$ on the sphere of $L^2_{\gamma_r}$ is similar to finding then one of $\langle \Av_\Lambda g, g\rangle$ on the sphere of $L^2_{\gamma_q}$. Let $\|g\| = 1$ decomposed as a sum of the tensorized Hermite polynomials,
\begin{align*}
    \langle \Av_\Lambda g, g\rangle = \sum_\beta \alpha_\beta^2 \prod_i \lambda_i^{\beta_i} \leq \lambda_1 \sum_\beta \alpha_\beta^2 = \lambda_1,
\end{align*}
with equality for $g(x_1, \dots, x_q) = h_1(x_1)$. Hence, $\|\Av_M\| =\|\Av_\Lambda\| = \lambda_1 = \|M\|$.

For the point (ii), thanks to the SVD of $A_M$, we have $\Av_M f = \sum_\beta \lambda_\beta \langle f, H_\beta(V) \rangle H_\beta(U)$. Now the Hilbert-Schmidt norm is equal to:
\begin{align*}
    \|\Av_M\|_{\mathrm{HS}}^2 &= \sum_\beta \|\Av_M H_\beta (V)\|^2 \\
    &= \sum_\beta \lambda_\beta^2 \\
    &= \sum_{\beta_1, \dots, \beta_q} \lambda_{1}^{2\beta_1} \dots \lambda_{q}^{2\beta_q} \\
    &= \sum_{\beta_1, \dots, \beta_q \in \N} \lambda_{1}^{2\beta_1} \dots \lambda_{q}^{2\beta_q} \\
    &= \left(\sum_{i \in \N} \lambda_{1}^{2i} \right) \dots \left(\sum_{i \in \N} \lambda_{q}^{2i} \right) \\
    &= \prod_{i = 1}^{q} \frac{1}{ 1 - \lambda_i^2}~,
\end{align*}
    which proves the result. 

    Finally for point (iii), the result follows from the fact that $ (\Av_M \circ \S^s) (f) = \sum_{|\beta| \leq s} \lambda_\beta \langle f, H_\beta(V) \rangle H_\beta(U)$, so that the smaller singular value corresponds to $\lambda_q^s$.
\end{proof}

%% file: Appendix/App_critical_points.tex
\begin{theorem}[Critical points of $L$]
The critical points of $L(W)$ in $\mathcal{G}(d,r)$ defined in \eqref{eq:critical_points} are given by 
\begin{equation}
    \mathrm{Crit}(L) = \left\{W \in \mathcal{G}(d,r)\,;\, \mathrm{Sp}(G_W) \in \Gamma_{\tau}(f) \text{ and } \|\mathsf{A}_{\mathrm{Sp}(G_W)}f\| = \|\mathsf{A}_{\mathrm{Jt}(G_W)}f\| \right\}~.
\end{equation}
Moreover:
\begin{enumerate}[label=(\roman*)]
    \item The only local maxima is hence global and corresponds to $W = W_*$. 
    \item All local minima are also global, given by $$\{W; \mathrm{Sp}(G_W) \in \Gamma_\tau^0(f) \text{ and } \|\mathsf{A}_{\mathrm{Sp}(G_W)}f\| = \| \mathsf{A}_{\mathrm{Jt}(G_W)}f\| \}~.$$
    \item Therefore, all the saddle points are given by 
    $$\bigcup_{\tau = 1}^{q-1} \left\{W ; \mathrm{Sp}(G_W) \in \Gamma_{\tau}(f)\setminus \Gamma_\tau^0(f) \text{ and } \|\mathsf{A}_{\mathrm{Sp}(G_W)}f\| = \|\mathsf{A}_{\mathrm{Jt}(G_W)}f\| \right\}~,$$
    where for each $ \tau \in \llbracket 1 , q-1 \rrbracket$, each set in the union has $\tau$ ascent directions and $ q - \tau$ descent ones.
\end{enumerate}
\end{theorem}
This result can be interpreted as follows. In the positive semi-definite cone section $\mathcal{C}_q$, there are special `corners', given by $\Gamma_{\tau}(f)$ (and whose eigenvalues are either $0$ or $1$). These corners are stratified according to the dimension of the subspace (or equivalently $\text{rank}(G_W) = \sum_i \lambda_i$), which plays a role similar to the index of the critical point -- although in general these saddles won't be strict. In that sense, it is interesting to note that the underparametrised landscape (corresponding to $k < q$) is in fact controlling the saddle point structure of the overparametrised problem ($r \geq q$). 

Hence, in the $(\lambda, V)$ space, the eigenvalues $\lambda$ belongs to the polytope $[0,1]^q$, and Theorem~\ref{thm:critical_points} states that the natural critical points are given by \textbf{all the vertices of this polytope}, that is $\lambda \in \{0,1\}^q$, while $V$ is to be a critical point of the energy in the corresponding Grassmann space.

\begin{proof}
    Let us denote for convenience 
    \begin{align*}
        \Omega &=  \mathrm{Crit}_{\mathcal{G}(r,d)}(L) ~,\\
        \Theta &= \left\{W \in \mathcal{G}(r,d)\, ;\, \mathrm{Sp}(G_W) \in \Gamma_{\tau}(f^*) \text{ and } \|\mathsf{A}_{\mathrm{Sp}(G_W)}f^*\| = \|\mathsf{A}_{\mathrm{Jt}(G_W)}f^*\|  \right\}~. 
    \end{align*}

\paragraph{\underline{First step: Inclusion $\mathbf{\Omega \subseteq \Theta}$}.}
We first establish that $\Omega \subseteq \Theta$ by proving that if $W \notin \Theta$ then $W \notin \Omega$, i.e. if either  $\mathrm{Sp}(G_W) \notin \Gamma_\tau(f^*)$, or $\|\mathsf{A}_{\mathrm{Sp}(G_W)}f^*\| < \|\mathsf{A}_{\mathrm{Jt}(G_W)}f^*\|$, then $W \notin \Omega$. \\ 

\noindent \underline{\textit{Case 1: $\mathrm{Sp}(G_W) \notin \Gamma_\tau(f^*)$}}. Hence, there exists a smooth curve $t \mapsto Z(t) \in \mathcal{G}(q,\tau)$, $t\in (-1,1)$, such that $Z(0) = \mathrm{Sp}(G_W)$ and $\frac{d}{dt} L_\tau(Z(t))\neq 0$. The aim is to build a smooth curve $t \mapsto W(t) \in \G(d,r)$ such that $W(0) = W$, and $\frac{d}{dt} L(W(t))\neq 0$. We build this curve via natural representations of the Grassmaniann manifold~\cite{bendokat2020grassmann}, recalled in Appendix~\ref{subsecapp:representation_special}. Indeed, thanks to Lemma~\ref{lem:grasm_representation}, there exists $t \mapsto Q(t) \in \mathcal{O}_q$ a smooth curve satisfying $Q(0)=I_q$, such that $Z(t) =  \mathrm{Sp}(G_W) Q(t)$. 
\begin{lemma}
Define $\bar{W}_* = [W_* ; W_*^\perp] \in \mathcal{O}_d$, where $W_*^\perp$ is an orthonormal complement of $W_*$, and
\begin{align*}
    W(t):= \bar{W}_* \begin{pmatrix} Q(t)^\top & 0 \\ 0 & I_{d-q} \end{pmatrix} \bar{W}_*^\top W ~,
\end{align*}
then we have that $  G_{W(t)} = Q(t)^\top G_{W} Q(t)$, and in particular $\mathrm{Sp}(G_{W(t)}) = \mathrm{Sp}(G_W) Q(t)$.
\end{lemma}
\begin{proof}
Indeed, let us calculate, 
\begin{align*}
G_{W(t)} &=  W_*^\top W(t) W(t)^\top W_* = W_*^\top \bar{W}_* \begin{pmatrix} Q(t)^\top & 0 \\ 0 & I_{d-q} \end{pmatrix} \bar{W}_*^\top W W^\top \bar{W}_* \begin{pmatrix} Q(t) & 0 \\ 0 & I_{d-q} \end{pmatrix} \bar{W}_*^\top W_*,
\end{align*}
and as $\bar{W}_*^\top W_* = [I_q, 0_{d-q}]$, we have that 
\begin{align*}
G_{W(t)} &= [Q(t)^\top, 0_{d-q}] \bar{W}_*^\top W W^\top \bar{W}_* \begin{pmatrix} Q(t)  \\ 0_{d-q}  \end{pmatrix} = Q(t)^\top W_*^\top W W^\top W_* Q(t) = Q(t)^\top G_{W} Q(t), 
\end{align*}
that is implying that $\mathrm{Sp}(G_{W(t)}) = \mathrm{Sp}(G_W) Q(t)$. 
\end{proof}

We conclude by noting that $L(W(t)) = L_\tau(Z(t))$, and therefore we built a smooth curve such that $W(0) = 0$ and $\frac{d}{dt} L(W(t))  =\frac{d}{dt} L_\tau(Z(t)) \neq 0$. Hence, $W \notin \Omega$.
\\ 

\noindent \underline{\textit{Case 2: $\|\mathsf{A}_{\mathrm{Sp}(G_W)}f^*\| < \|\mathsf{A}_{\mathrm{Jt}(G_W)}f^*\|$}}. Define the sets $A_W = \{ j;  \lambda_j = 1\}$ and $B_W = \{ j; 0< \lambda_j < 1\}$. For $\beta \in \mathbb{N}^q$ and $A \subseteq [q]$, we denote by $|\beta|_A:= \sum_{j \in A} \beta_j$. The energy deficiency assumed in this case is equivalent to 
\begin{align}
    \label{eq:wash2}
        0 & < \sum_{\substack{\beta; \text{supp}(\beta)\subseteq A_W \cup B_W \\ |\text{supp}(\beta) \cap B_W| > 0}} \langle \mathsf{P}_{V} f^*, H_\beta \rangle^2 = \sum_{k=1}^\infty \sum_{\substack{\beta; \text{supp}(\beta)\subseteq A_W \cup B_W \\ |\beta|_{B_W} =k}} \langle \mathsf{P}_{V} f^*, H_\beta \rangle^2  : = \sum_{k=1}^\infty \eta_k.
\end{align}
    Let $k^* = \min\{ k; \eta_k >0\}$. 

From its SVD, $W_*^\top W = V \Lambda U^\top$, and it follows that $\bar{U} =  W U =[\bar{u}_i]_{i\leq r}\in \mathcal{S}(d,r)$ and $\bar{V} =  W_* V =[\bar{v}_j]_{j \leq q} \in \mathcal{S}(d,q)$ are a collection of $r + q$ vectors in $\mathbb{R}^d$ satisfying the orthogonality relationships
\begin{equation}
    \label{eq:green}
    \langle \bar{u}_i, \bar{u}_{i'} \rangle = \delta_{i-i'}~,~\langle \bar{v}_j, \bar{v}_{j'} \rangle = \delta_{j-j'}~,~\langle \bar{u}_i, \bar{v}_{j} \rangle = \lambda_j \delta_{i-j}~.    
\end{equation}
For each $j \in B_W$, let us consider the two-dimensional subspace spanned by $\bar{u}_j$ and $\bar{v}_j$, and $\bar{z}_j \in \text{span}(\bar{u}_j,\bar{v}_j)$ a vector orthogonal to $\bar{v}_j$. We consider $\bar{u}_j(t) := \cos(\pi \alpha_j t) \bar{z}_j + \sin(\pi \alpha_j t) \bar{v}_j$, with $\alpha_j = \frac{2\arcsin(\lambda_j)}{\pi}$ satisfying $0 < \alpha_j < 1$ for all $j\in B_W$, since $0< \lambda_j < 1$.  We then define
\begin{equation}
\label{eq:path_rotation}
    W(t):= \text{span}( \bar{u}_j;\,j\notin B_W;\, \bar{u}_j(t)\,;\, j \in B_W )~.    
\end{equation}
We verify that for $j \in B_W$, $\bar{u}_j = \bar{u}_j(\frac12)$, and thus $W = W(\frac12)$. From (\ref{eq:green}), the SVD decomposition of $M(t) = W_*^\top W(t) $ is given by $M(t) = V \Lambda(t) U^\top$, with 
    $\lambda_j(t) = \lambda_j$ for $j \notin B_W$, and $\lambda_j(t) = \sin(\pi \alpha_j t)$ for $j\in B_W$. 

\begin{lemma}
   Let the loss $\bar{L}(t):= L(W(t))$, where $W(t)$ is defined in equation~\eqref{eq:path_rotation}, then we have that $\bar{L}'\left(\frac12\right) > 0$.
\end{lemma}
\begin{proof}
Indeed, let us now evaluate the loss $\bar{L}(t)= L(W(t))$. Defining  $G(t) = M(t) M(t)^\top$, we have
\begin{align*}
    \bar{L}(t) =  \langle \mathsf{A}_{G(t)} f^* , f^* \rangle = \sum_\beta \left(  \prod_{j \in B_W} \sin^{2\beta_j}(\pi \alpha_j t) \cdot \prod_{i \notin B_W} \lambda_i^{2\beta_i}  \right) \langle \mathsf{P}_{V} f^* , H_\beta \rangle^2~.
\end{align*}
Furthermore, we verify that $\bar{L}'(\frac{1}{2}) \neq 0$, proving that $W$ cannot be a critical point. Indeed, 
\begin{align}
    \bar{L}'(t) & =  2 \pi \sum_{\beta} \sum_{j \in B_W} \alpha_j \beta_j \sin^{2\beta_j - 1}(\pi \alpha_j t) \cos(\pi \alpha_j t) \prod_{j' \in B_W, j'\neq j} \sin^{2\beta_{j'}}(\pi \alpha_{j'} t) \left(\prod_{i \notin B_W} \lambda_i^{2\beta_i}\right) \langle \mathsf{P}_{V} f^* , H_\beta \rangle^2 ~,
\end{align}
so that we have,
\begin{align}
\bar{L}'\left(\frac12\right) &= 2 \pi \sum_{\beta} \left(\sum_{j \in B_W} \alpha_j \beta_j \cot(\pi \alpha_j /2)\right) \prod_{j' \in B_W} \lambda_{j'}^{2\beta_{j'}} \left(\prod_{i \notin B_W} \lambda_i^{2\beta_i}\right) \langle \mathsf{P}_{V} f^* , H_\beta \rangle^2 \nonumber \\
&= 2\pi \sum_{\beta} \left(\sum_{j \in B_W} \alpha_j \beta_j \cot(\pi \alpha_j /2)\right) \left(\prod_{i} \lambda_{i}^{2\beta_{i}}\right)  \langle \mathsf{P}_{V} f^* , H_\beta \rangle^2~.
\end{align}
Since $\cot(t) > 0$ for $0<t<\pi/2$ and $0 < \alpha_j < 1$ for all $j \in B_W$, we have 
$$0< \min_{j \in B_W} \alpha_j \cot(\pi \alpha_j/2):= C~,$$
and therefore, denoting $\bar{\lambda} = \min_{j \in B_W} \lambda_j$, we have 
\begin{align}
    \bar{L}'\left(\frac{1}{2}\right) &\geq 2\pi C \sum_{\beta} \left(\sum_{j \in B_W} \beta_j \right) \left(\prod_{i} \lambda_i^{2\beta_i}\right) \langle \mathsf{P}_{V} f^* , H_\beta \rangle^2 \nonumber \\
    & = 2 \pi C \sum_{\beta; \text{supp}(\beta)\subseteq A_W \cup B_W} \left(\sum_{j \in B_W} \beta_j \right) \left(\prod_{j \in B_W} \lambda_j^{2\beta_j}\right) \langle \mathsf{P}_{V} f^* , H_\beta \rangle^2 \nonumber \\
    & \geq 2 \pi C \sum_{k=0}^\infty \sum_{\substack{\beta; \text{supp}(\beta)\subseteq A_W \cup B_W \\ |\beta|_{B_W} = k}} k \bar{\lambda}^{2 k}\langle \mathsf{P}_{V} f^* , H_\beta \rangle^2  \nonumber\\ 
    & \geq 2 \pi C \sum_{k=1}^\infty \sum_{\substack{\beta; \text{supp}(\beta)\subseteq A_W \cup B_W \\ |\beta|_{B_W} = k}} \bar{\lambda}^{2 k}\langle \mathsf{P}_{V} f^* , H_\beta \rangle^2  \nonumber\\ 
    & \geq 2 \pi C \bar{\lambda}^{2k^*} \eta_{k^*} \nonumber \\
    & > 0~,
\end{align}
where we used  (\ref{eq:wash2}) and the definition of $k^*$ and $\eta^*$. 
\end{proof}
A straightforward consequence of the lemma is that  $W$ cannot be a critical point. \\

\paragraph{\underline{Second step: Inclusion $\mathbf{\Theta \subseteq \Omega}$}.} Let $W \in \Theta$. 
Since $W \in \Theta$, by definition we have that $\mathrm{Sp}(G_W) \in \mathcal{G}(q,\tau)$ is a critical point of $L_\tau(W)$. 
Let us first verify that $\mathrm{Jt}(G_W) = \mathrm{Ess} (G_W) \oplus  \mathrm{Sp} (G_W) $ is also a critical point of $L_{\tau+\tau'}$:
\begin{lemma}
\label{lem:critical_composition_joint}
    If $B \in \mathcal{G}(q,\tau)$ is a critical point of $L_\tau$ and $B' \in \mathcal{G}(q,\tau+\tau')$ is such that $B \subset B'$ and 
    $\| \mathsf{A}_{B} f^* \| = \| \mathsf{A}_{B'} f^* \| $, then $B'$ is a critical point of $L_{\tau+\tau'}$.
\end{lemma}

Furthermore, we have the following useful fact.
\begin{lemma}
\label{fact:lastclaim}
We decompose $\overline{V}=[\overline{V}_{\mathrm{Sp}}; \overline{V}_{\mathrm{Ess}}; \overline{V}_0]$, where $\overline{V}_{\mathrm{Sp}} \in \mathbb{R}^{q \times \tau}$ and $\overline{V}_{\mathrm{Ess}} \in \mathbb{R}^{q \times \tau'}$. Then $\overline{V}_{\mathrm{Ess}} = 0~.$    
\end{lemma}

Let us conclude assuming for now Lemma~\ref{fact:lastclaim}. Observe that if $W \in \Theta$, then  the loss $L(W)$ only depends on the two subspaces $\mathrm{Sp}(G_W)$ and $\mathrm{Ess}(G_W)$.  In other words, we claim that if $W,W'$ are such that $\mathrm{Ess}(G_W)=\mathrm{Ess}(G_{W'})$, $\mathrm{Sp}(G_W) = \mathrm{Sp}(G_{W'})$, and $\| \mathsf{A}_{\mathrm{Sp}(G_W)} f^* \| = \| \mathsf{A}_{\mathrm{Jt}(G_W)} f^* \|$, then $L(W) = L(W')$. 
Indeed, from $\| \mathsf{A}_{\mathrm{Sp}(G_W)} f^* \| = \| \mathsf{A}_{\mathrm{Jt}(G_W)} f^* \|$, we have 
\begin{align}
\label{eq:future}
    L(W) &= \sum_{\beta; \text{supp}(\beta)\subseteq B_W\cup A_W}  \left(\prod_{j \in B_W} \lambda_j^{2\beta_j}\right) \langle \mathsf{P}_{V} f^*, H_\beta \rangle^2 \nonumber \\
    &= \sum_{\substack{\beta; \text{supp}(\beta)\subseteq B_W\cup A_W \\ |\text{supp}(\beta) \cap B_W| = 0}}  \left(\prod_{j \in B_W} \lambda_j^{2\beta_j}\right) \langle \mathsf{P}_{V} f^*, H_\beta \rangle^2 + \sum_{\substack{\beta; \text{supp}(\beta)\subseteq B_W\cup A_W \\ |\text{supp}(\beta) \cap B_W| >0}}  \left(\prod_{j \in B_W} \lambda_j^{2\beta_j}\right) \langle \mathsf{P}_{V} f^*, H_\beta \rangle^2 \nonumber \\
    &= \sum_{\beta; \text{supp}(\beta)\subseteq A_W }  \langle \mathsf{P}_{V} f^*, H_\beta \rangle^2~,
\end{align}
which shows in particular that there is no dependency on $\lambda_i$, $i \in B_W$, namely
$\partial_{\lambda_i} \ell(V, \Lambda) = 0~,~i\in B_W$. 
Moreover, we verify directly from the definition  
\begin{align*}
    \partial_{\lambda_i} \ell(V, \Lambda) &= 2\sum_{\beta} \beta_i \lambda_i^{2\beta_i - 1} \prod_{j \neq i} \lambda_j^{2\beta_j} \langle \mathsf{P}_{V} f^*, H_\beta \rangle^2 ~,
\end{align*}
that $\partial_{\lambda_i} \ell(V, \Lambda) = 0~,~i > \tau + \tau'$, 
and thus that 
\begin{equation}
\label{eq:blint}
\overline{\Lambda}_{i,i} = 0 ~\text{ for }~ i > \tau~.     
\end{equation}
From now, we have all the calculation to show the following result
\begin{lemma}
\label{lem:calculations_critical}
Denoting $X := ( I - W W^\top) W_* V $, we have the calculation $X^\top X = I - \Lambda^2$ and the equivalence 
\begin{align}
\label{eq:bing}
\nabla_W^{\mathcal{G}(r,d)} L(W) = 0  \Leftrightarrow  X^\top X \Xi \Xi^\top  = 0~.    
\end{align}
\end{lemma}
Recall that we have $\Xi = \overline{\Lambda}  + (F \circ \overline{V}) \Lambda$, with an abuse of notation (only in this appendix!!) that confuses  $\overline{V} \to  V^\top \overline{V} -  \overline{V}^\top V$ for the sake of shorter calculations.
\begin{align}
\label{eq:kip}
\Xi \Xi^\top &= (\overline{\Lambda}  + (F \circ \overline{V}) \Lambda) ( \overline{\Lambda} - \Lambda (F \circ \overline{V}^\top) )  \nonumber \\
&= \overline{\Lambda}^2 - (F \circ \overline{V})\Lambda^2 (F \circ \overline{V}^\top) - \overline{\Lambda}\Lambda (F \circ \overline{V}^\top) \overline{\Lambda} +  (F \circ \overline{V}) \Lambda \overline{\Lambda}~.
\end{align}
Therefore from Lemma~\ref{lem:calculations_critical} we deduce that 
\begin{align*}
  \Xi^\top \Xi X X^\top & = (\overline{\Lambda}  + (F \circ \overline{V}) \Lambda) ( \overline{\Lambda} - \Lambda (F \circ \overline{V}^\top) ) (1 - \Lambda^2) \\
  & = \overline{\Lambda}^2(1 - \Lambda^2) - (F \circ \overline{V})\Lambda^2 (F \circ \overline{V}^\top)(1 - \Lambda^2) - \overline{\Lambda}\Lambda (F \circ \overline{V}^\top) \overline{\Lambda}(1 - \Lambda^2) +  (F \circ \overline{V}) \Lambda \overline{\Lambda}(1 - \Lambda^2) \\
  &= 0 ~,
\end{align*}
since  $\overline{\Lambda}(I - \Lambda^2) = 0$ and $\Lambda(I - \Lambda^2)$  are diagonal matrices with non-zero entries only for $j \in B_W$, which are exactly the columns where $\overline{V}_j = 0$ thanks to Lemma~\ref{fact:lastclaim}. This proves that $\nabla_W^{\mathcal{G}} L(W)=0$ and thus $W \in \Omega$. 
We conclude this second part of the proof by establishing all the intermediate results (Lemmas ~\ref{lem:critical_composition_joint}, \ref{fact:lastclaim} and   \ref{lem:calculations_critical}).  

\begin{proof}[Proof of Lemma \ref{lem:critical_composition_joint}]
Let us argue by contradiction. 
Assume there is a smooth path $\gamma(t) \in \mathcal{G}(q,\tau + \tau')$, $t \in (-1, 1)$, such that $\gamma(0) = B'$ and $\ell(t) = L_{\tau + \tau'}(\gamma(t))$ satisfies $\ell'(0) \neq 0$. 
We consider a projection $P_\tau$ such that $P_\tau B' = B$, and we decompose 
\begin{align}
    L_{\tau+\tau'}(\gamma(t)) &= \| \mathsf{A}_{P_\tau \gamma(t)} f\|^2 + \| \mathsf{A}_{\gamma(t)} f - \mathsf{A}_{P_\tau \gamma(t)} f\|^2 ~.
\end{align}
It follows that 
\begin{align}
    \ell'(0) &= \frac{d}{dt}L_{\tau+\tau'}(\gamma(t)) |_{t=0} =  \frac{d}{dt} \| \mathsf{A}_{P_\tau \gamma(t)} f\|^2 + \frac{d}{dt}\left[ \| \mathsf{A}_{\gamma(t)} f - \mathsf{A}_{P_\tau \gamma(t)} f\|^2 \right]_{|t=0}
\end{align}
and since the last term in the RHS is zero at $t=0$ and is non-negative, its derivative must vanish, implying that $\frac{d}{dt} \| \mathsf{A}_{P_\tau \gamma(t)} f\|^2 $ is non-zero at $t=0$. But this contradicts the fact that $B$ is a critical point of $L_\tau$. 
\end{proof}

\begin{proof}[Proof of Lemma~\ref{fact:lastclaim}]
Defining $\nabla_V^\beta a_\beta(V) := \nabla_V (\langle \mathsf{P}_{V} f^*, H_\beta \rangle)$, we have thanks to equation~\eqref{eq:future} that for $W \in \Theta$,
\begin{align}
\label{foop}
 \nabla_V \ell(V, \Lambda) = 2 \sum_{\beta; \text{supp}(\beta)\subseteq  A_W} \langle \mathsf{P}_{V} f^*, H_\beta \rangle \nabla_V^\beta a_\beta(V) ~,
\end{align}
But observe that the RHS of (\ref{foop}) does not depend on $\Lambda$, which means that it must also hold for $\lambda_j = 1$ for all $j \in B_W$. This is precisely $\mathrm{Jt}(G_W)$, and from Lemma \ref{lem:critical_composition_joint} we know that $\nabla^{\mathcal{G}} L_{\tau+\tau'}(\mathrm{Jt}(G_W)) = 0$. Finally, since for $\lambda_i=\{0,1\}$ the tangent 
space of $(\mathcal{O}_q / \sim)$ at $(\Lambda, V)$ and the tangent space of $\mathcal{G}(q,\tau+\tau')$ at $\mathrm{Jt}(G_W)$ coincide, we conclude that $\overline{V}_{\mathrm{Jt}}=0$, proving fact~\ref{fact:lastclaim}. 
\end{proof}

\begin{proof}[Proof of Lemma \ref{lem:calculations_critical}]
For the first calculation we verify that
\begin{align*}
    X^\top X &= V^\top W_*^\top (I - W W^\top)^2 W_* V \\
    &= V^\top W_*^\top (I - W W^\top) W_* V \\ 
    &= V^\top( I - G ) V \\
    &=  I - \Lambda^2 ~,
\end{align*}
and for the critical points characterization part, we have 
\begin{align}
    \nabla_W^{\mathcal{G}(r,d)} L(W) = 0 & \Leftrightarrow  X\Xi = 0 \Leftrightarrow \mathrm{Tr}( \Xi X X^\top \Xi^\top) = 0 \Leftrightarrow X^\top X \Xi \Xi^\top   = 0~,
\end{align}
that concludes the proof of the lemma.
\end{proof}

\paragraph{\underline{Last step: Saddle points characterization}.}
Finally, we establish that $W \in \Theta$ is a saddle point whenever $0< \tau < q$ and 
$\mathrm{Sp}(G_W) \in \Gamma_{\tau}(f^*)\setminus \Gamma_\tau^0(f^*)$. 
Similarly as before, we will construct two paths $W^+(t)$ and $W^-(t)$ such that $W^+(0) = W^-(0) = W$ and 
such that the associated loss $\bar{L}^{+}(t) = L(W^{+}(t))$ (resp. $\bar{L}^{-}(t) = L(W^{-}(t))$) is strictly increasing (resp. decreasing) at $t=0$. 

Following (\ref{eq:green}), let $\bar{z}_1, \ldots \bar{z}_\tau$ be a collection of orthonormal vectors in $\mathbb{R}^d$, orthogonal to both $W$ and $W_*$.
We consider first for $j =1\ldots  \tau$ the two-dimensional subspaces $B_{j}$ spanned by $\bar{u}_{j}$ and $\bar{z}_j$. We define $\bar{u}_j(t) = \cos(\pi t) \bar{u}_{j} + \sin(\pi t) \bar{z}_j$, and $W^{-}(t) := \text{span}( \bar{u}_j \text{ for } j > \tau; \bar{u}_j(t) \text{ for } j \leq \tau)$. We verify that $W^{-}(0) = \text{span}( \bar{U})= W$, and that the SVD of $M(t) = W_*^\top W(t) $ is given by $M(t) = V(t) \Lambda(t) U(t)^\top$, with $\lambda_j(t) = \lambda_j$ for $j>\tau$ and $\lambda_{j}(t) = \cos(\pi t)$ for $j=1\ldots \tau$.  
The loss $\bar{L}^{-}(t)$ is given by
\begin{align*}
    \bar{L}^{-}(t) &= \sum_\beta \prod_i \lambda_i(t)^{2 \beta_i} \langle \mathsf{P}_{V} f^*, H_\beta \rangle^2 \\ 
    &= L(W) + \sum_\beta \left[\left( \prod_{j \leq \tau}\cos(\pi t)^{2 \beta_{j}}\right)-1\right] \left[\prod_{i > \tau} \lambda_i^{2\beta_i} \right]\langle \mathsf{P}_{V} f^*, H_\beta \rangle^2 \\ 
    &= L(W) + \sum_{\beta; \text{supp}(\beta)\subseteq A_W\cup B_W} \left[\cos(\pi t)^{2 |\beta|_{A_W}}-1\right] \left[\prod_{i \in B_W} \lambda_i^{2\beta_i} \right] \langle \mathsf{P}_{V} f^*, H_\beta \rangle^2 \\
    &= L(W) + \sum_{\beta; \text{supp}(\beta)\subseteq A_W} \left[\cos(\pi t)^{2 |\beta|_{A_W}}-1\right] \langle \mathsf{P}_{V} f^*, H_\beta \rangle^2 \\
    & \leq L(W) + \left[\cos(\pi t)^{2}-1\right] \sum_{\beta; \text{supp}(\beta)\subseteq A_W}  \langle \mathsf{P}_{V} f^*, H_\beta \rangle^2 \\
    &= L(W) + \left[\cos(\pi t)^{2}-1\right] L_\tau(\mathrm{Sp}(G_W))~,
\end{align*}
which is strictly smaller than $L(W)$ for any $t>0$. Above we have used 
$\| \mathsf{A}_{\mathrm{Sp}(G_W)} f^* \| = \| \mathsf{A}_{\mathrm{Jt}(G_W)} f^* \|$ in the fourth line, and $\mathbb{E} f^* =0$ in the fifth line.

On the other hand, since $0< \tau < q$,  $\| \mathsf{A}_{\mathrm{Sp}(G_W)} f^* \| = \| \mathsf{A}_{\mathrm{Jt}(G_W)} f^* \|$ and $f^*$ has intrinsic dimension $q$, the orthogonal complement $Z=\mathrm{Jt}(G_W)^\perp \in \mathcal{G}(n,q)$ with $0<n = q-\tau-\tau'$ satisfies 
\begin{equation}
\label{eq:grup}
\| \mathsf{A}_{\mathrm{Jt}(G_W)} f^* \| < \| \mathsf{A}_{\mathrm{Jt}(G_W) \oplus Z} f^* \| = \|f^*\|~.    
\end{equation}
Let $C_W = \{ j \in [q]; \lambda_j = 0\}$. Thus, from (\ref{eq:grup}) we have
\begin{align}
    0 & < \sum_{\beta; |\beta|_{C_W}>0} \langle \mathsf{P}_{V} f^*, H_\beta \rangle^2 \nonumber \\
    & = \sum_{k} \sum_{\beta; |\beta|_{C_W}>0; |\beta|=k} \langle \mathsf{P}_{V} f^*, H_\beta \rangle^2 := \sum_{k} \eta_k~.
\end{align}
We define $k^* = \min \{ k; \eta_k >0 \}$.

Let $\tilde{Z} = Z^\top W_*$ and consider for $j=1\ldots t$ 
$\tilde{u}_j(t) = \cos(\pi t) \bar{u}_{\tau+\tau'+j} + \sin(\pi t) \tilde{z}_j$. 
Defining now $W^{+}(t) := \text{span}( \bar{u}_i \text{ for } i \leq \tau+\tau'; \tilde{u}_j(t) \text{ for } j \leq t)$, we verify again that 
$W^{+}(0) = W$, and that $G(t) = V \Lambda^2(t) V^\top$, with $\lambda_j(t) = \lambda_j$ for $j<\tau+\tau'$ and $\lambda_{j}(t) = \sin(\pi t)$ for $j=1+\tau+\tau'\ldots t+\tau+\tau'$.  
Let $\bar{\lambda} = \min_{j \in B_W} \lambda_j$. The loss $\bar{L}^{+}(t)$ is given by
\begin{align*}
    \bar{L}^{+}(t) &= \sum_\beta \prod_i \lambda_i(t)^{2 \beta_i} \langle \mathsf{P}_{V} f^*, H_\beta \rangle^2 \\ 
    &= L(W) + \sum_{\beta; |\beta|_{C_W}>0} \prod_{j} \lambda_j(t)^{2\beta_j} \langle \mathsf{P}_{V} f^*, H_\beta \rangle^2 \\ 
    &\geq L(W) + \sum_{\beta; |\beta|_{C_W}>0} \prod_{j\in B_W} \lambda_j^{2\beta_j} \sin(\pi t)^{2|\beta|_{C_W}} \langle \mathsf{P}_{V} f^*, H_\beta \rangle^2 \\ 
    &\geq L(W) + \bar{\lambda}^{2k^*} \sin(\pi t)^{2k^*} \eta_{k^*}~,
\end{align*}
which satisfies $L^{+}(t) > L(W)$ for $t>0$. 
We have just shown that there exist both strictly increasing and strictly decreasing smooth paths at such critical points, and therefore they are saddle points. The proof is now complete.

\end{proof}

%% file: Appendix/App_gradient_flow.tex
\subsection{Gradient and flow with respect to \texorpdfstring{$G$}{G}}
\label{subsecapp:gradient_flow_G}

We first prove Proposition~\ref{prop:gradient_loss_G}. Let us recall that $g : W \to G_W = M  M^\top= W_*^\top W W^\top W_* $, and define $\ell : G \to \ell(G)$ such that $ L(W) = \ell (g(W))$.
Before proving the proposition, let us state two useful identities that we use during the derivation:
\begin{fact}
    We have the following calculations: 
    \begin{enumerate}[label=(\roman*)]
        \item For all $f \in H^1_{\gamma_q}$, all $z \in \R^q$, we have $\nabla_z \Av_G f(z)  = G  \Av_G [\nabla_z f](z)$,
        \item For all $F = (f_1, \dots, f_q) \in \left(H^1_{\gamma_q}\right)^q$, we have  $\E_{\gamma_q}[z F(z)^\top] = \E_{\gamma_q}[\nabla F (z)]$.
    \end{enumerate}
\end{fact}
\begin{proof}
    The fact (i) follows simply from the right of changing expectations and derivative for $H^1_{\gamma_q}$ functions. Indeed, for all $z \in \R^q$, $\nabla_z \Av_G f(z) = \nabla_z \left[\E_y \left(f(Gz + \sqrt{I - G^2} y)\right)\right] = G \E_y \left( \nabla f(Gz + \sqrt{I - G^2} y)\right) =  G  \Av_G [\nabla_z f](z) $.
    
    The second fact (ii) is a consequence of Stein's identity. Indeed recall that in dimension $1$, if $x$ is distributed according to the standard Gaussian distribution, we know that $\E_{x} (x f(x)) = \E_{x} (f'(x))$. Now for $F = (f_1, \dots, f_q) \in \left(H^1_{\gamma_q}\right)^q$ and all $i,j \in \llbracket  1, q\rrbracket$,  
    $$\E_{z}[z_i f_j(z)] = \E_{z}[\partial_i f_j(z)] = \E_{\gamma_q}[\nabla F (z)]_{ij}.$$
    This concludes the proof of this fact.
\end{proof}
\begin{proof}[Proof of Proposition \ref{prop:gradient_loss_G}]
Recall first that $\ell(G) = \langle f , \Av_G f \rangle$, where, in this proof, we take the notational freedom that $f$ is  $f^*$. Hence, by linearity, $\nabla_G \ell(G) = \langle f^* , \nabla_G \Av_G f^* \rangle$. Now let us calculate $\nabla_G \Av_G f$ for any $f \in H^2_{\gamma_q} $. Let $z \in \R^q $, 
\begin{align*}
    \nabla_G \Av_G f(z) &=  \nabla_G \left[ \E_y \left( f (G z + (I - G^2)^{1/2} y \right)\right] \\
    &=   \E_y \left( ( z - (I - G^2)^{-1/2} G y) \nabla f (G z + (I - G^2)^{1/2} y )^\top \right) \\
    &=  z \Av_G[\nabla f](z)^\top - (I - G^2)^{-1/2} G\,  \E_y \left( y \nabla f (G z + (I - G^2)^{1/2} y)^\top \right). 
\end{align*}
But then, by the property (ii) of the above Fact, we have that the second term writes  
\begin{align*}
    \E_y \left( y \nabla f (G z + (I - G^2)^{1/2} y)^\top \right) &=  (I - G^2)^{1/2} \E_y \left(   \nabla^2 f (G z + (I - G^2)^{1/2} y)^\top \right) \\
    &=  (I - G^2)^{1/2}\Av_G \left[   \nabla^2 f \right](z).
\end{align*}
The $ (I - G^2)^{1/2}$ terms simplify and we finally have the expression:
\begin{align*}
    \nabla_G \Av_G f(z) &=  z \Av_G[\nabla f](z)^\top -  G \Av_G \left[   \nabla^2 f \right](z). 
\end{align*}
Now, we use this identity and 
\begin{align*}
      \nabla_G \ell(G) &= \E_z \left[f(z) \nabla_G \Av_G f (z)\right]  \\
      &= \E_z \left[z f(z) \Av_G[\nabla f](z)^\top \right] - G \E_z \left[f(z) \Av_G \left[ \nabla^2 f \right](z) \right].
\end{align*}
Let us work on the first term applying the integration by part formula.
\begin{align*}
      \E_z \left[z f(z) \Av_G[\nabla f](z)^\top \right] &= \E_z \left[\nabla_z \left(f(z) \Av_G[\nabla f](z)^\top \right) \right]  \\
      &= \E_z \left[\nabla_z f(z) \Av_G[\nabla f](z)^\top \right] + \E_z \left[ f(z)  \nabla_z \Av_G[\nabla f](z)^\top \right]   \\
      &= \E_z \left[\nabla_z f(z) \Av_G[\nabla f](z)^\top \right] + G \E_z \left[ f(z)  \Av_G[\nabla^2 f](z) \right].
\end{align*}
Hence finally, 
\begin{align*}
      \nabla_G \ell(G) &=  \E_z \left[\nabla_z f(z) \Av_G[\nabla f](z)^\top \right] + G \E_z \left[ f(z)  \Av_G[\nabla^2 f](z) \right] - G \E_z \left[f(z) \Av_G \left[ \nabla^2 f \right](z) \right] \\
      & = \E_z \left[\nabla_z f(z) \Av_G[\nabla f](z)^\top \right].
\end{align*}
The fact that we can rewrite $ \nabla_G \ell(G)  = \E_z \left[\nabla_z f(z) \Av_G[\nabla f](z)^\top \right] =  \E_z \left[\Av_M [\nabla f](z) \Av_M[\nabla f](z)^\top \right] \succcurlyeq 0 $  concludes the proof of the proposition.
\end{proof}

\begin{remark}[Generalization of Proposition \ref{prop:gradient_loss_G}]
\label{rem:gradient_loss_M}
    Proposition \ref{prop:gradient_loss_G} is a particular case of the general property $\nabla_M \langle f, \mathsf{A}_M g \rangle = \nabla f \otimes \mathsf{A}_M \nabla g$, valid for any $M \in \mathbb{R}^{q \times q}$ and $f,g \in H^1_{\gamma_q}$. The proof is analogous.  
\end{remark}

Now we prove Lemma \ref{lem:dotGlemma}.
\begin{proof}[Proof of Lemma \ref{lem:dotGlemma}]
    For $t \geq 0$, we have $ \dot{G}_t = \J g (W_t) \left[ (I - W_t W_t^\top) \J g (W_t)^\top [\overline{G}]\right],$ with the explicit forms:  
    \begin{align*}
    \forall H \in \R^{d \times q}, \qquad \J g (W_t)(H) &= W_*^\top ( H W_t^\top + W_t  H^\top ) W_*  \qquad \text{ and }\\
    \forall G \in \C^{q \times q}, \qquad  \J g(W_t)^\top[G] &=  2 W_* G M_t,
    \end{align*}
when $H'$ is symmetric. Hence, 
    \begin{align*}
    \dot{G}_t &= 2 \J g (W_t) \left[ (I - W_t W_t^\top) W_* \overline{G}_t M_t \right]\\
     &= 2 W_*^\top\left[ (I - W_t W_t^\top) W_*  \overline{G}_t M_t W_t^\top + \text{sym}  )\right] W_* \\
     &=  2\left( I - G_t \right)  \overline{G}_t G_t  + \text{sym}\\
    &=  2 G_t  \overline{G} + 2 \overline{G} G_t - 4 G_t \overline{G} G_t~.
    \end{align*}
    
\end{proof}

Let us now express (\ref{eq:dotGeq}) in terms of the eigenbasis $V$. Abusing notation, we define $\dot{G}_V = V \dot{G} V^\top$ 
and $\overline{G}_V = V \overline{G} V^\top$ (and we will drop the subindex $V$ when the context is clear). We have 
\begin{align}
   \frac12 \dot{G}_V  &= \Lambda^2 \overline{G}_V (I - \Lambda^2) + (I - \Lambda^2) \overline{G}_V \Lambda^2~.
\end{align}

From 
$$\overline{G} = \mathbb{E}_z \left[ \mathsf{P}_V[\nabla f] \mathsf{A}_{\Lambda^2}[\mathsf{P}_V[\nabla f]]^\top \right] $$
and $\nabla \mathsf{P}_V [f] = V \mathsf{P}_V[\nabla f]$ we observe that 
\begin{align}
    \overline{G}_V &= \mathbb{E} \left[ \nabla \mathsf{P}_V[f] \mathsf{A}_{\Lambda^2}[\nabla \mathsf{P}_V[f]]^\top \right]~.
\end{align}
From now on we drop the $V$ subindex and use $\dot{G}$ and $\overline{G}$ instead. 
Since $\Lambda^2$ and $I - \Lambda^2$ are diagonal, defining $B_{i,j} = \lambda_i^2 (1 - \lambda_j^2) + \lambda_j^2 ( 1 - \lambda_i^2)$ we have $\dot{G} = \overline{G} \circ B$. 

Let us now study $\overline{G}$. We consider the Hermite decomposition $\mathsf{P}_V[f] = \sum_\beta a_V(\beta) H_\beta$ and denote as usual $s = \inf\{ |\beta|; a_V(\beta) \neq 0\}$ the information exponent. Denote by $\Omega = \{ \beta; a_V(\beta) \neq 0; |\beta|=s\}$ the support of the lowest harmonics. For $\beta \in \mathbb{N}^q$, we define $\beta^{(i)} := (\beta_1, \ldots, \beta_{i-1}, \beta_i + 1, \beta_{i+1}, \ldots, \beta_q)$, and similarly we define $\Omega^{(i)} := \{ \beta = \tilde{\beta}^{(i)}; \tilde{\beta} \in \Omega\}$. 
From basic definitions of Hermite polynomials, we have that 
\begin{align}
\partial_{x_i} \mathsf{P}_V[f] &= \sum_\beta \sqrt{\beta_i} a_V(\beta) h_{\beta_i -1 }(x_i) \prod_{j \neq i} h_{\beta_j}(x_j) \nonumber \\
&= \sum_{\beta} \sqrt{\beta_i+1} a_V(\beta^{(i)}) H_{\beta} ~,
\end{align}
thus 
\begin{align}
\label{eq:random_G}
    \overline{G}_{i,j} &= \langle \partial_{x_i} \mathsf{P}_V[f] , \mathsf{A}_{\Lambda^2} \partial_{x_j} \mathsf{P}_V[f] \rangle \nonumber \\
    &= \sum_\beta \langle \partial_{x_i} \mathsf{P}_V[f], H_\beta \rangle \langle \partial_{x_j} \mathsf{P}_V[f], H_\beta \rangle \prod_{l} \lambda_l^{2\beta_l} \nonumber \\
    &= \sum_{\beta} \sqrt{(\beta_i+1)(\beta_j + 1)} a_V(\beta^{(i)}) a_V(\beta^{(j)}) \prod_{l} \lambda_l^{2\beta_l} ~.
\end{align}

\subsection{Summary statistics}
\label{subsecapp:gradient_flow_lambda_V}

We now turn into proving the equations of motion of the summary statistics written in Lemma~\ref{lem:sumstatdyn}.
\begin{proof}[Proof of Lemma~\ref{lem:sumstatdyn}]
Let us treat $\lambda$ and $V$ separately.  Recall the results of Lemma~\ref{lem:chainrule_grassmann}: we have the dynamics of $W$ that writes $$\dot{W_t} =  (I - W_t W_t^\top) W_* V \Xi U^\top~.$$

\noindent \emph{Dynamics of $\lambda$}. As $\J_{\Lambda}(W)(H) = \textrm{diag} (V^\top  W_*^\top H U )$, we have
\begin{align*}
    \dot{\Lambda}_t =  \J_{\Lambda}(W)(\dot{W_t}) = \textrm{diag} (V^\top  W_*^\top \dot{W}_t U ).
\end{align*}
Let us pause and calculate the product that appears:
\begin{align*}
    V^\top  W_*^\top \dot{W}_t U = V^\top  W_*^\top (I - W_t W_t^\top) W_* V \Xi U^\top U = \Xi - \Xi V^\top G V = \Xi(I - \Lambda^2),
\end{align*}
and looking at its diagonal, the anti-diagonal part in $\Xi$ vanishes. That is 
\begin{align*}
    \dot{\Lambda}_t &= \textrm{diag}((I - \Lambda^2) \overline{\Lambda}),
\end{align*}
Translating this component-wise, $\forall i \in \llbracket 1, r\rrbracket$, 
$$\dot{\lambda}_i =  (1 - \lambda_i^2) \partial_{\lambda_i} \ell(\lambda, V).$$
\noindent \emph{Dynamics of $V$}. As $\J_{V}(W)(H) = V (F \circ (V^\top  W_*^\top H U \Lambda + \text{sym}))$, we have
\begin{align*}
    \dot{V}_t &=  \J_{V}(W)(\dot{W_t}) = V_t (F \circ ( V_t^\top W_*^\top \dot{W_t} U_t \Lambda + \text{sym})) .
\end{align*}
Furthermore, as previously calculated, we have $ V_t^\top W_*^\top \dot{W_t} U_t    = \Xi (I - \Lambda^2) $, so that, 
\begin{align*}
    \dot{V}_t = V_t \left[F \circ \left(\Lambda (I - \Lambda^2) \Xi +  \Xi (I - \Lambda^2) \Lambda \right)\right].
\end{align*}
But now, as $F$ is anti-diagonal, the $\overline{\Lambda}$ does play a role in the expression, and hence 
\begin{align*}
    \dot{V}_t = V_t \left[F \circ \left(\Lambda (I - \Lambda^2) (F \circ \overline{V}^a) \Lambda + \Lambda (F \circ \overline{V}^a) (I - \Lambda^2) \Lambda \right)\right],
\end{align*}
with $\overline{V}^a = V^\top \overline{V} -\overline{V}^\top V$, and this gives the expression announced in the lemma.
\end{proof}

%% file: Appendix/App_dynamics_description.tex
\subsection{Upper bound on the eigenvalues escaping initialization: proof of Lemma \ref{lem:sequential_escape}}
\label{sec:prooflemmaseqescape}

For the sake of clarity, we restate the result:  
\lemmaescape*

\begin{proof}

We recall that the loss in terms of $\Lambda, V$ is  
\begin{align}
    \ell( \Lambda, V) &= \sum_{\beta} \left(\prod_{i=1}^q \lambda_i^{2\beta_i}\right) \alpha_\beta(V)^2  ~,
\end{align}
where $\alpha_\beta(V) = \langle \mathsf{P}_V f, H_\beta \rangle$. 
For convenience, throughout the proof we denote $V_m = [v_1, \ldots v_m] \in \mathcal{G}(q,m)$. We  define for $k=1\ldots K$ the complements ${Y}_k := W_k \setminus W_{k-1}$.
We  denote for all $i \leq q$, $\delta_i = c_i / \sqrt{d}:=\lambda_i(0)$. From Lemma~\ref{lem:eigenvalues_init}, for all $\delta > 0$, we know that the probability that $\{c \geq \delta/n_q\}$ is larger than $1- \delta$, where $n_q = \Theta_d(1)$. Until the end, we place ourselves on such on event.

We proceed by induction over $k$. 

\vspace{0.25cm}

\paragraph{First subspace learning: case $\bm{k=1}$} For $t \leq \tau_1(1/2)$, and for $j \in \llbracket p_1, q \rrbracket$ we have 
\begin{align}
\label{eq:blink1}
    \frac12 \frac{d}{dt}[\lambda_{j}^2(t)] &= \lambda_{j}(t) \dot{\lambda}_{j}(t) \nonumber \\
      &= 2 (1 - \lambda_j^2)\sum_\beta \alpha_\beta(V)^2 \beta_{j} \prod_{i=1}^q \lambda_i^{2\beta_i} \nonumber \\
      &\geq \sum_\beta \alpha_\beta(V)^2 \beta_{j} \prod_{i=1}^q \lambda_i^{2\beta_i} \nonumber \\
        & \geq \sum_{|\beta|=s_1} \alpha_\beta(V)^2 \beta_{j} \prod_{i=1}^q \lambda_i^{2\beta_i}~ \nonumber \\ 
        & \geq \sum_{|\beta|=s_1; |\beta|_{[j+1:q]}=0, \beta_{j}>0} \alpha_\beta(V)^2 \prod_{i=1}^q \lambda_i^{2\beta_i}~ \nonumber \\   
    &\geq  \lambda_{j}^{2s_1} \xi_j(t)~,
\end{align}
where we have defined 
\begin{align}
    \xi_j(t) &:= \|\mathsf{A}_{V_{j}}  f_1\|^2 - \| \mathsf{A}_{V_{j-1}} f_1 \|^2 ~,
\end{align}
and recalling that $f_1=\Pi_{{s}_1} f$ has intrinsic dimension $p_1$ with support $W_1$. 
In order to quantify the escape time of $\lambda_j$, we need a suitable lower bound for $\xi_j(t)$. For that purpose, we have the following result.
\begin{claim}
\label{claim:xilower}
    Let $g$ have intrinsic dimension $l$ and order $s$, so that $g=\mathsf{S}^s g$. Then, for any nested family of subspaces $V_j$ of dimension $j=l,\ldots q$, the sequence $\xi_j = \| \mathsf{A}_{V_j} g \|^2 - \| \mathsf{A}_{V_{j-1}} g\|^2$ satisfies
    \begin{equation}
        \max_{j \in \llbracket l, q\rrbracket} \xi_j 
        \geq \mathcal{C}_g~.
    \end{equation}
\end{claim}
Before proving this claim, let us finish the argument of the case $k=1$ by applying it to $f_1$. 
Let us define 
\begin{align*}
    \chi_j = \int_0^{\tau_1} \mathbf{1}\left[\xi_j(t) \geq q^{-(1+s/2)} \mathcal{E}(g)\right] dt.
\end{align*} 
Thanks to Claim~\ref{claim:xilower}, we know that $\sum_{j=l_1}^q \chi_j \geq \tau_1$, so there exists 
$j^*$ with $\chi_{j^*} \geq q^{-1} \tau_1$. Recalling that $t \to \lambda^2_{j^*}(t)$ is a non-decreasing function, we can apply an adaptation of Gronwall comparison lemma stated in Lemma~\ref{lem:gromwall_adapted} on the set 
$\{ t; \xi_{j^*}(t) \geq c_q \mathcal{E}(g)\}$ to obtain that

\noindent \textbf{(i) Case $s_1 \geq 2$}, then, we have 
\begin{align}
    \lambda^2_{j^*}(\tau_1) &\geq \left[\frac{1}{ [\lambda^2_{j^*}(0)]^{s_1-1/2}} - (s_1-1/2) c_q \mathcal{E}(g)  \tau_1 \right]^{-\frac{1}{s_1-1/2}} \nonumber\\
     &\frac12 \geq \left[\frac{d^{ s_1-1/2}}{ {c}^{2 s_1-1}} - (s_1-1/2) c_q \mathcal{E}(g)  \tau_1 \right]^{-\frac{1}{s_1-1/2}}~,
\end{align}
and hence, we have the upper bound
\begin{align}
    \tau_1 \leq \frac{1}{ (s_1-1/2) c_q \mathcal{E}(g)  (\delta / n_q)^{2s_1-1}}  d^{ s_1-1/2}~.
\end{align}
Since $j^*\in \llbracket p_1, q\rrbracket$ and the eigenvalues are ordered, this implies that at least $p_1$ eigenvalues have escaped mediocrity in the timescale $\mathcal{C}_{f, \delta, s_1} d^{s_1-1/2}$.

\noindent \textbf{(ii) Case $s_1 = 1$}, then, we have 
\begin{align}
    \lambda^2_{j^*}(\tau_1) &\geq \lambda^2_{j^*}(0) e^{c_q \mathcal{E}(g)  \tau_1 } \nonumber\\
     &\frac12 \geq \frac{c^2}{d} e^{c_q \mathcal{E}(g)  \tau_1 } ~,
\end{align}
and hence, we have the upper bound
\begin{align}
    \tau_1 \leq \frac{1}{ c_q \mathcal{E}(g)}  [\log(d/\delta^2) - \log(n_q^2/2)] ~.
\end{align}
Like previously, this implies that at least $p_1$ eigenvalues have escaped mediocrity in the timescale $\mathcal{C}_{f, \delta} (\log(d) + \log(1/\delta))$. 

Now we can pursue the same reasoning for times $ t \in [\tau_1(1/2),  \tau_1(\eta)]$, except that equation \eqref{eq:blink1} becomes
\begin{align*}
     \frac{d}{dt}[1 - \lambda_{j}^2(t)] &= - 2 (1 - \lambda_j^2)\sum_\beta \alpha_\beta(V)^2 \beta_{j} \prod_{i=1}^q \lambda_i^{2\beta_i} \nonumber \\
      &\leq - 2  \lambda_{j}^{2s_1} \xi_j(t) (1 - \lambda_j^2) \nonumber \\
       &\leq - 2^{-2s_1 + 1} \xi_j(t) (1 - \lambda_j^2)~ .
\end{align*}
Hence, by a standard Grownwall inequality, for times $ t \in [\tau_1(1/2),  \tau_1(\eta)]$,
\begin{align*}
    1 - \lambda_{j}^2(t) &\leq \frac{3}{4} e^{\Cu (t - \tau_1(1/2))} .
\end{align*}
Thus, by a standard comparison argument, we have that $$\tau_1(\eta) \leq \tau_1(1/2) + \mathcal{C}_{f, \delta, s_1} \log(1 / \eta) \leq \mathcal{C}_{f, \delta, s_1} \left[ d^{s_1 - 1/2} \vee \log(d/\delta)  + \log( 1 / \eta) \right], $$
which concludes the first step. Now, it remains to prove the lower bound on $\xi_j$ claimed in~Claim \ref{claim:xilower}. 
\begin{proof}[Proof of Claim \ref{claim:xilower}]
Let $W=W[g]$ be the support of $g$, which has intrinsic dimension $l$. We can then write 
$g = \mathsf{P}_{W} \bar{g}$, with $\bar{g} \in L^2_{\gamma_{l}}$ of intrinsic dimension $l$. 
We first define, for all $j \in \llbracket 2, q\rrbracket$, the matrix 
\begin{align}
    M_j :=  W V_j V_j^\top W^\top \in  \C_{l}.
\end{align}
as well as $\eta_j := \lambda_{\text{min}}(M_j) = \sigma^2_{\text{min}}(W V_j)$. The sequence of these correlation matrices satisfy
\begin{align}
\label{eq:m_j_sequence}
    M_j = M_{j-1} + W v_j v_j^\top W^\top~,
\end{align}
so that, thanks to Weyl's inequality, $(\eta_j)_j$ is a (non-negative) non-decreasing sequence. Furthermore, noticing that $M_q = I_{l}$, and $\text{rank}(M_{l - 1}) \leq l - 1$ we have $\eta_q = 1$ and $\eta_{l-1}=0$, hence it follows that 
there exists $j^* \in \llbracket l,q\rrbracket$ such that 
$\eta_{j^*} - \eta_{j^*-1} \geq 1/q$. As a consequence, $M_{j^*}$ is invertible so that we can define
\begin{align}
T_{j^*}:=M_{j^*}^{-1/2}M_{{j^*}-1} M_{j^*}^{-1/2} \in  \C_{l}~.
\end{align}
Informally, $M_j$ measures the overlap between the subspaces $W_1$ and $V_j$, whereas $T_j$ quantifies the difference between two successive subspaces. We first rewrite $\xi_{j^*}$ in terms of these matrices. Indeed, we have
\begin{align}
\| \mathsf{A}_{V_{j^*}} g \|^2 &= \langle \mathsf{A}_{V_{j^*}} \mathsf{P}_{W} \bar{g}, \mathsf{A}_{V_{j^*}} \mathsf{P}_{W} \bar{g} \rangle \nonumber\\
&= \langle \mathsf{A}_{W V_{j^*}} \bar{g}, \mathsf{A}_{W V_{j^*}} \bar{g} \rangle \nonumber \\
&= \langle \mathsf{A}_{M_{j^*}} \bar{g}, \bar{g} \rangle~.
\end{align}
Let us define $\bar{g}_{j^*} := \mathsf{A}_{M_{j^*}^{1/2}} \bar{g}$, that has intrinsic dimension $l$  
and satisfies $\mathcal{E}(\bar{g}_{j^*}) \geq \lambda_{\text{min}}(M_{j^*})^{s/2}\mathcal{E}(g) \geq q^{-s/2} \mathcal{E}(g)$ thanks to Proposition~\ref{prop:intrinsic_averaging} and the fact that $\eta_{j^*}= \lambda_{\text{min}}(M_{j^*}) \geq 1/q$. 
We finally have
\begin{align}
\label{eq:queen1}
    \xi_{j^*} &=  \langle \mathsf{A}_{M_{j^*}} \bar{g}, \bar{g} \rangle - \langle \mathsf{A}_{M_{{j^*}-1}} \bar{g}, \bar{g} \rangle~ \nonumber \\
    &= \langle \mathsf{A}_{M_{j^*}} \bar{g}, \bar{g} \rangle - \langle \mathsf{A}_{T_{j^*}} \mathsf{A}_{M_{j^*}^{1/2}} \bar{g},\mathsf{A}_{M_{j^*}^{1/2}} \bar{g} \rangle ~ \nonumber \\
    & = \langle \bar{g}_{j^*}, \bar{g}_{j^*} \rangle - \langle \mathsf{A}_{T_{j^*}}  \bar{g}_{j^*}, \bar{g}_{j^*} \rangle ~.
\end{align}
The main idea of the proof is now to show that $\mathsf{A}_{T_{j^*}}$ will act as a shrinkage. To show this, we upper bound the smallest eigenvalue of $T_{j^*}$ and get a lower bound on $\xi_{j^*}$. By Ostrowski inequality~\cite[Theorem 4.5.9]{horn2012matrix}, we have
\begin{align*}
    \lambda_{\min}(T_{j^*}) = \lambda_{\min}(M_{j^*}^{-1/2}M_{j^*-1}M_{j^*}^{-1/2}) \leq \lambda_{\min}(M_{j^*-1}) \lambda_{\max}(M_{j^*}^{-1}) = \frac{\eta_{j^*-1}}{\eta_{j^*}}~,
\end{align*}
And as $\eta_{j^*-1} \leq \eta_{j^*} - 1/q$, we have
\begin{align}
\label{eq:upper_bound_T_j}
    \lambda_{\min}(T_{j^*}) \leq 1 - \frac{1}{q\eta_{j^*}} \leq 1 - \frac{1}{q} ~.
\end{align}
Let us develop the equality Eq.\eqref{eq:queen1} into the basis of Hermite polynomials adapted to the eigenvectors of $T_{j^*}$. With the slight overuse of the notation, we denote by $(a_\beta)_\beta$ the coefficient of $\bar{g}_{j^*}$ in this basis, and $(\lambda_i)_{i \leq l}$ the (sorted) eigenvalues of $T_{j^*}$. We get:
\begin{flalign*}
        && \xi_{j^*} &= \langle \bar{g}_{j^*}, \bar{g}_{j^*} \rangle - \langle \mathsf{A}_{T_{j^*}}  \bar{g}_{j^*}, \bar{g}_{j^*} \rangle   \\
        &&  &= \sum_\beta a_\beta^2 - \sum_\beta a_\beta^2 \left(\prod_{i = 1}^{l} \lambda_i^{\beta_i}\right)  \\
        &&  &\geq \sum_\beta a_\beta^2 - \sum_\beta a_\beta^2 \lambda_{l}^{\beta_{l}}  & \text{(as all $\lambda_i \leq 1$, for $i \leq l-1$)}  \\
        &&  &\geq \sum_\beta \left(1  - \lambda_{l}^{\beta_{l}} \right)a_\beta^2   \\
        &&  &\geq \sum_{\beta_{l > 0}} \left(1  - (1 - q^{-1})^{\beta_{l}} \right)a_\beta^2    & \text{(in virtue of Eq.\eqref{eq:upper_bound_T_j})} \\
        &&  &\geq  \left(1  - (1 - q^{-1})\right) \sum_{\beta_{l > 0}}a_\beta^2    \\
         &&  &\geq \frac{\mathcal{E}(\bar{g}_{j^*})}{q}  \\
        &&  & \geq \mathcal{C}_g~,
\end{flalign*}
thanks to Proposition \ref{prop:intrinsic_averaging}.

\end{proof}

\paragraph{Induction step: case $\bm{k>1}$} Let us now consider $k>1$. 
Suppose that after time $\tau_k(\eta)$ we have  $\ell_k \geq p_k$ eigenvalues which have reached the value $1 - \eta$. Our goal is now to show that after $\tau_k(\eta) + \Cu d^{2\max_{k'\leq k+1} s_{k'} - 1}$ at least $p_{k+1}$ eigenvalues will have reached level $1 - \eta$.

Let $\varepsilon>0$, that we will adjust later. If $\ell_k \geq  p_{k+1}$, then there the statement is already proven. Hence, we consider the case $\ell_k < p_{k+1}$. For $ m \in \llbracket 1,  k +1 \rrbracket$ and $j \in \llbracket \ell_k, q \rrbracket$ we define
\begin{align*}
    M_{m,j} = W_m V_{j} V_{j}^\top W_m^\top \in \R^{\ell_m \times \ell_m}~,
\end{align*}
which measure the correlation between the first $j$ eigenvectors and $W_m$. We define also $\eta_{m,j} := \lambda_{\min} (M_{m,j})$. When $m = 0$, we simply adopt the convention that $\eta_{0,j} := 1$ for all $j \in \llbracket \ell_k, q \rrbracket$.
\begin{lemma}
    Let $ k_\varepsilon^*:= \sup\,\{ m \in \llbracket 0, k \rrbracket    \, ;\, \eta_{m, \ell_k} \geq 1 - \varepsilon \}$, then there exists and $j_\varepsilon^* \in \llbracket \ell_k + 1, q \rrbracket$, such that 
    \begin{align}
    \label{eq:etamj_gap}
        \eta_{k_\varepsilon^* + 1 ,j_\varepsilon^*}-\eta_{k_\varepsilon^* + 1 ,j_\varepsilon^*-1}\geq q^{-1}\varepsilon~.
    \end{align}
\end{lemma}
\begin{proof}
    First, let us observe that the sequence $(\eta_{m,j})_m$ is non-increasing. Indeed, for all $m \in \llbracket 1,  k \rrbracket$
\begin{align*}
    \eta_{m,j} = \min_{ x \in \R^{\ell_m}} \frac{\langle W_m V_{j} V_{j}^\top W_m^\top x , x \rangle}{\|x\|^2} = \min_{ x \in \R^{\ell_m}} \frac{\langle V_{j} V_{j}^\top W_m^\top x , W_m^\top  x \rangle}{\|W_m W_m^\top x\|^2} = \min_{ y \in \mathrm{Ran}(W_m^\top) } \frac{\langle V_{j} V_{j}^\top y , y \rangle}{\|y\|^2}~,
\end{align*}
which obviously define a non-decreasing sequence as $(\mathrm{Ran}(W_m^\top))_m$ is a nested sequence of spaces. Also, as $\mathrm{dim}(W_{k+1}) > \ell_k$, we have $\eta_{k+1, \ell_k} = 0$. Let us drop the $\varepsilon$ subscript for the sake of clearness here. If $ k = k^*$, we have $ \eta_{k^* +  1,\ell_k} = 0 \leq 1 - \varepsilon $, else if  $ k < k^*$, we have that $ \eta_{k^* +  1,\ell_k} \leq 1 - \varepsilon = \eta_{k^* +  1, q} - \varepsilon $. Hence, either case, we have  $\eta_{k^* + 1,q} - \eta_{k^* + 1,\ell_k} \geq \varepsilon$, so again we can ensure that there will be $j^* \in \llbracket \ell_k+1,q\rrbracket$ such that $\eta_{k^*+1,j^*}-\eta_{k^*+1,j^*-1}\geq q^{-1}\varepsilon$.
\end{proof}
\noindent Let us drop the subscript $\varepsilon$ for the sake of clarity. For $j \in \llbracket \ell_k+1, q \rrbracket$, let us assume that $t \leq \tau_{j}(1/2)$, then we have, for any $z\geq s_{k^*+1}$, 

\begin{align}
\label{eq:blink1bis}
    \frac12 \frac{d}{dt}[\lambda_{j}^2(t)] &= \lambda_{j}(t) \dot{\lambda}_{j}(t) \nonumber \\
      &\geq \sum_\beta \alpha_\beta(V)^2 \beta_{j} \prod_{i=1}^q \lambda_i^{2\beta_i} \nonumber \\
               & \geq \sum_{\substack{|\beta|_{[1:\ell_{k}]}\leq {z+{s}_{k^*+1}}; |\beta|_{[j+1:q]}=0 \\|\beta|_{[\ell_k+1:q]}\leq {s}_{k^*+1}; \beta_{j}>0 }} \alpha_\beta(V)^2 \prod_{i=1}^q \lambda_i^{2\beta_i} \nonumber \\
        & \geq 2^{-2(z+{s}_{k^*+1})} \lambda_j^{2{s}_{k^*+1}} \xi_{j,k^*}(t)~,
\end{align}
with 
\begin{align}
    \xi_{j,k^*}(t) &:= \sum_{\substack{|\beta|_{[1:\ell_k]}\leq {z+s_{k^*+1}}; |\beta|_{[\ell_k+1:q]}\leq {s}_{k^*+1}; \\ |\beta|_{[j+1:q]}=0; \beta_{j}>0 }} \alpha_\beta(V)^2~.
\end{align}
We will now seek to lower-bound $\xi_{j,k^*}$ using the structural property that $f_{k^*+1}$ has intrinsic dimension $p_{k^*+1}$, and that it has energy contained in certain harmonics, as captured by the relative information and acquired exponents $s_{k^*+1}$ and $r_{k^*+1}$.  

Suppose first that $\eta_{k^*, \ell_k} = 1$, meaning that $W_{k^*}$ is exactly contained in the eigenspace $V_{\ell_k}$. 
In that case, by considering an orthonormal basis of $V_{\ell_k}$ of the form $[W_{k^*}; W_{k^*}^\perp]$, completed with $V_{\ell_k}^\perp$, 
we have
\begin{align}
    \xi_{j,k^*} = \sum_{\substack{|\beta|_{[1:\ell_k]}\leq {z+s_{k^*+1}}; |\beta|_{[\ell_k+1:q]}\leq {s}_{k^*+1}; \\ |\beta|_{[j+1:q]}=0; \beta_{j}>0 }}\!\!\!\!\!\! \alpha_\beta(V)^2 &\geq \sum_{\substack{|\beta|_{[k^*+1:q]}\leq  {s}_{k^*+1}; |\beta|\leq {z} + {s}_{k^*+1} \\ |\beta|_{[j+1:q]}=0 ; \beta_{j}>0 }}\!\!\!\!\!\! \alpha_\beta(V)^2 ~.\\ 
    &  \geq \sum_{\substack{|\beta|_{[\ell_{k^*}+1:q]}\leq  {s}_{k^*+1}; |\beta|\leq {z} + {s}_{k^*+1} \\ |\beta|_{[j+1:q]}=0 ; \beta_{j}>0 }}\!\!\!\!\!\! \alpha_\beta(V)^2 ~,
\end{align}
since $\ell_k \geq k^*$. 
Picking $z = r_{k^*+1}- s_{k^*+1}$, and recalling Eq (\ref{eq:acqexp}), plus the fact that $r_{k^*+1} \geq 2 s_{k^*+1}$, we thus obtain
\begin{equation}
      \xi_{j,k^*}  \geq \| \mathsf{A}_{V_j} \mathsf{S}^{r_{k^*+1}}f_{k^*+1} \|^2 - \| \mathsf{A}_{V_{j-1}} \mathsf{S}^{r_{k^*+1}} f_{k^*+1} \|^2 ~.
\end{equation}
Now, recalling that $ \mathsf{S}^{r_{k^*+1}} f_{k^*+1}$ has intrinsic dimension $p_{k^*+1}$, supported in $W_{k^*+1}$, using Claim \ref{claim:xilower} we conclude that the sequence $\xi_{j,k^*}(t)$ 
satisfies, for all $t>0$ such that $\eta_{k^*, \ell_k} = 1$,
\begin{align}    
    \max_{j\in [\ell_k+1,q]} \xi_{j,k^*}(t)&\geq \Cu~.
\end{align}

Now, we observe that $V \mapsto \xi_{j,k^*}(V_{\ell_k})$ is a smooth map, since it is in fact a homogeneous polynomial of degree ${r}_{k^*+1}$, by Lemma \ref{lem:hermitechange}. Hence, we can choose $\varepsilon > 0$ small enough, depending only of the dimension $q$ and $f$, such that $\xi_{j,k^*}(V_{\ell_k}) \geq \xi_{j,k}(V_{\ell_k}^+)/2$, where $V_{\ell_k}^+$ is a perturbation of $V_{\ell_k}$ such that $\lambda_{\min}(W_{k^*} V_{\ell_k}^+ (V_{\ell_k}^+)^\top W_{k^*}^\top) = 1$, 
since the two terms are equal in the limit $\varepsilon \to 0$. This means finally that for all $t > 0$
\begin{align}
\label{eq:lower_bound_iteration_lambda}
    \max_{j\in \llbracket \ell_k+1,q\rrbracket} \xi_{j,k^*}(t) \geq \Cu~.
\end{align}
Hence, from Eqs.\eqref{eq:blink1bis} and \eqref{eq:lower_bound_iteration_lambda}, we have that for all $t > 0$, there exists $j:=j_t \in \llbracket \ell_k + 1, q\rrbracket$, such that
\begin{align*}
    \frac{1}{2}\frac{d}{dt} [\lambda_{j_t}^2(t)] \geq \Cu \lambda_{j_t}^{2 s_{k^*+1}}
\end{align*}
Hence, we are in position to conclude. In fact similarly to the initialization step (see the detailed proof below the Claim~\ref{claim:xilower}), we can apply~Lemma~\ref{lem:gromwall_adapted} to conclude that the time, noted $\tau_{k + 1, \ell_k +1 }(1/2)$,  that at least $\ell_k + 1$ have escaped mediocrity is:
\begin{align*}
    \tau_{k + 1, \ell_k + 1}(1/2) \leq \tau_k(1/2) +  \mathcal{C}_{f,\delta, \tilde{s}_{k+1}} d^{s_{k^* + 1} - 1/2} \leq \tau_k(1/2) +  \Cu d^{ \max_{k' \leq k + 1 }s_{k'} - 1/2} 
\end{align*}
We can now iterate this argument at most $p_{k+1} - \ell_k$ times, to conclude that after a time $$\tau_{k+1}(1/2) \leq \tau_k(1/2) + \mathcal{C}_{f,\delta, \tilde{s}_{k+1}} d^{\max_{k'\leq k} s_{k'+ 1} -1/2}~,$$ at least $p_{k+1}$ eigenvalues have escaped mediocrity. Finally, the second part from level $1/2$ to level $1 - \eta$, being exponentially fast as detailed in the case $ k = 1$, we have that $$\tau_{k+1}(\eta) \leq \tau_k(\eta) + \mathcal{C}_{f,\delta, \tilde{s}_{k+1}}\left[d^{\max_{k'\leq k} s_{k'+ 1} \vee \log(d/\eta) -1/2} + \log(1 / \eta) \right]~,$$
and the proof is complete.
\end{proof}

\subsection{Exact number of eigenvalues escaping initialization: proof of Lemma \ref{lem:lowergroup}}
\label{sec:prooflemmalowergroup}

We restate here Lemma \ref{lem:lowergroup}:
\lemmaregrouped*

\begin{proof}
We now argue that there are exactly $\tilde{p}_{k}$ eigenvalues that could have reached a value $\Omega(d^{-\frac{1}{2}})$ in the time-scale $\tilde{\tau}_k$ given by step $k \geq 1$. 

We begin by establishing a matching upper bound for the correlation growth, as described in the following lemma:
\begin{lemma}[Correlation Upper bound]
\label{lem:correl_upperbound}
    For any $B \in \mathcal{G}(q,p)$, let $Z_B(t) := \| B^\perp G(t) \| $. Then $\dot{Z}_B \leq (2Z_B)^{s(f; B)}\|\nabla f\|_s$, where we recall that $s(f; B)$ is the information exponent of $f$ relative to the subspace $B$, and $\|f\|_\Xi$ is the Sobolev norm with spectral weights $\Xi(k; s,q) := k^s {c_{q,k}}$.  
\end{lemma}

Let us apply Lemma \ref{lem:correl_upperbound} with $B = \tilde{W}_k$, 
leading to 
\begin{equation}
\frac{d}{dt} Z_{\tilde{W}_k}(t) \leq \mathcal{C}_f Z_{\tilde{W}_k}^{\tilde{s}_{k+1}}(t)~.    
\end{equation}
Since $\text{dim}(\tilde{W}_k) = \tilde{p}_k$, we have that $\lambda_{\tilde{p}_k+1}^2(t) \leq Z_{\tilde{W}_k}(t)$, and  
\begin{align}
    \lambda_{\tilde{p}_k+1}^2(\tilde{\tau}_k) &\leq Z_{\tilde{W}_k}(\tilde{\tau}_k) \leq \left( Z_{\tilde{W}_k}(0)^{1-\tilde{s}_{k+1}} + \mathcal{C}_f \tilde{\tau}_k ( 1 - \tilde{s}_{k+1}) \right)^{1/(1-\tilde{s}_{k+1})}~,
\end{align}
where we have used the Gronwall comparison Lemma \ref{lem:gromwall_adapted}, Eq (\ref{eq:gronwall_upper}). Now, since the regrouped cascade satisfies $\tilde{s}_{k+1} > \tilde{s}_k$, observe that $\mathcal{C}_f \tilde{\tau}_k ( 1 - \tilde{s}_{k+1}) = O(d^{\tilde{s}_k-1}) \ll Z_{\tilde{W}_k}(0)^{1-\tilde{s}_{k+1}} = \Theta(d^{\tilde{s}_{k+1} - 1})$, which implies that 
$\lambda_{\tilde{p}_k+1}(\tilde{\tau}_k) = O( d^{-1/2})$, as claimed. 

Finally, let us establish (\ref{eq:alignment}). $G(t)$ has exactly $\tilde{p}_k$ eigenvalues greater than $(1 - \eta)^2$ for $\tilde{\tau}_k  \ll t \ll \tilde{\tau}_{k+1}$, and $\| \tilde{W}_k^\perp G(t) \| = O(d^{-1}) $ for $t \ll \tilde{\tau}_{k+1}$. Setting $\eta = 1/d$ leads to 
$\lambda_{\text{min}}( \tilde{W}_k G(t) ) \geq (1-\eta)^2 - O(d^{-1})=1-O(d^{-1})$.
We thus have 
\begin{align}
    \| V_{\tilde{p}_k}(t)V_{\tilde{p}_k}(t)^\top - \tilde{W}_k \tilde{W}_k^\top \| &\leq \| G(t) - \tilde{W}_k \tilde{W}_k^\top \| + \| G(t) - V_{\tilde{p}_k}(t)V_{\tilde{p}_k}(t)^\top \| \nonumber \\
    & \leq \sqrt{2(1-\lambda_{\text{min}}(\tilde{W}_k G(t) ) )} + O(d^{-1}) = O(d^{-1/2})~,
\end{align}
proving (\ref{eq:alignment}).

\end{proof}

\begin{proof}[Proof of Lemma \ref{lem:correl_upperbound}]
Recall from Lemma \ref{lem:dotGlemma} that 
$\dot{G} = 2( G \overline{G} + \overline{G} G - 2 G \overline{G} G)$. 
Fix $x \in B^\perp$, $\|x\|=1$,  and consider first $Z_x(t) = x^\top G(t) x$. 
We have 
\begin{align}
    \dot{Z_x} &= x^\top \dot{G}(t) x \nonumber \\
    &= 2 x^\top \left[(I-G) \overline{G} G + G \overline{G} (I-G) \right] x~. 
\end{align}

Since $\overline{G},G \succeq 0$, we have that 
\begin{align}
\label{eq:verm1}
\dot{Z_x} &= x^\top \left[\overline{G} G + G \overline{G} - 2 G \overline{G} G  \right]x \nonumber \\
&\leq x^\top \left[\overline{G} G + G \overline{G} \right]x \nonumber \\
&\leq 2 \| \overline{G} x \| \| G x\| \nonumber\\
&\leq 2 \| \overline{G} x\| Z_B~.    
\end{align}

It is thus sufficient to bound $\| \overline{G} x\|$.
For that purpose, let us consider an orthonormal basis of the form $[B ; \, B^\perp]$ and 
its associated Tensorised Hermite basis $\{H_\beta\}_\beta$. By definition of the relative information exponent, if $f = \sum_\beta \langle f, H_\beta \rangle H_\beta$, we have that 
$|\beta|_{p+1:q} \geq s(f; B)$ whenever $\alpha_\beta:= \langle f, H_\beta \rangle \neq 0$. 

We recall also from Proposition \ref{prop:gradient_loss_G} that 
$\overline{G} = ( \nabla f ) \otimes (\mathsf{A}_G \nabla f)$.
Expressing $f$ in coordinates leads to 
$$\overline{G} = \sum_{\beta, \beta'} \alpha_\beta \alpha_{\beta'} \nabla H_\beta \otimes \mathsf{A}_G \nabla H_{\beta'}~.$$

Suppose w.l.o.g. that $B$ is spanned by the first $p$ canonical vectors, so that $x_j=0$ for $j\in[1,B]$. We have the following coordinate-wise control of $\overline{G}x$:
\begin{align}
\label{eq:verm0}
    |(\overline{G} x)_i| &= \left| \langle \partial_{x_i} f, \mathsf{A}_G (\nabla f \cdot x) \rangle \right| \nonumber \\
    &\leq \| \partial_{x_i} f\| \| \mathsf{A}_G \partial_x f \|~,
\end{align}
where we slightly abuse notation and denote $\partial_x f = \nabla f \cdot x$. We now find a suitable upper bound of $\| \mathsf{A}_G \partial_x f \|$ using the basic properties of averaging operators on psd matrices. 

From the expansion of $f$ in the canonical tensorised Hermite decomposition, and leveraging that $x$ is of the form $x_j=0$ for $j\in[1,B]$, we have that 
\begin{align}
\label{eq:verm}
    \partial_x f &= \nabla f \cdot x \nonumber \\
    &= \sum_\beta \langle f, H_\beta \rangle (\nabla H_\beta \cdot x) \nonumber \\
    &= \sum_{\beta ; |\beta_{p+1:q}| \geq s} \langle f, H_\beta \rangle (\nabla H_\beta \cdot x) ~,
\end{align}
since $\nabla H_\beta \cdot x = 0$ whenever $\beta_i = 0$ for $i=p+1, \ldots q$. The function $\partial_x f$ thus has relative information exponent $s-1$ w.r.t. $B$.
We now use the following property:
\begin{claim}
\label{cla:dirpath}
    If $f$ has relative information exponent $s=s(f; B)$ and $\rho_\perp = \| B^\perp G \|$, then 
    \begin{equation}
        \| \mathsf{A}_G f \|^2 \leq \rho_\perp^{2s} \|f\|^2_{\Xi}~.
    \end{equation}
\end{claim}

From Claim \ref{cla:dirpath} we thus deduce that
\begin{align}
    \| \mathsf{A}_{G} \partial_x f \|^2 & \leq (Z_B)^{2(s-1)} \| \partial_x f \|^2_s ~;
\end{align}
therefore from Eq (\ref{eq:verm0}) we have 
\begin{equation}
    \| \overline{G} x\|^2 \leq \|\nabla f\|^2 (Z_B)^{2(s-1)} \| \partial_x f \|^2_s~,
\end{equation}
and thus from Eq (\ref{eq:verm1}) we obtain 
\begin{equation}
    \dot{Z}_x \leq Z_B^{s} \|\nabla f\|_s ~.
\end{equation}

Finally, observe that $Z_B = \sup_{x \in B^\perp; \|x\|=1} Z_x$. 
From the envelope theorem \cite{milgrom2002envelope}, we therefore have that 
$$\dot{Z}_B(t) \leq \sup_x \dot{Z}_x(t) \leq Z_B^s \|\nabla f\|_s ~,$$
which proves the Lemma.
\end{proof}

\begin{proof}[Proof of Claim \ref{cla:dirpath}]

We show that $\| \mathsf{A}_G f \|^2 \leq \rho_\perp^{2s} \|f\|^2_\Xi$, where $\| \cdot \|_\Xi$ is the Sobolev norm with spectral weights $\Xi(k;q,s) = k^s {c_{q,k}}$.

We first compute, with $\bar{G} = G^\top G$, $\| \mathsf{A}_G f \|^2$. 
\begin{align}
\label{eq:flatt}
   \| \mathsf{A}_G f \|^2 &=  \langle f, \mathsf{A}_{\bar{G}} f \rangle \nonumber \\
   &= \sum_{|\beta| = |\gamma|} \langle H_\beta, \mathsf{A}_{\bar{G}} H_\gamma \rangle \langle f, H_\beta \rangle \langle f, H_\gamma \rangle \nonumber \\
   &= \sum_{\substack{|\beta| = |\gamma| \\ |\gamma|_{p+1:q}\wedge |\beta|_{p+1:q}\geq s}} \langle H_\beta, \mathsf{A}_{\bar{G}} H_\gamma \rangle \langle f, H_\beta \rangle \langle f, H_\gamma \rangle~. 
\end{align}
We now compute $\langle \mathsf{A}_{\bar{G}} H_\gamma, H_\beta \rangle$ for $|\gamma|=|\beta|$ and $|\gamma|_{p+1:q}\wedge |\beta|_{p+1:q}\geq s$. 
Using Lemma \ref{lem:hermitechange} and Proposition \ref{prop:ave_closed} we have 
\begin{align}
\label{eq:cornell}
    \langle \mathsf{A}_{\bar{G}} H_\beta, H_\gamma \rangle &= \langle \mathsf{P}_{\bar{G}} H_\beta, H_\gamma \rangle \nonumber \\
    &= \sum_{{\Tau} \in \Pi(\beta, \gamma)} Q({ \Tau}; \beta, \gamma)^{1/2} \left(\prod_{i,j=1}^q \bar{G}_{i,j}^{\Tau_{i,j}}\right) ~.
\end{align}
Let $\tilde{\beta}=( |\beta|-|\beta|_{p+1:q}; \beta_{p+1:q})$ and $\tilde{\gamma}$ defined analogously. 
We claim that 
\begin{align}
    \left|\langle \mathsf{A}_{\bar{G}} H_\beta, H_\gamma \rangle\right| &\lesssim \rho_\perp^{2s} \binom{|\tilde{\beta}|}{\tilde{\beta}}^{1/2} \binom{|\tilde{\gamma}|}{\tilde{\gamma}}^{1/2}~.  
\end{align}
Indeed, from \eqref{eq:cornell}, observe that 
for $i,j \in [p+1,q] $ we have $|\bar{G}_{i,j}| = |G_i \cdot G_j| \leq \rho^2$, 
and similarly for $(i,j) \in [p+1,q] \times [1,p]$ we have $|\bar{G}_{i,j}| \leq \rho$. Since $|\beta|_{p+1:q} \geq s$ and $|\gamma|_{p+1:q} \geq s$, this implies that for any ${\Tau} \in \Pi(\beta, \gamma)$ we have 
$$\frac{\prod_{i,j=1}^q \bar{G}_{i,j}^{\Tau_{i,j}} }{\prod_{i,j=1}^p \bar{G}_{i,j}^{\Tau_{i,j}} } \leq \rho^{|\beta|_{p+1:q} + |\gamma|_{p+1:q}} \leq \rho^{2s}~.$$ 
Moreover, for large $k=|\gamma|=|\beta|$, using Laplace's approximation, the term $\prod_{i,j=1}^p \bar{G}_{i,j}^{\Tau_{i,j}}$ will concentrate the sum over $\Tau$ onto the largest entries of $\bar{G}$, thus 
\begin{align}
\label{eq:cornell2}
\left| \langle \mathsf{A}_{\bar{G}} H_\beta, H_\gamma \rangle\right| &\leq 
\sum_{{\Tau} \in \Pi(\beta, \gamma)} Q({ \Tau}; \beta, \gamma)^{1/2} \prod_{i,j=1}^p \bar{G}_{i,j}^{\Tau_{i,j}} \frac{\prod_{i,j=1}^q \bar{G}_{i,j}^{\Tau_{i,j}} }{\prod_{i,j=1}^p \bar{G}_{i,j}^{\Tau_{i,j}} } \nonumber \\
&\leq \rho^{|\beta|_{p+1:q} + |\gamma|_{p+1:q}} \sum_{{\Tau} \in \Pi(\beta, \gamma)} Q({ \Tau}; \beta, \gamma)^{1/2} \prod_{i,j=1}^p \bar{G}_{i,j}^{\Tau_{i,j}} \nonumber \\
& \lesssim \rho^{|\beta|_{p+1:q} + |\gamma|_{p+1:q}} \sum_{{\tilde{\Tau}} \in \Pi(\tilde{\beta}, \tilde{\gamma})} Q({ \tilde{\Tau}}; \tilde{\beta}, \tilde{\gamma})^{1/2}~.
\end{align}
Now, observe that 
\begin{align}
\label{eq:cornell3}
    \sum_{{\tilde{\Tau}} \in \Pi(\tilde{\beta}, \tilde{\gamma})} Q({ \tilde{\Tau}}; \tilde{\beta}, \tilde{\gamma})^{1/2} &= q^k \langle H_{\tilde{\gamma}}, \mathsf{A}_{q^{-1} 11^\top} H_{\tilde{\beta}} \rangle \nonumber \\
    &= \binom{k}{\tilde{\beta}}^{1/2} \binom{k}{\tilde{\gamma}}^{1/2}~,
\end{align}
using again Lemma \ref{lem:hermitechange} and Proposition \ref{prop:ave_closed}.

Now, consider the operator $\mathsf{G}_k: \mathsf{\Pi}_k \to \mathsf{\Pi}_k$ 
with kernel 
$$ [\mathsf{G}_k]_{\beta,\gamma} = \langle H_\beta, \mathsf{A}_{\bar{G}} H_\gamma \rangle \cdot {\mathbf{1}}( |\gamma|_{p+1:q} \wedge |\beta|_{p+1:q} \geq s)~.$$ 
Equations \eqref{eq:cornell2} and \eqref{eq:cornell3} therefore provide pointwise control of the form
$$| [\mathsf{G}_k]_{\beta,\gamma} | \leq \rho^{|\beta|_{p+1:q} + |\gamma|_{p+1:q}} \binom{k}{\tilde{\beta}}^{1/2} \binom{k}{\tilde{\gamma}}^{1/2} \cdot {\mathbf{1}}( |\gamma|_{p+1:q} \wedge |\beta|_{p+1:q} \geq s) ~,$$
which by Schur's lemma implies that 
has norm $\| \mathsf{G}_k \| \leq \rho^{|\beta|_{p+1:q} + |\gamma|_{p+1:q}}_\perp \Xi_{k,(|\beta|_{p+1:q} + |\gamma|_{p+1:q})/2,q}$, 
where $\Xi_{k,l,q}=k^l {c_{q,k}}$, and where we recall that $c_{q,k}=\text{dim}(\mathsf{\Pi}_k)$. From (\ref{eq:flatt}), we thus conclude that 
\begin{align}
    \| \mathsf{A}_G f\|^2 &\leq \sum_{k \geq s} \| \mathsf{G}_k \| \| \mathsf{\Pi}^k f \|^2 \nonumber \\
    &\leq \rho_\perp^{2 s} \sum_{k \geq s}   \Xi_{|\beta|,s,q}  \| \mathsf{\Pi}^k f \|^2 \nonumber \\
    & := \rho_\perp^{2 s} \|f \|_\Xi^2~,
\end{align}
as claimed. 

\end{proof}

%% file: Appendix/App_planted_case.tex
\subsection{Stiefel Gradient flow}
\label{subsecapp:stiefel_equations}

We consider the gradient flow dynamics
\begin{align}
    \dot{W} &= \nabla_W^{\mathcal{S}} L(W)~,
\end{align}
where $\nabla_W^{\mathcal{S}}$ is the Stiefel Gradient.
Using Eq (\ref{eq:stiefelgrad}) and Lemmas \ref{lem:towsend}, \ref{lem:towsend2} we can now derive the explicit dynamics of the summary statistics $U, \Lambda, V$. 

\begin{lemma}
\label{lem:stiefel_summary}
    Let $\ell(U, \Lambda, V) = \sum_\beta \alpha_\beta(V) \alpha_\beta(U) \lambda_\beta$, and 
    $L(W) = \ell( U, \Lambda, V)$, where $ V \Lambda U^\top$ is the Singular Value Decomposition of $M = W_*^\top W$, and consider the gradient flow dynamics $\dot{W} = \nabla_W^{\mathcal{S}}L(W)$. We have
\begin{align}
    \nabla_W^{\mathcal{S}} L(W) &=  W_* V  \Xi U^\top  - W U \Xi^\top \Lambda U^\top ~,
\end{align}
where we defined 
    \begin{align}
        \label{eq:xi}
        \Xi :=  \overline{\Lambda} +  \left( F_\lambda \circ \overline{U}^a \right) \Lambda + \Lambda \left( F_\lambda \circ {\overline{V}^a} \right)~,
    \end{align}
    with notation that are defined in Lemma~\ref{lem:towsend}. 
Moreover, the summary statistics $U(t), V(t)$ and $\Lambda(t)$ and $M(t)$ follow the dynamics
\begin{align}
\label{eq:motion_stiefel_lambda}
    \dot{\Lambda} &= \left( I - \Lambda^2\right) \overline{\Lambda} ~,\\
\label{eq:motion_stiefel_U}    
    \dot{U} &=  U \left[ B_\lambda \circ \overline{U}^a + C_\lambda \circ \overline{V}^a  \right]    ~, \\
\label{eq:motion_stiefel_V} \dot{V} &=  V \left[ C_\lambda \circ \overline{U}^a + B_\lambda \circ \overline{V}^a  \right] ~, 
 \end{align}
where we have defined the matrices for $i \neq j$, $B_\lambda[ij] := \frac{\lambda_j^2 ( 1 - \lambda_i^2) + \lambda_i^2 ( 1 - \lambda_j^2)}{(\lambda_j^2 - \lambda_i^2)^2}$, $C_\lambda[ij] := \frac{\lambda_i \lambda_j ( 2 - (\lambda_i^2 + \lambda_j^2))}{(\lambda_j^2 - \lambda_i^2)^2}$, and $D_\lambda[ij] := \frac{\lambda_j ( 1 - \lambda_i^2)}{\lambda_j^2 - \lambda_i^2}$ with zeros in the diagonal.
\end{lemma}
\begin{proof}
    Let $M = W_*^\top W $. Using Lemma \ref{lem:towsend} we obtain 
    \begin{align*}
    \overline{M} &= V \Xi U^\top ~,
    \end{align*}
    Notice that the diagonal of $\Xi$ coincides with $\overline{\Lambda}$, while the off-diagonal coincides with the two other terms that are symmetric matrices with zero diagonal. 
    It follows that $\overline{W} = W_* \overline{M} $, and, applying (\ref{eq:stiefelgrad}), we obtain
    \begin{align}
        \nabla_W^{\mathcal{S}} L(W) &= W_* \overline{M}  - W  \overline{M}^\top W_*^\top   W~ \nonumber \\
        &= W_* V \Xi U^\top - W U \Xi^\top V^\top V \Lambda U^\top \nonumber \\
        &= W_* V \Xi U^\top  - WU \Xi^\top \Lambda U^\top~. 
    \end{align}
Let us now compute $\frac{d}{dt} \Lambda$, $\frac{d}{dt} U$ and $\frac{d}{dt} V$ from 
$\frac{d}{dt} W$ that we just computed. 
We have 
\begin{align*}
\dot{M} & =  W_*^\top \dot{W}  \\
& =   V \Xi U^\top  - M U \Xi^\top \Lambda U^\top \\
& =  V \Xi U^\top - V \Lambda \Xi^\top \Lambda V^\top \\
& =  V \left[\Xi - \Lambda \Xi^\top \Lambda \right] U^\top.
\end{align*}
Using Lemma \ref{lem:towsend2} we finally obtain for the dynamics of $\Lambda$:
\begin{align*}
        \dot{\Lambda} &=  I \circ \left[U^\top U \left[\Xi - \Lambda \Xi^\top \Lambda \right] V^\top V \right]  \\
         &= I \circ \left[ \Xi - \Lambda \Xi^\top \Lambda \right]   \\
         &= \left( I - \Lambda^2\right) \overline{\Lambda}.
\end{align*}
We also obtain the dynamics over the left eigenvectors:
\begin{align*}
        \dot{v} &=  V \left[  F \circ \left(  \left[\Xi - \Lambda \Xi^\top \Lambda \right]\Lambda  + \text{sym}  \right)   \right] \\
         &= V \left[  F \circ \left(  \left[\left( F \circ \overline{U}^a \right) \Lambda + \Lambda \left( F \circ {\overline{V}^a} \right) - \Lambda \left(\left( F \circ \overline{U}^a \right) \Lambda + \Lambda \left( F \circ {\overline{V}^a} \right) \right)^\top \Lambda \right] \Lambda  + \text{sym}  \right)   \right] \\
         &= V \left[  F \circ \left(  \left[\left( F \circ \overline{U}^a \right) \Lambda - \Lambda^2 \left( F \circ \overline{U}^a \right) \Lambda + \Lambda \left( F \circ {\overline{V}^a} \right) - \Lambda  \left( F \circ {\overline{V}^a} \right)  \Lambda^2 \right] \Lambda  + \text{sym}  \right)   \right].
\end{align*}
If we define as in the statement of the Lemma, the matrices $B_\lambda[ij] := \frac{\lambda_j^2 ( 1 - \lambda_i^2) + \lambda_i^2 ( 1 - \lambda_j^2)}{(\lambda_j^2 - \lambda_i^2)^2}$ and $C_\lambda[ij] := \frac{\lambda_i \lambda_j ( 2 - (\lambda_i^2 + \lambda_j^2))}{(\lambda_j^2 - \lambda_i^2)^2}$, with zeros in the diagonal, then 
\begin{align*}
        \dot{V} &=  V \left[ B_\lambda \circ \overline{U}^a + C_\lambda \circ \overline{V}^a  \right].
\end{align*}
Similarly, we obtain the dynamics over the right eigenvectors:
\begin{align*}
        \dot{U} &=  U \left[  F \circ \left( \Lambda  \left[\Xi - \Lambda \Xi^\top \Lambda \right]   + \text{sym}  \right)   \right] \\
         &= U \left[  F \circ \left( \Lambda \left[\left( F \circ \overline{U}^a \right) \Lambda + \Lambda \left( F \circ {\overline{V}^a} \right) - \Lambda \left(\left( F \circ \overline{U}^a \right) \Lambda + \Lambda \left( F \circ {\overline{V}^a} \right) \right)^\top \Lambda \right]   + \text{sym}  \right)   \right] \\
         &= U \left[  F \circ \left( \Lambda \left[\left( F \circ \overline{U}^a \right) \Lambda - \Lambda^2 \left( F \circ \overline{U}^a \right) \Lambda + \Lambda \left( F \circ {\overline{V}^a} \right) - \Lambda  \left( F \circ {\overline{V}^a} \right)  \Lambda^2 \right]  + \text{sym}  \right)   \right].
\end{align*}
That is, with the same matrices as for the left eigenvectors, we have
\begin{align*}
        \dot{U} &=  U \left[ C_\lambda \circ \overline{U}^a + B_\lambda \circ \overline{V}^a  \right].
\end{align*}
\end{proof}

\subsection{Study of the radial case: proof of Theorem \ref{thm:planted_radial_case}}
\label{app:radial_case}
First, recall that the loss function writes as a function of $\Lambda$: $\ell(\Lambda) = \sum_\beta \lambda_\beta \alpha_\beta^2$, where  for all $\beta \in \N^q$, we have $\lambda_\beta = \prod_i \lambda_i^{\beta_i}$.
We recall that we have the following dynamics on $\Lambda$: for all $i \in \llbracket 1 ,q \rrbracket$,
\begin{align}
\label{eq:movementlambdaraddial}
\dot{\lambda_i^2}(t) &= 2 (1 - \lambda_i(t)^2) \sum_\beta \beta_i\alpha_\beta^2 \prod_{k = 1}^q \lambda_k^{\beta_k}~,
\end{align}
where $\lambda$ are distributed according to the joint law written in Eq.\eqref{eq:multivariate_law_eigenvalues}. 

\paragraph{Point (i) of Theorem \ref{thm:planted_radial_case}: convergence.} As $(W_t)_{t\geq 0}$ is bounded, up to extraction, it converges. From the fact that $W$ follows a gradient flow, the only possible points of accumulation are the critical points of $L$. But now, from equation \eqref{eq:movementlambdaraddial}, we see that the only critical point of the loss landscape corresponds to $\lambda \in \{0,1\}^q$. As the $\lambda_i$, for all $i$ are non-decreasing functions of time, all the critical points corresponding the some coordinates being zero cannot been attained by the gradient flow. Hence the only critical point meeting the gradient flow is $\Lambda = I_q$, this is why $\Lambda_t  \to I_q$. This corresponds to $L(W_t) \to 0$ and $W_tW_t^\top \to W_* W_*^\top$ as announced is the Theorem. 
\paragraph{Point (ii) of Theorem \ref{thm:planted_radial_case}: rate of convergence.} Recall that the probability of the event that $\lambda_q \geq \frac{b}{\sqrt{d}}$ is larger than $1-n_q b$, as per Lemma \ref{lem:eigenvalues_init}. We place ourselves on this event. Let us fix any $\eta > 0$, and $\tau_{1/2}$ the time that $\lambda_q$ has reached $1/2$ and $\tau_{\eta}$ the time it has reached $1 - \eta$. From equation \eqref{eq:movementlambdaraddial}, for any $t \leq \tau_{1/2}$, we have the following inequality
\begin{align*}
\dot{\lambda_q^2}(t) &= 2 (1 - \lambda_i(t)^2) \sum_\beta \beta_q \alpha_\beta^2 \prod_{k = 1}^q \lambda_k^{\beta_k} \\
&\geq \sum_{|\beta| = s} \beta_q \alpha_\beta^2 \prod_{k = 1}^q \lambda_k^{\beta_k}  \\
&\geq \lambda_q^{s}\sum_{|\beta| = s, \beta_q > 0} \alpha_\beta^2   \\
&\geq \lambda_q^{s}\, \mathcal{E}(\S^s f)~.
\end{align*}
Hence, from Lemma~\ref{lem:gromwall_adapted}, we have that 

\noindent \textbf{Case $ s \geq 3$.}
\begin{align}
    \lambda^2_{q}(\tau_{1/2}) &\geq \left[\frac{1}{ [\lambda^2_{q}(0)]^{s/2-1}} - (s/2-1)  \mathcal{E}(\S^s f)  \tau_{1/2} \right]^{-\frac{1}{s/2-1}} \nonumber\\
     &\frac12 \geq \left[\frac{d^{ s/2-1}}{ b^{s-2}} - (s/2-1) \mathcal{E}(\S^s f)  \tau_{1/2} \right]^{-\frac{1}{s/2-1}}~,
\end{align}
and hence, we have the upper bound
\begin{align}
    \tau_{1/2} \leq \Cu \delta^{2-s} d^{ s/2-1}:= \mathcal{C}_{f,\delta} d^{ s/2-1} ~,
\end{align}
holding with probability greater than $1 - \delta$. 

\noindent \textbf{Case $ s = 2$.}
\begin{align}
    \lambda^2_{q}(\tau_{1/2}) &\geq \lambda^2_{q}(0) e^{ \mathcal{E}(\S^s f)  \tau_{1/2} } \nonumber\\
     &\frac12 \geq \frac{b^2}{d}e^{ \mathcal{E}(\S^s f) \tau_{1/2}}  ~,
\end{align}
and hence, we have the upper bound
\begin{align}
    \tau_{1/2} \leq \mathcal{E}(\S^s f)^{-1} \log(d/(2 \delta^2)) \leq (\Cu + \log(1/\delta))  \log(d)~,
\end{align}
holding with probability greater than $1 - \delta$.

Now we can pursue the same reasoning for times $ t \in [\tau_{1/2},  \tau_{\eta}]$, except that equation 
\begin{align*}
     \frac{d}{dt}[1 - \lambda_{q}^2(t)] &\leq - (1 - \lambda_{q}^2(t))\, \mathcal{E}(\S^s f)~ .
\end{align*}
Hence, by a standard Grownwall inequality, for times $ t \in [\tau_{1/2},  \tau_{\eta}]$,
\begin{align*}
    1 - \lambda_{q}^2(t) &\leq \frac{3}{4} e^{\Cu (t - \tau_{1/2})} .
\end{align*}
Thus, by a standard comparison argument, we have that $$\tau_{\eta} \leq \tau_{1/2} + \Cu \log(1/\eta) \leq \mathcal{C}_{f,\delta,s} \left[ d^{s/2 - 1}  + \log(1/\eta) \right]~. $$
To conclude the proof of the theorem, we observe that after time $t\geq \tau_\eta$ we have that all eigenvalues are greater than $1-\eta$, and therefore  
\begin{align*}
    \|W(t) W(t)^\top - W_* W_*^\top \|^2 & \leq 2 - 2 \lambda_{\text{min}}(W(t) W(t)^\top W_* W_*^\top) \nonumber \\
    &= 2( 1- \lambda_{\text{min}}( W(t)^\top W_* W_*^\top W(t)) ) \nonumber \\
    &= 2( 1 - \lambda_{q}^2(t)) \nonumber \\
    &\leq 4 ( 1- \lambda_q(t)) \nonumber \\
    & \leq 4 \eta~,
\end{align*} 
and this concludes the proof of the theorem.

\subsection{Study of the failure case: proof of Theorem \ref{thm:planted_failure}}
\label{sec:app_failure_stiefel}

We begin by detailing some basic properties of the function~$\varphi$, as well as deriving basic facts about the gradient dynamics on~$f^*$.

\paragraph{Properties of~$\varphi$.} Recall that~$N$ is odd and~$s$ is even. We have:
\begin{align*}
    \varphi(u) &= 4 \cos(u)^s + 2 \sum_{j=1}^{N-1} \cos(u + \frac{2\pi j}{N})^s + 2 \sum_{j=1}^{N-1} \cos(-u + \frac{2\pi j}{N})^s+ \sum_{j,j'=1}^{N-1} \cos(u + \frac{2\pi}{N}(j - j'))^s \\
    &= 4 \cos(u)^s + 2 \sum_{j=1}^{N-1} \cos(u + \frac{2\pi j}{N})^s + 2 \sum_{j=1}^{N-1} \cos(u - \frac{2\pi j}{N})^s+ \sum_{j,j'=1}^{N-1} \cos(u + \frac{2\pi}{N}(j - j'))^s \\
    &= 4 \cos(u)^s + 4 \sum_{j=1}^{N-1} \cos(u - \frac{2\pi j}{N})^s + \sum_{j,j'=1}^{N-1} \cos(u + \frac{2\pi}{N}(j - j'))^s \\
    &= 4 \cos(u)^s + 4 \sum_{j=1}^{N-1} \cos(u - \frac{\pi j}{N})^s + \sum_{j,j'=1}^{N-1} \cos(u + \frac{\pi}{N}(j - j'))^s,
\end{align*}
where the last equality follows from the $\pi$-periodicity of~$\cos(\cdot)^s$ for even~$s$ and the fact that~$N$ is odd.
We have that~$\varphi$ is $\pi$-periodic, even, and achieves its global maximum at~$0$. Further, since~$\cos(u)^s$ converges to~$\mathbf{1}_{u=0}$ for~$s \to \infty$, for some fixed~$0 < \gamma \ll \frac{\pi}{2N}$, say~$\gamma = \frac{\pi}{10N}$, we may choose~$s$ large enough such that~$\cos(\gamma)^s \leq 1/10N^2$, and we have the following properties:
\begin{itemize}
    \item The maximal value of~$\varphi$ is achieved at~$u_0^+ = 0$ and satisfies~$\varphi(0) \geq N+3$, since all terms are negligible except the first term and the~$N-1$ elements with~$j=j'$ in the double sum;
    \item $\varphi$ has $N-1$ suboptimal peaks at locations~$u_k^+$, $k = 1, \ldots, N-1$ with~$|u_k^+ - \frac{k\pi}{N}| \leq \gamma$ and a value that satisfies~$|\varphi(u_k^+) - (N + 2)| \leq 1/3$, since all terms are negligible except the term $4\cos(u - \pi k/N)^2$ in the first sum as well as the terms in the double sum with~$j - j' = -k \mod N$, of which there are~$N-2$ when~$k \ne 0$;
    \item Defining~$u_k^- = \frac{2\pi k + 1}{2N}$ for $k = 0, \ldots, N-1$, which satisfies~$u_k^- \in [u_k^+, u_{k+1}^+]$ (where~$u_N^+$ is identified with~$u_0^+$ due to $\pi$-periodicity), we have~$\varphi(u_k^-) < 1$ for each~$k$, since all terms in~$\varphi$ are upper bounded by~$\cos(\gamma)^s \leq 1/10N^2$.
    In particular, we will define~$\mu := u_0^-$, and note that~$u_{N-1}^- = -\mu \mod \pi$, so that~$\pm \mu$ are both ``barriers'' with low correlation.
    \item Since~$\varphi(\mu) < 1$ and~$\varphi(u_1^+) \geq N+2 - 1/3$, by the intermediate value theorem we may choose~$\nu \in [\mu, \frac{\pi}{N} + \gamma]$ such that~$\varphi(\nu) = \varphi(\mu) + 1$.
\end{itemize}

\paragraph{Loss specification and equations of motion}

 Since $f$ and $g$ have energy in disjoint harmonic orders, we have the following decomposition, denoting~$W_*^\top W  = V \Lambda U^\top$:
    \begin{align*}
    L(W) &= \sum_{|\beta|=2} \alpha_\beta^f(U)  \alpha_\beta^f(V)  \lambda_\beta +\epsilon\sum_{|\beta|=s} \alpha_\beta^g(U)\alpha_\beta^g(V) \lambda_\beta \nonumber \\
    &= \sum_{|\beta|=2} (\alpha_\beta^f)^2 \lambda_\beta + \epsilon \sum_{|\beta|=s} \alpha_\beta^g(U)\alpha_\beta^g(V) \lambda_\beta \nonumber \\
    &= \frac{1}{2}\left(\lambda_1^2 + \lambda_2^2\right) + \epsilon\sum_{|\beta|=s} \alpha_\beta^g(U)\alpha_\beta^g(V) \lambda_\beta \nonumber \\
    &:= \ell_f(\Lambda) + \epsilon \ell_g(U, \Lambda, V)~,    
    \end{align*}
    since $f$ is radial. Therefore, thanks to equation \eqref{eq:motion_stiefel_lambda}, we have for $i=1,2$, 
\begin{align*}
    \dot{\lambda}_i &= (1-\lambda_i^2) \left( \partial_{\lambda_i} \ell_f + \epsilon \partial_{\lambda_i} \ell_g \right)~\nonumber \\
    &= (1-\lambda_i^2) \left(\lambda_i + \epsilon \partial_{\lambda_i} \ell_g(U, \Lambda, V)\right)~.
\end{align*}
We will now derive appropriate bounds for the `perturbation' terms $\partial_{\lambda_i} \ell_g(U, \Lambda, V)$. For $i=1,2$, we have 
\begin{align*}
    \partial_{\lambda_i} \ell_g(U, \Lambda, V) &= \partial_{\lambda_i} \sum_{|\beta|=s} \alpha_\beta^g(U) \alpha_\beta^g(V) \lambda_\beta \\
    &=  \begin{cases}
            \sum_{k=1}^s k \alpha_{(k,s-k)}^g(U) \alpha_{(k,s-k)}^g(V) \lambda_1^{k-1} \lambda_2^{s-k}~, & \text{ if } i=1 ~,\\
           \sum_{k=1}^s k \alpha_{(s-k,k)}^g(U) \alpha_{(s-k,k)}^g(V) \lambda_2^{k-1} \lambda_1^{s-k}~, & \text{ if } i=2~. 
        \end{cases} 
\end{align*}
It follows that, by Cauchy-Schwartz inequality, we have for $i = 1,2$,
\begin{equation}
  |\partial_{\lambda_i} \ell_g(U, \Lambda, V)| \leq \lambda_1^{s-1} s \|g\|^2= c \lambda_1^{s-1}~,  
\end{equation}
with $c = s \|g\|^2$. Therefore, we obtain the following sandwich inequalities
\begin{align}
\label{eq:odes1}
(1-\lambda_1^2)( \lambda_1 - c \epsilon \lambda_1^{s-1} ) &\leq \dot{\lambda}_1 \leq (1-\lambda_1^2)( \lambda_1 + c \epsilon \lambda_1^{s-1} ) ~, \\
\label{eq:odes2}
(1-\lambda_2^2)( \lambda_2 - c \epsilon \lambda_1^{s-1} ) &\leq \dot{\lambda}_2 \leq (1-\lambda_2^2)( \lambda_2 + c \epsilon \lambda_1^{s-1} )  ~.
\end{align}
with $\lambda_1(0) = \delta$ and $\lambda_2 = a \delta$, with $a \in (0,1)$. The equations of motion for $(U, V)$ are not written here for the sake of concision but can be deduced from Eqs.\eqref{eq:motion_stiefel_U}-\eqref{eq:motion_stiefel_V}.  

\paragraph{Phase-I: escaping mediocrity}

The aim of this phase is to show that we can follow very closely the movement of the eigenvalues and show that they will escape mediocrity, thanks to the radial part, according to the following movement: $\lambda_1 \sim \delta e^{t}$ and $\lambda_2 \sim a \delta e^{t}$.

\begin{lemma}[Mediocrity Phase]
\label{lem:mediocrity_planted}
    For all $ t \leq \tau_1 :=  \log(1/\delta) - \log(2)/2$, the following inequalities hold true:
    \begin{align}
    \label{eq:sandwich_all}
    \frac{1}{\sqrt{ \left( \delta^{-2} - 2 \right)e^{-2t} + 2}}  &\leq \lambda_1 \leq \frac{1}{\sqrt{ \left( \delta^{-2} + 1 \right)e^{-2t} - 1}} \leq \sqrt{2} \delta e^t~, \\
     \frac{ a \delta e^t}{\sqrt{1 + a^2}} \leq \frac{1}{\sqrt{ \left( (a\delta)^{-2} - 2 \right)e^{-2t} + 2}}  &\leq \lambda_2 \leq \frac{1}{\sqrt{ \left( (a\delta)^{-2} + 1 \right)e^{-2t} - 1}}~.
\end{align}

\end{lemma}
Before providing the lemma's proof, let us comment on it. This lemma shows that up to a constant factor, and until a time $\Theta(\log(1/\delta))$, the dynamics of $\lambda_1$ and $\lambda_2$, are close to exponential with rate $1$. Note that at the end of this phase, $\lambda_1(\tau_1) \geq 1/2$, and $\lambda_2(\tau_1) \geq a/\sqrt{2(1 + a^2)}$. This means that they have reached a constant level: ``mediocrity" has been escaped.
\begin{proof}    
Let us first sandwich the $\lambda_1$ dynamics. In fact, as $(1-\lambda_1^2)( \lambda_1 - c \epsilon \lambda_1^{s-1} ) \leq \dot{\lambda}_1 \leq (1-\lambda_1^2)( \lambda_1 + c \epsilon \lambda_1^{s-1} )$, we have in fact that
\begin{align*}
    \lambda_1-\lambda_1^3  - c \epsilon \lambda_1^{s-1}  &\leq \dot{\lambda}_1 \leq \lambda_1 + c \epsilon \lambda_1^{s-1}
\end{align*}
Now using that  $c \epsilon \lambda_1^{s-1} \leq \lambda_1^3$, for $\epsilon \leq 1/c$ and $s \geq 4$, we have
\begin{align*}
    \lambda_1- 2\lambda_1^3  &\leq \dot{\lambda}_1 \leq \lambda_1 + \lambda_1^{3}~.
\end{align*}
Now, thanks to Gronwall inequality and Lemma~\ref{lem:bernoulli_ODE}, this corresponds to, 
\begin{align*}
   \frac{1}{\sqrt{ \left( \delta^{-2} - 2 \right)e^{-2t} + 2}}  &\leq \lambda_1 \leq \frac{1}{\sqrt{ \left( \delta^{-2} + 1 \right)e^{-2t} - 1}}~,
\end{align*}
for all $t < \frac{1}{2} \log\left( 1 + \delta^{-2} \right)$. Now let's do the same sandwich for the dynamics of $\lambda_2$. We have 
\begin{align*}
    \lambda_2-\lambda_2^3  - c \epsilon \lambda_1^{s-1}  &\leq \dot{\lambda}_2 \leq \lambda_2 + c \epsilon \lambda_1^{s-1}
\end{align*}
Now define $\bar{\tau}:= \inf\{ t \geq 0\, ;\,c \epsilon \lambda_1^{s-1} \geq \lambda_2^3 \}$. We have that $\bar{\tau} > 0$ by continuity and, for all $t \leq \bar{\tau}$,
\begin{align*}
    \lambda_2- 2\lambda_2^3  &\leq \dot{\lambda}_1 \leq \lambda_2 + \lambda_2^{3}~.
\end{align*}
so that similarly as for $\lambda_1$ for all $t \leq \frac{1}{2} \log\left( 1 + \delta^{-2} \right) \wedge \bar{\tau}$, 
\begin{align}
\label{eq:sandwich_lambda_2}
   \frac{1}{\sqrt{ \left( (a\delta)^{-2} - 2 \right)e^{-2t} + 2}}  &\leq \lambda_2 \leq \frac{1}{\sqrt{ \left( (a\delta)^{-2} + 1 \right)e^{-2t} - 1}}~,
\end{align}
From basic analysis we can show that $\bar{\tau} \geq \log(1/\delta) - \frac{1}{2} \log(2)$ and conclude. Now, the inequalities with respect to the ``simple" functions, lower bound for $\lambda_2$ and upper bound for $\lambda_1$, follow from standard inequalities. 

\end{proof}

\paragraph{Phase-II: Fast increase.} Let $\eta \in (0, 1)$ and define $\tau_2 = \inf \{ t \geq 0\, ;\, \lambda_2(t) \geq 1-\eta \}$. We want to show that until $\tau_2$ the increase of $\lambda_1, \lambda_2$ towards one is still exponential. We show that meanwhile the left and right eigenvectors move only by a small amount.

\begin{lemma}
\label{lem:fast_increase}
    For $t \in [\tau_1, \tau_2)$, the functions $t \to \lambda_1(t)$ and $ t \to \lambda_2(t)$ are non-decreasing functions. Furthermore we have the upperbound: $\tau_2 \leq \tau_1 + \frac{2}{a} \log(1/\eta) $. 
\end{lemma}
Before proving the result, let us state the main consequence of this second phase: $(\lambda_1, \lambda_2)$ are in the interval $(1 - \eta, 1)$ after a time $\log(1/\delta) + 2 \log(1/\eta) / a - \log(2)/2$.
\begin{proof}
    From the lower bound presented in Eqs.\eqref{eq:odes1}-\eqref{eq:odes2}, we have $ \dot{\lambda}_1 \geq (1 - \lambda_1^2)(\lambda_1 - c \epsilon \lambda_1^{s-1})$. In $t = \tau_1$, $\dot{\lambda}_1(\tau_1) \geq 3/4 (1/2 - c \epsilon /2^{s-1}) > 0$. Hence, defining $\bar{\tau}_1 = \inf\{t \geq \tau_1 ; \dot{\lambda}_1(t) < 0\}$, we see that $\bar{\tau}_1 = + \infty$, since if $\bar{\tau}_1 < + \infty$, this would mean that $ 1/2 = \lambda_1{(\tau_1)} \leq  \lambda_1(\bar{\tau}_1) \leq c \epsilon \lambda_1^{s-1}(\bar{\tau}_1) \leq c \epsilon$, which is a contradiction since we take $\epsilon \leq  1/(4c)$. 
Similarly, $t \to \lambda_2(t)$ is non-decreasing. Let us upper bound $\tau_2$ now. For all $t \geq \tau_1$, define $v(t) = 1 - \lambda_2 (t)$, 
    \begin{align*}
        \dot{v}(t) &\leq -(1 - \lambda_2^2)(\lambda_2 - c \epsilon \lambda_1^{s-1}) \\
        &\leq -v(t)(\lambda_2(\tau_1) - c \epsilon ) \\
        &\leq -v(t)\left(\frac{a}{\sqrt{3}} - c \epsilon \right) \\
        &\leq -\frac{a}{2}v(t),
    \end{align*}
    if $\epsilon < a/(20c)$. Then we have $v(t) \leq v(\tau_1) e^{-a(t-\tau_1)/2} \leq  e^{-a(t-\tau_1)/2}$, that is $\tau_2 \leq \tau_1 + \frac{2}{a} \log(1/\eta)$.    
\end{proof}
We also prove that until time $t \leq \tau_2$, the torsion, defined as $Q_t := U_t V_t^\top $, will have only moved slightly.
\begin{lemma}
\label{lem:upperbound_general_torsion}
For all $t \geq 0$, we have the following upper bound on the torsion $Q_t = U_t V_t^\top$,
\begin{align*}
\| Q(t) - Q(0) \| &\leq 32 \epsilon s \|g\|  \|\nabla g\| \int_0^t \frac{\lambda_1^s(t')}{\lambda_2^2(t')} dt'~,
\end{align*}    
\end{lemma}
Before proving this Lemma, let us quantify the amount of movement of $Q$ thanks to the bounds on $\lambda_1, \lambda_2$.
\begin{lemma}
\label{lem:upperbound_quantitative_torsion}
We have the following upper bound
\begin{align*}
\| Q(\tau_2) - Q(0) \| &\leq \frac{250 \epsilon  s \|g\| \|\nabla g\|}{a^3}\left(1 + \log(1/\eta)\right)  ~.
\end{align*}    
\end{lemma}
\begin{proof}[Proof of Lemma~\ref{lem:upperbound_quantitative_torsion}]
We divide the torsion movement in the two phases $t \in (0, \tau_1)$ and $t \in (\tau_1, \tau_2)$.  We have
\begin{align*}
\| Q(\tau_2) - Q(0) \| &\leq 32 \epsilon s \|g\|  \|\nabla g\| \int_0^{\tau_1} \frac{\lambda_1^s(t')}{\lambda_2^2(t')} dt' + 32 \epsilon s \|g\|  \|\nabla g\| \int_{\tau_1}^{\tau_2} \frac{\lambda_1^s(t')}{\lambda_2^2(t')} dt'~.
\end{align*}  
For the first term, we use the bounds given by Lemma~\ref{lem:mediocrity_planted}, i.e. $\lambda_1 \leq \sqrt{2} \delta e^t$ and $ \frac{ a \delta e^t}{\sqrt{1 + a^2}} \leq  \lambda_2 $. Hence, 
\begin{align*}
\int_0^{\tau_1} \frac{\lambda_1^s(t')}{\lambda_2^2(t')} dt' &\leq  \frac{2 2^{s/2}}{a^2} \delta^{s-2}  \int_0^{\tau_1}e^{(s-2)t'} dt' \\
&\leq  \frac{2 2^{s/2}}{a^2(s-2)} \delta^{s-2}  e^{(s-2)(\log(1/\delta) - \log(2)/2))} \\
&\leq  \frac{2 2^{s/2}}{a^2(s-2)} \delta^{s-2}  2^{-(s-2)/2} \\
&\leq  \frac{1}{a^2(s-2)}.
\end{align*}  
For the second integral, we use that $\lambda_1 \leq 1$, and $\lambda_2 \geq \frac{a}{\sqrt{2 (1+a^2)}}$, hence 
\begin{align*}
\int_{\tau_1}^{\tau_2} \frac{\lambda_1^s(t')}{\lambda_2^2(t')} dt' &\leq \frac{4}{a^2} (\tau_2 - \tau_1)   \\
&\leq  \frac{8}{a^3} \log(1/\eta).
\end{align*}  
Adding these two bounds concludes.
\end{proof}
We now prove Lemma~\ref{lem:upperbound_general_torsion}.
\begin{proof}[Proof of Lemma~\ref{lem:upperbound_general_torsion}]

First, we begin by the observation that, thanks to the triangular inequality, we have 
$$\| Q(\tau) - Q(0) \| \leq \int_0^\tau \| \dot{Q}(t) \| dt $$
with $\dot{Q} = \dot{U} V^\top + U \dot{V}^\top$. Then, denoting $D_\lambda[ij] = (B_\lambda - C_\lambda)[ij] = \frac{1+\lambda_i \lambda_j}{(\lambda_i + \lambda_j)^2} $, we have the time evolution 
\begin{align}
    \dot{Q} &= U[ D_\lambda \circ (\overline{U}^a - \overline{V}^a)] V^\top~,
\end{align}
where $\overline{U}^a$, $\overline{V}^a$ are the antisymmetric components of $\nabla_U L$, $\nabla_V L$ respectively; see Lemma \ref{lem:stiefel_summary}. 
Thus, we can naturally upper bound the torsion for all $t \geq 0$, by 
\begin{align*}
\| Q(t) - Q(0) \| &\leq \int_0^t \| D_\lambda(t') \circ (\overline{U}^a(t') - \overline{V}^a(t')) \| dt'~.
\end{align*}
Now standard norm comparisons gives that  
$$\| D_\lambda(t) \circ (\overline{U}^a(t) - \overline{V}^a(t)) \| \leq 2 \| D_\lambda(t)\|_\infty \| \overline{U}^a(t) - \overline{V}^a(t) \|  \leq \frac{4}{\lambda_2^2} \| \overline{U}^a(t) - \overline{V}^a(t) \| \leq \frac{4}{\lambda_2^2} (\| \overline{U}^a(t)\| +  \|\overline{V}^a(t) \|).  $$
Let us bound  $\| \overline{U}^a(t)\|$. We have
\begin{align*}
    \| \overline{U}^a(t)\| &\leq 2 \| \overline{U}(t)\| \\
    &\leq 2\epsilon \left\|\sum_{|\beta|=s} \nabla_U[\alpha_\beta^g(U)]\alpha_\beta^g(V) \lambda_\beta\right\| \\
    &\leq 2\epsilon \lambda_1^s \sum_{|\beta|=s} \left\|\nabla_U[\alpha_\beta^g(U)]\right\| \left|\alpha_\beta^g(V)\right| \\
    &\leq 2\epsilon \lambda_1^s \sqrt{ \sum_{|\beta|=s} \left\|\nabla_U[\alpha_\beta^g(U)]\right\|^2 } \sqrt{\sum_{|\beta|=s} \left|\alpha_\beta^g(V)\right|^2}.
\end{align*}
The second square root is equal to $\|g\|$. Let us calculate the first square root, for this we calculate, for $i,j=1,2$,
\begin{flalign*}
     && \nabla_U[\alpha_\beta^g(U)][ij]  &=   \left|\int [\partial_i g] (Ux) x_j h_{\beta_1}(x_1)h_{\beta_2}(x_2) d \gamma(x) \right| \\
     && &=   \beta_j \int [\partial_i g] (Ux)  h_{\beta_j - 1}(x_j) h_{\beta_{\bar{j}}}(x_{\bar{j}}) d \gamma(x)  & \text{(where $\bar{j} = j + 1 \text{ in } \mathbb{Z}/2\mathbb{Z}$)}\\
     && &= \beta_j  \alpha_\beta^{\partial_i g}(U)  ~.
\end{flalign*}
Hence, upper bounding the spectral norm with the Frobenius norm, we have \begin{align*}
    \sum_{|\beta|=s} \left\|\nabla_U[\alpha_\beta^g(U)]\right\|^2 &\leq  \sum_{|\beta|=s} 2 s^2 \left([\alpha_\beta^{\partial_1 g}(U)]^2 + [\alpha_\beta^{\partial_2 g}(U)]^2  \right) \\
    &\leq 2 s^2 \|\nabla g\|^2 ~,
\end{align*}
and hence coming back to the bound on  $\| \overline{U}^a(t)\|$, we have: 
\begin{align*}
    \| \overline{U}^a(t)\| &\leq 4 \epsilon s \lambda_1^s \|g\| \|\nabla g\|.
\end{align*}
Finally, as the same calculation holds for the right eigenvectors, we have
\begin{align*}
\| Q(t) - Q(0) \| &\leq 8 \int_0^t \frac{1}{\lambda_2^2(t')}\| \overline{U}^a(t')\|  dt'  \\
& \leq 32 \epsilon s \|g\|  \|\nabla g\| \int_0^t \frac{\lambda_1^s(t')}{\lambda_2^2(t')} dt'~,
\end{align*}
which ends the proof of the Lemma.
\end{proof}

Let us conclude the end of the second phase: the dynamics is at a point where for all $t \geq \tau_2$,  $\lambda_1(t), \lambda_2(t) \in (1-\eta, 1)$, and the eigenvectors have not moved much, ie quantitatively $\|Q_{\tau_2} - Q_0\| \leq \frac{\epsilon K}{a^3}(1 + \log(1/\eta))$, with $K > 0$, independent of $d, \epsilon, \eta$.

\paragraph{Phase III: trapped in spurious valleys}

We can now leverage the presence of spurious valleys in the correlation function $\varphi$ to show that gradient flow on $L(W)$ will be trapped in spurious valleys too:

\begin{lemma}[Spurious Valleys away from mediocrity]
\label{lem:spurious_valleys}
Set $\eta = \frac{\epsilon}{ 4\|f\|^2 + 2s \epsilon \|g\|^2}$. 
Consider the set $\Gamma = \{A R_\theta : A \in \{I_2, \text{diag}(1, -1)\}, |\theta| \leq \pi/N + \gamma\} \subset \mathcal O_2$.\footnote{Recall from the beginning of this section that we may consider~$\gamma = \frac{\pi}{10N}$.}

 Then, the set $\mathcal{A}=\{ M = W W_*^\top = U \Lambda V^\top ; \, UV^\top \notin \Gamma; \lambda_i \geq 1-\eta \}$ is a spurious valley. In other words, gradient flow initialized in $\mathcal{A}$ converges to a sub-optimal local maxima. 
\end{lemma}

\begin{proof}

Consider any path~$W(t)$ with~$W(0) \in \mathcal A$ and of non-decreasing correlation, with~$M(t) = U(t) \Lambda(t) V(t)^\top$, and non-decreasing~$\lambda_i(t)$ for~$i=1,2$ (this is the case for any gradient flow initialized in~$\mathcal A$, per Lemma~\ref{lem:fast_increase}), and
assume by contradiction that~$U(\infty) V(\infty)^\top = I $, i.e., that we found the global maximum.
Below, we define~$\tilde L(M) = L(W)$ the correlation as a function of~$M = W W_*^\top$, and similarly $\tilde L_f(\lambda_1, \lambda_2)$ the correlation of~$f$, which only depends on~$\Lambda$ since~$f$ is radial.

Let us show that when $\Lambda = \text{diag}(\lambda_1, \lambda_2)$ with $\lambda_1, \lambda_2 \geq 1 - \eta$, then
\begin{equation}
    |\tilde L(U \Lambda V^\top) - \tilde L(U V^\top)| \leq C_\epsilon \eta,
\end{equation}
with~$C_\epsilon := 2\|f\|^2 + s \epsilon \|g\|^2$. Indeed, we have
\begin{align*}
    |\tilde L(U V^\top) - \tilde L(U \Lambda V^\top)| &= \left|\sum_{|\beta|=2} (1 - \lambda_\beta) \alpha_\beta^f(U) \alpha_\beta^f(V) + \epsilon \sum_{|\beta| = s} (1 - \lambda_\beta) \alpha_\beta^g(U) \alpha_\beta^g(V)\right| \\
    &\leq (1 - (1 - \eta)^2) \left|\sum_{|\beta|=2} \alpha_\beta^f(U) \alpha_\beta^f(V)\right| + \epsilon (1 - (1 - \eta)^s) \left|\sum_{|\beta|=s} \alpha_\beta^g(U) \alpha_\beta^g(V)\right| \\
    &\leq \eta (2\|f\|^2 + s \epsilon \|g\|^2).
\end{align*}

Now note that we may write $U(t) = A R_{\theta(t)}$ and $V(t) = R_{\theta'(t)}$, with~$A \in \{I_2, \text{diag}(1, -1)\}$, and~$R_{\theta}$, $R_{\theta'}$ rotation matrices. Note that~$A$ cannot change with~$t$, otherwise the path~$W(t)$ would not be continuous (since the eigenvalues of~$M(t)$ would need to switch sign, while being $\geq 1 - \eta > 0$).
Then, using the fact that~$g$ is invariant under reflections, we have
\begin{align*}
\tilde L(U(t) V(t)^\top) &= \tilde L_f(1, 1) + \epsilon \tilde L_g(U(t) V(t)^\top) \\
    &= 2 + \epsilon \E_{z}[g(U(t)z) g(V(t)z)] \\
    &= 2 + \epsilon \E_z[g(R_{\theta(t)}z) g( R_{\theta'(t)}z)]  \\
    &= 2 + \epsilon \varphi(\theta(t) - \theta'(t)).
\end{align*}
Recall from the beginning of this section that~$\varphi$ is even, has its global maximum at~$0$, and that there exists~$0 < \mu < \nu \leq \pi/N + \gamma$ such that~$\varphi(\mu) < 1$ and $\varphi(\nu) = \varphi(\mu) + 1$.

Since~$\eta < \frac{\epsilon}{2 C_\epsilon}$ by assumption, we thus have
\[2 + \epsilon \varphi(\nu) > 2 + \epsilon \varphi(\mu) + 2 C_\epsilon \eta.\]

Then, defining $u(t) = \theta(t) - \theta'(t)$, we have by assumption $|u(0)| > \nu$, since~$U(0)V(0)^\top = A R_{\theta(0) - \theta'(0)} \notin \Gamma$, and $u(\infty) = 0$. By continuity of $u(t)$, this implies that there is $t_1 < t_2$ with $|u(t_1)| = \nu$ and~$|u(t_2)| = \mu$. Yet, we have $L(W(t_1)) \geq 2 + \epsilon \varphi(\nu) - C_\epsilon \eta > 2 + \epsilon \varphi(\mu) + C_\epsilon \eta \geq L(W(t_2))$, which contradicts the fact that the correlation~$L(W(t))$ is non-decreasing.
\end{proof}

Given the three phases described above, we are now ready to finish the proof of Theorem~\ref{thm:planted_failure}.

\begin{proof}[Proof of Theorem~\ref{thm:planted_failure}]

Let~$\delta_0 = 1/N\sqrt{d}$ and~$a_0 = 1/2N\sqrt{\log N}$. By Lemma~\ref{lem:eigenvalues_init} (applied with~$a=2\sqrt{\log N}$ and~$b=1/N$) we have that at initialization,
$\lambda_1(0) = \delta \geq \delta_0$ and~$\lambda_2(0) = a \lambda_1(0)$ with~$a \geq a_0$ are satisfied with probability~$1 - O(1/N)$. Denote this event~$\mathcal E_1$.

Consider the event
\[
\mathcal E_2 = \{M(0) = W(0) W_*^\top = U \Lambda V^\top, U V^\top = AR_{\theta} \text{ with } A \in \{I, \text{diag}(1, -1)\}, |\theta| > \pi / N + 2 \gamma \}
\]
Recall that we may take~$\gamma = \pi/10N = O(1/N)$ in this section. By rotational invariance, the angle~$\theta$ above is uniformly distributed in~$[0, 2\pi]$ for either choice of~$A$, so that the event~$\mathcal E_2$ holds with probability~$1 - O(1/N)$.

From now on, we assume that the initialization is such that~$\mathcal E_1$ and~$\mathcal E_2$ both hold, which happens with probability at least $1 - O(1/N)$ by a union bound.

Let~$0 < \epsilon < 1/s\|g\|^2$ to be fixed later, and consider the associated gradient flow~$W(t)$, with~$M(t) = W(t) W_*^\top = U(t) \Lambda(t) V(t)^\top$. We also write $U(t) V(t)^\top =: A R_{\theta(t)}$, where~$A \in \{I_2, \text{diag}(1, -1)\}$ does not depend on~$t$, since by Lemma~\ref{lem:mediocrity_planted} and~\ref{lem:fast_increase}, $\lambda_i(t) > 0$ for all~$t$, so that~$\det(M(t))$ cannot change sign.

Set~$\eta = \frac{\epsilon}{4 C_\epsilon}$ in order to satisfy Lemma~\ref{lem:spurious_valleys}.
Note that since~$\mathcal E_2$ holds, we have~$|\theta(0)| > \pi / N + 2 \gamma$.
In order to apply Lemma~\ref{lem:spurious_valleys}, we want to ensure~$|\theta(\tau_2)| > \pi/N + \gamma$. A sufficient condition is to show~$|\theta(\tau_2) - \theta(0)| \leq \gamma$.
Note that by Lemma~\ref{lem:upperbound_quantitative_torsion}, we have
\[
\|A R_{\theta(\tau_2)} - A R_{\theta(0)}\| \leq \frac{\epsilon K}{a^3} (1 + \log(1/\eta)) \leq \frac{\epsilon K}{a_0^3} \paren{1 + \log\paren{\frac{1 +2\|f\|^2 + s \|g\|^2}{\epsilon}}} =: \psi(\epsilon)
\]
Noting that
\[
\|A R_{\theta(\tau_2)} - A R_{\theta(0)}\| = \|R_{\theta(\tau_2) - \theta(0)} - I_2\| \geq \frac{1}{\sqrt{2}}(1 - \cos(\theta(\tau_2) - \theta(0))),
\]
it suffices to choose~$\epsilon$ small enough to satisfy
\[
\arccos(1 - \sqrt{2} \psi(\epsilon)) \leq \gamma.
\]
With such a choice of~$\epsilon$ and~$\eta$, we may then apply all the phases outlined above to show that under events~$\mathcal E_1$ and~$\mathcal E_2$, the gradient flow will converge to spurious valleys of the landscape, with value at most $N+2 + 1/3$, which is the desired result.

\end{proof}

\subsection{Proofs of Section \ref{sec:planted_failure2}}
\label{app:planted_failure2}

\subsubsection{Proof of Fact \ref{fact:discrete_unitary_correl}}
\discretecontcorrel*
\begin{proof}
    By definition, we have 
    \begin{align}
        \langle \mathsf{P}_U g_Z, g_Z \rangle &= \sum_{j,j'} Z_j Z_{j'} \langle \mathsf{P}_U \mathsf{P}_{w_j} h_s, \mathsf{P}_{w_{j'}} h_s \rangle \nonumber \\
        &= \sum_{j,j'} Z_j Z_{j'} (w_j^\top U^\top w_{j'})^s~.
    \end{align}
    Assume first that $\det U = 1$. Denoting $w_j = e^{i \theta_j}$, so that $\theta_j = 2\pi j /N$, observe that 
    $w_j^\top U^\top w_{j'} = \cos( \theta_j - \theta_{j'} - \theta)$, hence 
    \begin{align}
        \sum_{j,j'} Z_j Z_{j'} (w_j^\top U^\top w_{j'})^s &= \sum_{l} \sum_j Z_j Z_{j+l} (\cos( 2\pi l /N - \theta) )^s \nonumber \\
        &= \sum_l (\cos( 2\pi l /N - \theta) )^s \varphi_Z(l) \nonumber \\
        &= (\varphi_Z \ast \cos^s)(\theta)~.
    \end{align}
    The case $\det U = -1$ is treated analogously. 
\end{proof}

\subsubsection{Proof of Lemma \ref{lem:negauto}}

\negativecorrel*
\begin{proof}
    Consider $\hat{Z}(\omega) = \sum_l Z_l e^{-2\pi i \omega l / N}$ the discrete Fourier transform of $Z$. By elementary Fourier analysis, we have that 
    $\hat{\varphi}_Z(\omega) = | \hat{Z} (\omega)|^2$ and $\hat{\overline{\varphi}}_Z(\omega) = \hat{Z}(\omega)^2$. We are going to impose that $Z^*$ is even, which (together with the fact that it is real) implies that $\hat{Z}(\omega)$ is even and real, and hence that $\varphi_Z$ and $\overline{\varphi}_Z$ are the same. 
    
    Let us design $Z^*$ from its Fourier coefficients $\hat{Z}(\omega)$, in fact from the squares $P_\omega := (\hat{Z}(\omega))^2$. 
    We want that 
    \begin{equation}
        \forall~k\neq 0~,~\varphi_Z(k) = \sum_\omega (\hat{Z}(\omega))^2 e^{2 \pi i \omega k/ N  } = \sum_\omega P_\omega \cos( 2\pi \omega k/N) = P \cdot C(k) < 0~, 
    \end{equation}
    where we have written $P=(P_0,\ldots, P_{(N-1)/2}) \in \mathbb{R}_{+}^{N/2}$ and $C(k) = (\cos( 2\pi \omega k/N) )_{\omega=0,\ldots (N-1)/2} $, and used the fact that $P_\omega$ is even. 
    In other words, we need to find a vector $P$ in the positive orthant that is negatively correlated with all the cosine vectors $C_1, \ldots, C_{(N-1)/2}$, that we can assume have unit norm without loss of generality from now on.
    
    Let $M = N/2$. Since $\{ C_i \}_{i=0, \ldots M-1}$ forms an orthonormal basis, we can build such $P$ explicitly as follows. Consider $0 < \eta < (2M)^{-1}$ and let $Y=(1-\eta, -\delta, \ldots, -\delta)$ with 
    $\delta = -\sqrt{(M-1)^{-1}( 2\eta - \eta^2)}$, and $P = \mathbf{C}^\top Y$, where $\mathbf{C} = [C_0, \ldots, C_{M-1}]$. 
    We have $\| C_0 - P \|^2 = \| Y - e_1 \|^2 = 2\eta$, so since $\eta < (2M)^{-1}$ we are guaranteed that $P$ is in the positive orthant. Moreover, observe that $P \cdot C_i = -\delta < 0$ for $i=1, \ldots M-1$.     
\end{proof}

\subsubsection{Proof of Proposition \ref{prop:badmaxima}}

\badmaxim*
\begin{proof}
     Suppose first that $U, V$ are such that $\det(U V^\top)=1$, and write $U = R_\theta$, $V = R_\eta$. The loss writes
     \begin{align}
       L(W) &= \ell(\theta, \eta, \lambda_1, \lambda_2) = \sum_{j,j'} Z_j Z_{j'} \left( w_j^\top M w_{j'} \right)^s \nonumber \\
    &= \sum_{j,j'} Z_j Z_{j'} \left[ \lambda_2 w_j^\top R_{\theta-\eta} w_{j'} + (\lambda_1-\lambda_2) ( w_j \cdot \theta) (w_{j'} \cdot \eta) \right]^s \nonumber \\
    &= \sum_{k=0}^s \binom{s}{k} \lambda_2^k (\lambda_1-\lambda_2)^{s-k} \left[\sum_{j,j'} Z_j Z_{j'}  \left(w_j^\top R_{\theta-\eta} w_{j'} \right)^k  ( w_j \cdot \theta)^{s-k} (w_{j'} \cdot \eta)^{s-k} \right]~\nonumber \\
    &:= \sum_{k=0}^s \binom{s}{k} \lambda_2^k (\lambda_1-\lambda_2)^{s-k} \chi_{k}(\theta, \eta)~.
     \end{align}
When $\det(U V^\top) = -1$, writing $V = \tilde{R} R_\eta$, where $\tilde{R}=\begin{bmatrix}
    1 & 0 \\ 0 & -1
\end{bmatrix}$, we have the analogous representation
\begin{align*}
    L(W) &= \sum_{k=0}^s \binom{s}{k} \lambda_2^k (\lambda_1-\lambda_2)^{s-k} {\chi}_{k}(\theta, \eta)~,
\end{align*}
since $g$ is invariant to reflection, $g = \mathsf{P}_{\tilde{R}}g$,  thanks to the fact that $Z$ is even. 

 Let $a^*\neq b^*$ such that $Z_{a^*} Z_{b^*} > 0$ and let $\theta = w_{a^*}$, $\eta = w_{b^*}$. We fix $\lambda_1 = 1$ and study the loss along the $\lambda_2$ direction:
    \begin{equation}
        \tilde{\ell}( \lambda ) = \ell( w_{a^*},  w_{b^*}, 1, \lambda) = \sum_{k=0}^s \binom{s}{k} \lambda^k (1-\lambda)^{s-k} \chi_{k}(w_{a^*}, w_{b^*})~.
    \end{equation}
We have 
\begin{align}
\tilde{\ell}(0)  &=  \chi_0( w_{a^*}, w_{b^*}) \nonumber \\
&= \left(\sum_j Z_j (w_j \cdot w_{a^*})^s \right)\left(\sum_{j'} Z_{j'} (w_{j'} \cdot w_{b^*})^s \right) \nonumber \\
&= Z_{a^*} Z_{b^*} - O( N \cos(2\pi/N)^s) > 0~,
\end{align}
while 
\begin{align}
    \tilde{\ell}(1) &= \chi_s( w_{a^*}, w_{b^*})  = \varphi_{g}( w_{a^*} - w_{b^*}) < 0~,
\end{align}
so the maxima of $\tilde{\ell}(\lambda)$ must be in $[0,c)$ for some $c<1$ since $\tilde{\ell}$ is continuous. 
Let $\lambda^* \in [0,c)$ be one such local maxima, and consider $M^*=(w_{a^*}, w_{b^*}, 1, \lambda^*)$. 
We claim the following:
\begin{lemma}[Topological Obstruction]
\label{claim:spurious_valleys_bis}
    There does not exist a continuous path $\gamma: t \mapsto \gamma(t) $ such that $\gamma(0) = M^*$, $\gamma(1) \in \mathcal{O}_2$, and such that $t \mapsto \ell( \gamma(t))$ is non-decreasing. 
\end{lemma}
Before proving this Lemma, let us show how it allows us to establish the existence of a bad local maxima with $\lambda_2 < 1$. Indeed, if all local maxima were at $\lambda_2 = 1$, then for any point $M \notin \mathcal{O}_2$ we could build a non-decreasing path starting at $M$ and arriving at $\mathcal{O}_2$. But this is ruled out by Lemma \ref{claim:spurious_valleys_bis}, and so we conclude that there must exist a bad local maxima with $\lambda_2 < 1$.

\end{proof}
\begin{proof}[Proof of Lemma \ref{claim:spurious_valleys_bis}]
Let $\xi > 0$ and consider $B_\xi:= \{ (\theta, \eta, \lambda_1, \lambda_2); \max( |\theta \,\text{mod} 2\pi/N|, |\eta \,\text{mod} 2\pi/N| ) \leq \xi \}$. 
Observe that if $M=( \theta, \eta, \lambda_1, \lambda_2) \notin B_\xi$, then $|\chi_k(\theta, \eta)| \leq \|Z\|_1^2 (1-\xi)^s$, so that
\begin{align}
    L(M) &= \sum_{k=0}^s \binom{s}{k} \lambda_2^k (\lambda_1-\lambda_2)^{s-k} \chi_{k}(\theta, \eta) \leq \lambda_1^s \|Z\|_1^2 (1-\xi)^s~.
\end{align}
Therefore, if $Z_{a^*} Z_{b^*} - O(N \cos(2\pi/N)^s) > \|Z\|_1^2 (1-\xi)^s$, any path exiting $B_\xi$ cannot be non-decreasing. 

Now, observe that $B_\xi$ consists of $N^2$ disconnected components, corresponding to the pairs of $N$-th roots of unity $w_j$. The connected component associated with $w_{a^*}, w_{b^*}$ intersects $\mathcal{O}_2$ in rotations of the form $R_{w_a^* - w_b^* + \xi'}$, with $|\xi'| \leq \xi$. Since $\varphi_g(\theta)$ is only positive for $|\theta| \leq N^{-1}$ and $w_a^* - w_b^* \neq 0$, we conclude that paths staying in this connected component of $B_\xi$ cannot be non-decreasing either. Finally, the only option for a path to switch continuously between connected components is to go through $\lambda_1 = \lambda_2 = 0$, but this point satisfies $L(0)=0$, which would again decrease the loss. We therefore conclude that there are no non-decreasing continuous paths from $M^*$ to the boundary $\mathcal{O}_2$. 
\end{proof}

%% file: Appendix/App_Grassmannians.tex
In this section we recall basic facts on the Stielel and Grassmaniann manifold. We refer to~\cite{edelman1998geometry,bendokat2020grassmann} for gentle introductions on this topic. 

\subsection{Statistics on special manifold}
\label{subsecapp:statistics_special}

For a detailed presentation of this, we refer to the book~\cite{chikuse2003statistics}.

\subsubsection{Proof of Lemma~\ref{lem:eigenvalues_init}}
\label{subsubsecapp:proof_initialization}

We first prove~\ref{lem:eigenvalues_init} on the estimation of the singular vectors of $M = W_*^\top W$. Recall that, if $W\sim\mathrm{Unif}(\G(d,q))$ they follow the following multivariate distribution
\begin{align}
    p(\lambda_1, \cdots, \lambda_q) = Z_q^{-1} \prod_{i < j} |\lambda_i^2 - \lambda_j^2| \prod_{ i = 1}^q  (1 - \lambda_i^2)^{( d- 2q - 1)/2} \mathds{1}_{ 0 \leq \lambda_q \leq \dots \leq \lambda_1 \leq 1}~, 
\end{align}
with normalization constant $Z_q = \frac{\Gamma_q^2(q/2)}{\pi^{q^2 / 2}} \frac{\Gamma_q((d-q)/2)}{\Gamma_q(d/2)}$. 

\paragraph{Upper bound on the spectrum.} For any $a > 0$, we calculate an estimate of  $\mathbb{P}(\lambda_1 \geq a / \sqrt{d})$. We have
\begin{align*}
    \mathbb{P}(\lambda_1 \geq a / \sqrt{d}) &=  Z_q^{-1}\int_{\lambda_1 \geq \frac{a}{\sqrt{d}} } \int_{\lambda_2} \dots \int_{\lambda_q}  \prod_{i < j} |\lambda_i^2 - \lambda_j^2| \prod_{ i = 1}^q  (1 - \lambda_i^2)^{( d- 2q - 1)/2} \mathds{1}_{ 0 \leq \lambda_q \leq \dots \leq \lambda_1 \leq 1} d\lambda_1 \dots d \lambda_q~.
   \end{align*}
We remark that any differences $|\lambda_1^2 - \lambda_j^2| \leq \lambda_1^2$ for $j \in \llbracket 2, q \rrbracket$, thanks to the fact that the eigenvalues are sorted. Hence, we can separate the integrals and have
\begin{align*}
    \mathbb{P}(\lambda_1 \geq a / \sqrt{d}) & \leq  Z_q^{-1} \int_{\frac{a}{\sqrt{d}} } \lambda_1^{2 (q-1)} (1 - \lambda_1^2)^{( d- 2q - 1)/2}  \\
    & \hspace{2cm} \int_{\lambda_2} \dots \int_{\lambda_q}  \prod_{1<i < j} |\lambda_i^2 - \lambda_j^2| \prod_{ i = 2}^q(1 - \lambda_i^2)^{( d- 2q - 1)/2} \mathds{1}_{ 0 \leq \lambda_q \leq \dots \leq \lambda_2 \leq \lambda_1 \leq 1} d\lambda_1 \dots d \lambda_q~.
\end{align*}
We remark that as $\lambda_1$ is larger that the other eigenvalues, we have  $$\prod_{ i = 2}^q(1 - \lambda_i^2)^{( d- 2q - 1)/2} = \prod_{ i = 2}^q(1 - \lambda_i^2)^{( d- 2(q-1) - 1)/2} \prod_{ i = 2}^q(1 - \lambda_i^2)^{-1} \leq (1 - \lambda_1^2)^{-(q - 1)}  \prod_{ i = 2}^q(1 - \lambda_i^2)^{( d- 2(q-1) - 1)/2} .$$
Hence, we can upperbound the probability as follows, 
\begin{align*}
    \mathbb{P}(\lambda_1 \geq a / \sqrt{d}) & \leq  Z_q^{-1} Z_{q-1}\int^1_{\frac{a}{\sqrt{d}} }  \lambda^{2 (q-1)} (1 - \lambda^2)^{( d- 4q + 1)/2} d\lambda.
\end{align*}
Hence, we first begin by estimating the ratio of the normalizing constants. Recall that we have the formula of the multivariate $\Gamma$ function \cite[p.62]{muirhead2009aspects}, for all $a > 0$,
\begin{align*}
    \Gamma_q(a/2) = \pi^{q(q-1)/4} \prod_{i = 1}^q \Gamma\left( \frac{a - i + 1}{2} \right).
\end{align*}
Hence, we have the calculation
\begin{align*}
    \frac{Z_{q-1}}{Z_q} &= \frac{\Gamma_{q-1}^2((q-1)/2)}{\pi^{(q-1)^2 / 2}} \frac{\Gamma_{q-1}((d-q+1)/2)}{\Gamma_{q-1}(d/2)} \frac{\pi^{q^2 / 2}}{\Gamma_q^2(q/2)} \frac{\Gamma_q(d/2)}{\Gamma_q((d-q)/2)} \\
    &= \pi^{(q - 1/2 )} \left(\frac{\Gamma_{q-1}((q-1)/2)}{\Gamma_{q}(q/2)}\right)^2 \frac{\Gamma_{q-1}((d-q+1)/2)}{\Gamma_q((d-q)/2)} \frac{\Gamma_q(d/2)}{\Gamma_{q-1}(d/2)} \\
    &= \frac{\pi^{(q - 1/2 )}}{\Gamma^2(q/2)} \frac{\Gamma^2((d-q + 1)/2)}{\Gamma((d-2(q-1))/2) \Gamma((d-2q + 1)/2)}.
\end{align*}

Then, recall that we have the following inequality, provided by \cite[equality 3.2]{laforgia2013some} -dating back at least to \cite{gautschi1959some}, for all $a > 0$
\begin{align}
\label{eq:inequality_gautschi}
     \frac{\Gamma(a)}{\Gamma(a-1/2)} \leq \sqrt{a}~.
\end{align}
Hence,  we have that $$ \frac{\Gamma^2((d-q + 1)/2)}{\Gamma((d-2(q-1))/2) \Gamma((d-2q + 1)/2)} \leq \frac{\Gamma^2((d-q + 1)/2)}{\Gamma^2((d-2(q-1))/2) } \sqrt{\frac{d}{2}}.$$
We estimate the latter ratio, taking $a = d/2$ and $b = (q-1)/2$, we have  
\begin{align*}
    \frac{\Gamma(a - b)}{\Gamma(a - 2b) } &=  \frac{\Gamma(a - b) \Gamma(a - b - 1/2) \Gamma(a - b - 1) \cdots \Gamma(a - b - (b-1))  \Gamma(a - b - (b-1) - 1/2) }{\Gamma(a - b - 1/2) \Gamma(a - b - 1)  \cdots \Gamma(a - b - (b-1))  \Gamma(a - b - (b-1) - 1/2) \Gamma(a - 2b)} \\
    &\leq \sqrt{(a-b)(a-b-1/2)(a-b-1) \cdots (a-b -(b-1)) (a-b -(b-1) - 1/2)} \\
    &\leq a^{(2b)/2} \\
    &\leq \left(\frac{d}{2}\right)^{(q-1)/2}~.
\end{align*}
Hence, finally, 
\begin{align*}
    \frac{Z_{q-1}}{Z_q} \leq \frac{\pi^{q-1/2}}{\Gamma^2(q/2)}  \left(\frac{d}{2}\right)^{q-1}\sqrt{\frac{d}{2}} \\
    \frac{Z_{q-1}}{Z_q} \leq \frac{\pi^{q}}{2^{q}\Gamma^2(q/2)}  d^{q-1/2}.
\end{align*}
It remains to estimate the integral
\begin{align*}
   \int^1_{\frac{a}{\sqrt{d}} }  \lambda^{2 (q-1)} (1 - \lambda^2)^{( d- 4q + 1)/2} d\lambda &= \frac{1}{d^{q-1/2}} \int^{\sqrt{d}}_a  t^{2 (q-1)} \left(1 - \frac{t^2}{d}\right)^{( d- 4q + 1)/2} d t  \\
   &\leq \frac{1}{d^{q-1/2}} \int^{\sqrt{d}}_a  t^{2 (q-1)} \exp\left(- \frac{ d- 4q + 1}{2d} t^2\right) d t  \\
   &\leq \frac{1}{d^{q-1/2}} \int^{\sqrt{d}}_a  t^{2 (q-1)} \exp\left(- \frac{t^2}{2}\right)  \exp\left(\frac{ 4q - 1}{2d} t^2\right) d t  \\
   &\leq \frac{e^{2q - 1/2}}{d^{q-1/2}} \int^{\infty}_a  t^{2 (q-1)} \exp\left(- \frac{ t^2}{2}\right) d t  ~.
\end{align*}
But for $a \geq 4q$, we have the inequality $t^{2 (q-1)} \exp\left(- \frac{ t^2}{2}\right) \leq t \exp\left(- \frac{ t^2}{4}\right)$. Hence, 
\begin{align*}
   \int^1_{\frac{a}{\sqrt{d}} }  \lambda^{2 (q-1)} (1 - \lambda^2)^{( d- 4q + 1)/2} d\lambda &\leq  \frac{e^{2q - 1/2}}{d^{q-1/2}} \int^{\infty}_a t \exp\left(- \frac{ t^2}{4}\right) d t \\
    &\leq  \frac{4 e^{2q - 1/2}}{d^{q-1/2}}  e^{-a^2/4}.
\end{align*}
Hence, finally we have
\begin{align*}
    \mathbb{P}(\lambda_1 \geq a / \sqrt{d}) & \leq  b_q  e^{-a^2/4}~,
\end{align*}
where $b_q = \frac{\pi^q e^{2q}}{2^{q}\Gamma^2(q/2)}$.

\paragraph{Lower bound on the spectrum.} For any $b > 0$, we now calculate an estimate of  $\mathbb{P}(\lambda_q \leq  b/ \sqrt{d})$. We have like for the previous calculation,
\begin{align*}
    \mathbb{P}(\lambda_q \leq  b/ \sqrt{d}) &=  Z_q^{-1}\int_{\lambda_1} \dots \int_{\lambda_q \leq \frac{b}{\sqrt{d}}}  \prod_{i < j} |\lambda_i^2 - \lambda_j^2| \prod_{ i = 1}^q  (1 - \lambda_i^2)^{( d- 2q - 1)/2} \mathds{1}_{ 0 \leq \lambda_q \leq \dots \leq \lambda_1 \leq 1} d\lambda_1 \dots d \lambda_q~.
   \end{align*}
We adopt the same technique as previously, giving a special treatment this time on $\lambda_1$ and $\lambda_q$. 
\begin{align*}
    \mathbb{P}(\lambda_q \leq  b/ \sqrt{d}) &=  Z_q^{-1}\int_{\lambda_1} \dots \int_{\lambda_q \leq \frac{b}{\sqrt{d}}}  \prod_{i < j} |\lambda_i^2 - \lambda_j^2| \prod_{ i = 1}^q  (1 - \lambda_i^2)^{( d- 2q - 1)/2} \mathds{1}_{ 0 \leq \lambda_q \leq \dots \leq \lambda_1 \leq 1} d\lambda_1 \dots d \lambda_q \\
    &\leq  Z_q^{-1}\int_{\lambda_1} \dots \int_{\lambda_q \leq \frac{b}{\sqrt{d}}}  \prod_{2\leq i < j\leq q} |\lambda_i^2 - \lambda_j^2| \lambda_1^{2(q-1) + 2 (q-2)} \prod_{ i = 1}^q  (1 - \lambda_i^2)^{( d- 2q - 1)/2} \mathds{1}_{ 0 \leq \lambda_q \leq \dots \leq \lambda_1 \leq 1} d\lambda_1 \dots d \lambda_q \\
    &\leq  Z_q^{-1}\left[\int_{\lambda_1}\lambda_1^{2(q-1) + 2 (q-2)} (1 - \lambda_1^2)^{( d- 4q + 1)/2} d\lambda_1 \right] \left[\int_{\lambda_q \leq \frac{b}{\sqrt{d}} }(1 - \lambda_q^2)^{( d- 2q - 1)/2} d\lambda_q \right]  \\
    & \hspace{1.1cm} \left[ \int_{\lambda_2} \dots \int_{\lambda_{q-1} }   \prod_{2\leq i < j\leq q} |\lambda_i^2 - \lambda_j^2| \prod_{ i = 2}^{q-1}  (1 - \lambda_i^2)^{( d- 2(q-1) - 1)/2} \mathds{1}_{ 0 \leq \lambda_{q-1} \leq \dots \leq \lambda_2 \leq 1} d\lambda_2 \dots d \lambda_{q-1} \right] \\
     &\leq  Z_q^{-1} Z_{q-2} \left[\int_{\lambda_1}\lambda_1^{2(2q-3)} (1 - \lambda_1^2)^{( d- 4q + 1)/2} d\lambda_1 \right] \left[\int_{\lambda_q \leq \frac{b}{\sqrt{d}} }(1 - \lambda_q^2)^{( d- 2q - 1)/2} d\lambda_q \right]  \\
     &\leq  Z_{q-1}Z_q^{-1} Z_{q-1}^{-1}Z_{q-2} \left[\int_{\lambda_1}\lambda_1^{2(2q-3)} (1 - \lambda_1^2)^{( d- 4q + 1)/2} d\lambda_1 \right] \left[\int_{\lambda_q \leq \frac{b}{\sqrt{d}} }(1 - \lambda_q^2)^{( d- 2q - 1)/2} d\lambda_q \right].
   \end{align*}
Let us bound these three terms separately. For the ratios recall that $\frac{Z_{q-1}}{Z_q} \leq \frac{\pi^{q}}{2^{q}\Gamma^2(q/2)}  d^{q-1/2}$ is also valid for $q-1$, hence, we have finally that $$Z_{q-1}Z_q^{-1} Z_{q-1}^{-1}Z_{q-2} \leq \frac{\pi^{q}}{2^{q}\Gamma^2(q/2)}  d^{q-1/2} \frac{\pi^{q-1}}{2^{q-1}\Gamma^2((q-1)/2)}  d^{q - 3/2} \leq\frac{\pi^{2q}}{4^q \Gamma^4((q-1)/2)} d^{2(q-1)}. $$
Second, 
\begin{align*}
    \int_{0}^1\lambda^{2(2q-3)} (1 - \lambda^2)^{( d- 4q + 1)/2} d\lambda &\leq \int_{0}^1\lambda^{2(2q-3)} \exp(-( d- 4q + 1)\lambda^2/2)  d\lambda \\
    &\leq \int_{0}^1\lambda^{2(2q-3)} \exp(-d \lambda^2/2)  \exp((4q-1)\lambda^2/2)  d\lambda \\
    &\leq \frac{e^{2q}}{d^{2q-3}} \frac{1}{\sqrt{d}} \int_{0}^{\sqrt{d}}t^{2(2q-3)} \exp(-t^2/2)  dt \\
    &\leq \frac{e^{2q} 2^{2q} \Gamma(2q) }{d^{2q-3 + 1/2}}~,
\end{align*}
where the last bound follows from the classic bound on Gaussian moments. And finally, 
\begin{align*}
     \int_{0}^{\frac{b}{\sqrt{d}}}  (1 - \lambda^2)^{( d- 2q - 1)/2} d\lambda &\leq \int_{0}^{\frac{b}{\sqrt{d}}} \exp(- \lambda^2( d- 2q - 1)/2) d\lambda \\
     &\leq \int_{0}^{\frac{b}{\sqrt{d}}} \exp(- \lambda^2 d/2) \exp((2q + 1)/2) d\lambda \\
     &\leq \frac{2 e^{q}}{\sqrt{d}}\int_{0}^{b} \exp(- u^2 /2) du \\
     &\leq \frac{2 e^{q}}{\sqrt{d}} b~.
\end{align*}


Mixing every thing, we have,
\begin{align*}
    \mathbb{P}(\lambda_q \leq  b/ \sqrt{d}) &\leq \frac{\pi^{2q}}{4^q \Gamma^4((q-1)/2)} d^{2(q-1)} \frac{e^{2q} 2^{2q} \Gamma(2q) }{d^{2q-3 - 1/2}}   \frac{2 e^{q}}{\sqrt{d}} b ~,
    &\leq c_q b~,
   \end{align*}
where $c_q = 2 \frac{\pi^{2q}  e^{3q} \Gamma(2q) }{ \Gamma^4((q-1)/2)}$.

\paragraph{Final bound} Let recall that we defined $b_q = \frac{\pi^q e^{2q}}{2^{q}\Gamma^2(q/2)}$ and $c_q = 2 \frac{\pi^{2q}  e^{3q} \Gamma(2q) }{ \Gamma^4((q-1)/2)}$ $c_q$, then we have
\begin{align*}
\mathbb{P}\left( \frac{b}{ \sqrt{d}} \leq \lambda_q \leq \lambda_1 \leq \frac{a}{\sqrt{d}}\right) & \geq 1 - \mathbb{P}\left(\lambda_q \leq  \frac{b}{ \sqrt{d}}  \right) - \mathbb{P}\left( \lambda_1 \geq \frac{a}{\sqrt{d}}\right) \\
& \geq 1 - c_q b - b_q e^{-a^2/4}~,
\end{align*}
which proves the Lemma.

\clearpage
\subsection{Representation of the Grassmaniann manifold}
\label{subsecapp:representation_special}
\begin{lemma}
    \label{lem:grasm_representation}
    Any smooth curve $t \mapsto Z(t) \in \mathcal{G}(q,\tau)$ such that $Z(0) = \mathrm{Sp}(G_W)$ can be represented as
    \begin{equation}
    \label{eq:sux}
        Z(t) = \mathrm{Sp}(G_W) Q(t)~,
    \end{equation}
    with $t \mapsto Q(t) \in \mathcal{O}_q$ a smooth curve satisfying $Q(0)=I_q$. 
\end{lemma}
\begin{proof}
Indeed, the Grassmann manifold $\mathcal{G}(q,\tau)$ has an algebraic quotient structure $\mathcal{G}(q,\tau) \cong \mathcal{O}_q /\!\! \sim\,$, where 
    $U \sim U'$ iff $U' = R U$ for $R \in \mathcal{H}_{q,\tau}$, where $\mathcal{H}_{q,\tau}$ is the subgroup given by
\begin{align}
\mathcal{H}_{\tau, q} = \left\{ R \in \mathcal{O}_q; R = \begin{pmatrix} R_\tau & 0 \\ 0 & R_{q-\tau}\end{pmatrix};\, R_\tau \in \mathcal{O}_\tau, R_{q-\tau} \in \mathcal{O}_{q-\tau} \right\} \cong \mathcal{O}_\tau \times \mathcal{O}_{q-\tau}~.
\end{align}
Given any $Z \in \mathcal{G}(q,\tau)$, represented as a $q \times \tau$ matrix, 
consider $Z^\perp$ an orthonormal complement of $Z$ and $\bar{Z} = [Z ; Z^\perp] \in \mathcal{O}_q$. We can thus 
identify $Z$ with the equivalence class $\{\!\{ \bar{Z} \}\!\} := \{ R \bar{Z}; R \in \mathcal{H}_{q,\tau} \}$. 
Moreover, thanks to this quotient structure, we can embed the tangent space of $\mathcal{G}(q,\tau)$ at any point $Z$ with the horizontal space at $\bar{Z}$, a subspace of the tangent space of $\mathcal{O}_q$ consisting in matrices of the form
$$\begin{pmatrix}
  0 & -B^\top \\ B & 0  
\end{pmatrix} \bar{Z}~.$$
As a result, after fixing the representative $\bar{Z}$ at $t=0$, we can embed the smooth curve $Z(t) \in \mathcal{G}(q,\tau)$ into a smooth curve in $\mathcal{O}_q$ by only performing horizontal motion, which yields (\ref{eq:sux}). 
\end{proof}

\subsection{Gradient Computations on Stiefel and Grassmann Manifolds}
\label{appsec:gradients}

From \cite{townsend2016differentiating} we have the following useful facts:
\begin{lemma}[Chain Rule for Singular Value Decompositions, Reverse Mode] 
\label{lem:towsend}
Let $A$ be a square matrix,
$F(A) = f( U, \Lambda, V)$ where $A = U \Lambda V^\top$ is the singular-value decomposition of $A$. Then $\overline{A}= \nabla_A F$ is given by 
\begin{equation}
    \overline{A} = U \left[ \overline{\Lambda} +  \left( F \circ \overline{U}^a \right) \Lambda + \Lambda \left( F \circ {\overline{V}^a} \right)   \right] V^\top ~,
\end{equation}
    where $F_{ij} = (\lambda_j^2 - \lambda_i^2)^{-1}$ for $i \neq j$ and $0$ otherwise,  
    $\overline{Z}^{a} = Z^\top \overline{Z} - \overline{Z}^\top Z$ and $\overline{\Lambda}_{ii} = \partial_{\lambda_i} f(U, \Lambda, V)$.     
\end{lemma}

\begin{lemma}[Chain Rule for Singular Value Decomposition, Forward Mode]
\label{lem:towsend2}
Under the same notations as above, and assuming again that $A$ is a squared matrix, the differentials $dU$, $dV$ and $d\Lambda$ are given by 
\begin{align}
    dU &= U\left( F \circ ( U^\top dA V \Lambda + \Lambda V^\top dA^\top U) \right)~,\\
    dV &= V\left( F \circ ( \Lambda U^\top dA V + V^\top dA^\top U \Lambda) \right)~,\\
    d\Lambda &= I \circ ( U^\top dA V )~.
\end{align}
\end{lemma}

\paragraph{Stiefel Setting}
\label{appsec:stiefelgrad}

Let $F(Y)$ be a smooth function defined on the Stiefel manifold $\mathcal{S}(d,q)$, and $\overline{F} \in \mathbb{R}^{d \times q }$ the gradient of the Euclidean retraction, ie $\overline{F}_{ij} = \frac{\partial F}{\partial Y_{ij}}$. Then the Stiefel gradient is given by \cite[Eq.(2.53)]{edelman1998geometry} 
\begin{equation}
\label{eq:stiefelgrad}
\nabla^{\mathcal{S}}_Y F = \overline{F} - Y \overline{F}^\top Y~.    
\end{equation}

\paragraph{Grassman Setting}
\label{appsec:grassgrad}
The analog for the Grassman setting is given by \cite[Eq.(2.70)]{edelman1998geometry}
\begin{equation}
\label{eq:grassgrad}
\nabla^{\mathcal{G}}_Y F = (I - Y Y^\top) \overline{F} ~.    
\end{equation}

%% file: Appendix/App_ODE_technical.tex
\subsection{A technical differential inequality \textit{à la Gronwall}} 

\begin{lemma}[Gronwall Lemma for non-decreasing functions]
\label{lem:gromwall_adapted}
Let $t \to u_t$ be a non-decreasing positive function satisfying, for all $t \geq 0$, the inequality 
$\dot{u_t} \geq u_t^a v_t, $
with $a > 1$ and $(v_t)_t$ a positive function such that $\int_0^T \mathbf{1}(v_t \geq C) \dd t \geq \tau$ for some $T, C, \tau > 0$. Then we have
\begin{align}
    u_T \geq \left[\frac{1}{u_0^{a-1}} - (a-1) C  \tau \right]^{-\frac{1}{a-1}}~.
\end{align}
Moreover, in the case where $a = 1$, we have 
\begin{align}
    u_T \geq u_0 e^{C \tau}~.
\end{align}
Finally, if $ \dot{u}_t \leq \tilde{C} u_t^a$, then
\begin{align}
\label{eq:gronwall_upper}
    u_T &\leq \left[\frac{1}{u_0^{a-1}} - (a-1)\tilde{C}T \right]^{-1/(a-1)}~
\end{align}
for all $0<T< u_0^{1-a} \tilde{C}^{-1} (a-1)^{-1}$.

\end{lemma}
\begin{proof}
    Let us apply to the differential inequality the classical Gronwall Lemma. We have for $T > 0$, 
\begin{align*}
    u_T \geq \left[\frac{1}{u_0^{a-1}} - (a-1) \int_0^T v_s \dd s \right]^{-\frac{1}{a-1}}.
\end{align*}
Noticing that the right term is a non-decreasing function of $\int_0^T v_s \dd s$  and that we have the inequality
\begin{align*}
   \int_0^T v_s \dd s \geq  \int_0^T v_s \mathbf{1}(v_t \geq C) \dd s \geq C \int_0^T \mathbf{1}(v_t \geq C) \dd s \geq C \tau, 
\end{align*}
we prove the claimed inequality.
The proof for the case $a = 1$ and of the reverse inequality follows exactly the same argument.
\end{proof}

\subsection{Closed-form intergration of Bernoulli's ODE}
\begin{lemma}
    \label{lem:bernoulli_ODE}
    Let $a\in \R$, $\delta > 0$ and $s \in \N^*$. If $y(0) = \delta$ and $(y_t)_{t \geq 0}$ verifies the dynamics
    \begin{align}
        \dot{y}(t) = y + a y^s,
    \end{align}
    Then we have the solution, $\forall t \in (0, \tau)$, 
    \begin{align}
        y(t) = \left[\left(\frac{1}{\delta^{s-1}} + a\right) e^{-(s-1)t}- as\right]^{-\frac{1}{s-1}},
    \end{align}
    where $\tau$ is the maximal time of existence the ODE.
\end{lemma}
\begin{proof}
    As $y \equiv 0$ is a solution to the ODE, thus by Cauchy-Lipschitz theorem, $y > 0$ and there exists a solution up to a certain exploding time $\tau$. Then, starting with the ODE $ \dot{y}(t) = y + a y^s$ and multiplying it by $y^{-s}$, we have $ y^{-s}\dot{y}(t) = y^{1 - s} + a$. Set $u = y^{1-s}$, we have $$ \frac{1}{1-s} \dot{u} = u + a~,$$
    that integrates in close form as 
    $$ u(t) =  \left(\frac{1}{\delta^{s-1}} + a\right) e^{-(s-1)t}- a ~.$$
    Going back to $y$ gives the solution announced in the lemma.
\end{proof}